\documentclass{article} 
\usepackage[utf8]{inputenc}
\usepackage[final]{colm2025_conference}

\usepackage{microtype}
\usepackage{hyperref}
\usepackage{url}
\usepackage{booktabs}
\usepackage[normalem]{ulem}
\usepackage{lineno}

\usepackage[linesnumbered,ruled,vlined]{algorithm2e}
\usepackage{amsmath} 
\usepackage{amssymb} 
\usepackage{amsfonts}
\usepackage{wrapfig}
\usepackage[most]{tcolorbox}
\newcount\AlgoSel
\newcommand{\IMPROVE}{\operatorname{\mathtt{IMPROVE}}}
\usepackage{amsthm}
\usetikzlibrary{arrows.meta, positioning, fit, calc, shapes.geometric}

\usepackage{tikz}
\usetikzlibrary{mindmap}
\usepackage{multirow}
\usepackage{xcolor}
\usepackage[edges]{forest}
\usepackage{csquotes}
\usepackage{fontawesome5}
\usepackage[T1]{fontenc}

\usepackage[dvipsnames]{xcolor} 
\definecolor{hidden-red}{RGB}{205, 44, 36}
\definecolor{hidden-blue}{RGB}{194,232,247}
\definecolor{hidden-orange}{RGB}{243,202,120}
\definecolor{hidden-green}{RGB}{34,139,34}
\definecolor{hidden-pink}{RGB}{255,245,247}
\definecolor{hidden-black}{RGB}{20,68,106}
\definecolor{lightBlue}{RGB}{220,235,255}

\definecolor{sia-color}{HTML}{002060}
\definecolor{fm-blue}{HTML}{9CC3E6}
\definecolor{sf-green}{HTML}{A9D18E}
\definecolor{eval-purple}{HTML}{D8B4E2}
\definecolor{tab-blue}{HTML}{220090}
\definecolor{roseviolet}{HTML}{B76C8F}
\definecolor{VOrange}{HTML}{ED7D31}
\definecolor{app-purple}{RGB}{182,182,244}

\definecolor{darkblue}{rgb}{0, 0, 0.5}
\hypersetup{colorlinks=true, citecolor=darkblue, linkcolor=darkblue, urlcolor=darkblue}
\usepackage[framemethod=TikZ]{mdframed}
\definecolor{algobg}{RGB}{245,245,245}

\definecolor{ChipBack}{RGB}{242,245,250}  
\definecolor{ChipEdge}{RGB}{210,220,230}
\newtcbox{\flowchip}{
  on line, tcbox raise base,
  boxrule=0pt, arc=5pt, outer arc=5pt,
  colback=ChipBack, colframe=ChipEdge, coltext=sia-color,
  left=1pt, right=1pt, top=1pt, bottom=1pt,
  fontupper=\ttfamily 
}
\definecolor{mcwarm}{RGB}{150, 90, 60}

\usepackage{enumitem}

\newtcolorbox{descbox}[1][]{
  enhanced, breakable,
  drop shadow,
  colback=gray!5,
  colframe=white,
  boxrule=0pt, arc=4pt,
  left=6pt, right=6pt, top=2pt, bottom=2pt,
  #1
}

\newlist{fdesc}{description}{1}
\setlist[fdesc]{style=nextline, leftmargin=0pt, labelsep=0pt, itemsep=0.8ex}

\newlength{\desclabelwd}

\newtcolorbox{ttcolorbox}[1][]{colframe=sia-color, colback=sia-color!4!white, title=#1}

\newcommand{\tagbox}[2][teal!70!black]{%
  \tcbox[
    on line,
    enhanced,
    arc=2.5pt,         
    boxrule=0pt,      
    colback=#1,         
    colframe=#1,        
    coltext=white,     
    boxsep=0.2ex,      
    left=0.5ex,right=0.5ex, top=0.1ex,bottom=0.1ex,
    tcbox raise base   
  ]{\sffamily\bfseries #2}
}

\usepackage{makecell}
\usepackage[table]{xcolor}
\usepackage{tabularx}
\newcolumntype{Y}{>{\centering\arraybackslash}X}
\usepackage{adjustbox}  

\DeclareFontShape{T1}{ppl}{m}{scit}{<->ssub * ppl/m/sc}{}

\newcommand{\cmark}{\ensuremath{\checkmark}}
\newcommand{\omark}{\ensuremath{\circ}}
\newcommand{\nmark}{--}

\newcommand{\fcirc}[2]{%
\tikz[baseline=(c.base)]\node[circle, fill=#1, inner sep=1.2pt, minimum size=1.35em] (c)
{\textcolor{white}{\scriptsize\bfseries #2}};%
}

\usepackage{longtable}
\usepackage{array}

\definecolor{NotationHeader}{RGB}{232,232,232}
\definecolor{NotationGroup}{RGB}{244,246,248}

\colorlet{red}{black}
\makeatletter
\let\oldtextcolor\textcolor
\renewcommand{\textcolor}[2]{%
  \def\temp{#1}%
  \def\red{blue}%
  \ifx\temp\red
    \oldtextcolor{black}{#2}%
  \else
    \oldtextcolor{#1}{#2}%
  \fi
}
\makeatother

\title{Self-Improvements in Modern Agentic Systems: A Survey}

\author{%
\parbox{\textwidth}{\centering
\bfseries\normalsize
Zhe Ren$^{1}$, Yimeng Chen$^{2}$\thanks{Corresponding authors}, Dandan Guo$^{1,2}$\footnotemark[1], Guowei Rong$^{1}$, Tonghui Li$^{1}$,\\
R.B. Xiong$^{3}$, Qingfeng Lan$^{4}$,  Wenyi Wang$^{2}$, Li Nanbo$^{2}$, Yibo Yang$^{2}$,\\
 Mingchen Zhuge$^{2}$, J\"urgen Schmidhuber $^{2,5}$}\\[1.60em] %
\parbox{\textwidth}{\centering
\normalsize
$^{1}$School of Artificial Intelligence, Jilin University \\
$^{2}$King Abdullah University of Science and Technology (KAUST) \\
$^{3}$Independent Researcher \quad
$^{4}$University of Alberta \\
$^{5}$The Swiss AI Lab IDSIA/USI/SUPSI
}\\[2.60em] %
\parbox{\textwidth}{\centering
\ttfamily\footnotesize
renzhe25@mails.jlu.edu.cn, yimeng.chen@kaust.edu.sa, guodandan@jlu.edu.cn, \\
\{ronggw25, lith\}@mails.jlu.edu.cn, rbxiong1@outlook.com, qlan3@ualberta.ca, \\
\{wenyi.wang, nanbo.li, yibo.yang, mingchen.zhuge, juergen.schmidhuber\}@kaust.edu.sa \\
}
}

\begin{document}

\ifcolmsubmission
\linenumbers
\fi

\maketitle

\vspace{-2.5em}

\begin{center}
    \small
    \setlength{\tabcolsep}{10pt}
    \begin{tabular}{cc}
        \href{https://github.com/selfimproving-agent/awesome-Self-Improving-Agents}{
            \faGithub\ \textsf{Self-Improving-Agents}
        } & 
        \href{https://selfimproving-agent.github.io/}{
            \textcolor{blue!70!black}{\faGlobe}\ \textsf{Project Page}
        }
    \end{tabular}
\end{center}
\vspace{2.5em}

\begin{figure*}[ht]
    \centering
    \includegraphics[width=1\linewidth]{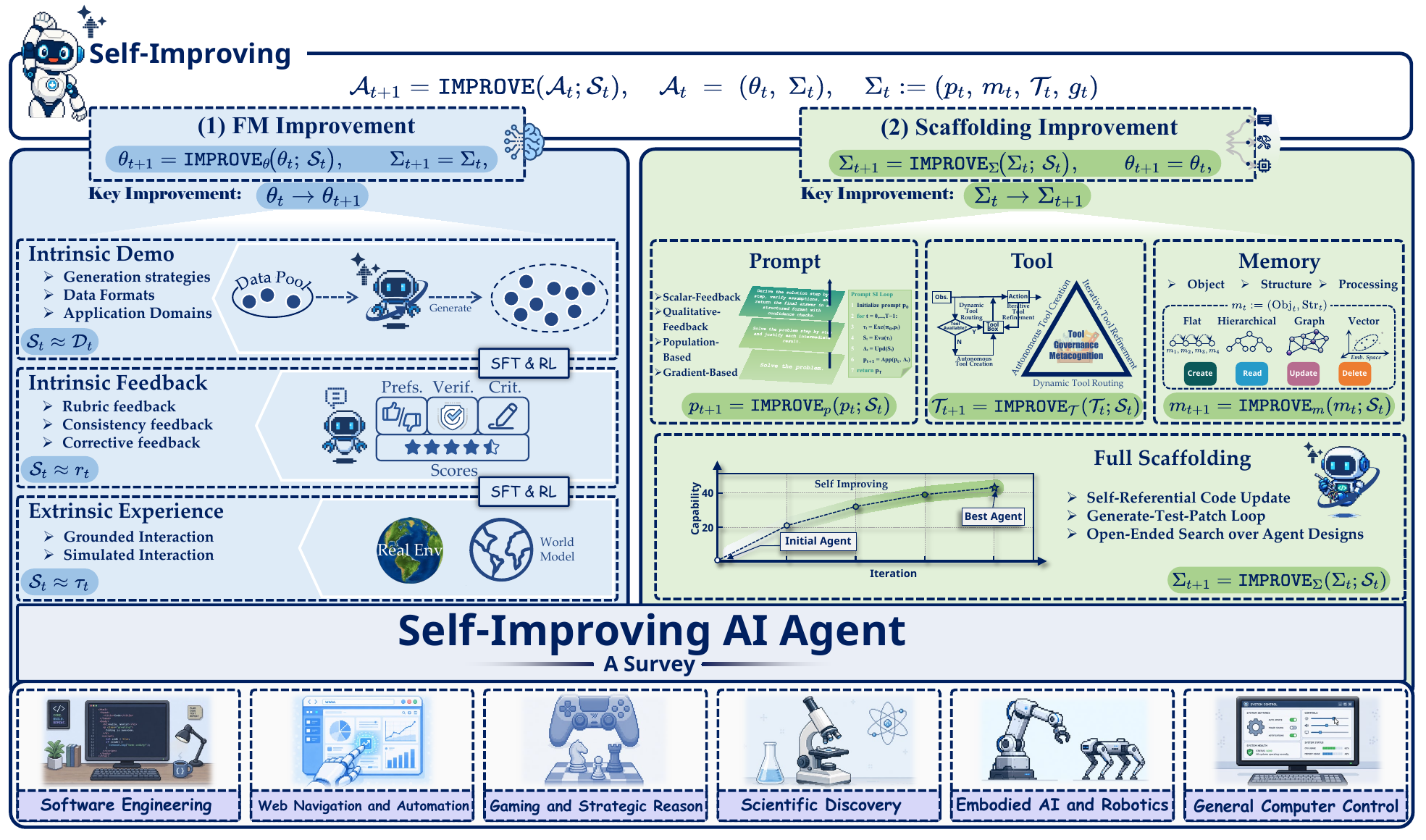}
    \caption{\textbf{Overview of self-improvement paradigms for modern AI agents.} We categorize existing methods into two primary pathways according to what is modified. The first pathway is \textbf{Foundation Model Improvement}, where the model parameters are updated from $\theta_t$ to $\theta_{t+1}$ using intrinsic generative demonstrations $\mathcal{D}_t$, intrinsic evaluative feedback $e_t$, or extrinsic exploratory experience $\tau_t$. The second pathway is \textbf{Scaffolding Improvement}, where the operational scaffold is updated from $\Sigma_t$ to $\Sigma_{t+1}$ through non-parametric changes. Across scaffold components, a generic update signal $\mathcal{S}_t$ is instantiated to drive improvements in prompts $p_t$, memory $m_t$, tools $\mathcal{T}_t$, or the full scaffolding $\Sigma_t$.}
    \label{fig:mainfig}
\end{figure*}

\newpage
\begin{abstract}
Self-improving autonomous agents are moving from research prototypes to deployed systems. The primary goal is controllable evolution, or adaptation, from experience with minimal or even no human input. This survey frames modern self-improving agents as adaptive systems that convert experience into accumulated capability gains. We offer a system-level framework that represents a modern agent as a configuration coupling a foundation model with an operational scaffold of prompts, memory, tools, and control logic. Within this framework, self-improvement is formalized as a self-induced update operator that obtains and commits updates to model parameters or scaffold components. 
We organize prior work by update target and by the signals that drive change, then review applications and discuss evaluation, before closing with open problems and future directions.
For convenience, we track technical updates on \href{https://github.com/selfimproving-agent/awesome-Self-Improving-Agents}{this GitHub page}.
\end{abstract}

\newpage
{
  \hypersetup{linkcolor=Plum, linktoc=page}
  \tableofcontents
}
\clearpage

\clearpage
\section{Introduction}
\label{sec:Introduction}

\begin{tcolorbox}[colback=brown!20!white, colframe=brown!20!white]

{\centering
\textit{"The first ultraintelligent machine is the last invention that man need ever make."}\par
}
\noindent\hfill--- \hyperlink{cite.good1966speculations}{I. J. Good (1966)}\par

\end{tcolorbox}

The development of artificial intelligence (AI) technology has driven a paradigm shift in agentic systems \citep{shoham1993agent, maes1994agents, wooldridge1995intelligent, yao2022react, Wang_2024}, from earlier narrow systems built around task-specific models or hand-engineered modules to modern agentic systems powered by \textbf{foundation models} (FMs), including large language models (LLMs) and vision-language models (VLMs), where natural language serves as a shared interface for representation, reasoning, and control.
Progress in foundation models has produced a qualitative shift in generalization, yielding striking successes across a wide range of domains, most notably code generation~\citep{chen2021evaluating}, language understanding~\citep{hendrycks2020measuring}, and mathematical and formal reasoning~\citep{wei2022chain}. These advances have brought a long-standing question to the foreground: the prospect of AI systems that improve themselves. Fundamentally, self-improvement is an inherently self-referential process. It defines a system's capacity to autonomously inspect, evaluate, and deliberately modify its own underlying optimization mechanisms and operational logic. Good articulated possible consequences of machine self-improvement~\citep{good1966speculations}, describing the possibility of an ``Intelligence Explosion'' once machines acquire the capacity to design more capable successors. Early work on concrete self-improvement algorithms dates back to Schmidhuber's self-referential learning framework~\citep{schmidhuber1987selfreferential}, which introduced mechanisms in which a system generates and evaluates modified descendant versions of itself. Establishing the theoretical ceiling of this pursuit, the G\"odel Machine \citep{schmidhuber2003godel} introduced a fully self-referential algorithm designed to rewrite its own code whenever it can mathematically prove an expected-utility improvement. While a persistent lineage of research successfully demonstrated neural networks (NNs) learning to program other networks via fast weights \citep{schmidhuber1992learning}, advancing to self-referential systems capable of modifying themselves \citep{schmidhuber1993self, irie2022modern}, and acting as meta-learning systems to discover their own learning algorithms \citep{hochreiter2001learning, schmidhuber2004optimal, kirsch2021meta, kirsch2022eliminating, irie2025metalearningcontinuallearningalgorithms}, parallel efforts introduced incremental self-improvement, establishing the ability to enforce long-term reward accelerations through the success-story algorithm for undoing policy self-modifications through backtracking \citep{Schmidhuber1994OnLH, schmidhuber1996simple, schmidhuber1997shifting}. However, scaling these visionary mechanisms into open-ended agents was historically constrained. Traditionally, systems were forced to search through vast, low-level spaces of assembly-like code or raw synaptic weights. Today, modern FMs alleviate this historical bottleneck by introducing natural language as a unified, highly capable semantic medium for reasoning, policy execution, and self-modification. By drastically reducing the search space for viable modifications, this language-native paradigm has drawn intense recent attention, with growing evidence demonstrating that FM-based self-improving agents can yield substantial empirical gains \citep{yuan2024self, acikgoz2025selfimprovingllmagentstesttime, qi2025webrl, zhang2025darwingodelmachineopenended}.

\begin{figure*}[ht]
    \centering
    \includegraphics[width=1\linewidth]{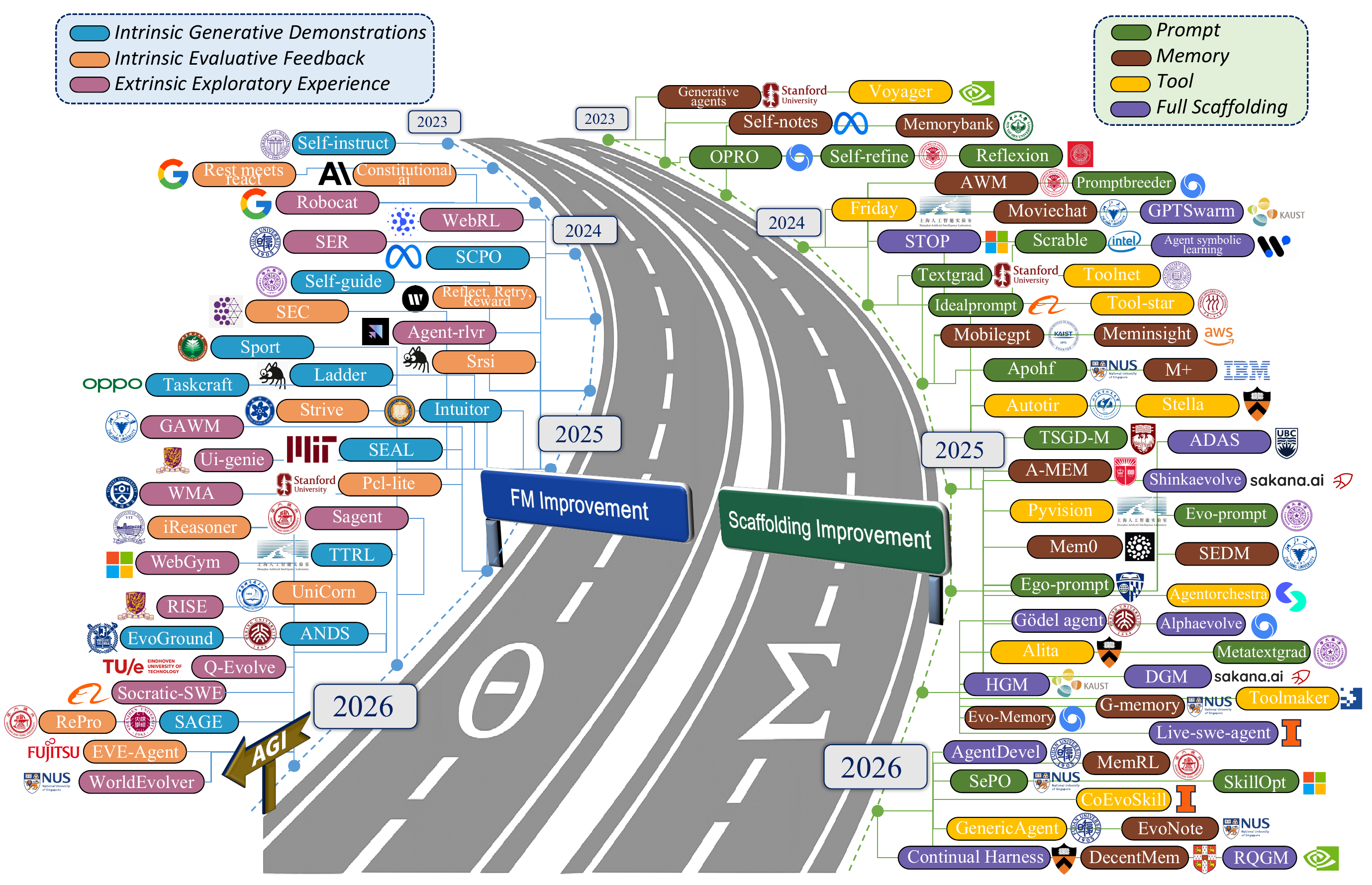}
    \caption{Timeline and taxonomy of self-improvement in foundation-model-based agents (2023–2026). Representative works are positioned by publication year. The left lane denotes foundation-model improvement ($\theta$), while the right lane denotes scaffolding improvement ($\Sigma$). The AGI signpost highlights the field’s long-term aspiration toward increasingly general agentic intelligence.}
    \label{fig:agent_timeline}
\end{figure*}

 To function autonomously in concrete environments, the FM serving as the cognitive core is typically enveloped by an operational {scaffold}, a structured framework comprising instruction schemes \citep{yuksekgonul2024textgradautomaticdifferentiationtext, guo2025evopromptconnectingllmsevolutionary}, memory systems \citep{chhikara2025mem0buildingproductionreadyai, zhang2026memrlselfevolvingagentsruntime}, tool interfaces \citep{qiu2025alitageneralistagentenabling, wölflein2025llmagentsmakingagent}, and control logic \citep{hong2023metagpt,10.5555/3692070.3694667,xiong2025beyond}. Recently, such an operational scaffold is often also referred to as an agent {harness} \citep{rajasekaran2026harnessdesign, trivedy2026agentharness, zhang2026selfharnessharnessesimprove, zhong2026aiharnessengineeringruntime}, while this survey uses the term scaffold to emphasize the modifiable structures surrounding the foundation model. Operationally, this scaffold acts as a controller that constructs context, selects actions, and enforces constraints. This modern architecture was conceptually foreshadowed by the "learning to think" framework \citep{schmidhuber2015learning}, where a general-purpose controller learned to dynamically query, or "prompt," a predictive world model to generate reasoning sequences akin to a modern "chain of thought" \citep{wei2022chain}.
Building upon this core-and-scaffold architecture, the integration of these classical control principles has crystallized into an emergent paradigm, which we refer to as FM-based self-improving agents. Because modern agentic systems comprise both the aforementioned neural core and scaffold, we organize this survey around a unified taxonomy with two primary pathways, as illustrated in Fig.~\ref{fig:mainfig} and summarized chronologically in Fig.~\ref{fig:agent_timeline}. The first pathway, which we call \textbf{foundation model improvement}, aims to achieve a slower but more stable form of long-term consolidation by updating the underlying model, thereby amortizing capability gains across varied tasks \citep{huang2023large, zhao2024selfguidebettertaskspecificinstruction, wang2025improvingmodelalignmentcollective, xiao2025uigenie, bougie2026alignuserhumanalignedllmagents}. The second pathway, which we call \textbf{scaffold improvement}, is typically faster and more easily reversible; it improves the agent by updating structural components, including prompts, memory, tool interfaces, and end-to-end control logic, to reshape the agent’s effective observation and action semantics \citep{fernando2024promptbreeder, chhikara2025mem0buildingproductionreadyai, wölflein2025llmagentsmakingagent, zhang2025darwingodelmachineopenended, borthwick2026robophdselfimprovingtexttosqlautonomous}. Although these pathways differ in their targets of modification, both are fundamentally driven by learning signals extracted during interaction. Accordingly, our taxonomy further refines these methods by separating \textit{what} is updated from \textit{where} the improvement signals originate, providing a common language for comparing disparate methods on a consistent footing. Fig.~\ref{fig:self_improving_taxonomy} provides a unified taxonomy of this survey.

\begin{wraptable}{r}{0.55\textwidth}
    \vspace{-\intextsep}
    \centering
    \footnotesize
    \setlength{\tabcolsep}{3pt}
    \renewcommand{\arraystretch}{1.08}
    \renewcommand{\tabularxcolumn}[1]{m{#1}}

    \newcommand{\badgeC}{\textcolor{green!55!black}{\bfseries\cmark}}
    \newcommand{\badgeO}{\textcolor{orange!85!black}{\bfseries\omark}}
    \newcommand{\badgeN}{\textcolor{red!70!black}{\bfseries\nmark}}

    \rowcolors{3}{cyan!3}{white}

    \begin{tabularx}{\linewidth}{
        >{\columncolor{black!4}\hsize=2.25\hsize\raggedright\arraybackslash}X
        *{4}{>{\hsize=0.6875\hsize\centering\arraybackslash}X}
    }
    \toprule
    \rowcolor{cyan!8}
    \textbf{Dimension} & \textbf{Ours (2026)} & \citet{gao2025surveyselfevolvingagentspath} & \citet{fang2025comprehensivesurveyselfevolvingai} & \citet{tao2024surveyselfevolutionlargelanguage} \\
    \midrule
    
    \textbf{Agent formulation}     & \badgeC & \badgeC & \badgeC & \badgeO \\
    \textbf{Definition scope}       & \badgeC & \badgeC & \badgeO & \badgeO \\
    \textbf{Historical roots}     & \badgeC & \badgeN & \badgeO & \badgeO \\
    \textbf{Signal lens}            & \badgeC & \badgeO & \badgeC & \badgeC \\
    \textbf{Update substrate}       & \badgeC & \badgeC & \badgeC & \badgeO \\
    \textbf{Evaluation lens}        & \badgeC & \badgeC & \badgeC & \badgeC \\
    \textbf{Domain coverage}        & \badgeC & \badgeO & \badgeO & \badgeN \\
    \textbf{Outlook \& issues}      & \badgeC & \badgeO & \badgeC & \badgeO \\
    
    \bottomrule
    \end{tabularx}

    \vspace{2pt}
    {\scriptsize
    \textcolor{green!55!black}{\bfseries\cmark} primary \quad
    \textcolor{orange!85!black}{\bfseries\omark} secondary \quad
    \textcolor{red!70!black}{\bfseries\nmark} not focus
    }

    \caption{Comparison of organizing emphases across related surveys.}
    \vspace{-8pt}
    \label{tab:survey-compare}
\end{wraptable}

Despite the rapid proliferation of such empirical frameworks, the broader research landscape remains highly fragmented in both terminology and scope. Closely related ideas are often described under different labels (e.g., self-correction, meta-prompting, or self-play), obscuring underlying mechanistic similarities. Recent surveys have provided valuable perspectives by organizing self-evolving agents along dimensions such as what to evolve, when to evolve, and how to evolve \citep{gao2025surveyselfevolvingagentspath, fang2025comprehensivesurveyselfevolvingai}, or by focusing specifically on the autonomous learning capabilities of static LLMs \citep{tao2024surveyselfevolutionlargelanguage}. 
However, existing reviews often treat foundation model fine-tuning and agent scaffolding as isolated topics, lacking a unified formal perspective. Furthermore, few trace the conceptual roots of self-improvement back to classical AI. To bridge this gap, our survey provides a rigorous and systematic positioning of these modern agentic systems. We offer a comprehensive taxonomy under a unified formalization, clarifying their historical evolution and providing a clear, forward-looking roadmap for self-improvement. Concretely, our main contributions are as follows:
\begin{itemize}
    \item \textbf{Historical Context and Evolution.} We trace the evolutionary roots of self-improving systems from classical AI to modern FM-based agents, establishing a clear trajectory for how foundational learning mechanisms have adapted to the agentic era.
    \item \textbf{A Unified Formalization and Systematic Taxonomy.} Under our proposed formulation, we systematically categorize mechanisms into two distinct pathways: foundation model improvement and  scaffold improvement (covering prompt optimization, memory evolution, tool governance, and full-scaffold redesign).    
    \item \textbf{Empirical Landscape, Evaluation Paradigms, and Frontiers.} We integrate the classification system with practical applications (such as software engineering, web navigation, and scientific discovery) and critically analyze current evaluation protocols. By comparing mechanism-level benchmarking and end-to-end evaluation, we highlight the security factors and challenges that need to be considered in achieving reliable, continuous self-improvement.
\end{itemize}

The remainder of this survey is organized as follows. Section~\ref{sec:Historical_Context_and_Theoretical_Foundations} traces the historical context and theoretical foundations of self-improving systems. Section~\ref{sec:Definitions} introduces the systems view of foundation-model-based agents and formalizes the problem setting. Section~\ref{sec:A_Taxonomy_of_Self_Improvement_Mechanisms} presents our unified taxonomy, outlining the two primary pathways for improvement. These pathways are subsequently surveyed in depth: Section~\ref{sec:Foundation_Model_Improvement} covers updates to the underlying foundation model, while Section~\ref{sec:Scaffolding_Improvement} focuses on structural modifications to the surrounding scaffold. Section~\ref{sec:Applications} reviews practical instantiations across representative domains. Section~\ref{sec:Evaluation} discusses protocols and benchmarks for evaluating these systems. Finally, Section~\ref{sec:Discussion} synthesizes open problems and safety considerations, followed by concluding remarks in Section~\ref{sec:Conclusion}.

\tikzstyle{my-box}=[
    rectangle,
    draw=hidden-black,
    rounded corners,
    text opacity=1,
    minimum height=1.5em,
    minimum width=5em,
    inner sep=2pt,
    align=center,
    fill opacity=.3,  
    line width=0.8pt,
]
\tikzstyle{leaf}=[my-box, minimum height=1.5em,
    fill=fm-blue, text=sia-color, align=left,font=\normalsize,
    inner xsep=2pt,
    inner ysep=4pt,
    line width=0.8pt, 
]
\tikzstyle{leaf1}=[my-box, minimum height=1.5em,
    fill=sf-green, text=sia-color, align=left,font=\normalsize,
    inner xsep=2pt,
    inner ysep=4pt,
    line width=0.8pt,
]
\tikzstyle{leaf2}=[my-box, minimum height=1.5em,
    fill=eval-purple, text=sia-color, align=left,font=\normalsize,
    inner xsep=2pt,
    inner ysep=4pt,
    line width=0.8pt,
]

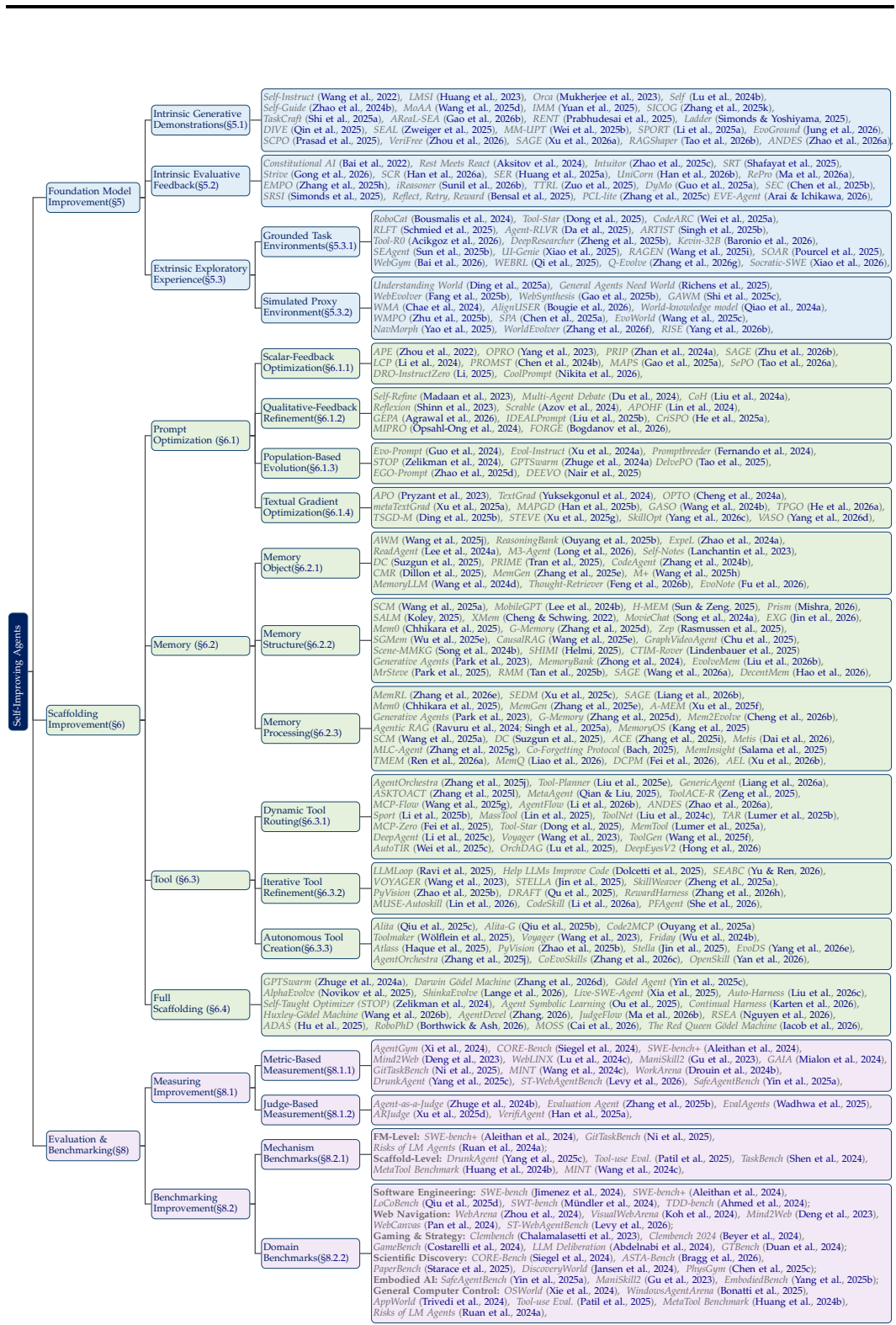
\begin{figure*}[t]
    \vspace{-2mm}
    \centering
    \resizebox{\textwidth}{!}
    {
        \begin{forest}
            forked edges,
            for tree={
                child anchor=west,
                parent anchor=east,
                grow'=east,
                anchor=west,
                base=left,
                font=\large,
                rectangle,
                rounded corners,
                align=left,
                edge+={darkgray, line width=1pt},
                s sep=3pt, 
                inner xsep=4pt,
                inner ysep=4pt,
                line width=0.8pt,
                ver/.style={
                    rotate=90, 
                    child anchor=north, 
                    parent anchor=south, 
                    anchor=center
                },
            },
            where level=0{font=\normalsize, text width=11em, text=white, fill=sia-color, fill opacity=1, draw=hidden-black}{},
            where level=1{font=\normalsize, text width=9em, fill=orange!20, fill opacity=0.4, draw=hidden-black}{},
            where level=2{font=\normalsize, text width=9.5em, fill=yellow!25, fill opacity=0.4, draw=hidden-black}{},
            where level=3{font=\normalsize, text width=9.5em, fill=blue!10, fill opacity=0.4, draw=hidden-black}{},
            where level=4{font=\normalsize, text width=16em, fill=blue!20, fill opacity=0.4, draw=hidden-black}{},
            [
                Self-Improving Agents, ver
                [
                    Foundation Model\\Improvement(\S \ref{sec:Foundation_Model_Improvement}) , leaf
                    [
                        Intrinsic Generative\\Demonstrations(\S \ref{sec:Intrinsic_Generative_Demonstrations}) , leaf
                        [
                            \textit{Self-Instruct}~\citep{wang2022self}{, }
                            \textit{LMSI}~\citep{huang2023large}{, }
                            \textit{Orca}~\citep{mukherjee2023orca}{, }
                            \textit{Self}~\citep{lu2023self}{, }\\
                            \textit{Self-Guide}~\citep{zhao2024selfguidebettertaskspecificinstruction}{, }
                            \textit{MoAA}~\citep{wang2025improvingmodelalignmentcollective}{, }
                            \textit{IMM}~\citep{yuan2025superficial}{, }
                            \textit{SICOG}~\citep{zhang2025will}{, }\\
                            \textit{TaskCraft}~\citep{shi2025taskcraft}{, }
                            \textit{AReaL-SEA}~\citep{gao2026selfevolvingsyntheticdataverifiablereward}{, }
                            \textit{RENT}~\citep{prabhudesai2025maximizing}{, }
                            \textit{Ladder}~\citep{simonds2025ladder}{, }\\
                            \textit{DIVE}~\citep{qin2025dive}{, }
                            \textit{SEAL}~\citep{zweiger2025selfadapting}{, }
                            \textit{MM-UPT}~\citep{wei2025unsupervised}{, }
                            \textit{SPORT}~\citep{li2025iterative}{, } \textit{EvoGround}~\citep{jung2026evogroundselfevolvingvideoagents}{, }\\
                            \textit{SCPO}~\citep{pmlr-v267-prasad25a}{, }
                            \textit{VeriFree}~\citep{zhou2026reinforcing}{, }
                            \textit{SAGE}~\citep{xu-etal-2026-sage}{, } \textit{RAGShaper}~\citep{tao2026ragshaperelicitingsophisticatedagentic}{, }
                            \textit{ANDES}~\citep{zhao2026andesagentnativedata}{, }\\
                            , leaf, text width=63em, text=darkgray!80                    
                        ]
                    ]
                    [
                        Intrinsic Evaluative\\Feedback(\S \ref{sec:Intrinsic_Evaluative_Feedback}) , leaf
                        [
                            \textit{Constitutional AI}~\citep{bai2022constitutional}{, }
                            \textit{Rest Meets React}~\citep{aksitov2023rest}{, }
                            \textit{Intuitor}~\citep{zhao2025learning}{, }
                            \textit{SRT}~\citep{shafayat2025can}{, }\\                            
                            \textit{Strive}~\citep{gong2025strive}{, }
                            \textit{SCR}~\citep{han2026structuredreasoninglargelanguage}{, }
                            \textit{SER}~\citep{huang2025selfevolved}{, }
                            \textit{UniCorn}~\citep{han2026unicornselfimprovingunifiedmultimodal}{, }
                            \textit{RePro}~\citep{ma2026retrospective}{, }\\
                            \textit{EMPO}~\citep{zhang2025right}{, }
                            \textit{iReasoner}~\citep{sunil2026ireasonertrajectoryawareintrinsicreasoning}{, }
                            \textit{TTRL}~\citep{zuo2025ttrl}{, }
                            \textit{DyMo}~\citep{guo2025sample}{, }
                            \textit{SEC}~\citep{chen2025self}{, }
                            \\
                            \textit{SRSI}~\citep{simonds2025self}{, }
                            \textit{Reflect, Retry, Reward}~\citep{bensal2025reflect}{, }
                            \textit{PCL-lite}~\citep{zhang2025adaptive}
                            \textit{EVE-Agent}~\citep{arai2026eveagentevidenceverifiableselfevolvingagents}{, }
                            \\
                            , leaf, text width=63em, text=darkgray!80                          
                        ]
                    ]
                    [
                        Extrinsic Exploratory\\Experience(\S \ref{sec:Extrinsic_Exploratory_Experience}) , leaf
                        [
                            Grounded Task\\Environments(\S \ref{sec:Grounded_Interaction_with_Verifiable_Environments}) , leaf
                            [
                                \textit{RoboCat}~\citep{bousmalis2023robocatselfimprovinggeneralistagent}{, }
                                \textit{Tool-Star}~\citep{dong2025tool}{, }
                                \textit{CodeARC}~\citep{wei2025codearc}{, }\\
                                \textit{RLFT}~\citep{schmied2025llms}{, }
                                \textit{Agent-RLVR}~\citep{da2025agent}{, }
                                \textit{ARTIST}~\citep{singh2025agenticreasoningtoolintegration}{, }
                                \\
                                \textit{Tool-R0}~\citep{acikgoz2026toolr0selfevolvingllmagents}{, }
                                \textit{DeepResearcher}~\citep{zheng2025deepresearcher}{, }
                                \textit{Kevin-32B}~\citep{baronio2025kevinmultiturnrlgenerating}{, }\\
                                \textit{SEAgent}~\citep{sun2025seagentselfevolvingcomputeruse}{, }
                                \textit{UI-Genie}~\citep{xiao2025uigenie}{, }
                                \textit{RAGEN}~\citep{wang2025ragenunderstandingselfevolutionllm}{, }
                                \textit{SOAR}~\citep{pourcel2025selfimproving}{, }
                                \\ 
                                \textit{WebGym}~\citep{bai2026webgymscalingtrainingenvironments}{, }
                                \textit{WEBRL}~\citep{qi2025webrl}{, }
                                \textit{Q-Evolve}~\citep{zhang2026selfevolvingllmagentsindistribution}{, }
                                \textit{Socratic-SWE}~\citep{xiao2026socraticsweselfevolvingcodingagents}{, }
                                \\
                                , leaf, text width=52em, text=darkgray!80                          
                            ]
                        ]
                        [
                            Simulated Proxy\\Environment(\S \ref{sec:Efficient_Interaction_with_Simulated_Environments}) , leaf
                            [
                                \textit{Understanding World}~\citep{ding2025understanding}{, }
                                \textit{General Agents Need World}~\citep{richens2025general}{, }\\
                                \textit{WebEvolver}~\citep{fang2025webevolver}{, }
                                \textit{WebSynthesis}~\citep{gao2025websynthesisworldmodelguidedmctsefficient}{, }
                                \textit{GAWM}~\citep{shi2025gawmglobalawareworldmodel}{, }\\
                                \textit{WMA}~\citep{chae2025webagentsworldmodels}{, }
                                \textit{AlignUSER}~\citep{bougie2026alignuserhumanalignedllmagents}{, }
                                \textit{World-knowledge model}~\citep{qiao2024agent}{, }\\
                                \textit{WMPO}~\citep{zhu2025wmpoworldmodelbasedpolicy}{, }
                                \textit{SPA}~\citep{chen2025internalizingworldmodelsselfplay}{, }
                                \textit{EvoWorld}~\citep{wang2025evoworldevolvingpanoramicworld}{, }\\
                                \textit{NavMorph}~\citep{yao2025navmorphselfevolvingworldmodel}{, }
                                \textit{WorldEvolver}~\citep{zhang2026selfevolvingworldmodelsllm}{, }
                                \textit{RISE}~\citep{yang2026riseselfimprovingrobotpolicy}{, }
                                , leaf, text width=52em, text=darkgray!80                              
                            ]
                        ]
                    ]
                ]
                [
                    Scaffolding\\Improvement(\S \ref{sec:Scaffolding_Improvement}) , leaf1
[
    Prompt\\Optimization (\S \ref{sec:Prompt_Optimization}) , leaf1
    [
        Scalar-Feedback\\Optimization(\S \ref{sec:Scalar_Feedback_Optimization}), leaf1
        [
    \textit{APE}~\citep{zhou2022large}{, }
    \textit{OPRO}~\citep{yang2023large}{, }
            \textit{PRIP}~\citep{zhan2024promptrefinementimagepivot}{, }
            \textit{SAGE}~\citep{zhu2026sage}{, }\\
            \textit{LCP}~\citep{li2024learningcontrastivepromptsautomated}{, }
            \textit{PROMST}~\citep{chen2024promptoptimizationmultisteptasks}{, }
            \textit{MAPS}~\citep{gao2025promptalchemistautomatedllmtailored}{, }
            \textit{SePO}~\citep{tao2026seposelfevolvingpromptagent}{, }\\
            \textit{DRO-InstructZero}~\citep{li2025droinstructzerodistributionallyrobustprompt}{, }
            \textit{CoolPrompt}~\citep{anonymous2025coolprompt}{, }
            , leaf1, text width=52em, text=darkgray!80
        ]
    ]
    [
        Qualitative-Feedback\\Refinement(\S \ref{sec:Qualitative_Feedback_Refinement}), leaf1
        [
    \textit{Self-Refine}~\citep{madaan2023self}{, }
    \textit{Multi-Agent Debate}~\citep{liang2023debate}{, }
    \textit{CoH}~\citep{liu2023chain}{, }\\
    \textit{Reflexion}~\citep{shinn2023reflexionlanguageagentsverbal}{, }
    \textit{Scrable}~\citep{azov2024selfimprovingcustomerreviewresponse}{, }
            \textit{APOHF}~\citep{lin2024promptoptimizationhumanfeedback}{, }\\
            \textit{GEPA}~\citep{agrawal2025gepareflectivepromptevolution}{, }
            \textit{IDEALPrompt}~\citep{liu2025boostingprivatedomainunderstanding}{, }
            \textit{CriSPO}~\citep{he2025crispomultiaspectcritiquesuggestionguidedautomatic}{, }\\
            \textit{MIPRO}~\citep{opsahlong2024optimizinginstructionsdemonstrationsmultistage}{, }
            \textit{FORGE}~\citep{Bogdanov_2026}{, }
            , leaf1, text width=52em, text=darkgray!80
        ]
    ]
    [
        Population-Based\\Evolution(\S \ref{sec:Population_Based_Evolution}), leaf1
        [
    \textit{Evo-Prompt}~\citep{guo2025evopromptconnectingllmsevolutionary}{, }
    \textit{Evol-Instruct}~\citep{xu2024wizardlm}{, }
    \textit{Promptbreeder}~\citep{fernando2024promptbreeder}{, }\\
    \textit{STOP}~\citep{zelikman2024self}{, }
    \textit{GPTSwarm}~\citep{10.5555/3692070.3694667}
    \textit{DelvePO}~\citep{tao2025delvepodirectionguidedselfevolvingframework}{, }\\
    \textit{EGO-Prompt}~\citep{zhao2025autooptimizepromptsdomaintasks}{, }
    \textit{DEEVO}~\citep{nair2025tournamentpromptsevolvingllm}
            , leaf1, text width=52em, text=darkgray!80
        ]
    ]
    [
        Textual Gradient\\Optimization(\S \ref{sec:Textual_Gradient_Optimization}), leaf1
        [
        \textit{APO}~\citep{pryzant2023automaticpromptoptimizationgradient}{, }
    \textit{TextGrad}~\citep{yuksekgonul2024textgradautomaticdifferentiationtext}{, }
            \textit{OPTO}~\citep{cheng2024traceautodiffgenerativeoptimization}{, }\\
    \textit{metaTextGrad}~\citep{xu2025metatextgradautomaticallyoptimizinglanguage}{, }
    \textit{MAPGD}~\citep{han2025mapgdmultiagentpromptgradient}{, }
    \textit{GASO}~\citep{wang2024correctlysemanticbackpropagationlanguagebased}{, }
    \textit{TPGO}~\citep{he2026learningevolveselfimprovingframework}{, }
    \\
            \textit{TSGD-M}~\citep{ding2025scalingtextualgradientssamplingbased}{, }
            \textit{STEVE}~\citep{xu2025pick}{, }
            \textit{SkillOpt}~\citep{yang2026skilloptexecutivestrategyselfevolving}{, }
            \textit{VASO}~\citep{yang2026vasoformallyverifiableselfevolving}{, }
            \\
            , leaf1, text width=52em, text=darkgray!80
        ]
    ]
]
[
    Memory (\S \ref{sec:Memory}) , leaf1
    [
        Memory\\Object(\S \ref{sec:Memory_Object}), leaf1
        [
    \textit{AWM}~\citep{wang2024agentworkflowmemory}{, }
    \textit{ReasoningBank}~\citep{ouyang2025reasoningbankscalingagentselfevolving}{, }
    \textit{ExpeL}~\citep{zhao2024expel}{, }\\
    \textit{ReadAgent}~\citep{lee2024humaninspiredreadingagentgist}{, }
    \textit{M3-Agent}~\citep{long2025seeinglisteningrememberingreasoning}{, }
    \textit{Self-Notes}~\citep{lanchantin2023learningreasonmemorizeselfnotes}{, }\\
    \textit{DC}~\citep{suzgun2025dynamiccheatsheettesttimelearning}{, }
    \textit{PRIME}~\citep{tran2025primeplanningretrievalintegratedmemory}{, }
    \textit{CodeAgent}~\citep{zhang-etal-2024-codeagent}{, }\\
    \textit{CMR}~\citep{dillon2025contextualmemoryreweavinglarge}{, }
    \textit{MemGen}~\citep{zhang2025memgenweavinggenerativelatent}{, }
    \textit{M+}~\citep{wang2025mextendingmemoryllmscalable}\\
    \textit{MemoryLLM}~\citep{wang2024memoryllmselfupdatablelargelanguage}{, }
    \textit{Thought-Retriever}~\citep{feng2026thoughtretrieverdontjustretrieve}{, }
    \textit{EvoNote}~\citep{fu2026betterexperienceselfevolvingllm}{, }
            , leaf1, text width=52em, text=darkgray!80
        ]
    ]
    [
        Memory\\Structure(\S \ref{sec:Memory_Structure}), leaf1
        [
    \textit{SCM}~\citep{wang2025scmenhancinglargelanguage}{, }
    \textit{MobileGPT}~\citep{lee2024exploreselectderiverecall}{, }
    \textit{H-MEM}~\citep{sun2025hierarchicalmemoryhighefficiencylongterm}{, }
    \textit{Prism}~\citep{mishra2026prismevolutionarymemorysubstrate}{, }
    \\
    \textit{SALM}~\citep{koley2025salmmultiagentframeworklanguage}{, }
    \textit{XMem}~\citep{cheng2022xmemlongtermvideoobject}{, }
    \textit{MovieChat}~\citep{song2024moviechatdensetokensparse}{, }
    \textit{EXG}~\citep{jin2026exgselfevolvingagentsexperience}{, }
    \\
    \textit{Mem0}~\citep{chhikara2025mem0buildingproductionreadyai}{, }
    \textit{G-Memory}~\citep{zhang2025gmemorytracinghierarchicalmemory}{, }
    \textit{Zep}~\citep{rasmussen2025zeptemporalknowledgegraph}{, }\\
    \textit{SGMem}~\citep{wu2025sgmemsentencegraphmemory}{, }
    \textit{CausalRAG}~\citep{wang2025causalragintegratingcausalgraphs}{, }
    \textit{GraphVideoAgent}~\citep{10.1145/3746027.3755537}{, }\\
    \textit{Scene-MMKG}~\citep{10531671}{, }
    \textit{SHIMI}~\citep{helmi2025decentralizingaimemoryshimi}{, }
    \textit{CTIM-Rover}~\citep{lindenbauer2025knowledgenoisectimroverpitfalls}\\
    \textit{Generative Agents}~\citep{park2023generativeagentsinteractivesimulacra}{, }
    \textit{MemoryBank}~\citep{zhong2024memorybank}{, }
    \textit{EvolveMem}~\citep{liu2026evolvememselfevolvingmemoryarchitectureautoresearch}{, }
    \\
    \textit{MrSteve}~\citep{park2025mrsteveinstructionfollowingagentsminecraft}{, }
    \textit{RMM}~\citep{tan2025prospectretrospectreflectivememory}{, }
    \textit{SAGE}~\citep{wang2026sageselfevolvingagenticgraphmemory}{, }
    \textit{DecentMem}~\citep{hao2026selfevolvingmultiagentsystemsdecentralized}{, }
    \\
            , leaf1, text width=52em, text=darkgray!80
        ]
    ]
    [
        Memory\\Processing(\S \ref{sec:Memory_Processing}), leaf1
        [
    \textit{MemRL}~\citep{zhang2026memrlselfevolvingagentsruntime}{, }
    \textit{SEDM}~\citep{xu2025sedmscalableselfevolvingdistributed}{, }
    \textit{SAGE}~\citep{liang2025selfevolvingagentsreflectivememoryaugmented}{, }\\
    \textit{Mem0}~\citep{chhikara2025mem0buildingproductionreadyai}{, }
    \textit{MemGen}~\citep{zhang2025memgenweavinggenerativelatent}{, }
    \textit{A-MEM}~\citep{xu2025amemagenticmemoryllm}{, }\\
    \textit{Generative Agents}~\citep{park2023generativeagentsinteractivesimulacra}{, }
    \textit{G-Memory}~\citep{zhang2025gmemorytracinghierarchicalmemory}{, }
    \textit{Mem2Evolve}~\citep{cheng2026mem2evolveselfevolvingagentscoevolutionary}{, }
    \\
    \textit{Agentic RAG}~\citep{ravuru2024agenticretrievalaugmentedgenerationtime, singh2025agenticretrievalaugmentedgenerationsurvey}{, }
    \textit{MemoryOS}~\citep{kang2025memory}\\
    \textit{SCM}~\citep{wang2025scmenhancinglargelanguage}{, }
    \textit{DC}~\citep{suzgun2025dynamiccheatsheettesttimelearning}{, }
    \textit{ACE}~\citep{zhang2025agenticcontextengineeringevolving}{, }
    \textit{Metis}~\citep{dai2026metisbridgingtextcode}{, }
    \\
    \textit{MLC-Agent}~\citep{zhang2025mlcagentcognitivemodelbased}{, }
    \textit{Co-Forgetting Protocol}~\citep{bach2025pbftbackedsemanticvotingmultiagent}{, }
    \textit{MemInsight}~\citep{salama2025meminsightautonomousmemoryaugmentation}
    \\
    \textit{TMEM}~\citep{ren2026scalingselfevolvingagentsparametric}{, }
    \textit{MemQ}~\citep{liao2026memqintegratingqlearningselfevolving}{, }
    \textit{DCPM}~\citep{fei2026memoryrecalldualprocesscognitive}{, }
    \textit{AEL}~\citep{xu2026aelagentevolvinglearning}{, }\\
            , leaf1, text width=52em, text=darkgray!80
        ]
    ]
]
[
    Tool (\S \ref{sec:Tool}) , leaf1
    [
        Dynamic Tool\\Routing(\S \ref{sec:Dynamic_Tool_Routing}), leaf1
        [
    \textit{AgentOrchestra}~\citep{zhang2025agentorchestraorchestratinghierarchicalmultiagent}{, }
    \textit{Tool-Planner}~\citep{liu2025toolplannertaskplanningclusters}{, }
    \textit{GenericAgent}~\citep{liang2026genericagenttokenefficientselfevolvingllm}{, }\\
    \textit{ASKTOACT}~\citep{zhang2025asktoactenhancingllmstool}{, }
    \textit{MetaAgent}~\citep{qian2025metaagentselfevolvingagenttool}{, }
    \textit{ToolACE-R}~\citep{zeng2025toolacermodelawareiterativetraining}{, }\\
    \textit{MCP-Flow}~\citep{wang2025mcpflowfacilitatingllmagents}{, }
    \textit{AgentFlow}~\citep{li2025intheflowagenticoptimizationeffective}{, }
    \textit{ANDES}~\citep{zhao2026andesagentnativedata}{, }\\
    \textit{Sport}~\citep{li2025iterativetoolusageexploration}{, }
    \textit{MassTool}~\citep{lin2025masstoolmultitasksearchbasedtool}{, }
    \textit{ToolNet}~\citep{liu2024toolnetconnectinglargelanguage}{, }
    \textit{TAR}~\citep{lumer2025tooltoagentretrievalbridgingtools}{, }\\
    \textit{MCP-Zero}~\citep{fei2025mcpzeroactivetooldiscovery}{, }
    \textit{Tool-Star}~\citep{dong2025tool}{, }
    \textit{MemTool}~\citep{lumer2025memtooloptimizingshorttermmemory}{, }\\
    \textit{DeepAgent}~\citep{li2025deepagentgeneralreasoningagent}{, }
    \textit{Voyager}~\citep{wang2023voyageropenendedembodiedagent}{, }
    \textit{ToolGen}~\citep{wang2025toolgenunifiedtoolretrieval}{, }\\
    \textit{AutoTIR}~\citep{wei2025autotirautonomoustoolsintegrated}{, }
    \textit{OrchDAG}~\citep{lu2025orchdagcomplextoolorchestration}{, }
    \textit{DeepEyesV2}~\citep{hong2025deepeyesv2agenticmultimodalmodel}
            , leaf1, text width=52em, text=darkgray!80
        ]
    ]
    [
        Iterative Tool\\Refinement(\S \ref{sec:Iterative_Tool_Refinement}), leaf1
        [
    \textit{LLMLoop}~\citep{11185878}{, }
    \textit{Help LLMs Improve Code}~\citep{dolcetti2025helpingllmsimprovecode}{, }
    \textit{SEABC}~\citep{yu2026autonomousevolutionedatools}{, }\\
    \textit{VOYAGER}~\citep{wang2023voyageropenendedembodiedagent}{, }
    \textit{STELLA}~\citep{jin2025stellaselfevolvingllmagent}{, }
    \textit{SkillWeaver}~\citep{zheng2025skillweaverwebagentsselfimprove}{, }\\
    \textit{PyVision}~\citep{zhao2025pyvisionagenticvisiondynamic}{, }
    \textit{DRAFT}~\citep{qu2025explorationmasteryenablingllms}{, }
    \textit{RewardHarness}~\citep{zhang2026rewardharnessselfevolvingagenticposttraining}{, }\\
    \textit{MUSE-Autoskill}~\citep{lin2026museautoskillselfevolvingagentsskill}{, }
    \textit{CodeSkill}~\citep{li2026codeskilllearningselfevolvingskills}{, }
    \textit{PFAgent}~\citep{she2026pfagenttractableselfevolvingpowerflow}{, }
            , leaf1, text width=52em, text=darkgray!80
        ]
    ]
    [
        Autonomous Tool\\Creation(\S \ref{sec:Autonomous_Tool_Creation}), leaf1
        [
    \textit{Alita}~\citep{qiu2025alitageneralistagentenabling}{, }
    \textit{Alita-G}~\citep{qiu2025alitagselfevolvinggenerativeagent}{, }
    \textit{Code2MCP}~\citep{ouyang2025code2mcptransformingcoderepositories}\\
    \textit{Toolmaker}~\citep{wölflein2025llmagentsmakingagent}{, }
    \textit{Voyager}~\citep{wang2023voyageropenendedembodiedagent}{, }
    \textit{Friday}~\citep{wu2024oscopilotgeneralistcomputeragents}{, }\\
    \textit{Atlass}~\citep{haque2025advancedtoollearningselection}{, }
    \textit{PyVision}~\citep{zhao2025pyvisionagenticvisiondynamic}{, }
    \textit{Stella}~\citep{jin2025stellaselfevolvingllmagent}{, }
    \textit{EvoDS}~\citep{yang2026evods}{, }
    \\
    \textit{AgentOrchestra}~\citep{zhang2025agentorchestraorchestratinghierarchicalmultiagent}{, }
    \textit{CoEvoSkills}~\citep{zhang2026coevoskillsselfevolvingagentskills}{, }
    \textit{OpenSkill}~\citep{yan2026openskillopenworldselfevolutionllm}{, }\\
            , leaf1, text width=52em, text=darkgray!80
        ]
    ]
]
                    [
                        Full\\Scaffolding (\S \ref{sec:Full_Scaffolding}) , leaf1
                        [
    \textit{GPTSwarm}~\citep{10.5555/3692070.3694667}{, }
    \textit{Darwin Gödel Machine}~\citep{zhang2025darwingodelmachineopenended}{, }
    \textit{Gödel Agent}~\citep{yin2025godel}{, }\\
    \textit{AlphaEvolve}~\citep{novikov2025alphaevolvecodingagentscientific}{, }
    \textit{ShinkaEvolve}~\citep{lange2025shinkaevolveopenendedsampleefficientprogram}{, }
    \textit{Live-SWE-Agent}~\citep{xia2025livesweagentsoftwareengineeringagents}{, }
    \textit{Auto-Harness}~\citep{liu2026adaptiveautoharnesssustainedselfimprovement}{, }\\
    \textit{Self-Taught Optimizer (STOP)}~\citep{zelikman2024self}{, }
    \textit{Agent Symbolic Learning}~\citep{ou2025symbolic}{, }
    \textit{Continual Harness}~\citep{karten2026continualharnessonlineadaptation}{, }
    \\
    \textit{Huxley-Gödel Machine}~\citep{wang2025huxleygodelmachinehumanlevelcoding}{, }
    \textit{AgentDevel}~\citep{zhang2026agentdevelreframingselfevolvingllm}{, }
    \textit{JudgeFlow}~\citep{ma2026judgeflowagenticworkflowoptimization}{, }
    \textit{RSEA}~\citep{nguyen2026recursiveselfevolvingagentsheldout}{, }\\
    \textit{ADAS}~\citep{hu2025automateddesignagenticsystems}{, }
    \textit{RoboPhD}~\citep{borthwick2026robophdselfimprovingtexttosqlautonomous}{, }
    \textit{MOSS}~\citep{cai2026mossselfevolutionsourcelevelrewriting}{, }
    \textit{The Red Queen G\"odel Machine}~\citep{iacob2026redqueengodelmachine}{, }
    \\
                                , leaf1, text width=63em, text=darkgray!80                                
                        ]
                    ]
                ]
                [
                    Evaluation \&\\Benchmarking(\S \ref{sec:Evaluation}) , leaf2
                    [
                        Measuring\\Improvement(\S \ref{sec:Measuring_Improvement}) , leaf2
                        [
                            Metric-Based\\Measurement(\S \ref{sec:Metric_based_measurement}) , leaf2
                            [
                                \textit{AgentGym}~\citep{xi2024agentgym}{, }
                                \textit{CORE-Bench}~\citep{siegel2024corebenchfosteringcredibilitypublished}{, }
                                \textit{SWE-bench+}~\citep{aleithan2024swebenchenhancedcodingbenchmark}{, }\\
                                \textit{Mind2Web}~\citep{deng2023mindweb}{, }
                                \textit{WebLINX}~\citep{lu2024weblinx}{, }
                                \textit{ManiSkill2}~\citep{gu2023maniskill2unifiedbenchmarkgeneralizable}{, }
                                \textit{GAIA}~\citep{mialon2023gaiabenchmarkgeneralai}{, }\\
                                \textit{GitTaskBench}~\citep{ni2025gittaskbenchbenchmarkcodeagents}{, }
                                \textit{MINT}~\citep{wang2024mintevaluatingllmsmultiturn}{, }
                                \textit{WorkArena}~\citep{drouin2024workarenacapablewebagents}{, }\\
                                \textit{DrunkAgent}~\citep{yang2025drunkagentstealthymemorycorruption}{, }
                                \textit{ST-WebAgentBench}~\citep{levy2025stwebagentbenchbenchmarkevaluatingsafety}{, }
                                \textit{SafeAgentBench}~\citep{yin2025safeagentbenchbenchmarksafetask}{, }
                                , leaf2, text width=52em, text=darkgray!80
                            ]
                        ]
                        [
                            Judge-Based\\Measurement(\S \ref{sec:Judge_based_measurement}) , leaf2
                            [
                                \textit{Agent-as-a-Judge}~\citep{zhuge2024agentasajudgeevaluateagentsagents}{, }
                                \textit{Evaluation Agent}~\citep{zhang-etal-2025-evaluation}{, }
                                \textit{EvalAgents}~\citep{wadhwa2025evalagentdiscoveringimplicitevaluation}{, }\\
                                \textit{ARJudge}~\citep{xu-etal-2025-learning}{, }
                                \textit{VerifiAgent}~\citep{han2025verifiagentunifiedverificationagent}{, }
                                , leaf2, text width=52em, text=darkgray!80
                            ]
                        ]
                    ]
                    [
                        Benchmarking\\Improvement(\S \ref{sec:Benchmarking_Improvement}) , leaf2
                        [
                            Mechanism\\Benchmarks(\S \ref{sec:Mechanism_Benchmarks}) , leaf2
                            [
                                \textbf{FM-Level:}
                                \textit{SWE-bench+}~\citep{aleithan2024swebenchenhancedcodingbenchmark}{, }
                                \textit{GitTaskBench}~\citep{ni2025gittaskbenchbenchmarkcodeagents}{, }\\
                                \textit{Risks of LM Agents}~\citep{ruan2024identifyingriskslmagents}{; }\\
                                \textbf{Scaffold-Level:}
                                \textit{DrunkAgent}~\citep{yang2025drunkagentstealthymemorycorruption}{, }
                                \textit{Tool-use Eval.}~\citep{patil2025the}{, }
                                \textit{TaskBench}~\citep{shen2024taskbenchbenchmarkinglargelanguage}{, }\\
                                \textit{MetaTool Benchmark}~\citep{huang2024metatoolbenchmarklargelanguage}{, }
                                \textit{MINT}~\citep{wang2024mintevaluatingllmsmultiturn}{, }
                                , leaf2, text width=52em, text=darkgray!80
                            ]
                        ]
                        [
                            Domain\\Benchmarks(\S \ref{sec:Domain_Benchmarks}) , leaf2
                            [
                                \textbf{Software Engineering:}
                                \textit{SWE-bench}~\citep{jimenez2024swebenchlanguagemodelsresolve}{, }
                                \textit{SWE-bench+}~\citep{aleithan2024swebenchenhancedcodingbenchmark}{, }\\
                                \textit{LoCoBench}~\citep{qiu2025locobenchagentinteractivebenchmarkllm}{, }
                                \textit{SWT-bench}~\citep{mündler2025swtbenchtestingvalidatingrealworld}{, }
                                \textit{TDD-bench}~\citep{ahmed2024tddbenchverifiedllmsgenerate}{; }\\
                                \textbf{Web Navigation:}
                                \textit{WebArena}~\citep{zhou2024webarena}{, }
                                \textit{VisualWebArena}~\citep{koh2024visualwebarenaevaluatingmultimodalagents}{, }
                                \textit{Mind2Web}~\citep{deng2023mindweb}{, }\\
                                \textit{WebCanvas}~\citep{pan2024webcanvas}{, }
                                \textit{ST-WebAgentBench}~\citep{levy2025stwebagentbenchbenchmarkevaluatingsafety}{; }\\
                                \textbf{Gaming \& Strategy:}
                                \textit{Clembench}~\citep{chalamalasetti-etal-2023-clembench}{, }
                                \textit{Clembench 2024}~\citep{beyer2024clembench2024challengingdynamiccomplementary}{, }\\
                                \textit{GameBench}~\citep{costarelli2024gamebench}{, }
                                \textit{LLM Deliberation}~\citep{abdelnabi2024llmdeliberation}{, }
                                \textit{GTBench}~\citep{duan2024gtbenchuncoveringstrategicreasoning}{; }\\
                                \textbf{Scientific Discovery:}
                                \textit{CORE-Bench}~\citep{siegel2024corebenchfosteringcredibilitypublished}{, }
                                \textit{ASTA-Bench}~\citep{bragg2025astabenchrigorousbenchmarkingai}{, }\\
                                \textit{PaperBench}~\citep{starace2025paperbenchevaluatingaisability}{, }
                                \textit{DiscoveryWorld}~\citep{jansen2024discoveryworldvirtualenvironmentdeveloping}{, }
                                \textit{PhysGym}~\citep{chen2025physgym}{; }\\
                                \textbf{Embodied AI:}
                                \textit{SafeAgentBench}~\citep{yin2025safeagentbenchbenchmarksafetask}{, }
                                \textit{ManiSkill2}~\citep{gu2023maniskill2unifiedbenchmarkgeneralizable}{, }
                                \textit{EmbodiedBench}~\citep{yang2025embodiedbenchcomprehensivebenchmarkingmultimodal}{; }\\
                                \textbf{General Computer Control:}
                                \textit{OSWorld}~\citep{xie2024osworldbenchmarkingmultimodalagents}{, }
                                \textit{WindowsAgentArena}~\citep{bonatti2024windowsagentarenaevaluating}{, }\\
                                \textit{AppWorld}~\citep{trivedi2024appworldcontrollableworldapps}{, }
                                \textit{Tool-use Eval.}~\citep{patil2025the}{, }
                                \textit{MetaTool Benchmark}~\citep{huang2024metatoolbenchmarklargelanguage}{, }\\
                                \textit{Risks of LM Agents}~\citep{ruan2024identifyingriskslmagents}{, }
                                , leaf2, text width=52em, text=darkgray!80
                            ]
                        ]
                    ]
                ]
            ]
        \end{forest}
    }
    \caption{A unified taxonomy of self-improving agents spanning foundation-model updates, scaffold updates, and evaluation benchmarks.}
    \label{fig:self_improving_taxonomy}
\end{figure*}
\section{Historical Context and Theoretical Foundations}
\label{sec:Historical_Context_and_Theoretical_Foundations}

\begin{figure*}[ht]
    \centering
    \includegraphics[width=\linewidth]{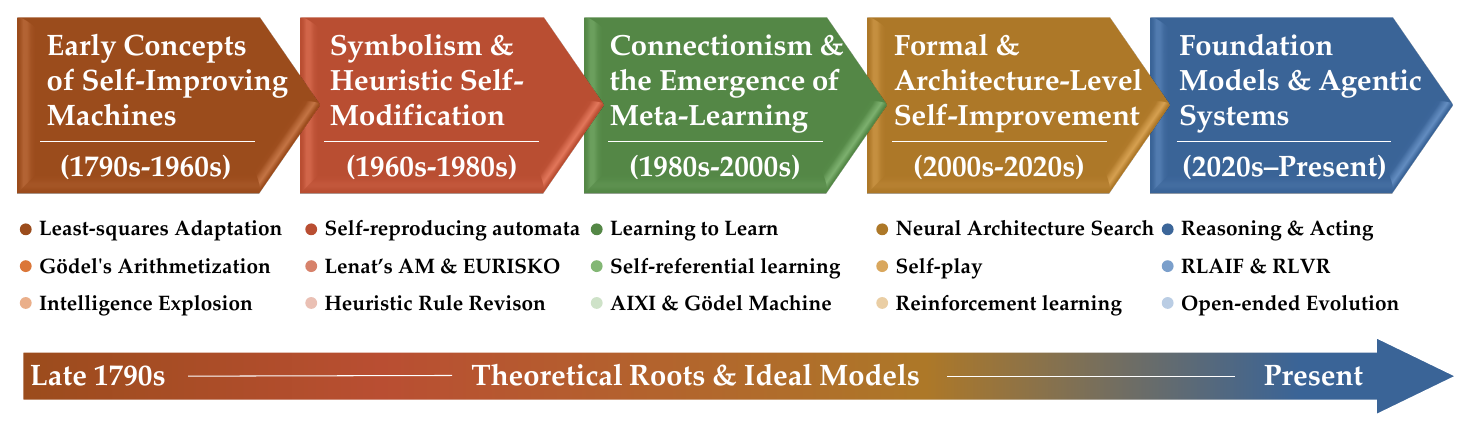}
    \caption{A timeline of theoretical roots and idealized models for self-improving agents, from the late 1790s to the present.}
    \label{fig:timeline}
\end{figure*}

Self-improvement is not unique to modern foundation models; it is a foundational objective of artificial intelligence. Whereas standard machine learning optimizes parameters within a fixed architecture, true self-improvement demands that a system explicitly inspect and rewrite its own operational logic, heuristics, or learning algorithms. As illustrated in Figure~\ref{fig:timeline}, the mechanisms have been successfully instantiated across several historical paradigms. By tracing this intellectual lineage, we clarify how earlier systems achieved self-improvement within  bounded domains, and how modern foundation models now provide a highly expressive, general-purpose substrate to scale these established principles into open-ended and real-world environments.

\subsection{Foundational Concepts (1790s-1960s)}
The mathematical roots of error-driven adaptation extend back to early optimization frameworks. The method of least squares \citep{legendre1805nouvelles, gauss1809theoria}\footnote{The first famous example of pattern recognition through an NN dates back over 200 years: the rediscovery of the dwarf planet Ceres in 1801 through Gauss, who collected noisy data points from previous astronomical observations, then used them to adjust the parameters of a predictor, which essentially learned to generalize from the training data to correctly predict the new location of Ceres~\citep{schmidhuber2022annotated}.} 
demonstrated how parameters of what's now known as ``linear neural networks'' can be systematically adjusted to minimize errors. As highlighted in historical retrospectives, this established a mechanism that is still heavily used today and serves as the very foundation of all neural networks, including deeper ones \citep{schmidhuber2022annotated}.

Throughout the mid-20th century, these concepts of adaptation continued to evolve, acting as conceptual precursors. Cybernetics popularized the feedback view of adaptive behavior in closed-loop systems \citep{wiener1948cybernetics}, while Ashby's work on homeostasis emphasized internal state adjustment to maintain viability under perturbations \citep{ashby1952design}. Alan Turing's ``child machine'' proposed a model whose internal configuration could be shaped through training and external feedback, such as reward and punishment \citep{Turing1948IntelligentMachinery, turing1950computing}. Early systems, from Rosenblatt's perceptron \citep{rosenblatt1958perceptron} to Samuel's checkers program \citep{samuel1959some}, provided concrete demonstrations of how performance feedback could change future behavior. However, while these mid-century milestones demonstrated that behavior could optimize over time, their underlying architectures and learning algorithms themselves remained strictly fixed. Consequently, they are best understood as advanced learning systems, rather than fully self-referential, self-improving systems. 

The formal basis for transcending fixed learning rules and enabling true self-reference followed a distinct trajectory. In the early 1930s, Kurt Gödel founded modern theoretical computer science and identified fundamental limits of any type of computation-based AI \citep{godel1931formal}. He introduced a universal coding language based on integers, using it to represent both data and programs in axiomatic form. By famously constructing formal statements that talk about the computation of other formal statements—especially self-referential statements—Gödel provided the theoretical blueprint for programs capable of manipulating their own code \citep{schmidhuber2022annotated}. In the 1960s, Good \citep{good1966speculations} introduced the visionary prospect of an ``Intelligence Explosion,'' hypothesizing that machines might one day acquire the capacity to autonomously design more capable successors. However, while this narrative highlighted self-improvement as an ultimate pursuit in AI, it remained a conceptual idea without a formal mathematical or algorithmic framework. Translating this speculative concept into actual deployable mechanisms would require the subsequent development of explicit structural and algorithmic self-modification frameworks.

\subsection{Symbolism and Heuristic Self-Modification (1960s–1980s)}
As symbolic AI rose to prominence, self-improvement was reinterpreted as the ability to manipulate explicit representations of knowledge and strategy. 
A key conceptual precursor was von Neumann's theory of self-reproducing automata. Originating in the late 1940s and later published in 1966, it distinguished a constructive mechanism from a symbolic description that can be copied and varied ~\citep{von1966theory}. \cite{myhill1964abstract} abstracted self-reproduction into a rigorous formal framework, emphasizing how systems can use symbolic descriptions to generate copies or variants of themselves. Together, these concepts clarified how a system can generate successive versions of increasing complexity by editing the descriptions it interprets. 
A similar approach in symbolic artificial intelligence is to treat problem-solving heuristics as first-class objects that can be examined, modified, and recombined.

Lenat’s Automated Mathematician (AM) system exemplified this approach by using heuristics to propose, extend, and evaluate candidate concepts, effectively searching over symbolic programs that encode mathematical ideas \citep{lenat11984automated}. EURISKO pushed further by representing heuristics themselves as manipulable Lisp objects, allowing the system to generate, modify, and test its own problem-solving strategies \citep{lenat1983eurisko, lenat1984and}. However, as emphasized by \cite{schmidhuber1987selfreferential}, the practical success of systems like EURISKO depended heavily on the user serving as an external evaluation signal, interpreting outputs and manually pruning unproductive heuristic drift, rather than on the system possessing a truly autonomous closed-loop credit assignment mechanism. This historical limitation underscores a challenge that persists in modern agentic systems, namely the need for robust internal evaluation signals and verification procedures to sustain autonomous recursive self-improvement.

\subsection{Connectionism and the Emergence of Meta-Learning (1980s–2000s)}

The resurgence of connectionism shifted the emphasis from hand-coded heuristics to learning dynamics as the substrate of improvement. 
Self-improvement was consequently reframed as explicitly optimizing the learning process itself, encompassing the optimizer, inductive biases, and adaptation rules. \citet{schmidhuber1987selfreferential} introduced self-referential learning frameworks in which evolutionary processes could optimize not only candidate solutions but also the learning procedures that generate them, making the learning system itself the object of search. Along a complementary line, \citet{bengio1990learning}  demonstrated that synaptic learning rules need not be fixed; by parameterizing these rules, the optimizer itself becomes a learnable object.

Schmidhuber introduced fast-weight programmers (FWPs) as a broad class of networks \citep{Schmidhuber:91fastweights, schmidhuber1992learning, schmidhuber1993self}. A representative early instance—retrospectively identified as the 1991 unnormalized linear Transformer (ULTRA) \citep{schlag2021linear}—features a “slow” neural network that learns by gradient descent to program the “fast weights” of another network through additive outer-product updates. The explicitly self-referential version followed in work on self-referential weight matrices, where a network learns to modify its own fast weights while receiving feedback signals such as errors or rewards as inputs, making the learning dynamics themselves part of what is optimized \citep{schmidhuber1993self}. This line is important for self-improvement because it moves self-modification into a continuous program space, rather than restricting it to symbolic code rewriting. Recent work has revisited this direction in scalable self-referential neural architectures, including modern self-referential weight matrices that learn to modify themselves \citep{irie2022modern}, recurrent fast-weight programmers and their self-referential extensions \citep{irie2023practicalcomputationalpowerlinear, irie2021goinglineartransformersrecurrent}, and self-referential networks that meta-learn continual learning algorithms \citep{irie2025metalearningcontinuallearningalgorithms}. Such mechanisms suggest a possible additional meta-level for self-improving foundation models, where the foundation model itself may become part of the modifiable substrate rather than remaining only a frozen component.

This period also consolidated \textit{learning to learn} as a framing for self-improvement. \textcolor{blue}{In reinforcement learning (RL), the 1994 paper introduced an RL machine as an early recursive self-improving agent that alter parts of their own learning strategy to shift inductive bias over a single lifelong interaction stream \citep{Schmidhuber1994OnLH}.} 
The same underlying idea was later described using related terminology, including environment-independent reinforcement acceleration and the success-story algorithm \citep{DBLP:conf/icml/WieringS96, schmidhuber1996simple, schmidhuber1996multi, schmidhuber1997shifting}. Thrun and Pratt likewise emphasized learning reusable inductive biases across task distributions \citep{thrun1998learning}. Related neuroevolution work such as NeuroEvolution of Augmenting Topologies (NEAT) improved neural architectures by expanding network topology under performance-driven selection \citep{stanley2002evolving}. AIXI formalized an uncomputable upper bound on optimal sequential decision-making in computable environments \citep{hutter2000theoryuniversalartificialintelligence}. Furthermore, \citet{schmidhuber2003godel} introduced the G\"odel Machine, a fully self-referential, self-improving machine that is theoretically optimal. A G\"odel Machine operates in a single-life reinforcement learning environment~\citep{Schmidhuber1994OnLH} and iteratively searches for candidate self-modifications, adopting a modification only when it can prove that doing so will improve its expected cumulative future reward. To establish this improvement, the proof must reason about future rewards while accounting for all possible subsequent self-modifications. The machine is fully self-referential in the sense that every component of the G\"odel Machine is itself subject to modification, including the self-modification generator and the theorem prover.

\subsection{Formal and Architecture-Level Self-Improvement (2000s-2020s)}

From the mid-2000s onward, research followed two main tracks: (i) analyzing the theoretical limits of self-improvement, and (ii) building practical mechanisms to automate architecture design. \textcolor{red}{Philosophical analyses began to explore the trajectory of these systems, debating the existential risks of the technological singularity~\citep{subramanian2010science}.}
In theory, Orseau and Ring pointed out that when agents interact with limited resources and irreversible actions, they face significant risks such as self-deception, reward tampering, and fatal errors \citep{orseau2011self}. Related research formalized how agents can safely model their environments and themselves without generating logical paradoxes. For instance, Fallenstein et al. \citep{fallenstein2015reflectiveoraclesfoundationclassical} introduced reflective oracles to reason about probabilistic programs that call themselves,  while logical induction offered a mathematical approach to updating beliefs under self-referential uncertainty \citep{garrabrant2016logical}. \textcolor{red}{Complementary work on proof-producing reflection in higher-order logic studied how reflective reasoning can be embedded in formal proof systems, providing a technical route for agents or theorem provers to reason about their own reasoning steps~\citep{fallenstein2015proof}.} 

Together, these frameworks clarified the gap between ideal models and practical deployments, demonstrating the  necessity of safe and constrained self-modification. \textcolor{red}{Early attempts to bridge this gap computationally, such as the endeavor to implement Gödel machines, sought architectures where agents could mathematically prove the optimality of their own code updates \citep{steunebrink2012towards}. At the architecture level, Nivel et al. proposed bounded recursive self-improvement, where reflective mechanisms and value-driven scheduling support self-modification under designer-imposed constraints~\citep{nivel2013bounded}.} 
In engineering, automated design has matured into Neural Architecture Search (NAS), where controller learning proposes architectures that can directly optimize task performance \citep{zoph2017neural}. Meanwhile, scalable algorithms such as self-play in reinforcement learning demonstrated how agents can generate increasingly difficult curricula from their own behavior, thus achieving continuous capability growth without human supervision \citep{silver2017mastering}. Ultimately, these theoretical and engineering advances have laid the foundation for modern agent systems centered on foundation models.

\subsection{Scalable Foundation Models and Agentic Systems (2020s--Present)}
The classical vision of autonomous agents---capable of perceiving, acting, and learning in open-ended environments---found a scalable, modern substrate in foundation models (FMs). Powered by the immense knowledge compressed during large-scale pretraining and the flexibility of a unified natural-language interface, FMs function as the cognitive engines of contemporary agentic systems. Crucially, the emergence of these systems fundamentally reshaped the paradigm of self-improvement, decoupling it into two distinct but complementary mechanisms: rapid, training-free adaptation and persistent parameter optimization.

Within this modern paradigm, the most immediate form of self-improvement occurs without computationally expensive weight updates. Because scalable sequence models facilitate broad generalization, agents can achieve rapid, training-free adaptation by dynamically revising their plans, prompts, and memory representations on the fly. This short-horizon self-improvement relies heavily on in-context learning, which effectively functions as a form of meta-learning. Mechanistically, this adaptation is mediated by the attention mechanism's key-value cache, acting as an associative memory that generates transient, context-dependent weight changes during sequence processing. This functional equivalence directly connects modern in-context learning to the 1991 fast-weight programmers (formally identified as unnormalized linear Transformers), in which networks similarly learned to induce rapid parameter changes during inference without requiring standard gradient descent \citep{Schmidhuber:91fastweights, schmidhuber1992learning, schmidhuber1993self, schlag2021linear}.

Complementing this fast, in-context adaptation are slow improvement loops designed to permanently internalize new capabilities. Reinforcement learning from human feedback (RLHF) and its variants achieved this by converting interaction traces and preference signals into persistent parameter updates \citep{ouyang2022training}. Parallel methodologies advanced partially self-supervised alignment by enabling models to critique their own outputs and provide AI feedback \citep{bai2022constitutional}. Meanwhile, frameworks like Reasoning and Acting (ReAct) and Reflexion build these loops on top of contextual execution, converting execution errors into natural-language reflections \citep{yao2022react, shinn2023reflexionlanguageagentsverbal}. Recent systems  pushed this further by maintaining skill libraries, curricula, and evaluation harnesses, spanning open-ended embodied learning \citep{wang2023voyageropenendedembodiedagent}, software-engineering agents with instrumented interfaces \citep{yang2024sweagentagentcomputerinterfacesenable, zhang2025darwingodelmachineopenended, wang2025huxleygodelmachinehumanlevelcoding}, and explicitly self-improving computer-use agents that iterate data, reward models, and policies over generations \citep{xiao2025uigenie, sun2025seagentselfevolvingcomputeruse}. Overall, the modern era reframes self-improvement as nested loops across parameters and scaffolding, renewing classical questions about control, evaluation, and safety at scale.

\section{Definitions}
\label{sec:Definitions}

\subsection{Formulation of Agentic Systems}
\label{subsec:Foundation_Model_Based_Agents}

\begin{figure*}[ht]
    \centering
    \includegraphics[width=1\linewidth]{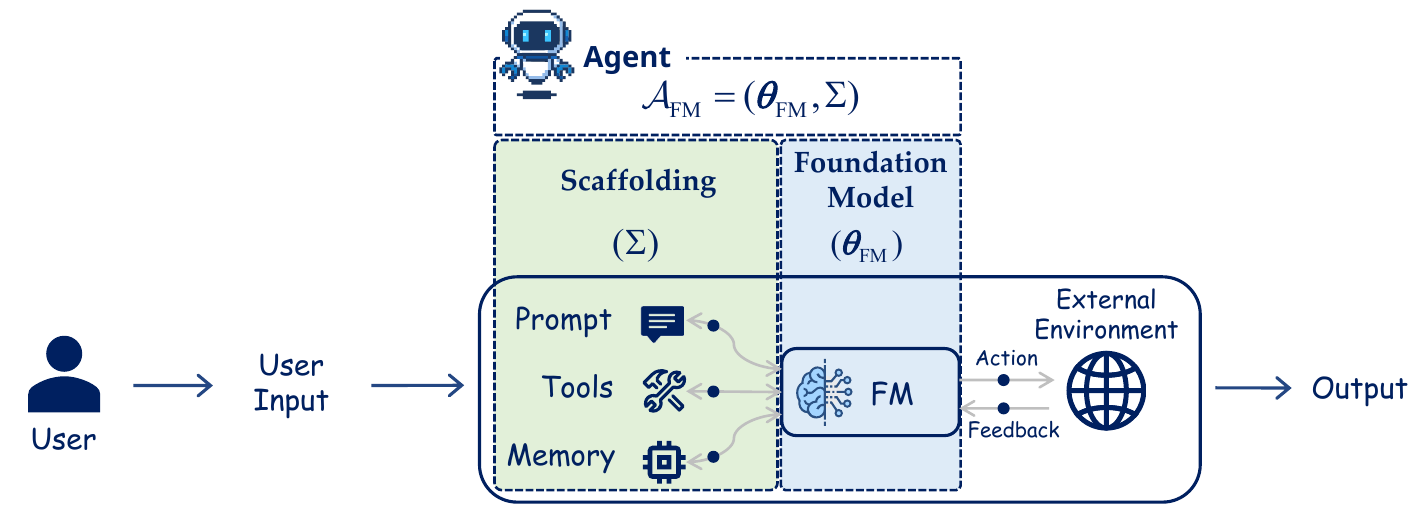}
    \caption{Schematic of an FM-based agent under our formalism.}
    \label{fig:agent_component}
\end{figure*}

We begin by clarifying the formal definition of an agent. The core of an autonomous agent lies in perceiving its environment, updating its internal state, and performing actions to achieve a specific goal. While such systems have driven decades of AI research, this survey focuses exclusively on a contemporary class: \textit{foundation-model-based agents}, which utilize large foundation models as their central cognitive engines. Crucially, they leverage natural language as a unified interface to integrate perception, reasoning, and tool manipulation. Because a foundation model fundamentally operates as a stateless inference engine, achieving autonomy requires coupling this cognitive core with a persistent, interactive scaffold. Therefore, we formally define the configuration of a foundation-model-based agent at time step $t$ as:
\begin{equation}
\label{eq:agent_formalism}
\mathcal{A}_t \;=\; (\theta_t,\; \Sigma_t),
\end{equation}
where $\theta_t$ encapsulates the neural parameters of the foundation model, and $\Sigma_t$ denotes the agent's dynamic operational scaffold. The scaffold $\Sigma_t$ specifies how the foundation model is conditioned, grounded, and connected to the external world. Concretely, it can be decomposed as
\begin{equation}
\label{eq:scaffold_formalism}
\Sigma_t := (p_t,\, m_t,\, \mathcal{T}_t,\, g_t),
\end{equation}
where $p_t$ denotes structured prompts or system instructions, $m_t$ denotes memory mechanisms and their retrieval and update policies, 
$\mathcal{T}_t$ denotes the set of external tools together with their invocation interfaces, and 
$g_t$ denotes additional control logic such as routing, scheduling, or safety constraints. The interaction between these components and the model's internal parameters is illustrated in Figure \ref{fig:agent_component}, which provides a schematic overview of an FM-based agent within our proposed formalism. Together, 
$\theta_t$ and $\Sigma_t$ determine how the agent reasons, plans, and acts.

During execution, to bridge this intrinsic configuration with the external environment, the agent maintains an ephemeral execution state $X_t$ (e.g., key-value caches, intermediate plans, or short-term working memory) that evolves as it processes an interaction stream. To connect this structural definition to observable behavior, it is beneficial to clarify how the agent's configuration generates its action selection mechanism. Although the foundation model parameters $\theta_t$ implement a general generative distribution, the agent’s realized behavior is jointly determined by both $\theta_t$ and its operational scaffold $\Sigma_t$. Accordingly, we denote the induced policy of the agent based on the foundation model as: 
\begin{equation}\label{policy}
\pi_{\theta_t, \Sigma_t}(A_t \mid X_t),
\end{equation}
where $A_t$ denotes the action produced by the system at time step $t$. Here, the conditioning on $X_t$ captures the transient context of the ongoing interaction. While $X_t$ may strongly influence immediate behavior, it is inherently ephemeral. It is typically discarded or reset once an immediate goal is reached or an external task boundary is crossed, and therefore does not constitute part of the agent's intrinsic architecture. In contrast, the parameters $(\theta_t, \Sigma_t)$ represent the agent's intrinsic configuration over time. While a standard agent may adapt to novel situations by dynamically updating its transient state $X_t$ (e.g., in-context examples), its underlying capabilities remain bounded by its fixed initial setup.

\subsection{Formal Definition of Self-Improvement}
\label{subsec:Formal_Definition_of_Self_Improvement}

Building upon the formal configuration $\mathcal{A}_t = (\theta_t, \Sigma_t)$, we formalize self-improvement in foundation-model-based agents (SI-FMA) as a process of persistent, endogenous adaptation. Rooted in the foundational principles of explicitly self-referential meta-learning \citep{schmidhuber1987selfreferential, schmidhuber1993self,schmidhuber1997shifting}, a self-improving agent actively leverages signals induced by its own execution---such as interaction outcomes, critiques, verification results, or proposed edits---to durably modify its underlying computational components.

\paragraph{Self-improvement as a self-induced operator.}
We conceptualize self-improvement through a \emph{self-induced} operator $\mathcal{U}$ that updates the agent’s intrinsic configuration. Let $\pi_{\theta_t,\Sigma_t}$ denote the agent’s induced policy. We formalize this process by factorizing the self-improvement step into an execution phase and an update phase:
\begin{equation}
\mathcal{A}_{t+1} \;=\; \mathcal{U}\Big(\mathcal{A}_{1:t},\; \mathcal{E}\big(\pi_{\theta_{t},\Sigma_{t}};\Sigma_{t},\mathcal{C}_{t}\big)\Big),
\end{equation}
where $\mathcal{E}$ denotes an \emph{agent-executed} procedure that produces a learning signal (e.g., interaction trajectories, reflections, critiques, or proposed edits) by running the induced policy against \textcolor{blue}{a} task context $\mathcal{C}_t$ (e.g., a task distribution, a user interaction stream, or a self-play environment). The scaffold $\Sigma_t$ is included as an explicit argument to $\mathcal{E}$ to permit direct self-inspection (e.g., critiquing prompt templates or auditing tool configurations). 

The system-level update rule $\mathcal{U}$ then applies this self-generated signal to modify the agent's intrinsic components ($\theta$ or $\Sigma$). Unlike the routine evolution of the transient state $X_t$ (e.g., merely accumulating dialogue history or working memory), the operator $\mathcal{U}$ commits durable changes. Crucially, in line with lifelong meta-learning principles, while these updates are not strictly irreversible and may be undone, they allow the agent to consolidate successful strategies into an increasingly stable policy over its interaction stream \citep{Schmidhuber1994OnLH, schmidhuber1995beyond, DBLP:conf/icml/WieringS96, schmidhuber1996simple}.

\paragraph{Two modes of self-reference in SI-FMA.}
SI-FMA may be self-referential in multiple, qualitatively distinct senses. In the first mode, the agent’s policy is executed to \emph{indirectly induce} improvement by generating experience or auxiliary artifacts that serve as learning signals. Self-generated trajectories, evaluations, preferences, or synthetic labels give rise to a learning objective that is subsequently consumed by an update rule, such as an external optimization procedure acting on the foundation model parameters $\theta_t$. This mode realizes self-improvement at the \emph{distributional} level: the agent’s behavior shapes the data and supervision that determine its future parameters.

In the second mode, the agent’s policy is executed to \emph{directly implement} improvements by modifying components of its own operational definition. Through its actions, the agent may edit prompt templates, manage or reorganize memory, reconfigure tool interfaces, or alter control logic, thereby explicitly updating the scaffolding $\Sigma_t$ that governs future execution. In this case, self-improvement is realized through direct, action-level self-modification.

These two modes correspond to distinct but complementary paradigms. \emph{Foundation model improvement} realizes self-improvement through self-induced learning in parameter space, whereas \emph{scaffolding improvement} realizes self-improvement through direct self-modification in execution space. Together, they capture the principal mechanisms by which FM-based agents may alter themselves over time.

\paragraph{Foundation model improvement.}
In foundation model improvement, self-improvement targets the parameters of the underlying foundation model while holding the agent-level scaffolding fixed. Starting from an induced policy $\pi_{\theta_t, \Sigma_t}$, the agent is repeatedly executed in the environment, generating self-induced experience through its own interaction trajectories.

We abstract this process by a self-induced update operator that maps the agent’s current configuration to an updated set of model parameters:
\begin{equation}
\theta_{t+1} = \mathcal{U}_\theta\bigl(\theta_{1:t},\; \mathcal{E}(\pi_{\theta_t, \Sigma_t};\Sigma_t,\mathcal{C}_t)\bigr), \quad \Sigma_{t+1} = \Sigma_t,
\end{equation}
where $\mathcal{C}_t$ is the task or deployment context defined above, $\mathcal{E}$ denotes the execution of the agent under its induced policy in $\mathcal{C}_t$, producing learning signals such as rewards, preference comparisons, synthetic labels, critiques, or verification outcomes, and $\mathcal{U}_\theta$ denotes a parameter-learning procedure that updates the foundation model parameters based on these signals.

Importantly, while the construction of these learning signals may involve rule-based evaluators, learned reward models, or auxiliary model invocations, the underlying data distribution is induced by the agent’s own policy. The operator $\mathcal{U}_\theta$ may instantiate policy-gradient methods, offline or online reinforcement learning, preference optimization, or other parameter-learning algorithms. Foundation model improvement typically operates on longer time scales, incurs substantial computational cost, and leads to stable, global changes in the agent’s internal representations, generalization behavior, and capabilities.

\paragraph{Scaffolding improvement.}
In \emph{scaffolding improvement}, self-improvement targets the agent-level structures surrounding the foundation model while keeping the model parameters $\theta$ fixed. The scaffolding $\Sigma_t$ governs how histories are converted into the conditioning context for the foundation model, how tool calls are specified and executed, how memory is stored and retrieved, and more generally how token sequences are constrained such that they correspond to valid, executable actions. Through these mechanisms, $\Sigma_t$ shapes the agent’s effective observation and action semantics, as well as the construction and evolution of the agent state $X_t$ during execution.

Formally, scaffolding improvement can be expressed as a self-induced update in which an \emph{agent-executed} procedure produces a meta-level signal that is then applied as an intrinsic update:
\begin{equation}
\Sigma_{t+1} \;=\; \mathcal{U}_\Sigma \Big(\Sigma_{1:t},\; \mathcal{E}\big(\pi_{\theta_t,\Sigma_t}; \Sigma_t, \mathcal{C}_t\big)\Big), \qquad \theta_{t+1}=\theta_t,
\end{equation}
where $\mathcal{E}$ denotes the execution of the agent (possibly under a meta-objective) to generate update-relevant artifacts, such as proposed prompt edits, memory reorganizations, tool-interface changes, or new control routines, and $\mathcal{U}_\Sigma$ denotes the system-level mechanism that commits these artifacts as structural scaffolding updates. Unlike the transient evolution of $X_t$, the update above is intended to endure beyond individual task boundaries.

These updates modify the induced policy $\pi_{\theta,\Sigma}$ by reshaping (i) the conditioning context supplied to the foundation model, (ii) the effective action space via admissibility constraints and tool schemas, and (iii) the execution semantics by which token sequences are parsed and grounded into environment actions. Compared to foundation model improvement, scaffolding improvement is typically faster, more reversible, and more context-dependent, yet it can produce substantial changes in the agent’s effective behavior and problem-solving strategies.

\paragraph{Skill as a reusable update.}
The two modes of self-reference above realize self-improvement either indirectly, through self-generated signals that an update rule consolidates into the parameters $\theta$, or directly, through actions that edit the scaffold $\Sigma$; a skill cuts across this distinction. We model a \emph{skill} as a reusable instance of the self-induced update operator $\mathcal{U}$: a named update to the agent's own configuration $\mathcal{A}_t$ that it retains and reuses. Acquiring a skill serializes this update into one of $\mathcal{A}_t$'s substrates: a tool and its calling convention ($\mathcal{T}$), an instruction or workflow ($p$), a memory entry ($m$), consolidated weights ($\theta$), or control logic ($g$). The skill's identity is the update it encodes; the substrate only names where it is stored. This is what makes ``skill'' orthogonal to the substrate axis of our taxonomy. A skill library is then a structured store of, and retrieval policy over, these serialized updates; the same store-and-retrieve structure recurs across substrates, which is why it surfaces in both tool routing and memory organization.

Reusability takes two forms. A skill may be invoked repeatedly, as a retained routine called many times across tasks, or applied once in the manner of an installer: a single update whose value lies in the persistent change it leaves behind, but which remains a portable artifact, reusable across agents and sessions. Either way, a skill is a first-class, serialized operator, in contrast to an ad hoc one-off action that leaves no retained trace.

We further distinguish two scopes according to what an invoked skill acts on. An \emph{object-level} skill acts on the task or world state: invoking it runs a (typically multi-step) routine through the execution operator $\mathcal{E}$ to carry out a sub-task (e.g., a learned \texttt{collect-wood} routine), the agentic analog of a temporally extended option in hierarchical reinforcement learning~\citep{sutton1999between, bacon2017option}. Here the $\mathcal{U}$ step lies in acquisition, which writes the routine into its substrate (e.g., $\mathcal{T}$), not in invocation. A \emph{meta-level} skill instead acts on the agent's own configuration $\mathcal{A}_t$: invoking it edits a component of $\Sigma_t$ or triggers a parameter update (e.g., writing a new tool, refactoring a prompt, consolidating experience into memory, or patching one's own scaffold). The installer-like skills above
are inherently meta-level. For self-improvement, the meta-level scope is the central one. Because a meta-level skill both acts on $\mathcal{A}_t$ and is itself serialized back into $\mathcal{A}_t$, the operator can become part of its own operand. Once such a skill is in turn improved, we recover the self-referential loop in which the improver evolves together with the system it improves~\citep{schmidhuber1987selfreferential, schmidhuber1993self,schmidhuber2003godel}.

Later sections describe how skills are serialized in recent works: as tools (Section~\ref{sec:Tool}), as memory workflows (Section~\ref{sec:Memory}); the application sections illustrate how acquired skills are invoked and reused across tasks.

\subsection{Connections to Related Learning Paradigms}
\label{subsec:Connections_to_Related_Learning_Paradigms}

Because the SI-FMA paradigm is deeply rooted in established learning paradigms, examining it through the lens of these foundational theories offers critical insights. We map SI-FMA onto classical frameworks to clarify where self-improving FM-based agents inherit existing assumptions, where they recombine familiar mechanisms, and where agent architecture makes these mechanisms operationally distinct in practice.

\textbf{Relation to Reinforcement Learning (RL).} 
Classical reinforcement learning provides a natural reference frame for SI-FMA: it formalizes how an agent improves its policy through interaction, and recent work on Agentic RL \citep{zhang2025landscape} extends this view to foundation-model-based agents. Mapping SI-FMA onto this frame clarifies which channels of self-improvement inherit standard RL machinery and which lie outside its formulation. Specifically, updates to $\theta$ correspond to standard policy optimization under a fixed decision process, whereas updates to $\Sigma$ lie outside the standard RL formulation, reshaping the decision process in which the policy operates.

\begin{itemize}
\item \textbf{Parameter updates ($\theta$)}.
When improvement targets the foundation model parameters using interaction trajectories, the process aligns with standard policy optimization. The foundation model acts as a large-scale policy network $\pi_{\theta}$, and techniques such as RLHF \citep{christiano2017deep, ouyang2022training} or self-play optimize $\pi_{\theta}$ via algorithms like Proximal Policy Optimization (PPO) \citep{schulman2017proximal} or Direct Preference Optimization (DPO) \citep{rafailov2023direct}.

\item \textbf{Scaffolding updates ($\Sigma$)}.
When improvement targets the scaffolding, the agent performs a form of structural meta-learning that has no direct counterpart in classical RL. First, whereas classical RL assumes a fixed action space and state representation, updating $\Sigma$ (e.g., adding tools or modifying memory) dynamically alters the effective state-action space and the observation processing logic, thereby reshaping the underlying Markov decision process (MDP) itself. Second, the optimization target itself differs in kind: $\Sigma$ comprises discrete, structured artifacts such as prompts, memory entries, and routing rules, which are typically updated through search, generation, or symbolic edits rather than gradient descent, and which remain explicit and inspectable rather than absorbed into network weights.
\end{itemize}

SI-FMA also departs from classical RL along a second axis: the source of the learning signal. While classical RL relies on external scalar rewards, self-improving agents increasingly utilize self-generated supervision, where the agent acts as its own critic to synthesize feedback signals from its own trajectories, judgments, or verifier modules.

\textbf{Relation to Online Learning.}
Online learning \citep{littlestone1988learning, hoi2021online} studies sequential decision or prediction under a data stream \citep{qi2025webrl}, where the learner updates each round and performance is assessed through cumulative loss or regret. SI-FMA intersects with this view when an agent is updated repeatedly during deployment and evaluation tracks an improvement trajectory rather than a single endpoint, though it is not restricted to the online setting—many self-improvement pipelines operate in batched or offline regimes. Where the paradigms meet, the two channels relate to online learning asymmetrically: $\theta$ updates recover the classical view of updating a hypothesis under a non-stationary stream, whereas $\Sigma$ updates enact a system-level analogue that shifts adaptation to explicit, inspectable components.

\begin{itemize}
    \item \textbf{Parameter updates ($\theta$).}
    These inherit the standard stability challenges of online learning, including distribution shift and catastrophic forgetting. Systems mitigate these risks through controlled update operators such as replay or rehearsal~\citep{robins1995catastrophic}, regularization~\citep{kirkpatrick2017overcoming,ramesh2026learning}, parameter-efficient tuning~\citep{hu2022lora}, and explicit versioning with rollback.

    \item \textbf{Scaffolding updates ($\Sigma$).}
    These provide a channel for rapid adaptation through prompting policies, memory read-write rules, tool routing, and orchestration logic. Externalizing adaptation in this way improves transparency and control, but does not eliminate forgetting and introduces its own failure modes such as memory poisoning, drift in tool semantics, and brittle template dependence.
\end{itemize}

\textbf{Relation to Active Learning.} Active learning studies how a learner selects queries under a labeling budget to maximize information gain, typically by requesting labels from an external oracle \citep{ren2021survey}. SI-FMA intersects with this view when an agent actively controls its data-acquisition process---for example, by targeting frequent failure modes, seeking environments that maximize verifier disagreement, or explicitly requesting human feedback. However, because modern agents often rely on self-generated verification or critiques rather than oracle labels, this behavior transcends standard active learning. It connects SI-FMA directly to classical artificial-curiosity frameworks, in which systems are explicitly designed to actively construct experiments that maximize learning progress, Bayesian surprise, or compression progress~\citep{schmidhuber1991possibility,storck1995reinforcement, schmidhuber2006developmental,schmidhuber2010formal,herrmann2026interestingness}. This perspective places self-improvement within a broader family of curiosity-driven exploration mechanisms, emphasizing that the key design choices are the query policy, the intrinsic objective, and the trustworthiness of the resulting feedback.

\section{A Taxonomy of Existing Approaches}
\label{sec:A_Taxonomy_of_Self_Improvement_Mechanisms}

Building on the definitions introduced in
Section~\ref{sec:Definitions}, we now introduce a taxonomy to organize existing approaches to self-improvement in FM-based agents. This section serves as a \emph{methodological classification} that provides a common reference frame for comparing a rapidly growing and heterogeneous literature. In the remainder of this section, we use $\IMPROVE_{target}(\,\cdot\,;\mathcal{S}_t)$ to denote an abstract self-improvement procedure in the sense of SI-FMA. Here $\mathcal{S}_t$ denotes an update signal produced through the agent’s own execution (the operator $\mathcal{E}$ in Section~\ref{sec:Definitions}), such as interaction trajectories, critiques, preferences, or other self-generated artifacts. This notation organizes existing approaches first by the target of modification, separating foundation-model updates from scaffolding updates. Within each target, approaches are further distinguished by the form of the self-induced learning signal that drives the update. For scaffolding improvement, we additionally group methods by the scaffold component being modified, such as prompts, memory, tools, or full scaffolding.

\subsection{Foundation Model Improvement}
\label{subsec:Foundation_Model_Improvement_(The_Parametric_Slow_Loop)}

Foundation model improvement targets the parameters of the underlying foundation model while leaving the agent-level scaffold unchanged:
\begin{equation}
    \theta_{t+1} = \IMPROVE_{\theta}(\theta_{1:t}; \mathcal{S}_{t}),
    \qquad
    \Sigma_{t+1} = \Sigma_{t}.
\end{equation}
In this paradigm, the agent’s own execution under its induced policy $\pi_{\theta_t,\Sigma_t}$ generates learning signals that are subsequently consumed by a parameter-update procedure. By committing the update directly into $\theta_t$, the agent internalizes the learned capabilities. This allows the adaptation cost to be amortized across future interactions, though such parametric updates typically operate on longer time scales and incur higher computational overhead. $\theta_{1:t}$ denotes the parameter history, enabling validation and rollback (e.g., reverting to a prior checkpoint) when a proposed update degrades performance or violates constraints.

\paragraph{Classification by signal form.}
As detailed in Section \ref{sec:Foundation_Model_Improvement}, we further categorize foundation model improvement according to the form of the self-induced signal $\mathcal{S}_t$:
\begin{itemize}
    \item \textbf{Intrinsic Generative Demonstrations} (primary update signal: $\mathcal{D}_t$):
    the agent synthesizes explicit training instances (e.g., demonstrations or augmented datasets) that are used for supervised-style learning
    ~\citep{wang2022self, lu2023self, zhao2024selfguidebettertaskspecificinstruction, shi2025taskcraft}.

    \item \textbf{Intrinsic Evaluative Feedback} (primary update signal: $e_t$):
    the agent constructs supervisory signals such as scalar rewards, preference pairs, or critiques to guide optimization and alignment
    ~\citep{gong2025strive, huang2025beyond, bai2022constitutional, bensal2025reflect, zhang2025adaptive}.

    \item \textbf{Extrinsic Exploratory Experience} (primary update signal: $\tau_t$):
    the agent learns from interaction trajectories and environment-grounded outcomes produced during execution
    ~\citep{wei2025codearc, dong2025tool, da2025agent, sun2025seagentselfevolvingcomputeruse}.
\end{itemize}

\subsection{Scaffolding Improvement}
\label{subsec:Scaffolding_Improvement_(The_Non_Parametric_Fast_Loop)}

Scaffolding improvement targets the agent’s operational scaffold while keeping the foundation model parameters fixed:
\begin{equation}
    \Sigma_{t+1} = \IMPROVE_{\Sigma}(\Sigma_{1:t}; \mathcal{S}_{t}),
    \qquad
    \theta_{t+1} = \theta_{t}.
\end{equation}
Here, $\IMPROVE$ commits structural changes into $\Sigma_t=(p_t,m_t,\mathcal{T}_t,g_t)$,
thereby reshaping how the frozen model is conditioned, grounded, and constrained during all subsequent executions. This paradigm typically yields fast, reversible, and task-specific adaptation without the risks of catastrophic forgetting.

\paragraph{Classification by scaffold component.}
Crucially, each of the following modifications represents a structural update to the agent's intrinsic configuration, distinguishing it from transient shifts in working memory. As detailed in Section~\ref{sec:Scaffolding_Improvement}, we further decompose scaffolding improvement according to which component of $\Sigma_t$ is targeted:
\begin{itemize}
    \item \textbf{Prompt Optimization ($p_t \rightarrow p_{t+1}$):}
     edits to structured prompts or in-context exemplars,
    $p_{t+1}=\IMPROVE_{p}(p_{1:t}; \mathcal{S}_t)$
    ~\citep{yuksekgonul2024textgradautomaticdifferentiationtext, guo2025evopromptconnectingllmsevolutionary, fernando2024promptbreeder, zhou2022large}.

    \item \textbf{Memory Evolution ($m_t \rightarrow m_{t+1}$):}
     updates to how experience is stored, consolidated, and retrieved,
    $m_{t+1}=\IMPROVE_{m}(m_{1:t}; \mathcal{S}_t)$
    ~\citep{chhikara2025mem0buildingproductionreadyai, wang2024agentworkflowmemory, zhang2025memgenweavinggenerativelatent, zhong2024memorybank}.

    \item \textbf{Tool Governance ($\mathcal{T}_t \rightarrow \mathcal{T}_{t+1}$):}
     refinement or expansion of the agent’s action space through tool creation or selection,
    $\mathcal{T}_{t+1}=\IMPROVE_{\mathcal{T}}(\mathcal{T}_{1:t}; \mathcal{S}_t)$
    ~\citep{haque2025advancedtoollearningselection, wang2025toolgenunifiedtoolretrieval, dong2025tool, zhang2025agentorchestraorchestratinghierarchicalmultiagent}.

    \item \textbf{Full Scaffolding Update ($\Sigma_t \rightarrow \Sigma_{t+1}$):}
    holistic reconfiguration of the agent’s operational architecture,
    $\Sigma_{t+1}=\IMPROVE_{\Sigma}(\Sigma_{1:t}; \mathcal{S}_t)$
    ~\citep{hu2025automateddesignagenticsystems, zhang2025darwingodelmachineopenended, novikov2025alphaevolvecodingagentscientific, xia2025livesweagentsoftwareengineeringagents}.
\end{itemize}

\section{Foundation Model Improvement}
\label{sec:Foundation_Model_Improvement}

\AlgoSel=3 
\ifnum\AlgoSel=3
\begin{wrapfigure}{r}{0.52\columnwidth}
\vspace{-\intextsep}
\centering

\scalebox{0.52}{%
\begin{minipage}{\columnwidth}
\begin{mdframed}[
  backgroundcolor=algobg,
  roundcorner=5pt,
  linewidth=0pt,
  innerleftmargin=4pt,
  innerrightmargin=4pt,
  innertopmargin=4pt,
  innerbottommargin=4pt
]
\begingroup
\SetAlFnt{\large}   
\SetAlCapFnt{\large}  
\SetAlCapNameFnt{\large} 
\begin{algorithm}[H]
\caption{Foundation\mbox{-}Model Improvement}
\label{alg:foundation_model_improvement}

\SetKwInOut{Input}{Input}
\SetKwInOut{Output}{Output}
\Input{Initial agent state $\mathcal{A}_0=(\theta_0,\Sigma)$; maximum iterations $T$}
\Output{Improved agent $\mathcal{A}_T=(\theta_T,\Sigma)$}

\BlankLine
\For{$t \leftarrow 0$ \KwTo $T-1$}{

  \textcolor{RoyalBlue}{\tcp{1) Construct or collect learning signal $\mathcal{S}_t$}}
  $\mathcal{S}_t \leftarrow \emptyset$

  \If{subcategory includes intrinsic generative demonstrations}{%
    $\mathcal{S}_t \leftarrow \mathcal{S}_t \cup \textsc{GenerateDemonstrations}(\mathcal{A}_t)$
    \tcp*{\S \ref{sec:Intrinsic_Generative_Demonstrations}}
  }
  \If{subcategory includes intrinsic evaluative feedback}{%
    $\mathcal{S}_t \leftarrow \mathcal{S}_t \cup \textsc{GenerateEvaluativeFeedback}(\mathcal{A}_t)$
    \tcp*{\S \ref{sec:Intrinsic_Evaluative_Feedback}}
  }
  \If{subcategory includes extrinsic exploratory experience}{%
    $\mathcal{S}_t \leftarrow \mathcal{S}_t \cup \textsc{CollectExperience}(\mathcal{A}_t,\text{env})$
    \tcp*{\S \ref{sec:Extrinsic_Exploratory_Experience}}
  }

  \textcolor{gray}{\tcp{Optional: quality control / weighting of signals}}
  $\mathcal{S}_t \leftarrow \textsc{FilterOrWeight}(\mathcal{S}_t)$

  \textcolor{RoyalBlue}{\tcp{2) Parameter update}}
  $\theta_{t+1} \leftarrow \textsc{Update}(\theta_{1:t},\mathcal{S}_t)$

  \textcolor{RoyalBlue}{\tcp{3) State update}}
  $\mathcal{A}_{t+1} \leftarrow (\theta_{t+1},\Sigma)$

  \If{\textsc{Converged}($\mathcal{A}_{t+1}$)}{break}
}
\Return{$\mathcal{A}_T$}
\end{algorithm}
\endgroup
\end{mdframed}
\end{minipage}%
}

\vspace{-10pt}
\end{wrapfigure}
\fi

\ifnum\AlgoSel=4
\begin{wrapfigure}{r}{0.52\columnwidth}
\vspace{-\intextsep}
\centering

\scalebox{0.52}{%
\begin{minipage}{\columnwidth}
\begin{mdframed}[
  backgroundcolor=algobg,
  roundcorner=5pt,
  linewidth=0pt,
  innerleftmargin=4pt,
  innerrightmargin=4pt,
  innertopmargin=4pt,
  innerbottommargin=4pt
]
\begingroup
\SetAlFnt{\large}   
\SetAlCapFnt{\large}  
\SetAlCapNameFnt{\large} 
\begin{algorithm}[H]
\caption{Scaffolding Improvement}
\label{alg:scaffolding_improvement}

\SetKwInOut{Input}{Input}
\SetKwInOut{Output}{Output}
\Input{Initial agent state $\mathcal{A}_0=(\theta,\Sigma_0)$; max iterations $T$}
\Output{Improved agent $\mathcal{A}_T=(\theta,\Sigma_T)$}

\BlankLine
\For{$t \leftarrow 0$ \KwTo $T-1$}{

  \textcolor{RoyalBlue}{\tcp{1) Collect scaffolding learning signal $\mathcal{S}_t$}}
  $\mathcal{S}_t \leftarrow \textsc{InteractAndEvaluate}(\mathcal{A}_t,\text{env})$
  \tcp*{\footnotesize traces, critiques, success/failure, cost}

  \textcolor{gray}{\tcp{Optional: quality control / weighting of signals}}
  $\mathcal{S}_t \leftarrow \textsc{FilterOrWeight}(\mathcal{S}_t)$

  \textcolor{RoyalBlue}{\tcp{2) Scaffolding update (keep $\theta$ fixed)}}
  \uIf{subcategory = Full scaffolding}{%
    $\Sigma_{t+1} \leftarrow \IMPROVE_{\Sigma}(\Sigma_{1:t};\mathcal{S}_t)$
    \tcp*{\S~\ref{sec:Full_Scaffolding}}
  }
  \Else{%
    $p_{t+1} \leftarrow p_t$; $m_{t+1} \leftarrow m_t$; $\mathcal{T}_{t+1} \leftarrow \mathcal{T}_t$

    \If{subcategory includes Prompt-based}{%
      $p_{t+1} \leftarrow \IMPROVE_{p}(p_t;\mathcal{S}_t)$
      \tcp*{\S~\ref{sec:Prompt_Optimization}}
    }
    \If{subcategory includes Memory-based}{%
      $m_{t+1} \leftarrow \IMPROVE_{m}(m_t;\mathcal{S}_t)$
      \tcp*{\S~\ref{sec:Memory}}
    }
    \If{subcategory includes Tool-based}{%
      $\mathcal{T}_{t+1} \leftarrow \IMPROVE_{\mathcal{T}}(\mathcal{T}_t;\mathcal{S}_t)$
      \tcp*{\S~\ref{sec:Tool}}
    }

    $\Sigma_{t+1} \leftarrow (p_{t+1},\, m_{t+1},\, \mathcal{T}_{t+1})$
  }

  \textcolor{RoyalBlue}{\tcp{3) State update}}
  $\mathcal{A}_{t+1} \leftarrow (\theta,\Sigma_{t+1})$
}
\Return{$\mathcal{A}_T$}
\end{algorithm}
\endgroup
\end{mdframed}
\end{minipage}%
}

\vspace{-16pt}
\end{wrapfigure}
\fi

Foundation-model-based self-improvement constitutes one of the most direct pathways for enhancing an agent’s intrinsic capabilities. This approach targets the agent’s core, namely the parameter set $\theta_{t}$ of its underlying FM, and updates these parameters to internalize new behaviors and reasoning patterns~\citep{wang2022self, bai2022constitutional, shafayat2025can, bensal2025reflect, dong2025tool, wang2025ragenunderstandingselfevolutionllm, zhou2025selfchallenging}. Such updates are typically achieved through gradient-based optimization, resulting in stable changes to the model weights. In this sense, the FM is treated as a continually learnable system that stores improvements in its parametric memory, rather than relying solely on transient execution states. Because the model parameters compress the agent's learned representations and behavioral priors, updating them allows the agent to fundamentally refine its policy, reduce systematic errors, and improve alignment with target objectives. In practice, the agent effectively serves as its own source of supervision: it uses its own execution to generate training signals—such as demonstrations, evaluations, or interaction trajectories—and applies learning algorithms to these self-induced signals to systematically improve its capabilities over successive updates.

To systematically analyze FM-based self-improvement, we classify existing methods by the nature of how the learning signal $\mathcal{S}_t$ is used to update the foundation-model parameters under our formalism. Consistent with the history-aware operator in Section~\ref{sec:Definitions}, we write the transition as $\mathcal{A}_{t+1}=\text{IMPROVE}(\mathcal{A}_{1:t};\mathcal{S}_t)$ with $\theta_{t+1}=\text{IMPROVE}_{\theta}(\theta_{1:t};\mathcal{S}_t)$ and $\Sigma_{t+1}=\Sigma_t$, allowing validation and rollback to prior checkpoints when necessary. This taxonomy yields three principal subcategories:
(i) \textbf{Intrinsic generative demonstrations} (\S \ref{sec:Intrinsic_Generative_Demonstrations}), where the learning signal is instantiated as $\mathcal{S}_t \approx \mathcal{D}_t$: the FM-based agent generates explicit examples, demonstrations, or task--solution pairs that are used for parameter updates.
(ii) \textbf{Intrinsic evaluative feedback} (\S \ref{sec:Intrinsic_Evaluative_Feedback}), where the learning signal is instantiated as $\mathcal{S}_t \approx e_t$: the FM-based agent or an internal evaluator judges candidate behavior through scores, preferences, confidence estimates, consistency signals, critiques, or revisions, which are then used to update the foundation model.
(iii) \textbf{Extrinsic exploratory experience} (\S \ref{sec:Extrinsic_Exploratory_Experience}), where the learning signal is instantiated as $\mathcal{S}_t \approx \tau_t$: the agent collects trajectories, rewards, observations, or executable outcomes through interaction with external or simulated environments. These subcategories are not mutually exclusive in deployed systems. Many practical pipelines mix intrinsic demonstrations, intrinsic evaluations, and exploratory experience within a single improvement cycle; for clarity, we categorize methods by the dominant signal source and objective used in their update operator. Algorithm~\ref{alg:foundation_model_improvement} summarizes the generic update loop of foundation-model improvement under self-generated learning signals.

\begin{figure*}[ht]
    \centering
    \includegraphics[width=\linewidth]{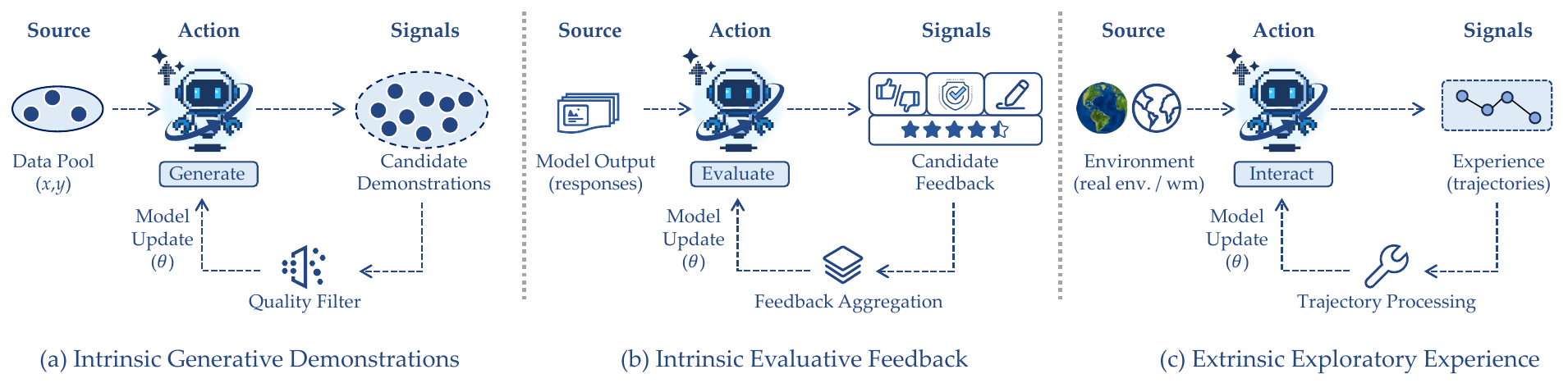}
    \caption{Overview of foundation model improvement under agent-induced learning signals. The agent improves the foundation model by generating intrinsic demonstrations, producing intrinsic evaluative feedback, or collecting extrinsic exploratory experience, each forming a distinct parameter-update loop.}
    \label{fig:FMI}
\end{figure*}

\subsection{Intrinsic Generative Demonstrations}
\label{sec:Intrinsic_Generative_Demonstrations}
\label{sec:Self_Generated_Data}

The remarkable capabilities of modern foundation models are fundamentally driven by their massive parameter scales, which inherently demand vast quantities of high-quality training data to continuously optimize and align~\citep{zhao2024selfguidebettertaskspecificinstruction, gan2025theoreticalunderstandingsyntheticdata}. The methods in this category fall into the group of internal generative processes, where FMs act simultaneously as the cognitive learner and the data synthesizer~\citep{zhao2024selfguidebettertaskspecificinstruction, zweiger2025selfadapting}. Leveraging the  semantic priors already compressed within their weights, FM-based agents can autonomously construct intrinsic generative demonstrations in forms such as instruction-response pairs, reasoning trajectories, and execution logs, without requiring new external observations~\citep{zhao2024selfguidebettertaskspecificinstruction, shi2025taskcraft, simonds2025ladder}. By casting the agent as its own data provider, this paradigm shifts the bottleneck of parameter updates from the costly acquisition of human annotations to the design of effective generation strategies and internal filtering mechanisms~\citep{qin2025dive}. 

Formally, as illustrated in Fig. \ref{fig:FMI} (a), under foundation-model improvement we treat these intrinsic generative demonstrations as the learning signal, i.e., $\mathcal{S}_t \approx \mathcal{D}_t$. At iteration $t$, the agent state $\mathcal{A}_t=(\theta_t,\Sigma_t)$ induces a generative distribution $\mathcal{P}_{\mathrm{gen}}(x,y \mid \mathcal{A}_t)$ and samples an intrinsic dataset
\begin{equation}
\mathcal{D}^{\mathrm{gen}}_t=\{(x_i,y_i)\}_{i=1}^{n_t}\sim \mathcal{P}_{\mathrm{gen}}(\cdot\mid \mathcal{A}_t).
\end{equation}
Optionally, a quality-control operator $\Phi_t$ filters or weights examples~\citep{jiang2025importance} to yield
$\tilde{\mathcal{D}}^{\mathrm{gen}}_t=\Phi_t(\mathcal{D}^{\mathrm{gen}}_t)$. The effective training set (and hence the learning signal) is then
$\mathcal{D}_t=\mathcal{D}^{\mathrm{base}}_t\cup \tilde{\mathcal{D}}^{\mathrm{gen}}_t$,
where $\mathcal{D}^{\mathrm{base}}_t$ denotes any existing data.

The parameter update can be written compactly as $\theta_{t+1}=\IMPROVE_{\theta}(\theta_{1:t};\mathcal{D}_t)$, where $\IMPROVE_{\theta}$ denotes the outcome of optimizing an empirical objective on $\mathcal{D}_t$ (typically initialized at the current active checkpoint $\theta_t$). We write this objective as $\mathcal{L}(\theta;\mathcal{D}_t)$, with $\Omega(\theta,\theta_0)$ an optional regularizer around an initialization $\theta_0$ and $\lambda\ge 0$ its coefficient:
\begin{equation}
\theta_{t+1}=\arg\min_{\theta}\;\mathcal{L}(\theta;\mathcal{D}_t)+\lambda\,\Omega(\theta,\theta_0).
\end{equation}
In practice, $\IMPROVE_{\theta}$ is implemented via iterative fine-tuning with a gradient-based optimizer. Let $\theta_t^{(0)}=\theta_t$, and let $\mathcal{B}_t^{(k)}$ denote a minibatch drawn from $\mathcal{D}_t$ according to the weighting induced by $\Phi_t$. Using step size $\eta_k$ at inner step $k$ and the gradient operator $\nabla_{\theta}$ with respect to $\theta$, for $k=0,\dots,K_t-1$,
\begin{equation}
\theta_t^{(k+1)}=\theta_t^{(k)}-\eta_k\,\nabla_{\theta}\mathcal{L}\big(\theta_t^{(k)};\mathcal{B}_t^{(k)}\big),
\end{equation}
and the outer-loop update is $\theta_{t+1}=\theta_t^{(K_t)}$. To systematically explore this approach, we consider three key aspects: the strategy used to generate these intrinsic demonstrations, the format of the generated data, and the challenges that arise in closing the self-improvement loop.

\paragraph{\textcolor{Plum}{Generation strategies.}}
The strategy of data generation is a critical factor for the quality and scalability of a self-improvement loop. 
The underlying techniques start with a small set of example seeds and then build a larger instruction-output pair corpus.\citep{wang2022self, zhao2024selfguidebettertaskspecificinstruction}. Methods such as Evol-Instruct \citep{xu2024wizardlm} go beyond simple volume expansion. They drive complexity evolution by using LLM to rewrite instructions and gradually increase the difficulty of the instructions, thereby overcoming the limitations of humans in building highly complex scenes. To improve data quality, more sophisticated methods have added a refining or filtering stage.  \cite{huang2023large} employs self-consistency to filter high-confidence inference paths, while \cite{singh2023beyond} uses external verifiers like unit tests to select only correct solutions from the model’s attempts. In addition to bootstrapping and filtering, some approaches conceptualize data generation as a form of curriculum learning. \cite{simonds2025ladder} create a learning curriculum by recursively decomposing complex problems into simpler sub-problems, and \cite{qin2025dive} explicitly introduce mechanisms to expand the sample pool to counteract the risk of diminished output diversity over time. More recently, Test-Time Self-Improvement (TT-SI) \citep{acikgoz2025selfimprovingllmagentstesttime} shifts the paradigm from massive offline generation to highly sample-efficient, on-the-fly adaptation. By detecting weak cases at inference time via uncertainty estimation, the agent self-generates targeted training examples for its specific blind spots and performs targeted low-rank adaptation (LoRA) fine-tuning; this yields immediate, per-instance improvements with a fraction of the data cost.

Each of these strategies has distinct strengths and failure modes. For example, self-consistency filtering rests on the assumption that correct reasoning tends to be self-agreeing. It can break down when the model is confidently wrong, because repeated reasoning may repeatedly converge to the same incorrect conclusion \citep{chen2024failuresselfconsistencymultistepreasoning, anonymous2025complementing}. 
Curriculum-based generation can significantly improve learning performance for complex tasks, but it becomes vulnerable when the chosen decomposition method is flawed or ignores necessary contextual information. For example, this can occur when the subproblems are constructed in a way that no longer preserves the constraints of the original task \citep{zhou2023leasttomost, xu2024wizardlm}.
Interactive refinement also depends on the model's ability to detect and correct its own errors. When certain types of errors are beyond the model's perception range, the improvement process may exacerbate these errors rather than eliminate them \citep{shinn2023reflexionlanguageagentsverbal, huang2024largelanguagemodelsselfcorrect}. Therefore, choosing an appropriate generative strategy requires matching the method to the model's current capabilities while avoiding mechanisms that focus on or amplify existing blind spots in the model.

\paragraph{\textcolor{Plum}{Data formats.}} These intrinsic demonstrations take various forms, ranging from simple input-output pairs to complex, structured artifacts designed to supervise specific cognitive abilities. Basic forms include simple instruction-response pairs, often with explicit reasoning traces added (e.g., chain-of-thought explanations) to make the model's intermediate reasoning steps transparent and trainable \citep{huang2023large}. 
However, in scenarios involving multi-step decision-making or tool usage, agents often struggle with long-term tasks because suboptimal behaviors accumulate, eventually causing them to deviate from the correct path. To address this issue, the generated data should capture structured trajectories, rather than the final result. For example, tool application programming interface (API) calls or code execution sequences, combined with intermediate environment feedback and validation labels \citep{shi2025taskcraft, singh2023beyond}, can provide fine-grained supervision, teaching agents how to handle dependencies and recover from errors.

This inherent generative paradigm also extends to multimodal scenarios, where agents synthesize training samples, combining visual inputs such as images or videos with textual descriptions and reasoning to improve performance on tasks including visual reasoning \citep{zhang2025will}. At a higher level of abstraction, these demonstrations can take the form of metacognitive artifacts. For example, decomposed problem trees \citep{simonds2025ladder} and self-editing instructions \citep{zweiger2025selfadapting}, both of which can support complex planning and self-modification. More generally, the choice of data format should be consistent with the target capability, as different formats encode different types of supervisory information. Inference trajectories are suited for logical reasoning, code combined with tests is well-suited for programming, and structured action logs provide supervision for tool usage.

\paragraph{\textcolor{Plum}{Challenges and safeguards.}}
This intrinsic generative paradigm still faces considerable challenges despite its success. First, recursively training the model on the generated corpus introduces the risk of model collapse and forgetting \citep{shumailov2024ai,shumailov2024curserecursiontraininggenerated}. If the model generates defective, biased, or erroneous samples, it will internalize these errors during fine-tuning, creating a negative feedback loop that degrades model performance \citep{zhao2024selfguidebettertaskspecificinstruction, wang2022self}. Insufficient diversity in generated demonstrations also narrows the solution space and leads to pattern collapse during iterations. In fact, agents may get trapped in knowledge bubbles and repeatedly generate data, which reinforces their existing biases and traits instead of generating genuine new capabilities \citep{wu2025progress, yuan2025superficial}.

To make these inherent generation loops more stable and reliable, several safeguards have been proposed. A simple and effective safeguard is to retain artificially generated benchmark data and accumulate generated demonstration data on top of it, which can prevent collapse under repeated training \citep{gerstgrasser2024is}. Another work strengthens quality control by using more robust checkers to verify generated samples. These checkers include external validators specifically designed for model-generated content \citep{yi2025escapingmodelcollapsesynthetic} and formal systems for inference verification, such as theorem provers \citep{Leang_2025}. Complementary data-centric safeguards focus on selecting and filtering the training corpus to remove low-quality or harmful examples. To address diversity loss and pattern collapse, diversity-aware pooling expansion and selection mechanisms have been shown to maintain exploratory nature during the iterative process \citep{qin2025dive}. Furthermore, shifting from large-scale offline generation to proactive, uncertainty-driven generation during testing naturally mitigates the accumulation of redundant data by focusing computational resources only on out-of-distribution or challenging samples \citep{acikgoz2025selfimprovingllmagentstesttime}. Finally, because automated evaluators and validation processes can be unreliable or drift over time, recent research emphasizes human auditing and human-computer interaction to improve evaluation criteria as practical safeguards \citep{10.1145/3731120.3744588, do2025generateevaluateiteratesynthetic}.

\subsection{Intrinsic Evaluative Feedback}
\label{sec:Intrinsic_Evaluative_Feedback}
\label{sec:Self_Generated_Supervision}

Just as massive parameter scales demand vast quantities of demonstration data, aligning and refining these capabilities requires high-quality supervisory signals. The bottleneck in traditional supervisory signal acquisition lies in the high cost of manually labeled preferences and human evaluation. To overcome this limitation, the methods in this category drive a paradigm shift by reframing the acquisition of supervision as an internal evaluative process. In this regime, FMs act not only as the output generator and the self-evaluator, but ultimately as the cognitive learner that internalizes its own judgments. Leveraging their language-native capabilities to follow rubrics, compare alternatives, and articulate reasoning, FM-based agents can autonomously produce \textit{intrinsic evaluative feedback}, such as scalar scores, preference pairs, consistency signals, and natural language critiques, without requiring new environment interaction. Unlike intrinsic generative demonstrations (Section~\ref{sec:Self_Generated_Data}) that provide examples, or extrinsic exploratory experience (Section~\ref{sec:Self_Generated_Experience}) derived from environment-grounded outcomes, the dominant learning signal here is the agent's endogenous judgment of its own candidate behaviors. By treating the agent as a critic of itself, this paradigm shifts the bottleneck of parameter updates to the design of robust internal evaluation rubrics and critique mechanisms.

Formally, let $\mathcal{A}_t=(\theta_t,\Sigma_t)$ denote the current agent configuration. Given an input or task context $x$, the agent samples a set of candidate outputs
\begin{equation}
\mathcal{Y}_t(x)=\{y_t^{(1)},\ldots,y_t^{(K)}\}, \qquad 
y_t^{(k)} \sim \pi_{\theta_t,\Sigma_t}(\cdot \mid x).
\end{equation}
An intrinsic evaluator $\phi_t$, which may be implemented by the current model, an auxiliary judge model, a learned reward model, or a fixed scaffolded critique procedure, maps the task, candidates, and evaluation criteria $\kappa_t$ into an evaluative signal:
\begin{equation}
e_t=\phi_t(x,\mathcal{Y}_t(x);\kappa_t).
\end{equation}
Here, $\phi_t$ is the source of feedback rather than the target of the update in this subsection. When an auxiliary evaluator is itself trained, we treat it as part of the feedback-construction pipeline; the method is classified here because the resulting evaluative signal is used to update the foundation model. The signal $e_t$ may instantiate a scalar reward $r_t$, a preference relation $y^+\succ y^-$, a confidence or uncertainty score, a textual critique $c_t$, or a refined target $y_t^\ast$. Under foundation-model improvement, this feedback becomes the update signal, $\mathcal{S}_t\approx e_t$, and the parameter update is written as
\begin{equation}
\theta_{t+1}=\IMPROVE_{\theta}(\theta_{1:t};e_t), \qquad \Sigma_{t+1}=\Sigma_t.
\end{equation}
Depending on the form of $e_t$, the update may be implemented through reinforcement learning, preference optimization, reward-model training, critique-conditioned fine-tuning, or supervised fine-tuning on revised outputs.

\paragraph{\textcolor{Plum}{Rubric feedback.}}
A first family of methods derives evaluative feedback by asking a model to judge candidate outputs against explicit criteria. The criteria may be task instructions, grading rubrics, safety principles, constitutional rules, or domain-specific preferences. In this setting, the evaluation standard $\kappa_t$ serves as the context for judgment: candidate outputs are scored, ranked, or compared according to how well they satisfy the stated criteria. The evaluator may also produce a brief rationale for its judgment, but the primary learning signal is the score, ranking, or preference relation rather than an open-ended revision.

Constitutional AI is a representative early example of this pattern. Instead of relying only on direct human preference labels, the model critiques and ranks outputs according to a set of written principles, producing AI feedback that can be used to train a preference model and subsequently optimize the policy through reinforcement learning~\citep{bai2022constitutional}. Recent self-improvement methods extend this judge-based loop more directly into parameter-updating systems. Meta-Rewarding trains a model to act, judge its own responses, and further evaluate its own judgments, turning both judgments and meta-judgments into preference pairs for iterative alignment improvement~\citep{wu-etal-2025-meta}. \cite{simonds2025self} show that LLM judges can provide reward signals without reference solutions, enabling reinforcement learning from model-generated judgments in reasoning domains where programmatic rewards are difficult to specify. Self-Evolved Reward Learning further studies a learned reward model that labels additional data for its own improvement and supports reinforcement learning from self-feedback~\citep{huang2025selfevolved}. Together, these methods illustrate how explicit principles, rubrics, or judge prompts can transform natural-language evaluation standards into scalable supervision for foundation-model improvement. The main advantage is flexibility, since the same model-based evaluator can be conditioned to assess helpfulness, harmlessness, factuality, reasoning quality, or format compliance. The main limitation is reliability, since ambiguous criteria or biased judges may reward superficial compliance rather than genuine improvement.

\paragraph{\textcolor{Plum}{Consistency feedback.}}
A second family of methods exploits the stochastic behavior of foundation models to derive feedback signals for self-improvement.
When ground-truth labels or reliable external verifiers are unavailable, the agent can generate multiple candidate solutions for the same task and use agreement among them as an intrinsic signal. The underlying assumption is not that consistency guarantees correctness, but that agreement, entropy, or self-certainty can provide a useful weak signal for ranking or weighting candidates.

Concretely, the agent samples $\mathcal{Y}_t(x)=\{y_t^{(1)},\ldots,y_t^{(K)}\}$ and applies an aggregation operator $C$ to produce a confidence or reward-like signal:
\begin{equation}
e_t=C(\mathcal{Y}_t(x)).
\end{equation}
For example, majority voting can identify a consensus answer, predictive entropy can be used as a confidence score, and agreement among independent reasoning paths can be used to construct preference or reward signals. TTRL~\citep{zuo2025ttrl} uses majority voting over multiple generated answers at test time to produce reward signals for reinforcement learning. SRT~\citep{shafayat2025can} similarly investigates whether self-consistency can support self-improvement without ground-truth labels. EMPO~\citep{zhang2025right} encourages reasoning behavior through entropy-based signals, while INTUITOR~\citep{zhao2025learning} uses self-certainty as an intrinsic reward. In these methods, the agent does not primarily learn from newly generated demonstrations; rather, it learns from evaluative structure extracted from its own distribution over answers.

The strength of this family is that it can be applied broadly, including in domains where labels are expensive and formal verification is unavailable. However, consistency is only a proxy for correctness. If a model is systematically biased or confidently wrong, repeated sampling may amplify the same error. Confidence-derived feedback also depends on calibration: a model's expressed certainty may not align with actual correctness. These issues make consistency-based self-improvement methods, while useful, quite fragile, especially when model errors are correlated across different samples.~\citep{tan-etal-2025-consistent, feng2025rethinkingllmuncertaintymultiagent, till2025teamingllmsdetectmitigate, anonymous2025complementing}.

\paragraph{\textcolor{Plum}{Corrective feedback.}}
A third family of methods treats natural-language critique or modification itself as the core evaluation outcome. Unlike rubric-based judging (whose main output is a score or preference under a stated criterion), error-correcting feedback methods require the model to identify specific flaws in candidate outputs and propose modifications. Therefore, this feedback is corrective, not merely comparative.

Let $R$ denote a critique-and-revision operator. Given a candidate output $y_t$, the agent produces
\begin{equation}
e_t=R(x,y_t), 
\end{equation}

where $e_t$ is composed of $c_t$ and $y_t^\ast$, $c_t$ is a critique and $y_t^\ast$ is a revised output. The pair $(y_t,y_t^\ast)$ can instantiate a preference signal $y_t^\ast \succ y_t$, while the critique $c_t$ can be used as explanatory supervision or as a natural-language training signal. ``ReST meets react''~\citep{aksitov2023rest} combines agentic reasoning and self-training to improve multi-step problem solving. SELF~\citep{lu2023self} uses language feedback to iteratively refine generated answers, transforming model-produced critiques and revisions into signals for self-improvement. RISE~\citep{qu2024recursive} uses interaction-based recursive self-reflection and fine-tunes based on improved predictions. Reflect, Retry, Reward~\citep{bensal2025reflect} follows a similar loop in which the model reflects on failures, retries the task, and uses the resulting feedback to guide reinforcement learning. The AlphaAllM approach~\citep{tian2024toward} further integrates search and critique, using model-generated evaluations during the search process to construct stronger training signals. These methods exploit the ability of foundation models to  provide semantic, actionable, and informative feedback that is richer than simple scalar rewards. A critique can identify missing constraints, unsupported claims, incorrect reasoning steps, or unsafe implications, thus providing a richer signal for parameter updates. This flexibility also introduces failure modes since the model may produce plausible but incorrect critiques, or learn to satisfy the surface form of the critique rather than its underlying goals.  Therefore, critique-based feedback is more reliable for parameter-level self-improvement when combined with safeguards such as comparing the original and modified outputs, filtering low-quality critiques, using heterogeneous critique models, or combining internal critiques with external validation when available.

\paragraph{\textcolor{Plum}{Trade-offs and safeguards.}}
The advantage of intrinsic evaluation feedback lies in reducing reliance on human annotation and converting the agent's own output into a learning signal that can be directly used by the parameter update algorithm. Its main risk is that the evaluator is often tightly coupled with the policy to be improved, so this loop may reinforce common blind spots, reward outputs that conform to the model's existing preferences, overfit superficial evaluation criteria, or become biased with repeated use and reinterpretation of the evaluation criteria. Signals based on confidence and consistency increase vulnerability when the model is poorly calibrated or when there are correlations between its samples. Therefore, reliable use of this paradigm requires safeguards such as separating the generator and evaluator at different checkpoints or model families, maintaining external anchors by retaining human annotations or context-based validation, treating disagreements between evaluators as signals of uncertainty, and regularly reviewing the scoring criteria, rewarding the model, and evaluating quality. In practice, intrinsic evaluation feedback can be viewed as a component of a broader improvement loop, supplemented by demonstrations, external validation, and exploratory experiences, rather than as the only source of supervision.

\subsection{Extrinsic Exploratory Experience}
\label{sec:Extrinsic_Exploratory_Experience}
\label{sec:Self_Generated_Experience}

The preceding two subsections focused on intrinsic self-improvement signals, where the agent improves the foundation model using its own generated demonstrations (\S \ref{sec:Intrinsic_Generative_Demonstrations}) or evaluative judgments (\S \ref{sec:Intrinsic_Evaluative_Feedback}). Extrinsic exploratory experience differs in that the learning signal is grounded in what happens after the agent acts. Under our foundation-model improvement formalism, the learning signal is instantiated as experience, $\mathcal{S}_t \approx \tau_t$, where $\tau_t$ denotes trajectories collected by executing the policy $\pi_{\theta_t, \Sigma_t}$ in a task environment or its learned proxy. The parameter update can be written as $\theta_{t+1} = \text{IMPROVE}_\theta(\theta_{1:t}; \tau_t)$, with $\Sigma_{t+1} = \Sigma_t$. While this view connects self-improvement back to the classical reinforcement learning framework \citep{silver2025welcome, zhao2024expel}, experience for foundation-model agents is not just a stream of state-action-reward tuples $(s,a,r,s')$. A trajectory may contain web pages, screenshots, code logs, compiler errors, tool calls, and intermediate reasoning traces. Because these artifacts are readable by foundation models, the same experience can be flexibly reused across reinforcement learning, supervised fine-tuning, preference construction, and failure analysis. However, acquiring and utilizing this rich experience introduces distinct difficulties: interaction can be slow or costly, rewards may be sparse or delayed, verifiers can be gamed, and learned world models \citep{schmidhuber1990making, schmidhuber2015learning,nanbo2025facts} may produce plausible but counterfactual transitions.

To address these challenges, we organize existing methods by how the exploratory experience is obtained. For interaction with grounded task environments, the agent acts in real, sandboxed, or rule-based task settings, and the learning signal comes directly from the task environment (e.g., state changes, unit tests, or task-specific verifiers). For interaction with simulated proxy environments, a learned world model serves as a proxy for the task environment, generating predicted states, rollouts, or outcomes for policy improvement. Notably, these two modes are not mutually exclusive. In fact, as demonstrated by early controller-world-model architectures \citep{schmidhuber1990making}, a system may use grounded interaction to collect data and update its world model, and subsequently use simulated interaction to plan, explore, or improve its policy.

\subsubsection{Interaction with Grounded Task Environments.}
\label{sec:Grounded_Interaction_with_Verifiable_Environments}
\label{sec:Environment_Interactive}

In this mode, the agent learns by directly acting in a task environment. Typical settings include code interpreters for coding agents, web APIs for web agents, mobile user interfaces (UIs) for GUI agents, and physical environments for embodied agents. The learning signal comes from the environment's response—such as state changes, execution traces, or unit tests. Training then proceeds by collecting these interaction trajectories and updating the foundation-model policy using reinforcement learning methods like PPO~\citep{schulman2017proximal}, or preference-based objectives like DPO~\citep{rafailov2023direct} when successful and unsuccessful trajectories can be contrasted. A useful way to organize this line of work is by the source of feedback:

\textit{(1) Programmatic verifiers} provide the clearest form of grounded signal. When a task admits an executable check, the agent can be trained without learning a separate reward model. Code generation is the canonical case, since unit tests provide direct pass or fail feedback for proposed programs. \textit{Agent-RLVR}~\citep{da2025agent}, for example, uses unit-test outcomes to  guide policy optimization, contrasting successful programs against failed attempts. The same pattern extends beyond code whenever an output can be checked by an external procedure, including structured query language (SQL) execution, theorem proving, tool use with checkable post-conditions, and structured tasks with deterministic success criteria.

\textit{(2) Learned reward models} evaluate trajectories collected from the task environment. In this setting, the task environment naturally produces the trajectory, while the learned evaluator only scores its success. \textit{WebRL}~\citep{qi2025webrl} trains an outcome-supervised reward model to label web-navigation trajectories automatically, reducing reliance on hand-crafted success criteria or human annotation. \textit{UI-Genie}~\citep{xiao2025uigenie} couples policy learning with reward-model refinement, using validated trajectories to train the agent and step-level labels from both successful and failed trajectories to improve the evaluator. Related work such as \textit{MobileGUI-RL}~\citep{shi2025mobilegui} adapts Group Relative Policy Optimization (GRPO) to mobile GUI navigation with trajectory-aware advantages and rewards that combine task success with execution efficiency. Across these methods, the feature is that feedback is computed over rich linguistic or visuolinguistic trajectories, which makes it natural to construct rewards, preferences, and step-level supervision from the same interaction record.

A third approach leverages \textit{(3) self-generated tasks} to acquire grounded experience. In these systems, the agent may propose new tasks or goals, but the learning signal remains extrinsic because candidate solutions are accepted or rejected by execution, verification, or environmental feedback. \textit{Absolute Zero}~\citep{zhao2025absolute} uses self-play to generate tasks and solutions in an open-ended environment, while execution-based validation determines which solutions are retained for learning. \textit{ETO}~\citep{song2024trial} learns more conservatively from contrasts between successful and failed trajectories in fixed environments. These methods blur the boundary between intrinsic and extrinsic signals at the level of task selection, but not at the level of supervision. The agent may choose what to try, while the environment determines whether the attempt succeeds.

Finally, standardized platforms have emerged to facilitate research across these grounded interaction modes. For example, \textit{AgentGym}~\citep{xi2024agentgym} provides unified APIs for interaction, evaluation, and training across multiple agent tasks. Such platforms significantly reduce the engineering cost of iterative experience collection, making it easier to compare reward sources, training objectives, and update procedures under shared environmental feedback.

\subsubsection{Interaction with Simulated Proxy Environments}
\label{sec:Efficient_Interaction_with_Simulated_Environments}
\label{sec:World_Models}

The simulated-interaction mode equips the agent with an internal predictive model of the environment. A typical pipeline first collects interaction trajectories from a task environment, then trains a world model to predict the environment's response to the agent's actions \citep{schmidhuber1990making, ha2018world}. Abstractly, this can be written as a learned dynamics model $W(s_{k+1}, r_k \mid s_k, a_k)$, which predicts the next state and reward from the current state and action at step $k$ of a trajectory. The policy can then obtain additional experience by interacting with this learned proxy instead of repeatedly querying the original task environment. This internal simulation improves sample efficiency and reduces the cost or risk of exploration, especially when direct interaction is slow, expensive, or unsafe~\citep{ding2025understanding, richens2025general}.

While classical world models successfully predict transitions over both compact state vectors and raw pixel spaces~\citep{ha2018world}, what distinguishes foundation-model agents is the extensive prior knowledge embedded in the learned dynamics. In FM-based systems, the world model is typically a generative language, vision-language, or video model that predicts high-dimensional observations—such as the next web page or video frame—in the exact format consumed by the policy. This design offers two practical advantages. First, large-scale generative pretraining provides a powerful prior over environment dynamics, significantly reducing the task-specific interaction needed to fit a useful proxy environment. Second, because generated observations lie in the same representation space as the policy input, simulated experience can be used directly without a separate representation-alignment step. This holds true for both linguistic web models and pixel-space video models like WMPO~\citep{zhu2025wmpoworldmodelbasedpolicy}. Together, these advantages make FM-based world models effective for planning, trajectory synthesis, and failure analysis.

This pattern is most developed in web navigation, where direct interaction can be slow and search over action sequences can be expensive. \textit{WebEvolver}~\citep{fang2025webevolver} trains a coevolving world model to predict next web observations and uses simulated rollouts to refine the agent policy. \textit{WebSynthesis}~\citep{gao2025websynthesisworldmodelguidedmctsefficient} uses a learned web world model for reversible, search-based trajectory synthesis, while \textit{WebDreamer}~\citep{gu2025is} leverages a web transition model for model-based planning to guide action selection. Beyond web navigation, \textit{SPA}~\citep{chen2025internalizingworldmodelsselfplay} learns explicit state-estimation and transition models through self-play fine-tuning, using them to initialize and stabilize downstream policy optimization. In embodied control, \textit{WMPO}~\citep{zhu2025wmpoworldmodelbasedpolicy} learns a pixel-space world model and optimizes the policy over imagined rollouts to avoid costly physical trial-and-error. A related direction uses structured memories of prior interaction rather than full generative dynamics. \textit{GLoW}~\citep{kim2025dualscaleworldmodelsllm} maintains a dual-scale textual world memory—consisting of a global frontier of high-value discoveries and local advantage reflections—to guide a Go-Explore-style agent in text-based games, achieving strong performance with 100--800$\times$ fewer real environment interactions than RL baselines.

\paragraph{\textcolor{Plum}{Challenges.}} Across extrinsic exploratory experience, foundation-model agents inherit classical difficulties of RL, including sparse and delayed rewards, low-throughput real interaction, overfitting to imperfect proxy environments. They also introduce failure modes specific to this setting. \textit{Reward hacking through language} is more readily available than in classical RL: a foundation-model agent can satisfy the literal condition of a verifier (e.g., exploiting prompt loopholes in an LLM judge) without solving the underlying task, because the verifier's specification is itself a linguistic object that the agent can manipulate. \textit{Capability regression} arises because extensive RL updates on narrow extrinsic rewards can erode the broader competencies the foundation model acquired in pretraining, a tension absent in agents trained from scratch. \textit{Hallucinated dynamics} pose a distinctive risk for world-model approaches, since generative simulators can fabricate plausible but incorrect transitions that the policy then learns to exploit. Finally, the linguistic, multimodal nature of trajectories creates a \textit{trajectory-length and context-window tension} unique to this setting: long-horizon experience must be compressed or summarized to fit within the model's context before it can be used as a learning signal. Current research therefore emphasizes more reliable extrinsic verifiers and reward models, world models with calibrated uncertainty over their own predictions, and training procedures that update the foundation model on extrinsic experience without sacrificing its general capabilities.

\begin{ttcolorbox}[Takeaway]
This section reviewed parameter-centric FM-based self-improvement, where agents refine their foundation models via intrinsic demonstrations, intrinsic evaluative feedback, or extrinsic experience. This paradigm allows learned behaviors to be broadly encoded into foundation model weights. However, success depends on managing the reliability of self-generated signals, mitigating distribution shift, and balancing the computational demands of iterative training.
\end{ttcolorbox}

\section{Scaffolding Improvement}
\label{sec:Scaffolding_Improvement}

This section analyzes the scaffolding improvement paradigm for FM-based agents, where we approach this by decomposing the agent's operational scaffold into core components. This workflow begins with receiving and interpreting instructions, then contextualizes them using internal knowledge and memory, proceeds to executing actions in the world, and ultimately evolves the overall operational structure. 

Formally, as established in Section ~\ref{sec:Definitions}, scaffolding improvement corresponds to transitions that modify the operational scaffold while explicitly keeping the foundation-model parameters frozen ($\theta_{t+1}=\theta_{t}$). Driven by an execution-derived learning signal $\mathcal{S}_{t}$ (e.g., task outcomes, critique, or execution errors), the update is formalized as $\Sigma_{t+1}=\IMPROVE_{\Sigma}(\Sigma_{1:t};\mathcal{S}_{t})$. By maintaining a version history ($\Sigma_{1:t}$), the system inherently supports validation and rollback against harmful modifications across all components. We structure the discussion by increasing depth of architectural intervention. Concretely, we instantiate the scaffolding as $\Sigma_{t}:=(p_{t},m_{t},\mathcal{T}_{t},g_{t})$, where $p_{t}$ denotes the prompt template, $m_{t}$ the internal memory, $\mathcal{T}_{t}$ the external tool set, and $g_{t}$ the control logic. Algorithm~\ref{alg:scaffolding_improvement} summarizes the generic improvement loop across these components:

\begin{itemize}[leftmargin=*]
    \item \textbf{Prompt-based improvement} (Section~\ref{sec:Prompt_Optimization}) targets the agent's input and instruction layer, formalized as $p_{t+1}=\IMPROVE_{p}(p_{1:t};\mathcal{S}_t)$.  As the primary semantic interface through which the agent perceives its tasks, optimizing the prompt provides a direct way to refine task communication, objectives, and constraints without altering the internal parameters of the foundation model.

    \item \textbf{Memory-based improvement} (Section~\ref{sec:Memory}) equips the agent with an evolving internal cognitive resource, updated via $m_{t+1}=\IMPROVE_{m}(m_{1:t};\mathcal{S}_t)$. 
    By dynamically storing, pruning, and retrieving past experiences, the agent shifts from memoryless execution to cumulative learning, supporting long-horizon reasoning and trustworthy adaptation.

    \item \textbf{Tool-based improvement} (Section~\ref{sec:Tool}) strengthens the agent's execution interface through the update $\mathcal{T}_{t+1}=\IMPROVE_{\mathcal{T}}(\mathcal{T}_{1:t};\mathcal{S}_t)$. 
    By autonomously refining and managing external callable modules (e.g., web search, code interpreters), the agent translates internal decisions into precise, executable actions, effectively extending its capabilities beyond the native limits of the foundation model.

    \item \textbf{Full scaffolding improvement} (Section~\ref{sec:Full_Scaffolding}) represents the most profound level of architectural intervention, formalized as $\Sigma_{t+1}=\IMPROVE_{\Sigma}(\Sigma_{1:t};\mathcal{S}_t)$. 
    By treating the entire codebase and operational logic as a mutable substrate, the agent dynamically reconfigures how its perception, reasoning, and execution faculties are integrated into a coherent whole.
\end{itemize}

\textbf{Crucially, these intervention types are compositional rather than mutually exclusive:} a single update can edit multiple scaffold components simultaneously, and full-scaffolding methods naturally subsume component-level edits while adding archive-based exploration and stronger acceptance tests.

\AlgoSel=4 

\subsection{Prompt}
\label{sec:Prompt_Optimization}
The prompt serves as the agent's core behavioral prior, defining how the foundation model parses its environment. Prompt optimization is therefore a central and highly accessible form of scaffolding improvement. While early methods relied on manual heuristic tuning, the field is rapidly developed toward automated, signal-driven improvement loops. Central to this development is the transition from scalar-based feedback to rich, structured linguistic critiques. By leveraging natural-language gradients, these systems now facilitate targeted, iterative updates that mirror the precision of gradient-based optimization in higher-dimensional strategy spaces.

In this section, prompt primarily refers to structural instruction layers that are reused across interactions, such as system prompts or stable policy templates. A closely related line of work optimizes context construction, including exemplar selection, retrieval assembly, and the maintenance of long-term playbooks. Although both are scaffold-level updates, they differ in target: optimizing a system prompt alters the agent’s core behavioral prior, whereas context optimization refines the dynamic conditioning of specific interactions. When discussing representative systems, we will explicitly indicate which object is updated.

\begin{figure*}[ht]
    \centering
    \includegraphics[width=1\linewidth]{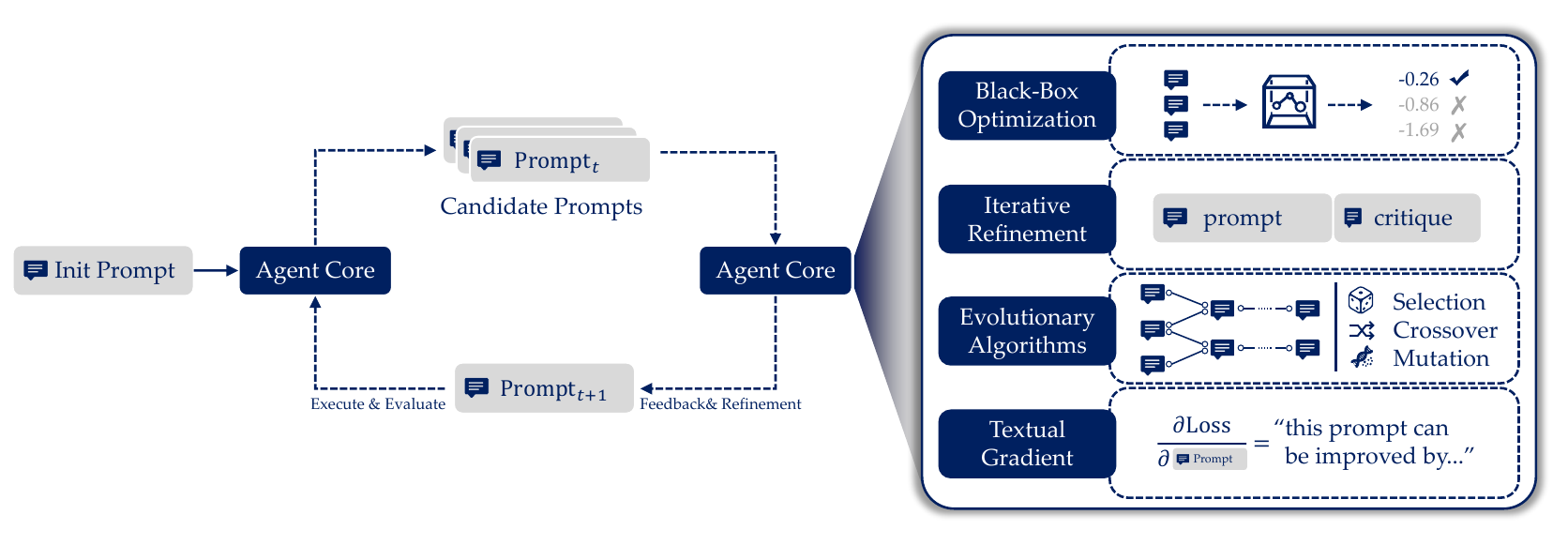}
    \caption{Prompt refinement as a self-improvement loop and its four paradigms, organized by the learning signal $\mathcal{S}_t$.}
    \label{fig:prompt}
\end{figure*}

As shown in Fig.~\ref{fig:prompt}, we categorize prompt refinement methods based on the form and richness of the learning signal $\mathcal{S}_t$. This yields four paradigms: (1) \textit{Scalar-Feedback Optimization}, where $\mathcal{S}_t$ is a scalar performance score such as accuracy or reward;
(2) \textit{Qualitative-Feedback Refinement}, where $\mathcal{S}_t$ is a natural-language critique or suggestion for providing interpretable revision guidance;
(3) \textit{Population-Based Evolution}, where $\mathcal{S}_t$ consists of population-level fitness evaluation and selection signals over a set of candidate prompts;
and (4) \textit{Textual Gradient Optimization}, where $\mathcal{S}_t$ is structured directional guidance, often expressed as a textual gradient that specifies how the prompt should be revised. As listed in Table~\ref{tab:prompt_optimization_paradigms}, we further summarize representative systems and trade-offs across paradigms.

\subsubsection{Scalar-Feedback Optimization}
\label{sec:Scalar_Feedback_Optimization}

Early approaches to prompt optimization usually rely on task-level quantitative metrics to evaluate candidates \citep{zhou2022large}. The core objective can be formally defined as finding an optimal prompt $p^*$ from a discrete space of possible prompts $\mathcal{P}$ that maximizes a scalar performance score (e.g., task accuracy or reward):
\begin{equation}p^* = \arg\max_{p \in \mathcal{P}} f(p),
\end{equation}
where $f(p)$ represents the evaluation function. In the context of our scaffolding improvement framework, this scalar score is used to constitutes the learning signal ($\mathcal{S}_t = f(p_t)$). Considering $\mathcal{S}_t$ is non-directional, providing only a magnitude of success without explanatory context, these methods typically rely on structured search algorithms to navigate the discrete text space. APE pioneered this paradigm by using an LLM to propose instruction candidates, subsequently selecting the best one based on empirical evaluation scores via simple search \citep{zhou2022large}. Building on this, OPRO formalized a more contextualized search by constructing a meta-prompt that includes a trajectory of previously evaluated prompts alongside their scalar scores, indirectly guiding the LLM to propose higher-scoring candidates in subsequent iterations \citep{yang2023large}.

Besides, researchers have adapted advanced derivative-free optimization algorithms to operate strictly on scalar feedback. For instance, RL-based frameworks like RLPrompt treat prompt generation as a discrete reinforcement learning problem, directly optimizing a reward signal derived from downstream task accuracy \citep{deng2022rlprompt}. Similarly, to improve the sample efficiency of exploring the discrete text space, methods like InstructZero map prompts into a continuous latent space, leveraging Bayesian Optimization to predict and maximize scalar task rewards efficiently \citep{cheninstructzero}. Recent systems like BPO (Black-Box Prompt Optimization) further extend this score-driven paradigm to human preference alignment, using scalar preference scores to optimize user inputs without altering the underlying model weights \citep{cheng2024black}.

\subsubsection{Qualitative-Feedback Refinement}
\label{sec:Qualitative_Feedback_Refinement}

Building upon scalar-driven methods, a more nuanced approach focuses on iterative refinement using qualitative, interpretable feedback. Instead of a single score, the learning signal ($\mathcal{S}_t$) manifests as a textual critique, error analysis, or natural-language suggestion, denoted as $c_t$. This richer signal allows for a targeted revision process, formally modeled as an iterative loop where each new prompt $p_{t+1}$ is a function of the previous prompt $p_t$ and its corresponding qualitative feedback $c_t$:
\begin{equation}p_{t+1} = \text{Refine}(p_t, c_t),
\end{equation}
where $c_t$ is typically generated by an evaluator or the LLM itself based on historical execution: $c_t = \text{Critique}(\text{Output}(p_t))$. While foundational paradigms like Self-Refine \citep{madaan2023self} and multi-agent debate \citep{liang2023debate} demonstrated the power of self-critique for transient output correction, recent scaffolding improvements persist this qualitative feedback to update the agent's prompting policy or instructional context. One prominent direction leverages error-driven qualitative analysis. Reflexion, for example, enables an agent to generate ``verbal introspection'' upon task failure, storing these qualitative reflections in memory to explicitly guide and constrain subsequent prompting attempts \citep{shinn2023reflexionlanguageagentsverbal}. Expanding on this diagnostic capability, MAPS introduces an automated, LLM-tailored prompt optimization framework that explicitly induces and validates reusable natural-language rules from failure cases. By iteratively injecting these qualitative insights into the prompt, MAPS substantially optimizes the policy for complex tasks like unit test generation \citep{gao2025promptalchemistautomatedllmtailored}. Similarly, inspired by Chain of Hindsight (CoH), agents can learn from textual contrasts by reviewing a history of past attempts accompanied by qualitative evaluations \citep{liu2023chain}. Another critical direction involves evolving the instructional context. To manage the growing complexity of text-based critiques, ACE introduces an agentic context engineering framework with a modular \textit{Generator–Reflector–Curator} pipeline. ACE treats prompts and memory as evolving text playbooks, actively curating qualitative feedback while mitigating brevity bias and context collapse \citep{zhang2025agenticcontextengineeringevolving}. In specific domain applications, systems like Scrable utilize continuous qualitative evaluation to iteratively refine the structural system prompt for customer review generation, halting only when the textual quality reaches a predefined standard \citep{azov2024selfimprovingcustomerreviewresponse}.

\newcommand{\posbar}{\textcolor{green!55!black}{\rule{1.2pt}{8.0pt}}\;}
\newcommand{\negbar}{\textcolor{red!70!black}{\rule{1.2pt}{8.0pt}}\;}
\newcommand{\prog}{\textcolor{black!45}{\(\rightarrow\)}}

\begin{table*}[t]
\centering
\scriptsize
\setlength{\tabcolsep}{3pt}
\renewcommand{\arraystretch}{1.30}

\caption{Comparison of prompt optimization paradigms in prompt-based self-improvement.
As learning signals become more structured and informative, optimization becomes less heuristic and more automated.
\textbf{\textcircled{1}}: Scalar-feedback optimization; \textbf{\textcircled{2}}: Qualitative-feedback refinement; \textbf{\textcircled{3}}: Population-based evolution; \textbf{\textcircled{4}}: Textual-gradient optimization.}
\vspace{0.4em}

\rowcolors{2}{lightBlue!18}{white}

\begin{tabularx}{\textwidth}{
>{\raggedright\arraybackslash}p{0.62cm} 
>{\raggedright\arraybackslash}p{1.10cm} 
>{\raggedright\arraybackslash}p{1.65cm}
>{\raggedright\arraybackslash}p{4.05cm} 
>{\raggedright\arraybackslash}X         
>{\raggedright\arraybackslash}X         
}
\toprule
\textbf{ID} &
\textbf{Signal $\mathcal{S}_t$} &
\textbf{Objective} &
\textbf{Representative Systems} &
\textbf{Advantages} &
\textbf{Limitations} \\
\midrule

\renewcommand{\arraystretch}{2.10}

\textbf{\fcirc{Blue}{1}}\;\prog &
\cellcolor{lightBlue!12}\makecell[l]{Scalar\\score} &
\cellcolor{lightBlue!8}$\arg\max_{p\in\mathcal{P}} f(p)$ &
\cellcolor{lightBlue!6}\makecell[l]{RLPrompt~\citep{deng2022rlprompt}\\
BBT~\citep{sun2022black}\\
APE~\citep{zhou2022large}\\
OPRO~\citep{yang2023large}\\
Dspy~\citep{khattab2023dspy}} &
\makecell[l]{\posbar \textcolor{green!55!black}{\bfseries +}\, Model-agnostic\\
\posbar \textcolor{green!55!black}{\bfseries +}\, Simple to deploy\\
\posbar \textcolor{green!55!black}{\bfseries +}\, No internal access} &
\makecell[l]{\negbar \textcolor{red!70!black}{\bfseries --}\, Low interpretability\\
\negbar \textcolor{red!70!black}{\bfseries --}\, Sample-inefficient\\
\negbar \textcolor{red!70!black}{\bfseries --}\, Sensitive to search} \\
\addlinespace[2pt]

\textbf{\fcirc{Blue}{2}}\;\prog &
\cellcolor{lightBlue!12}\makecell[l]{Text\\critique} &
\cellcolor{lightBlue!8}$\text{Refine}(p_t, c_t)$ &
\cellcolor{lightBlue!6}\makecell[l]{Self-Refine~\citep{madaan2023self}\\
Reflexion~\citep{shinn2023reflexionlanguageagentsverbal}\\
Critic~\citep{gou2024critic}\\
ACE~\citep{zhang2025agenticcontextengineeringevolving}} &
\makecell[l]{\posbar \textcolor{green!55!black}{\bfseries +}\, Interpretable edits\\
\posbar \textcolor{green!55!black}{\bfseries +}\, Targeted correction\\
\posbar \textcolor{green!55!black}{\bfseries +}\, Reusable feedback} &
\makecell[l]{\negbar \textcolor{red!70!black}{\bfseries --}\, Critique can be noisy\\
\negbar \textcolor{red!70!black}{\bfseries --}\, May drift\\
\negbar \textcolor{red!70!black}{\bfseries --}\, Validator-dependent} \\
\addlinespace[2pt]

\textbf{\fcirc{Blue}{3}}\;\prog &
\cellcolor{lightBlue!12}\makecell[l]{Selection\\signal} &
\cellcolor{lightBlue!8}$\text{Evolve}(P_t,\text{Fit})$ &
\cellcolor{lightBlue!6}\makecell[l]{Promptbreeder~\citep{fernando2024promptbreeder}\\
STOP~\citep{zelikman2024self}\\
GPTSwarm~\citep{10.5555/3692070.3694667}\\
AutoDAN~\citep{liu2024autodan}\\
Evol-Instruct ~\citep{xu2024wizardlm}\\
GEPA~\citep{agrawal2025gepareflectivepromptevolution}} &
\makecell[l]{\posbar \textcolor{green!55!black}{\bfseries +}\, Strong exploration\\
\posbar \textcolor{green!55!black}{\bfseries +}\, Maintains diversity\\
\posbar \textcolor{green!55!black}{\bfseries +}\, Escapes local optima} &
\makecell[l]{\negbar \textcolor{red!70!black}{\bfseries --}\, Compute-heavy\\
\negbar \textcolor{red!70!black}{\bfseries --}\, Fitness is domain-tuned\\
\negbar \textcolor{red!70!black}{\bfseries --}\, Population drift} \\
\addlinespace[2pt]

\textbf{\fcirc{Blue}{4}}\;\prog &
\cellcolor{lightBlue!12}\makecell[l]{Textual\\gradient} &
\cellcolor{lightBlue!8}$p_t \oplus g(p_t)$ &
\cellcolor{lightBlue!6}\makecell[l]{APO~\citep{pryzant2023automaticpromptoptimizationgradient}\\
TextGrad~\citep{yuksekgonul2024textgradautomaticdifferentiationtext}\\
metaTextGrad~\citep{xu2025metatextgradautomaticallyoptimizinglanguage}\\
SkillOpt~\citep{yang2026skilloptexecutivestrategyselfevolving}} &
\makecell[l]{\posbar \textcolor{green!55!black}{\bfseries +}\, Directional updates\\
\posbar \textcolor{green!55!black}{\bfseries +}\, Often sample-efficient\\
\posbar \textcolor{green!55!black}{\bfseries +}\, Highly automated} &
\makecell[l]{\negbar \textcolor{red!70!black}{\bfseries --}\, Brittle gradients\\
\negbar \textcolor{red!70!black}{\bfseries --}\, Quality varies by LLM\\
\negbar \textcolor{red!70!black}{\bfseries --}\, Limited guarantees} \\

\bottomrule
\end{tabularx}

\renewcommand{\arraystretch}{1.15}

\label{tab:prompt_optimization_paradigms}
\end{table*}

\subsubsection{Population-Based Evolution}
\label{sec:Population_Based_Evolution}

To introduce more structured exploration and mitigate the risk of converging to local optima, researchers have adapted principles from evolutionary biology. In this paradigm, the learning signal $\mathcal{S}_t$ manifests as population-level fitness evaluations and selection pressures over a diverse pool of candidate prompts. Formally, these methods treat prompts as ``genes'' within a population $P_t = \{p_t^{(i)}\}_{i=1}^{N}$ at generation $t$. The evolutionary update leverages LLMs to apply semantic operators rather than simple string manipulations, following three key steps:
\begin{enumerate}
\item \textbf{Selection:} A subset of prompts survives based on a fitness evaluation function, which translates task performance into selection pressure ($\mathcal{S}_t$).
\item \textbf{Crossover:} The LLM intelligently merges the semantic strengths of two parent prompts, generating offspring: $p_{\text{child}} = \text{Crossover}(p_t^{(i)}, p_t^{(j)})$.
\item \textbf{Mutation:} The LLM introduces semantic variations to explore new instructional phrasing: $p' = \text{Mutate}(p)$.
\end{enumerate}
The subsequent generation, $P_{t+1}$, is formed from the fittest individuals and their offspring. Foundational frameworks like EvoPrompt \citep{guo2025evopromptconnectingllmsevolutionary} pioneered this by explicitly guiding LLMs to act as evolutionary operators, yielding crossover and mutation steps that are semantically meaningful and far surpass classical random character edits. Expanding the depth of this standard evolutionary search, frameworks like Promptbreeder \citep{fernando2024promptbreeder} introduces a profound self-referential mechanism where the LLM evolves not only the task prompts but also the "mutation prompts" (the instructions governing how new offspring are generated). Addressing scenarios where scalar fitness scores are unavailable, DEEVO creatively structures $\mathcal{S}_t$ through multi-agent debates, utilizing the win-rate of LLM-driven argumentation as the evolutionary fitness signal \citep{nair2025tournamentpromptsevolvingllm}. Crucially, demonstrating the compositional nature of scaffolding improvement, recent work integrates population-based search with qualitative feedback. In reflective prompt evolution frameworks like GEPA \citep{agrawal2025gepareflectivepromptevolution}, the evolutionary process is guided not merely by a scalar fitness score, but by a meta-level reflection step. After evaluating a generation, a reflector LLM analyzes successes and failures to generate textual critiques. This qualitative feedback explicitly informs the next generation's mutation and crossover operators, making the search highly targeted and sample-efficient. The success of such hybrid approaches highlights the distinct advantage of replacing opaque scalar rewards with directional, semantic guidance—directly paving the way for the formalized textual gradient methods discussed next.

\subsubsection{Textual Gradient Optimization}
\label{sec:Textual_Gradient_Optimization}
The most recent paradigm is distinguished by a formal, mathematically-inspired treatment of the feedback signal, drawing direct analogies to gradient descent in continuous optimization. Rather than relying on heuristic edits or scalar search, the learning signal ($\mathcal{S}_t$) is formalized as a \textit{textual gradient}—a structured, directional feedback message that explicitly diagnoses why an output is incorrect and prescribes a precise revision vector. This optimization process can be expressed with an analogous update rule:
\begin{equation}
p_{t+1} = p_t \oplus g(p_t),
\end{equation}
where the textual gradient $g(p_t)$ serves directly as our learning signal ($\mathcal{S}_t = g(p_t)$). The $\oplus$ operator denotes a textual update step, where an LLM acts as the optimizer to semantically apply the gradient's guidance and revise the prompt. While the foundational concept of a ``textual gradient'' was introduced in Automatic Prompt Optimization (APO) \citep{pryzant2023automaticpromptoptimizationgradient}, this direction has been significantly advanced by framing agentic workflows as computational graphs. TextGrad, for instance, operationalized this by introducing automatic differentiation via text, allowing qualitative gradients to backpropagate through complex, multi-component language systems \citep{yuksekgonul2024textgradautomaticdifferentiationtext}. Concurrently, semantic backpropagation frameworks have further formalized this process, executing first-order-like optimization on text nodes to achieve principled prompt refinement \citep{wang2024correctlysemanticbackpropagationlanguagebased}. The sophistication of this paradigm is profoundly underscored by meta-level frameworks like MetaTextGrad. Rather than solely optimizing task prompts, it uses an LLM to dynamically optimize the ``optimizer prompts'' (the instructions dictating how the textual gradients are computed and applied), effectively allowing the agent to self-improve its own improvement process \citep{xu2025metatextgradautomaticallyoptimizinglanguage}. Looking forward, this formal feedback mechanism could bridge the gap between scaffolding and parameter-based learning: a textual gradient might  be translated into low-rank parameter updates, seamlessly integrating prompt engineering with model fine-tuning.

\begin{ttcolorbox}[Takeaway]
Prompt-based self-improvement turns prompt optimization from an ad hoc practice into a signal-driven improvement loop. In existing approaches, the learning signal ($\mathcal{S}_t$) becomes progressively richer, evolving from scalar scores to qualitative critiques, and further to population-level selection and structured directional guidance. As feedback grows in informational content, prompt updates become less heuristic and more targeted, enabling increasingly automated and sample-efficient refinement without changing the foundation-model parameters.
\end{ttcolorbox}

\subsection{Memory}
\label{sec:Memory}

The memory system serves as the core cognitive scaffolding for long-horizon agentic behavior. Traditional memory architectures often rely on raw content logging alongside fixed schemas, resulting in rigid, static organizations. Such append-only designs quickly succumb to context-window constraints and retrieval degradation, rendering them ill-suited for dynamically changing environments \citep{modarressi2024retllmgeneralreadwritememory, he2024malmmmemoryaugmentedlargemultimodal, packer2024memgptllmsoperatingsystems}. In contrast, self-improving agents treat memory not merely as a passive storage mechanism, but as an actively evolving scaffold. By continuously assessing the value, relevance, and strength of stored information, these agents autonomously reconstruct and expand their memory representations based on the flow of experience \citep{du2025rethinkingmemoryaitaxonomy, xu2025sedm}. This shift from static storage to dynamic self-organization  enhances the agent's generality, establishing a foundation for open-ended autonomy \citep{xu2025amemagenticmemoryllm, sang2025pipelinessurveyparadigmshift, li2025memosmemoryosai}.

To systematically analyze this paradigm, we decompose memory-based improvement into three core dimensions: \textit{\textbf{Memory Objects}} (the units of stored information), \textit{\textbf{Memory Structure}} (the topological organization and indexing schema), and \textit{\textbf{Memory Processing}} (the mechanisms for creation, retrieval, updating, and forgetting). 
Formally, we conceptualize the memory scaffold as a dynamic state $m_t := (\text{object}_t, \text{structure}_t)$, specifying the currently stored objects and their overarching organization. Driven by an execution-derived learning signal $\mathcal{S}_t$ (e.g., retrieval failures, task feedback, or capacity limits), memory evolution is formalized as:
\begin{equation}
m_{t+1} = \text{IMPROVE}m(m_t; \mathcal{S}t).
\end{equation}
This update is instantiated through the memory processing module—a signal-driven family of operations (Write, Read, Update, Delete) parameterized by $\mathcal{S}_t$ that governs what to consolidate into $\text{object}_{t+1}$ and how to reorganize $\text{structure}_{t+1}$. Finally, it is crucial to note the scope of our discussion. While some literature considers knowledge internalized within the foundation model's weights as "parametric memory" \citep{wu2025humanmemoryaimemory, he2025humaninspiredperspectivessurveyai, zhang2025survey}, this section focuses on {non-parametric, externalized memory} embedded within the agent's scaffold, maintaining the core assumption of a frozen foundation model.

\begin{figure*}[ht]
    \centering
    \includegraphics[width=1\linewidth]{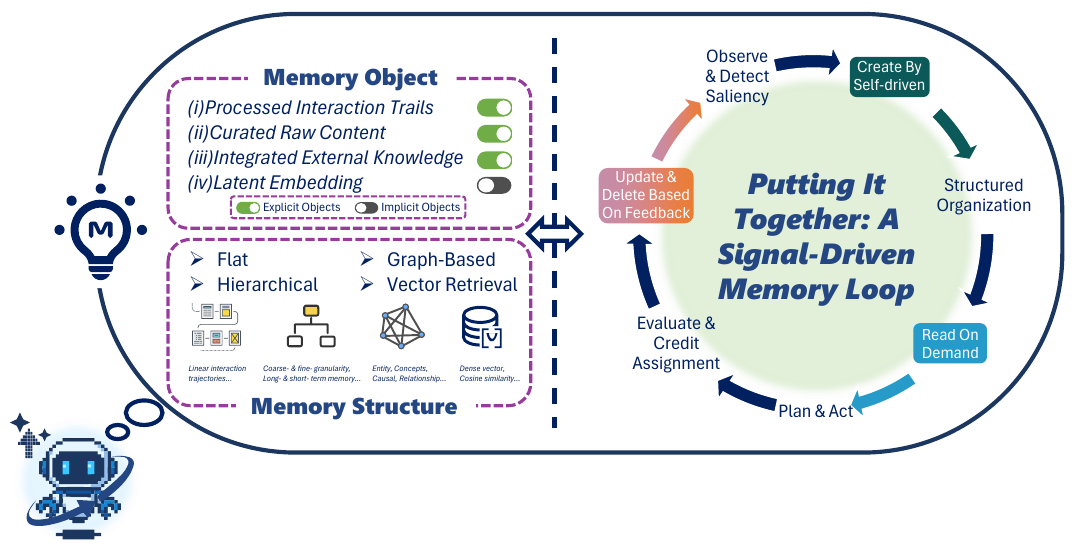}
    \caption{Overview of memory for self-improving agent.}
    \label{fig:agent_memory}
\end{figure*}

\subsubsection{Memory Object}
\label{sec:Memory_Object}

\makeatletter
\@ifundefined{rate}{
  \newcommand{\rate}[2]{\rate@impl{#1}{#2}}
}{
  \renewcommand{\rate}[2]{\rate@impl{#1}{#2}}
}
\newcommand{\rate@impl}[2]{
  \begingroup
  \def\filled{#1}\def\total{#2}
  \newcount\ratei \ratei=1
  \loop\ifnum\ratei<\numexpr\total+1\relax
    \ifnum\ratei>\filled
      \textcolor{black!20}{\rule{0.85ex}{0.85ex}}
    \else
      \textcolor{blue!70!black}{\rule{0.85ex}{0.85ex}}
    \fi
    \hspace{0.22ex}
    \advance\ratei by 1
  \repeat
  \endgroup
}
\makeatother

\newcommand{\bbull}{\textcolor{blue!70!black}{\raisebox{0.25ex}{\tiny$\bullet$}}\;}
\newcommand{\fbull}{\textcolor{black!55}{\raisebox{0.25ex}{\tiny$\blacktriangleright$}}\;}

\newcommand{\mini}[1]{{\scriptsize\color{black!70}#1}}

\begin{table*}[t]
\centering
\scriptsize
\setlength{\tabcolsep}{2pt} 
\renewcommand{\arraystretch}{1.25}

\caption{Memory-object scorecard (qualitative). Blue squares indicate an ordinal, literature-grounded assessment (1=low, 5=high) of typical tendencies for each memory object type, synthesized from representative system designs and reported failure analyses rather than from a single standardized benchmark.}
\vspace{0.35em}

\rowcolors{2}{lightBlue!18}{white}

\newcolumntype{L}[1]{>{\raggedright\arraybackslash}p{#1}}
\newcolumntype{R}{>{\centering\arraybackslash}m{1.05cm}} 

\begin{tabularx}{\textwidth}{
L{2.05cm}  
L{2.45cm}  
R R R R R  
X         
}
\toprule
\textbf{Object type} &
\makecell[l]{\textbf{Best-for}\\[-1pt]\textbf{persistence}} &
\textbf{Fidelity} &
\makecell[c]{\textbf{Interpre-}\\[-1pt]\textbf{tability}} &
\textbf{Compact} &
\makecell[c]{\textbf{Write}\\[-1pt]\textbf{cost}} &
\makecell[c]{\textbf{Audit-}\\[-1pt]\textbf{tability}} &
\makecell[l]{\textbf{Most common}\\[-1pt]\textbf{failure modes}} \\
\midrule

\makecell[l]{Processed\\trails} &
\makecell[l]{\bbull lessons\\[-1pt]\bbull routines\\[-1pt]\bbull summaries} &
\rate{3}{5} & \rate{5}{5} & \rate{4}{5} & \rate{3}{5} & \rate{5}{5} &
\makecell[l]{\fbull summary bias\\[-1pt]\fbull stale heuristics\\[-1pt]\fbull weak credit assignment} \\

\makecell[l]{Curated raw\\content} &
\makecell[l]{\bbull evidence\\[-1pt]\bbull exact artifacts} &
\rate{5}{5} & \rate{5}{5} & \rate{1}{5} & \rate{4}{5} & \rate{5}{5} &
\makecell[l]{\fbull context bloat\\[-1pt]\fbull retrieval noise\\[-1pt]\fbull privacy leakage surface} \\

\makecell[l]{Integrated external\\knowledge} &
\makecell[l]{\bbull shared factual state\\[-1pt]\bbull grounding} &
\rate{4}{5} & \rate{4}{5} & \rate{3}{5} & \rate{4}{5} & \rate{4}{5} &
\makecell[l]{\fbull grounding failure\\[-1pt]\fbull staleness / inconsistency\\[-1pt]\fbull tool brittleness} \\

\makecell[l]{Latent\\embeddings} &
\makecell[l]{\bbull associative carryover\\[-1pt]\bbull fast recall} &
\rate{3}{5} & \rate{1}{5} & \rate{5}{5} & \rate{2}{5} & \rate{1}{5} &
\makecell[l]{\fbull drift / contamination\\[-1pt]\fbull hard-to-debug retrieval\\[-1pt]\fbull silent corruption} \\

\bottomrule
\end{tabularx}

\label{tab:mem_object_scorecard}
\end{table*}

Recalling our formal decomposition of memory state, this subsection focuses on $\text{object}_t$, i.e., \emph{what} is stored within the agent's memory scaffold. The design of the memory object is paramount, as it directly governs storage efficiency, representation fidelity, and the capacity for cross-context knowledge transfer \citep{koley2025salmmultiagentframeworklanguage}. Moving beyond raw, exhaustive interaction trails (e.g., infinite chat histories), self-improving agents increasingly favor storing processed, high-density abstractions \citep{rasmussen2025zeptemporalknowledgegraph, deng2024humancenteredproactiveconversationalagents}. By utilizing execution feedback ($\mathcal{S}_t$) to filter or compress experiences, agents effectively mitigate context-window limitations and storage costs \citep{cao2025infiniteiclbreakinglimitcontext}. These memory objects can be fundamentally categorized into {explicit} and {implicit} representations.

\begin{itemize}[leftmargin=*]
    \item \textbf{Explicit Objects.}
    Explicit objects are human-readable and directly manipulable, which makes them the default choice when interpretability, attribution, and safety auditing are important. Their main advantage is controllability: developers can inspect, correct, and selectively expose them to the model. Their main limitation is scalability, since verbose or weakly curated memories tend to increase retrieval noise and context pressure as the memory grows.

    A first and widely used manifestation stores \textit{processed interaction trails} that compresses raw trajectories into reusable, semantically meaningful units (e.g., distilled routines, heuristics, or reflections). Driven by task success/failure signals, agents abstract generalizable strategies from experience or maintain note-like intermediate traces to support long-horizon reasoning \citep{wang2024agentworkflowmemory, ouyang2025reasoningbankscalingagentselfevolving, lanchantin2023learningreasonmemorizeselfnotes, lee2024humaninspiredreadingagentgist, long2025seeinglisteningrememberingreasoning}.

    A second manifestation retains \textit{curated raw content}. Certain scenarios require preserving exact surface details that are hard to summarize without loss (e.g., code snippets, formulas, or screenshots). Rather than storing everything, self-improving agents selectively write back only high-value artifacts validated during trial-and-error, distilling dynamic ``cheatsheets'' for future reuse \citep{zhao2024expel, suzgun2025dynamiccheatsheettesttimelearning}.

    A third manifestation integrates \textit{external knowledge}. Agents can integrate and  maintain facts from external repositories. Grounding memory in shared references rather than free-form recollection  enhances verifiability. Crucially, unlike static retrieval systems, self-improving agents utilize utility-based feedback to dynamically update, annotate, or prune these retrieved domain references (e.g., codebases or task-specific documents), thereby mitigating error propagation in downstream reasoning \citep{tran2025primeplanningretrievalintegratedmemory, zhang-etal-2024-codeagent, peng2023checkfactstryagain}.

    \item \textbf{Implicit Objects.} Implicit objects store memory in machine-native latent representations, including latent tokens, hidden states, and key--value cache augmentations. Their primary advantage is compactness and fast associative access, which makes them appealing under strict context limits or latency budgets. Their primary drawback is limited interpretability, which complicates debugging, targeted correction, and safety auditing, and can lead to representation drift when the memory is repeatedly rewritten or composed.

Recent advancements explore several mechanisms for implicit memory scaffolding without altering the base FM parameters. \textit{Generative latent memory} constructs latent sequences to enrich reasoning beyond text-based retrieval \citep{zhang2025memgenweavinggenerativelatent}. \textit{Latent state reconstruction} captures and reintegrates hidden representations to improve context retention \citep{dillon2025contextualmemoryreweavinglarge}. A related line augments the decoding process via offline coprocessors that inject latent embeddings directly into the KV cache to boost generation fidelity \citep{liu2024deliberationlatentspacedifferentiable, sun2025enhancinglatentcomputationtransformers}. Finally, maintaining self-updatable latent memory pools offers a practical compromise between persistent state tracking and strict capacity control \citep{wang2024memoryllmselfupdatablelargelanguage, wang2025mextendingmemoryllmscalable}.

\end{itemize}

\paragraph{Trade-offs.}
Explicit memory objects offer high interpretability and controllability, simplifying debugging and safety auditing, but require rigorous curation to avoid overwhelming the context budget. Among them, processed trails favor generalization, curated artifacts favor precision, and integrated external knowledge favors verifiability. Conversely, implicit memory objects provide compact, high-speed associative access and mitigate context length issues, but are difficult to inspect, correct, and are susceptible to representation drift over long horizons. We summarize these qualitative trade-offs and common failure modes in Table~\ref{tab:mem_object_scorecard}.

\newcommand{\Dfull}{\textcolor{blue!80!black}{\raisebox{0.15ex}{\large$\bullet$}}} 
\newcommand{\Dmid}{\textcolor{cyan!80}{\raisebox{0.15ex}{\large$\bullet$}}} 
\newcommand{\Dnone}{\textcolor{black!10}{\raisebox{0.15ex}{\large$\bullet$}}}

\newcommand{\Hstrut}{\rule{0pt}{3.2ex}}

\begin{table}[t]
\centering
\caption{Memory architecture and processing operators.
Checkmarks indicate the memory \emph{object} and \emph{structure} choices reported by each system.
Dots denote the relative emphasis of a mechanism as primary (\Dfull), secondary (\Dmid), or absent/unclear (\Dnone).
Processing is summarized by CRUD (Create, Read, Update, Delete).
We further characterize governance along two dimensions: \textbf{Select}, which determines what information is written and retrieved based on saliency and utility, and \textbf{Maintain}, which sustains memory quality over long horizons through consolidation, refresh, and forgetting.}
\setlength{\tabcolsep}{3pt}
\renewcommand{\arraystretch}{1.35}

\rowcolors{7}{gray!8}{white}
\resizebox{\linewidth}{!}{%
\begin{tabular}{%
  >{\raggedright\arraybackslash}m{3.0cm}
  *{12}{>{\centering\arraybackslash}m{0.92cm}}
}
\toprule
\rowcolor{tab-blue}
\multicolumn{1}{>{\centering\arraybackslash}m{3.0cm}}{%
  \multirow{2}{*}{\cellcolor{white}\centering
    \raisebox{-0.15\height}{\includegraphics[width=1.15cm,keepaspectratio]{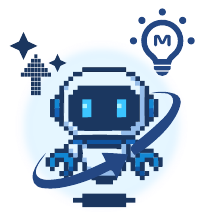}}%
  }%
} &
\multicolumn{2}{c}{\color{white}\bfseries Object\Hstrut} &
\multicolumn{4}{c}{\color{white}\bfseries Structure\Hstrut} &
\multicolumn{4}{c}{\color{white}\bfseries Processing\Hstrut} &
\multicolumn{2}{c}{\color{white}\bfseries Governance\Hstrut} \\
\cmidrule(lr){2-3}\cmidrule(lr){4-7}\cmidrule(lr){8-11}\cmidrule(lr){12-13}

& \cellcolor{blue!12}\thead{Explicit\\Objects\Hstrut} &
  \cellcolor{blue!12}\thead{Implicit\\Objects\Hstrut} &
  \cellcolor{blue!12}\thead{Flat\Hstrut} &
  \cellcolor{blue!12}\thead{Hier.\Hstrut} &
  \cellcolor{blue!12}\thead{Graph\Hstrut} &
  \cellcolor{blue!12}\thead{Vector\\Retr.\Hstrut} &
  \cellcolor{blue!12}\thead{Create\\\tagbox{C}\kern-0.35em\Hstrut} &
  \cellcolor{blue!12}\thead{Read\\\tagbox[cyan!60!teal]{R}\Hstrut} &
  \cellcolor{blue!12}\thead{Update\\\tagbox[roseviolet]{U}\Hstrut} &
  \cellcolor{blue!12}\thead{Delete\\\tagbox[VOrange]{D}\Hstrut} &
  \cellcolor{blue!12}\thead{Select\Hstrut} &
  \cellcolor{blue!12}\thead{Maint.\Hstrut} \\
\midrule

\textbf{\hyperlink{cite.lanchantin2023learningreasonmemorizeselfnotes}{\makecell[l]{Self-Notes\\(2023)}}}
& $\checkmark$ & -- & $\checkmark$ & -- & -- & --  & \Dmid  & \Dmid  & \Dnone & \Dnone & \Dmid  & \Dnone \\

\textbf{\hyperlink{cite.park2023generativeagentsinteractivesimulacra}{\makecell[l]{Generative\\Agents (2023)}}}
& $\checkmark$ & -- & $\checkmark$ & -- & -- & $\checkmark$ & \Dmid  & \Dfull & \Dnone & \Dmid  & \Dfull & \Dmid \\

\textbf{\hyperlink{cite.guan2024richelieu}{\makecell[l]{Richelieu\\(2024)}}}
& $\checkmark$ & -- & $\checkmark$ & -- & -- & --  & \Dmid  & \Dmid  & \Dnone & \Dnone & \Dmid  & \Dnone \\

\textbf{\hyperlink{cite.suzgun2025dynamiccheatsheettesttimelearning}{\makecell[l]{DC\\(2025)}}}
& $\checkmark$ & -- & $\checkmark$ & -- & -- & --  & \Dfull & \Dmid  & \Dfull & \Dfull & \Dfull & \Dfull \\

\textbf{\hyperlink{cite.liang2025selfevolvingagentsreflectivememoryaugmented}{\makecell[l]{SAGE\\(2025)}}}
& $\checkmark$ & -- & -- & $\checkmark$ & -- & -- & \Dfull & \Dmid  & \Dfull & \Dfull & \Dmid  & \Dfull \\

\textbf{\hyperlink{cite.chhikara2025mem0buildingproductionreadyai}{\makecell[l]{Mem0\\(2025)}}}
& $\checkmark$ & -- & -- & -- & $\checkmark$ & $\checkmark$ & \Dfull & \Dfull & \Dfull & \Dfull & \Dfull & \Dfull \\

\textbf{\hyperlink{cite.salama2025meminsightautonomousmemoryaugmentation}{\makecell[l]{MemInsight\\(2025)}}}
& $\checkmark$ & -- & -- & $\checkmark$ & -- & $\checkmark$ & \Dfull & \Dfull & \Dmid  & \Dnone & \Dfull & \Dmid \\

\textbf{\hyperlink{cite.zhang2025memgenweavinggenerativelatent}{\makecell[l]{MemGen\\(2025)}}}
& -- & $\checkmark$ & $\checkmark$ & -- & -- & -- & \Dfull & \Dfull & \Dmid  & \Dnone & \Dfull & \Dnone \\

\textbf{\hyperlink{cite.zhang2025agenticcontextengineeringevolving}{\makecell[l]{ACE\\(2025)}}}
& $\checkmark$ & -- & -- & -- & -- & $\checkmark$ & \Dfull & \Dfull & \Dfull & \Dmid  & \Dmid  & \Dfull \\

\textbf{\hyperlink{cite.xu2025amemagenticmemoryllm}{\makecell[l]{A-MEM\\(2025)}}}
& $\checkmark$ & -- & -- & -- & $\checkmark$ & $\checkmark$ & \Dfull & \Dfull & \Dfull & \Dnone & \Dfull & \Dfull \\

\textbf{\hyperlink{cite.wang2024agentworkflowmemory}{\makecell[l]{AWM\\(2024)}}}
& $\checkmark$ & -- & -- & $\checkmark$ & -- & -- & \Dfull & \Dmid  & \Dmid  & \Dnone & \Dmid  & \Dnone \\

\textbf{\hyperlink{cite.ouyang2025reasoningbankscalingagentselfevolving}{\makecell[l]{Reasoning\\Bank (2025)}}}
& $\checkmark$ & -- & -- & -- & -- & $\checkmark$ & \Dfull & \Dmid  & \Dmid  & \Dnone & \Dmid  & \Dnone \\

\textbf{\hyperlink{cite.lee2024humaninspiredreadingagentgist}{\makecell[l]{ReadAgent\\(2024)}}}
& $\checkmark$ & -- & $\checkmark$ & -- & -- & -- & \Dmid  & \Dfull & \Dnone & \Dnone & \Dmid  & \Dnone \\

\textbf{\hyperlink{cite.long2025seeinglisteningrememberingreasoning}{\makecell[l]{M3-Agent\\(2025)}}}
& $\checkmark$ & -- & -- & $\checkmark$ & -- & -- & \Dmid  & \Dmid  & \Dmid  & \Dmid  & \Dmid  & \Dmid \\

\textbf{\hyperlink{cite.zhao2024expel}{\makecell[l]{ExpeL\\(2024)}}}
& $\checkmark$ & -- & -- & -- & -- & $\checkmark$ & \Dfull & \Dmid  & \Dmid  & \Dnone & \Dfull & \Dmid \\

\textbf{\hyperlink{cite.tran2025primeplanningretrievalintegratedmemory}{\makecell[l]{PRIME\\(2025)}}}
& $\checkmark$ & -- & -- & -- & -- & $\checkmark$ & \Dnone & \Dfull & \Dnone & \Dnone & \Dfull & \Dnone \\

\textbf{\hyperlink{cite.zhang-etal-2024-codeagent}{\makecell[l]{CodeAgent\\(2024)}}}
& $\checkmark$ & -- & -- & -- & -- & $\checkmark$ & \Dnone & \Dfull & \Dnone & \Dnone & \Dfull & \Dnone \\

\textbf{\hyperlink{cite.dillon2025contextualmemoryreweavinglarge}{\makecell[l]{CMR\\(2025)}}}
& -- & $\checkmark$ & -- & $\checkmark$ & -- & -- & \Dfull & \Dfull & \Dfull & \Dnone & \Dmid  & \Dfull \\

\textbf{\hyperlink{cite.wang2024memoryllmselfupdatablelargelanguage}{\makecell[l]{MemoryLLM\\(2024)}}}
& -- & $\checkmark$ & $\checkmark$ & -- & -- & -- & \Dfull & \Dfull & \Dfull & \Dnone & \Dmid  & \Dfull \\

\textbf{\hyperlink{cite.wang2025mextendingmemoryllmscalable}{\makecell[l]{M+\\(2025)}}}
& -- & $\checkmark$ & $\checkmark$ & -- & -- & -- & \Dfull & \Dfull & \Dfull & \Dnone & \Dmid  & \Dfull \\

\textbf{\hyperlink{cite.sun2025hierarchicalmemoryhighefficiencylongterm}{\makecell[l]{H-MEM\\(2025)}}}
& $\checkmark$ & -- & -- & $\checkmark$ & -- & -- & \Dmid  & \Dfull & \Dfull & \Dmid  & \Dmid  & \Dmid \\

\textbf{\hyperlink{cite.koley2025salmmultiagentframeworklanguage}{\makecell[l]{SALM\\(2025)}}}
& $\checkmark$ & -- & -- & $\checkmark$ & -- & -- & \Dmid  & \Dfull & \Dmid  & \Dmid  & \Dmid  & \Dmid \\

\textbf{\hyperlink{cite.cheng2022xmemlongtermvideoobject}{\makecell[l]{XMem\\(2022)}}}
& -- & $\checkmark$ & -- & $\checkmark$ & -- & -- & \Dnone & \Dfull & \Dmid  & \Dnone & \Dnone & \Dnone \\

\textbf{\hyperlink{cite.song2024moviechatdensetokensparse}{\makecell[l]{MovieChat\\(2024)}}}
& -- & $\checkmark$ & -- & $\checkmark$ & -- & $\checkmark$ & \Dmid  & \Dfull & \Dmid  & \Dnone & \Dnone & \Dnone \\

\textbf{\hyperlink{cite.helmi2025decentralizingaimemoryshimi}{\makecell[l]{SHIMI\\(2025)}}}
& $\checkmark$ & -- & -- & $\checkmark$ & $\checkmark$ & -- & \Dmid  & \Dfull & \Dmid  & \Dmid  & \Dfull & \Dmid \\

\bottomrule
\end{tabular}%
}
\label{tab:memory_architecture}
\end{table}

\subsubsection{Memory Structure}
\label{sec:Memory_Structure}

Building upon the definition of memory objects ($\text{object}_t$), the memory structure ($\text{structure}_t$) specifies the organizational schema and relational topology through which these objects are indexed, linked, and retrieved. For self-improving agents operating in complex, dynamic environments, an adaptive memory structure acts as more than a passive database; it actively empowers the agent to consolidate long-horizon experiences, resolve context fragmentation, and perform robust relational reasoning \citep{zeng2024structuralmemoryllmagents}. Consequently, how this structure is designed and updated directly dictates the agent's ability to scale its cognitive baseline efficiently. Prevailing structural paradigms can be broadly categorized into the following  forms, each offering distinct trade-offs in scalability, expressiveness, and retrieval latency:

\begin{itemize}[leftmargin=*]
    \item \textbf{Flat Structure.}     Flat memories store entries in strict temporal order. This append-only design makes writing computationally cheap and  preserves the causal narrative of interaction. This topology is particularly advantageous for self-improving loops that rely on trajectory replay, as it maintains the unadulterated context necessary to diagnose failures ($\mathcal{S}_t$) and reproduce successful behaviors. The primary cost is retrieval scalability. As the stream expands, recall becomes overly dependent on truncation heuristics or coarse filtering. Consequently, agents often suffer from recency bias—surfacing recent but irrelevant interactions while missing older, decisive evidence. Furthermore, lacking inherent support for abstraction, flat structures tend to accumulate redundant low-level traces, exacerbating context pressure over long horizons \citep{antony2024causal, sun2025hierarchicalmemoryhighefficiencylongterm}. Representative systems operationalize this design by augmenting chronological logs with lightweight indexing. For instance, SCM maintains a temporal memory stream enriched with summaries and embeddings, bridging basic semantic search with chronological access \citep{wang2025scmenhancinglargelanguage}. Similarly, frameworks like Self-Notes write transient insights directly inline during long-context reasoning. This keeps the memory  aligned with the evolving cognitive state, allowing for immediate, feedback-driven course corrections while preserving the chronological flow of thought \citep{lanchantin2023learningreasonmemorizeselfnotes}.

\item \textbf{Hierarchical Structure.}
    Hierarchical memories organize objects across multiple abstraction levels, addressing a central bottleneck in self-improving agents: the need to compress long histories into stable, reusable knowledge while preserving interaction-level details. By allocating high-level summaries, mid-level plans, and low-level traces into distinct layers, this topology reduces retrieval noise and supports long-horizon coherence. However, the primary risk is structural brittleness. If the induced taxonomy misaligns with the task, strict hierarchies can fragment evidence and hinder cross-cutting retrieval, which severely impedes the agent's ability to recombine diverse experiences for continual improvement. Recent systems instantiate this hierarchy through different organizational lenses. For \textit{task-oriented} abstraction, MobileGPT employs a goal-to-subtask-to-action hierarchy for GUI tasks, supporting both plan reuse and fine-grained execution recall~\citep{lee2024exploreselectderiverecall}. For \textit{semantic} abstraction, H-MEM~\citep{sun2025hierarchicalmemoryhighefficiencylongterm} and SHIMI~\citep{helmi2025decentralizingaimemoryshimi} construct multi-level semantic nodes, enabling coarse-to-fine retrieval and top-down traversal from abstract intents to specific entities. Alternatively, SALM approaches hierarchy via \textit{tiered storage patterns} (short-term, long-term) to separate active context maintenance from durable reuse~\citep{koley2025salmmultiagentframeworklanguage}. Besides, similar principles extend to multimodal domains, where agents separate transient perceptual buffers from persistent representations for long-video understanding, drawing inspiration from classical human memory models~\citep{cheng2022xmemlongtermvideoobject, song2024moviechatdensetokensparse, atkinson1968human}.

    \item \textbf{Graph-Based Structure.} Graph memories represent objects as nodes linked by semantic, temporal, or causal edges, aligning naturally with the need for self-improving agents to generalize across continuous interactions. By explicitly encoding relations, this topology replaces recency-based recall with retrieval by association. It enables multi-hop evidence aggregation that flat streams or strict trees struggle to support, which is vital when agents must attribute failures to latent dependencies, track evolving entity states, or reuse abstracted insights. However, the primary trade-off is maintenance complexity. Continuous graph construction, edge updating, and conflict resolution introduce significant computational overhead and risk structural drift as the agent repeatedly edits its own memory.

       Recent literature explores this paradigm through different relational lenses. For {semantic and conversational} tracking, systems like Mem0~\citep{chhikara2025mem0buildingproductionreadyai} and SGMem~\citep{wu2025sgmemsentencegraphmemory} parse dialogues into fact- or sentence-level graphs to support highly structured recall. To support {temporal  and causal} reasoning, Zep provides a temporal-aware knowledge graph for continuous belief revision~\citep{rasmussen2025zeptemporalknowledgegraph}, while CausalRAG leverages explicit causal edges to prevent spurious correlations from contaminating the self-improvement loop~\citep{wang2025causalragintegratingcausalgraphs}. Some approaches even blend topologies; for instance, G-Memory introduces a {hybrid hierarchical graph} to separate reusable insights from fine-grained logs in multi-agent settings~\citep{zhang2025gmemorytracinghierarchicalmemory}. Beyond text, explicit graph structures are similarly crucial for multimodal agents—such as GraphVideoAgent and Scene-MMKG—where maintaining spatial-temporal state transitions and entity interactions is essential for grounded reasoning~\citep{10.1145/3746027.3755537, 10531671}.

    \item \textbf{Vector Retrieval Structure.}
    Vector-based memory indexes objects by dense embeddings and retrieves by similarity, which makes it a strong default when an agent must recall semantically related interactions under natural language queries. For self-improving agents, it serves as the primary engine for episodic recall and large-scale knowledge assimilation, supporting continual expansion without a hand-crafted schema.      
    Its dominant failure mode is misalignment between similarity and usefulness. The nearest neighbors under an embedding metric may be topically similar yet decision-irrelevant, and this retrieval noise can systematically bias the agent's subsequent learning updates.  As a result, self-improving systems often augment vector retrieval with re-ranking, hybrid signals, or adaptive controllers that dynamically gate the retrieval process.

    Building upon the classic RAG pipeline~\citep{lewis2021retrievalaugmentedgenerationknowledgeintensivenlp}, Agentic RAG integrates autonomous control over retrieval to manage dynamic episodic memories~\citep{ravuru2024agenticretrievalaugmentedgenerationtime, singh2025agenticretrievalaugmentedgenerationsurvey}. To mitigate the similarity-usefulness misalignment, systems explore various optimizations: \textit{hybrid heuristics}, such as Generative Agents combining dense relevance with recency and importance~\citep{park2023generativeagentsinteractivesimulacra}; and \textit{adaptive representations}, where RMM introduces differentiable ranking to improve recall under long-term dialogues~\citep{tan2025prospectretrospectreflectivememory}. For domain-specific trajectory reuse, systems like MemoryBank~\citep{zhong2024memorybank} and CTIM-Rover~\citep{lindenbauer2025knowledgenoisectimroverpitfalls} index episodic segments to robustly fetch past successes, whereas MIRIX routes queries across multiple specialized sub-stores to overcome the limits of a flat index~\citep{wang2025mirixmultiagentmemoryllmbased}. Finally, this vector-based topology readily accommodates multimodal experiences, enabling embodied agents like MrSteve to retrieve relevant past video frames for exploration~\citep{park2025mrsteveinstructionfollowingagentsminecraft}.

\end{itemize}

\subsubsection{Memory Processing}
\label{sec:Memory_Processing}
Given the memory state $m_t := (\text{object}_t,\text{structure}_t)$, memory processing refers to the family of signal-driven operations that act on $m_t$ throughout an agent's lifetime.  Formally, we view memory improvement as $m_{t+1} = \text{IMPROVE}_m(m_t; \mathcal{S}_t)$. Unlike traditional agents that rely on static, hard-coded indexing, self-improving agents use the learning signal $\mathcal{S}_t$ (e.g., task outcomes, utility, or internal critiques) to dynamically adjust what to write, how to retrieve, and when to forget \citep{xu2025sedmscalableselfevolvingdistributed, liang2025selfevolvingagentsreflectivememoryaugmented}. Concretely, we instantiate $\text{IMPROVE}_m$ as an adaptive CRUD (Create, Read, Update, Delete) operation family:

\paragraph{\tagbox{C}} Rather than appending raw logs exhaustively, memory creation is a selective distillation process guided by $\mathcal{S}_t$. Recent systems typically implement this through three recurring patterns: (1) \emph{Semantic compression}, which transforms raw interactions into structured metadata, summaries, or reusable schemas for efficient indexing~\citep{salama2025meminsightautonomousmemoryaugmentation, wang2024agentworkflowmemory, ouyang2025reasoningbankscalingagentselfevolving}; (2) \emph{Context-aware discrete decisions} (e.g., add, update, delete, or no-op), conditioned on retrieved neighbors to prevent redundant or conflicting entries~\citep{chhikara2025mem0buildingproductionreadyai, xu2025amemagenticmemoryllm}; and (3) \emph{Controlled boundary insertion}, which dynamically optimizes the write policy for downstream utility closer to generation time~\citep{zhang2025memgenweavinggenerativelatent}. Across all designs, creation is bounded by a trade-off: over-writing inflates retrieval noise, whereas under-writing sacrifices long-term capability.

\paragraph{\tagbox[cyan!60!teal]{R}}  Memory reading dictates the quality of an agent's downstream reasoning. To maintain precision as the memory bank scales, self-improving systems typically employ four key mechanisms: (1) \emph{Hybrid heuristics}, which blend semantic relevance with recency and importance scores to stabilize recall~\citep{park2023generativeagentsinteractivesimulacra}; (2) \emph{Structure-aware retrieval}, which performs coarse-to-fine traversal over hierarchical or graph topologies to isolate reusable abstractions from low-level traces~\citep{zhang2025gmemorytracinghierarchicalmemory}; (3) \emph{Retrieval gating}, which dynamically decides whether to query memory and how much context is sufficient, thereby optimizing token cost and minimizing distraction~\citep{wang2025scmenhancinglargelanguage}; and (4) \emph{Retrieval-driven adaptation}, where historical trajectories are fetched as actionable cases to guide behavior without requiring parametric model updates~\citep{zhou2025mementofinetuningllmagents}. Ultimately, ineffective reading policies will hinder performance: retrieving irrelevant noise or missing critical details will directly cause subsequent planning and execution to fail.

\paragraph{\tagbox[roseviolet]{U}} Memory updating is the primary mechanism for self-correction and knowledge evolution. Guided by feedback ($\mathcal{S}_t$), agents typically implement updates through four operational patterns: (1) \emph{Scheduled review and attenuation}, which periodically strengthens high-utility items and decays obsolete ones to stabilize memory growth~\citep{liang2025selfevolvingagentsreflectivememoryaugmented}; (2) \emph{Local refresh}, which dynamically updates topological neighbors during new insertions to maintain contextual consistency~\citep{xu2025amemagenticmemoryllm}; (3) \emph{Iterative distillation}, which synthesizes repeated successes into compact, reusable abstractions via continuous selection and replacement~\citep{suzgun2025dynamiccheatsheettesttimelearning, zhang2025agenticcontextengineeringevolving}; and (4) \emph{Offline aggregation}, which shifts computationally expensive compression away from the online execution loop while preserving recall quality~\citep{fang2025lightmemlightweightefficientmemoryaugmented}. Without effective update policies, agents suffer from memory decay: outdated facts remain, knowledge structures break down, and improper merging erases important details.

\paragraph{\tagbox[VOrange]{D}} Memory deletion acts as a systematic noise reduction and resource optimization mechanism. Guided by signal ($\mathcal{S}_t$), agents dynamically identify and remove obsolete or redundant entries through three main patterns: (1) \emph{Multi-stage pruning}, which filters low-value items at write time and periodically clears entries based on access frequency and relevance~\citep{zhang2025mlcagentcognitivemodelbased}; (2) \emph{Consensus-based eviction}, which uses collaborative voting in distributed setups to prevent the accidental deletion of critical shared knowledge~\citep{bach2025pbftbackedsemanticvotingmultiagent}; and (3) \emph{Tiered eviction}, which applies operating-system-inspired rules across different memory layers to bound size while preserving long-term consistency~\citep{kang2025memory}. Improper deletion policies create a direct dilemma: over-pruning loses critical knowledge, while under-pruning floods the system with outdated noise and slows down retrieval.

\paragraph{Putting it together: a signal-driven memory loop.} The dimensions of memory objects, structure, and processing culminate in a unified, signal-driven lifecycle (Figure~\ref{fig:agent_memory}). This loop proceeds in a continuous cycle: (i) Observe \textit{\&} Detect saliency from new interactions; (ii) Create compact objects; (iii) Organize them into the chosen topological structure; (iv) Read on demand via adaptive retrieval; (v) Plan \textit{\&} Act based on fetched context; (vi) Evaluate outcomes to derive the learning signal $\mathcal{S}_t$; and (vii) Update/Delete entries to consolidate high-value knowledge and prune noise. Together, these stages elevate memory from a passive cache to a self-governing engine that sustains open-ended autonomy.

\begin{ttcolorbox}[Takeaway]
The Self-Improvement Memory Loop transforms the conceptual questions of memory management into an actionable, signal-driven framework. By unifying memory writing, structural organization, adaptive retrieval, and feedback-based pruning, this continuous lifecycle effectively elevates memory from a static cache to a scalable engine for self-improvement.
\end{ttcolorbox}

\subsection{Tool}
\label{sec:Tool}

Basic agents have been constrained by their reliance on static, manually curated toolkits, lacking the autonomy to adapt, discover, or integrate new resources when confronted with novel challenges \citep{shapiro2023conceptual, zhang2025largelanguagemodelbrainedgui, sang2025pipelinessurveyparadigmshift, huang2025surveyfoundationmodelpoweredrecommender}. The transition from a basic executor to a genuinely self-improving agent necessitates a shift away from simple, pre-defined tool use toward \textit{Tool Governance Metacognition}. This dynamic paradigm empowers the agent to autonomously reason about the necessity, utility, and reliability of its tools, thereby continually advancing its capability boundaries rather than merely operating within them \citep{liu2025trulyselfimprovingagentsrequire, yin2024g}.

\begin{wrapfigure}{r}{0.37\textwidth}
  \vspace{-\intextsep}        
  \centering
  \includegraphics[width=\linewidth]{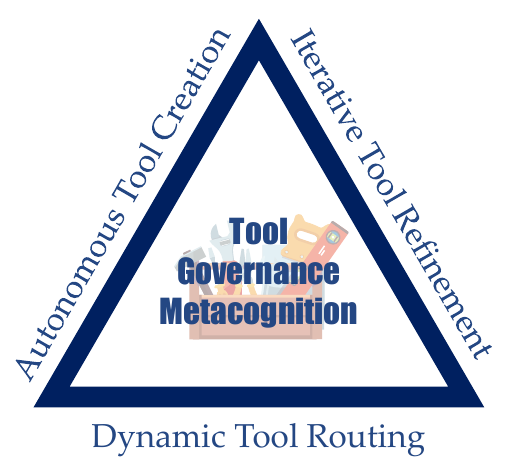}
  \caption{Tool Governance Metacognition.}
  \label{fig:agent_tool}
  \vspace{-1pt}             
\end{wrapfigure}

Formally, tool-based improvement can be written as
\begin{equation}
\mathcal{T}_{t+1}=\IMPROVE_{\mathcal{T}}(\mathcal{T}_t;\mathcal{S}_t),
\end{equation}
where $\mathcal{S}_t$ is the learning signal produced by tool governance metacognition. This metacognition represents the architectural shift essential for a self-improving agent, establishing a dynamic and self-directed mechanism for managing its operational tools \citep{wang2025theoryagentstoolusedecisionmakers}. It is instantiated through three core dimensions: Dynamic Tool Routing, which orchestrates the efficient selection and combination of tools from a growing pool; Iterative Tool Refinement, which adapts and debugs existing instruments to overcome execution failures or environmental shifts; and Autonomous Tool Creation, which invents novel tools to fundamentally expand the agent's capability boundaries. This governance structure moves tool use from a static look-up process to a generative, self-improving cycle.

\subsubsection{Dynamic Tool Routing}
\label{sec:Dynamic_Tool_Routing}

Dynamic tool routing concerns how an agent selects, sequences, and coordinates tools in a heterogeneous environment. As tool pools grow, routing becomes a primary bottleneck for self-improvement. A large tool set increases coverage but also increases error modes, including misrouting, compounding execution failures, and wasted compute. Routing methods therefore trade off coverage, reliability, and decision cost. Most existing systems can be understood through three routing paradigms, distinguished by what they treat as the ``retrieval unit'' and how they incorporate feedback over time.

\paragraph{\textcolor{Plum}{Retrieval- and graph-based routing.}}
Systems in this paradigm tackle the routing bottleneck by optimizing either the retrieval space or the representation of dependencies. To optimize the retrieval space and maintain scalability, systems dynamically manage the granularity and scope of the tool pool. For instance, while MemTool prevents routing degradation by actively pruning the tool set into a lightweight operational memory~\citep{lumer2025memtooloptimizingshorttermmemory}, TAR expands the retrieval unit itself, dynamically routing between atomic APIs and entire competent agents to handle higher-level tasks~\citep{lumer2025tooltoagentretrievalbridgingtools}. Crucially for self-improvement, this retrieval process is not static; it evolves by leveraging past successes as a learning signal. Systems like VOYAGER and MetaAgent index reusable tool-use trajectories into procedural memory, allowing the routing policy to continuously refine itself based on structural knowledge from prior executions~\citep{wang2023voyageropenendedembodiedagent, qian2025metaagentselfevolvingagenttool}. Conversely, to address the compositionality limitation of flat retrieval, recent works encode tool transitions as topological structures. Rather than one-shot relevance, systems like ToolNet and OrchDAG model tool dependencies as directed graphs, enabling routing that accounts for multi-step feasibility and preconditions~\citep{liu2024toolnetconnectinglargelanguage, lu2025orchdagcomplextoolorchestration}. Building on this, MassTool combines learned semantic matching with these graph structures, achieving high-precision navigation even within massive and complex tool topologies~\citep{lin2025masstoolmultitasksearchbasedtool}.

\paragraph{\textcolor{Plum}{Policy-learning routing.}} Policy-learning routing internalizes tool choice as sequential decision making. Instead of retrieving a tool from an external index, the agent learns invocation behavior from training signals, which can improve robustness in multi-turn settings where success depends on state, history, and error recovery. The main trade-off is data and feedback. Learned routing can overfit to training environments and can be brittle when tool interfaces or distributions drift. To establish a reliable routing baseline, systems like AUTOACT, MCP-Flow, Tool-Star, and DeepEyesV2 bootstrap policies through supervised fine-tuning on synthetic or mined trajectories~\citep{qiao2024autoactautomaticagentlearning, wang2025mcpflowfacilitatingllmagents, dong2025tool, hong2025deepeyesv2agenticmultimodalmodel}. However, to continually improve long-horizon behavior, routing policies must transcend static supervision. Works like AGENTFLOW and SPORT achieve this by applying sparse rewards or preference signals to shape planning and exploration, while AutoTIR and DeepAgent explicitly align tool selection with multi-objective compliance and action-level attribution~\citep{li2025intheflowagenticoptimizationeffective, li2025iterative, wei2025autotirautonomoustoolsintegrated, li2025deepagentgeneralreasoningagent}. Pushing this paradigm to its limit, ToolGen collapses retrieval, selection, and invocation into a single generative process via tool-token unification. While this elegantly reduces pipeline complexity, it places a heavy burden on the learned policy to remain calibrated under distribution shifts~\citep{wang2025toolgenunifiedtoolretrieval}.

\paragraph{\textcolor{Plum}{Proactive and interactive routing.}}
A third paradigm treats routing as an interactive process rather than a one-shot decision. This perspective is motivated by two recurring failure sources in self-improving agents. First, user intent is often underspecified, and misrouting can be avoided by clarification. Second, execution failures are common, and routing must support local repair without restarting the entire plan. Proactive routing therefore trades additional interaction and compute for higher reliability. Rather than failing passively, these systems embed metacognitive interventions to handle uncertainty. When confronting underspecified intents or functional gaps, agents like MCP-Zero and ASKTOACT proactively initiate tool discovery or prompt for clarification, transforming ambiguity into a learning signal for self-correction~\citep{fei2025mcpzeroactivetooldiscovery, zhang2025asktoactenhancingllmstool}. Conversely, when facing execution failures, dynamic repair becomes essential. Tool-Planner facilitates this by clustering tools into interchangeable kits, enabling localized, API-level repair without the prohibitive cost of global replanning~\citep{liu2025toolplannertaskplanningclusters}. Similarly, ToolACE-R dynamically calibrates its revision effort based on task difficulty, balancing routing reliability against practical compute budgets in real-world deployments~\citep{zeng2025toolacermodelawareiterativetraining}.

\subsubsection{Iterative Tool Refinement}
\label{sec:Iterative_Tool_Refinement}

Iterative tool refinement is the mechanism by which agents turn fragile programs into dependable skills. It is central to self-improvement because tool errors are not merely execution-time failures. If unreliable tools are stored and reused, they can corrupt the agent's future behavior through repeated retrieval and compounding errors. Refinement therefore serves both as debugging and as a gatekeeping process that controls what enters the structural skill repository~\citep{11185878, dolcetti2025helpingllmsimprovecode, petrovic2025surveygenaiautomotivesoftware}.

A canonical refinement loop alternates between generation, execution, and revision. VOYAGER establishes this baseline by executing generated code, capturing error traces and environment feedback as learning signals ($\mathcal{S}_t$), and feeding them back into subsequent revisions until a verifier confirms success~\citep{wang2023voyageropenendedembodiedagent}. Building on this foundation, recent systems strengthen the refinement process across three dimensions: (1) Critique Specialization. To enhance diagnostic accuracy, systems move beyond generic self-reflection. For instance, STELLA deploys a dedicated critic agent to assess intermediate results and provide targeted feedback, mitigating the tendency of monolithic models to overlook domain-specific failure modes~\citep{jin2025stellaselfevolvingllmagent}. (2) API Abstraction. Rather than merely revising raw action traces, several works focus on abstracting robust subroutines. SkillWeaver distills interactions into reusable web tools refined through execution feedback~\citep{zheng2025skillweaverwebagentsselfimprove}, while PyVision dynamically refines Python programs to stabilize multimodal tool use against noisy perception~\citep{zhao2025pyvisionagenticvisiondynamic}. This abstraction significantly improves skill transfer across tasks. (3) Interface Alignment. Addressing the semantic gap between agents and tools, systems like DRAFT iteratively refine tool documentation rather than the underlying code. This targets the critical observation that many execution failures stem from mismatches between natural language instructions and tool affordances, rather than from flawed  implementations~\citep{qu2025explorationmasteryenablingllms}.

\subsubsection{Autonomous Tool Creation}
\label{sec:Autonomous_Tool_Creation}

Autonomous tool creation expands the agent's capability boundary by synthesizing new executable functions when existing tools are insufficient. For self-improving agents, creation is most valuable when it converts one-off problem solving into reusable procedural knowledge. It also introduces its own risks. Newly created tools must be validated, documented, and integrated without destabilizing routing policies, otherwise tool growth can increase brittleness rather than autonomy~\citep{gao2025surveyselfevolvingagentspath, qiu2025alitageneralistagentenabling}.

Existing systems approach this challenge by advancing tool creation across three dimensions: (1) Synthesis Triggers. Tool creation can be driven either by immediate needs or open-ended curiosity. While systems like ATLASS and PyVision emphasize on-demand invention—synthesizing and refining tools  when current retrieved APIs fail~\citep{haque2025advancedtoollearningselection, zhao2025pyvisionagenticvisiondynamic}, agents like FRIDAY and STELLA shift toward self-directed exploration. They autonomously propose curricula or discover domain-specific resources (e.g., bioinformatics) to proactively accumulate tools outside a fixed task list~\citep{wu2024oscopilotgeneralistcomputeragents, jin2025stellaselfevolvingllmagent}. (2) Lifecycle Automation. Synthesizing raw code is only the first step; rendering it executable is the actual bottleneck. TOOLMAKER addresses this by automating the end-to-end tool lifecycle, extracting logic from scientific papers, installing dependencies, debugging in a closed loop, and producing robust callable interfaces~\citep{wölflein2025llmagentsmakingagent}. This transforms abstract function generation into tangible deployability~\citep{cai2024largelanguagemodelstool, yuan2024craftcustomizingllmscreating, qian2024creatortoolcreationdisentangling}. (3) Standardized Integration. To ensure that explosive tool growth does not destabilize existing routing policies, recent systems adopt protocol-driven architectures. Frameworks like Alita and Code2MCP automate the conversion of code repositories into standardized services via the Model Context Protocol (MCP)~\citep{qiu2025alitageneralistagentenabling, qiu2025alitagselfevolvinggenerativeagent, ouyang2025code2mcptransformingcoderepositories}. Similarly, AgentOrchestra enforces a rigorous pipeline of intent parsing, validation, and formal registration before a new tool enters the active pool~\citep{zhang2025agentorchestraorchestratinghierarchicalmultiagent}. This ensures that autonomous creation remains fully compatible with the agent's broader governance and retrieval constraints.

\begin{ttcolorbox}[Takeaway]
Based on the Tool Governance Metacognition framework, self-improving agents evolve tool usage into a growth-oriented architecture with dynamic routing, iterative refinement, and autonomous creation capabilities. This architecture transforms the system from a static toolkit into a dynamic skill repository with sustainably expandable capabilities.
\end{ttcolorbox}

\subsection{Full Scaffolding}
\label{sec:Full_Scaffolding}

Full scaffolding self-improvement represents the deepest level of architectural intervention. Rather than merely tuning isolated components (e.g., prompts, tools, or memory), the agent treats its entire operational logic and codebase as a mutable substrate, enabling fundamental structural reorganization. Formally, recalling the agent state $\mathcal{A}_t=(\theta_t,\Sigma_t)$, full scaffolding improvement corresponds to the most general \emph{scaffolding-only} transition:
\begin{equation}
\qquad \Sigma_{t+1}=\IMPROVE_{\Sigma}(\Sigma_t;\mathcal{S}_t),
\end{equation}
where $\mathcal{S}_t$ denotes the learning signal extracted from the agent's own execution traces and evaluations (e.g., task success/failure, unit tests, self-critique, cost signals).
Crucially, unlike settings with a fixed external meta-optimizer, the improvement procedure is itself implemented within the current scaffolding, making the update \emph{self-referential}:
\begin{equation}
\Sigma_{t+1} = \mathcal{I}_{\Sigma_t}(\Sigma_t;\mathcal{S}_t),
\end{equation}
where $\mathcal{I}_{\Sigma_t}$ highlights that the improver can evolve together with the agent it improves. In principle, these self-referential loops can discover unanticipated update strategies and enhance the agent's intrinsic evolvability~\citep{zhang2025darwingodelmachineopenended, gerhart2007theory, hendrikse2007evolvability, dawkins2019evolution}. While fully open-ended recursive self-improvement remains a grand challenge, current systems successfully operationalize this concept within defined objectives, benchmarks, and safety protocols.

\begin{figure*}[ht]
    \centering
    \includegraphics[width=1\linewidth]{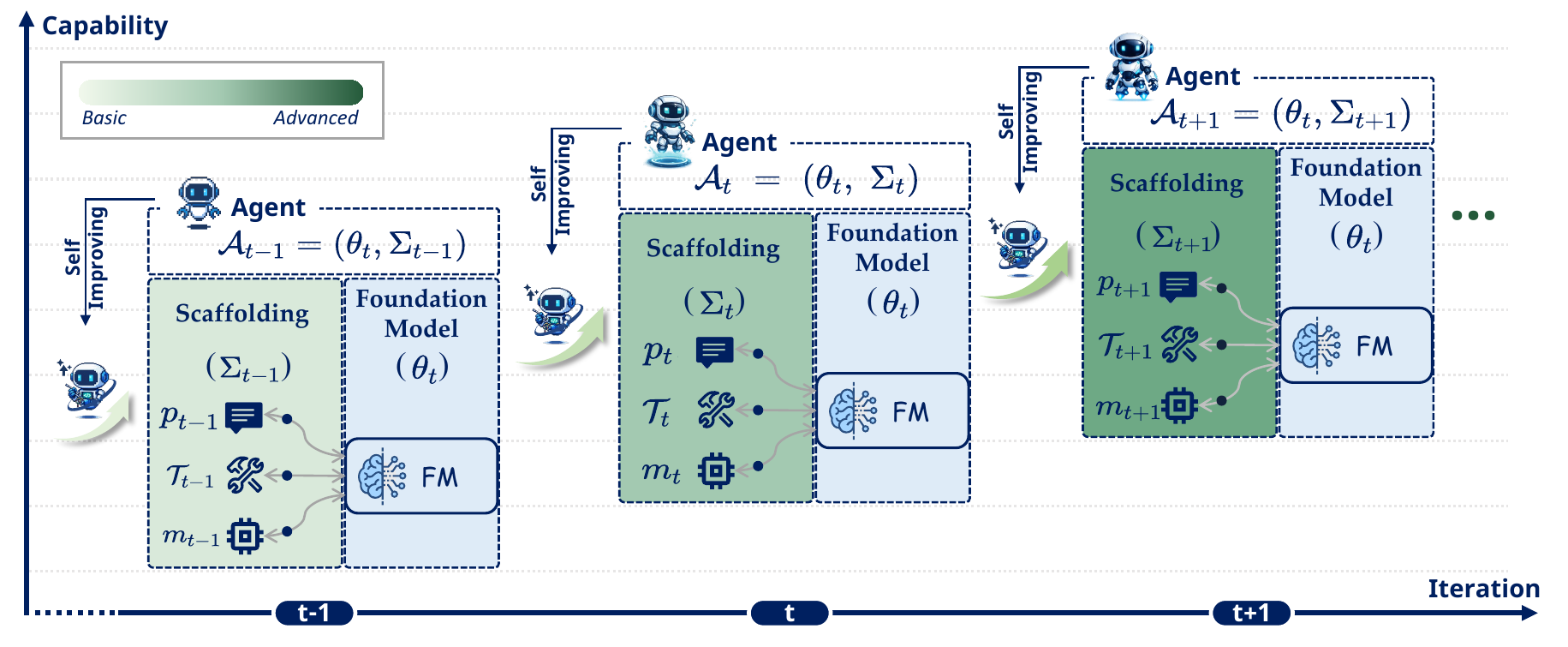}
    \caption{Full scaffolding self-improvement across iterations.}
    \label{fig:full_sacffolding}
\end{figure*}

Within this paradigm, agents' scaffolding is typically represented as conventional computer programs with mild constraints. {Formally,} let $\langle \Sigma_t \rangle$ denote a serializable encoding of the program that implements the agent, and let $exec(\cdot)$ denote execution in an environment (e.g., an interpreter, compiler, and sandbox).
A full-scaffolding update can be viewed as producing a candidate program via executing the current agent-as-improver on its own code:
\begin{equation}
\langle \tilde{\Sigma}_{t+1} \rangle = exec(\langle \Sigma_t \rangle;\mathcal{S}_t),
\end{equation}
often materialized as a patch $\Delta_t$ applied to the current scaffolding,
$\tilde{\Sigma}_{t+1} = \Sigma_t \oplus \Delta_t$.
In practice, a verifier (e.g., unit tests, regression suites, safety checks) gates the update:
\begin{equation}
\Sigma_{t+1}=
\begin{cases}
\tilde{\Sigma}_{t+1}, & \mathcal{V}(\tilde{\Sigma}_{t+1})=1,\\
\Sigma_t, & \text{otherwise}.
\end{cases}
\end{equation}

The universal representability of computer programs (i.e., Turing completeness) enables self-improving agents to explore an extensive space of self-improving strategies, bounded in principle only by the theoretical limits of computation. Early work on applying a computer program to improve itself dates back to self-referential program-search and meta-evolution frameworks \citep{schmidhuber1987selfreferential}.

AlphaEvolve \citep{novikov2025alphaevolvecodingagentscientific} is designed as a coding agent for open scientific problems. It adopts an evolutionary paradigm, continuously receiving feedback from one or more evaluators to iteratively improve its algorithms, thereby substantially accelerating new scientific discoveries and optimizing complex technical stacks. Similarly, ShinkaEvolve \citep{lange2025shinkaevolveopenendedsampleefficientprogram} is an LLM-driven program-evolution framework that enables efficient, open-ended program discovery and optimization across multiple tasks with very few evaluation samples, via exploration–exploitation–balanced parent sampling, novelty-based rejection sampling for code, and bandit-based adaptive selection of an integrated ensemble of LLMs.

ADAS \citep{hu2025automateddesignagenticsystems} searches within a design space for an agentic system that maximizes a given evaluation function using a dedicated search algorithm. EvoFlow continuously evolves a set of heterogeneous workflows that achieve favorable cost–performance trade-offs while maintaining structural diversity, yielding a Pareto set in an online manner. Self-Taught Optimizer \citep{zelikman2024self} abstracts an agent’s scaffolding as an improver: given its own program and a utility function, it repeatedly queries an LM to generate multiple candidate improved versions and selects the highest-scoring one as the output, forming a recursive self-improvement loop. Agent Symbolic Learning \citep{ou2025symbolic} views a language agent as a “symbolic network” whose weights are instantiated by prompts, tools, and their composition; it simulates backpropagation and gradient descent through natural-language losses/gradients, and uses a symbolic optimizer to jointly update prompts, tools, and pipelines, enabling sustained self-learning and evolution in real deployments.

Another representative line is the open-ended evolutionary framework inspired by the Gödel machine. \citet{yin2025godel} proposes a Gödel-machine-inspired self-referential agent framework, Gödel Agent, implemented via monkey patching. It autonomously performs self-awareness, self-modification, and recursive self-improvement, aiming to search the entire agent design space. Inspired by Darwinian evolution and open-ended research, \cite{zhang2025darwingodelmachineopenended} proposed the self-evolving coding agent Darwin Gödel Machine (DGM), which maintains an archive of newly generated coding agents and grows a continually expanding tree of diverse, high-quality agents through open-ended exploration, allowing parallel exploration of multiple paths in the search space. Subsequently, \cite{wang2025huxleygodelmachinehumanlevelcoding} proposed the Huxley-Gödel Machine (HGM), motivated by Huxley’s notion of clades; it introduces a \textit{clade-level metaproductivity} (CMP) metric to guide evolution as an approximation to the Gödel-machine ideal. Live-SWE-Agent \citep{xia2025livesweagentsoftwareengineeringagents} is the first real-time software agent that can autonomously and continuously evolve on the fly at runtime; by evolving its own scaffolding, especially its tool components, it achieves desired performance on software issue-solving tasks.

\begin{ttcolorbox}[Takeaway]
Full scaffolding improvement is the category most closely related to recursive self-improvement, because it treats the agent's own source code and operational logic as mutable substrates. While achieving truly open-ended recursive self-improvement remains a grand challenge, current systems successfully operationalize this concept as bounded, verifiable loops. Operating within human-designed objectives and safety protocols, these systems provide a principled and measurable path toward reliable self-improvement.
\end{ttcolorbox}

\section{Applications}
\label{sec:Applications}

The preceding sections organized self-improving agents by what is updated, namely model parameters or scaffolding, and by the signals that drive these updates. This section examines how these mechanisms appear across representative domains. Across applications, a common pattern is the use of sandboxed or otherwise controlled environments that provide feedback while limiting the cost of failure. \textbf{Software engineering} relies on compilers, tests, and continuous integration pipelines; \textbf{web automation} uses simulated or instrumented browsers; \textbf{games} provide resettable environments with explicit rules and outcomes; \textbf{scientific discovery} uses executable workflows, domain tools, and simulators; \textbf{embodied AI} relies on robotic simulators and limited real-world rollouts; and \textbf{general computer control} uses virtualized desktops or isolated operating-system environments. The fidelity, scalability, and cost of these environments shape each domain's dominant bottlenecks, improvement targets, and iteration modes, as summarized in Table~\ref{tab:applications_loops}.

\begin{figure*}[ht]
    \centering
    \includegraphics[width=1\linewidth]{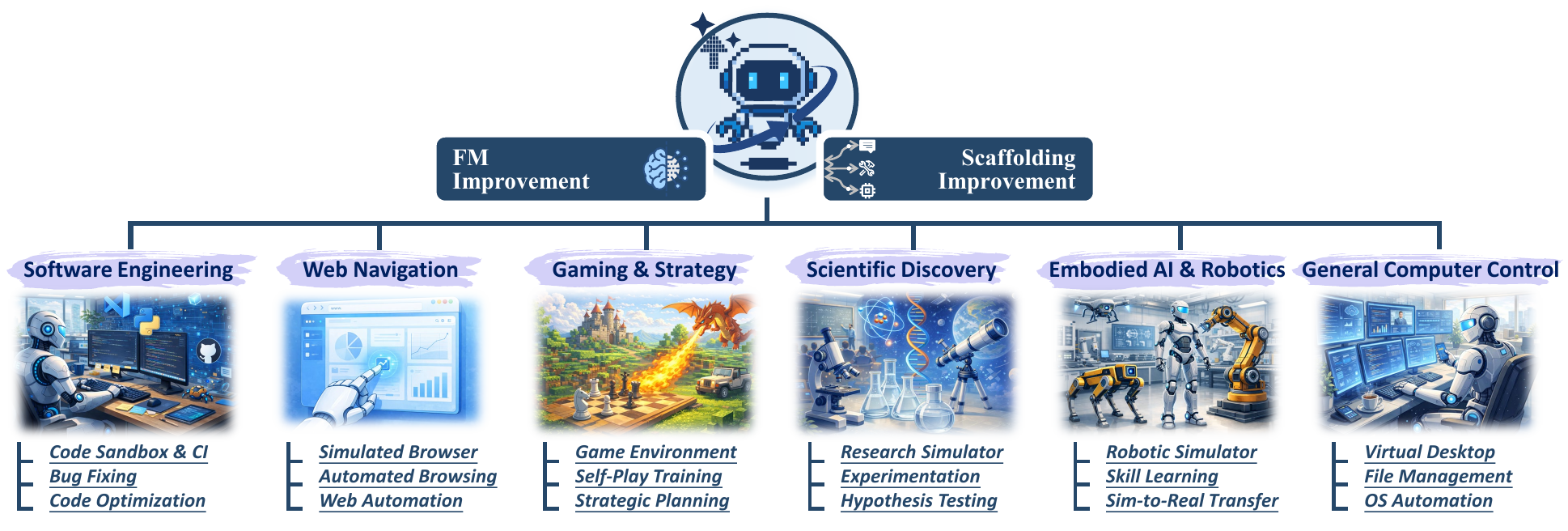}
    \caption{Representative application domains for self-improving agents.}
    \label{fig:application}
\end{figure*}

\subsection{Software Engineering}

Software engineering (SWE) is a useful testbed for self-improving agents because it provides dense and automatable feedback. Compilers, unit tests, linters, and continuous integration pipelines can turn many agent actions into checkable outcomes, making improvement easier to measure than in domains that rely on subjective evaluation. This property naturally supports both branches of our taxonomy: (i) \textbf{FM improvement}, where executable feedback is used to update model parameters $\theta$, and (ii) \textbf{scaffolding improvement}, where the model is fixed but the scaffold $\Sigma$ is revised. In SWE, the scaffold may include prompts, memory, tools, control logic, and, in some cases, the agent's own source code.

\paragraph{Evaluation substrate vs. self-improvement.}
It is important to distinguish evaluation environments from self-improving methods. SWE-bench provides a standardized suite of real GitHub issues, repositories, and test-based validation, enabling quantitative measurement of progress \citep{jimenez2024swebenchlanguagemodelsresolve}. Several related benchmarks have since been proposed \citep{deng2025swebenchproaiagents, aleithan2024swebenchenhancedcodingbenchmark, yang2025swebench}. However, systems such as SWE-agent \citep{yang2024sweagentagentcomputerinterfacesenable} and Agentless \citep{xia2024agentlessdemystifyingllmbasedsoftware} are better viewed as non-self-improving baselines under our definition. They may iterate within a task instance, but they do not, by default, perform persistent cross-interaction updates to $\theta$ or $\Sigma$. Their relevance here is methodological: they show that scaffold design can substantially affect performance, motivating the question of whether such scaffolds can be improved automatically rather than hand-engineered.

\paragraph{Scaffolding improvement via self-modifying software agents.}
A distinctive feature of SWE is that the agent itself is software, which makes direct scaffold or source-code modification feasible. The Darwin G"odel Machine operationalizes this idea by iteratively rewriting its own codebase and validating candidate modifications on coding benchmarks, producing a tree archive of improved descendants \citep{zhang2025darwingodelmachineopenended}. Building on this self-modification paradigm, the Huxley--G"odel Machine searches the space of self-modifications using an estimate of descendant performance, aiming to optimize long-term improvement potential rather than only immediate benchmark gains \citep{wang2025huxleygodelmachinehumanlevelcoding}.

Live-SWE-agent moves scaffold improvement from offline search to runtime self-evolution. Starting from a minimal scaffold, the agent edits and extends its own scaffold while solving real SWE tasks, so the current issue can inform future behavior \citep{xia2025livesweagentsoftwareengineeringagents}. SE-Agent studies a complementary trajectory-level mechanism, using cross-trajectory revision, recombination, and refinement to escape local optima in reasoning and action sequences \citep{lin2025seagentselfevolutiontrajectoryoptimization}. Viewed through $\Sigma=(p,m,T,g)$, these methods update increasingly rich scaffold components, from prompts and tool routines to full system code. They also highlight a SWE-specific lesson: performance depends not only on the base model, but also on the agent's executable interface to the codebase and its ability to preserve reusable problem-solving strategies across issues.

\paragraph{FM improvement from executable feedback.}
SWE also provides a direct route to FM improvement because test outcomes can serve as rewards or supervision. SWE-RL scales reinforcement learning to real-world software engineering tasks by using software-evolution data and executable signals to improve the underlying model \citep{wei2025swerladvancingllmreasoning}. Related work studies reinforcement learning in long-context and multi-turn SWE settings, with the goal of training models that can sustain extended tool-using interaction while remaining grounded in verification feedback \citep{golubev2025traininglongcontextmultiturnsoftware}. A practical limitation is that execution-based rewards can be sparse, noisy, or expensive, due to flaky tests, incomplete coverage, or costly environment setup. One complementary direction strengthens the executable signal itself by expanding the test suite: curiosity-driven planning can steer an LLM to generate tests that reach under-explored behaviors~\citep{amayuelas2026planning}. Execution-free feedback can therefore complement tests. SWE-RM explores reward-model-based scoring as a substitute for, or supplement to, unit tests, and analyzes which verifier properties transfer to reinforcement learning improvements \citep{shum2025swermexecutionfreefeedbacksoftware}.

\paragraph{Synthesis and open challenges.}
SWE exposes a useful contrast between the two improvement families. Scaffolding improvement is fast, modular, and model-agnostic, but it may overfit to benchmark tooling, repository conventions, or interaction protocols. FM improvement can internalize skills into $\theta$ and may transfer more broadly, but it is more expensive and more exposed to reward hacking and evaluation artifacts. Robust SWE self-improvement therefore requires (i) stronger evaluation against benchmark-specific search and verifier overfitting, such as benchmark mutation and stress testing, as well as (ii) hybrid loops in which scaffold-level discoveries are distilled into the base model only when they transfer beyond a fixed verifier or benchmark protocol \citep{garg2026savingswebenchbenchmarkmutation}.

\subsection{Web Navigation and Automation}

Web navigation and automation is a challenging domain for self-improving agents because the environment is dynamic, long-horizon, and often weakly verified. Web interfaces change frequently, the same user intent may be realized through different layouts or DOM structures, and failures often become visible only late in a trajectory. These properties make one-shot prompt engineering brittle and motivate improvement loops that persist across tasks. In our taxonomy, this domain supports both FM improvement, which updates $\theta$ from interaction data and feedback, and scaffolding improvement, which keeps $\theta$ fixed while updating $\Sigma$.

\paragraph{Evaluation and data substrate.}
A first line of work builds reproducible environments and demonstrations, which are prerequisites for systematic self-improvement. Mind2Web provides diverse instruction-following trajectories on real websites and supports supervised learning or imitation as an initial improvement step \citep{deng2023mindweb}. WebArena offers a self-hostable environment with long-horizon tasks and execution-based success criteria \citep{zhou2024webarena}. VisualWebArena extends evaluation to visually grounded web tasks and exposes failures in perception and grounding \citep{koh2024visualwebarenaevaluatingmultimodalagents}. WorkArena targets enterprise-style workflows and highlights the gap between current agents and routine knowledge-worker automation \citep{drouin2024workarenacapablewebagents}. BrowserGym further unifies evaluation across web-agent benchmarks, reducing experimental fragmentation and enabling more controlled comparisons of training and scaffold design \citep{workarena2024, chezelles2025browsergym}. These resources are not self-improving methods themselves, but they provide the data and measurement substrate needed to study improvement across interactions.

\paragraph{Scaffolding improvement with memory and grounding.}
In web environments, many improvements come from better scaffolding because robust grounding remains a central bottleneck. SeeAct couples visual understanding with action grounding on live websites, reducing brittle action selection in long-horizon execution \citep{zheng2024gptvision}. WebCoach is more directly aligned with self-improvement: it equips browsing agents with cross-session memory curated from new trajectories, allowing the agent to avoid repeated mistakes without retraining the base model \citep{liu2025webcoachselfevolvingwebagents}. ReAP similarly stores and reuses reflections over successful and failed trajectories to improve later web navigation \citep{azam2025reflectionbasedmemorywebnavigation}. These approaches instantiate scaffolding improvement by updating memory content, retrieval policies, and grounding-related structural artifacts that transfer across tasks.

\paragraph{FM improvement from interaction feedback.}
A complementary line updates $\theta$ using web interaction signals. \cite{patel2024largelanguagemodelsselfimprove} study self-improvement on WebArena by fine-tuning on mixtures of model-generated data and show that iterative training can improve web-agent capability. OpenWebVoyager proposes an exploration-and-feedback loop on real websites, improving its policy by learning from high-quality trajectories over multiple iterations \citep{he2024openwebvoyagerbuildingmultimodalweb}. Several works adopt reinforcement learning for long-horizon browser control. WebRL introduces a self-evolving online curriculum that generates new tasks from failures and combines it with outcome-supervised reward modeling \citep{qi2025webrl}. WebAgent-R1 trains web agents via end-to-end multi-turn reinforcement learning with binary success rewards and reports gains on WebArena-Lite \citep{wei2025webagentr1trainingwebagents}. Beyond navigation benchmarks, Agent Q studies learning from both successful and unsuccessful trajectories in WebShop and shows that such experience can improve generalization in multi-step web interaction \citep{putta2024agentqadvancedreasoning}. These FM-oriented methods also reveal a recurring difficulty: web feedback is often sparse, so improvement depends heavily on reward modeling, curriculum design, and the stability of the evaluation environment.

\paragraph{Synthesis and open challenges.}
Web automation faces a core challenge of self-improvement. This challenge comes from non-stationarity and weak observability. Page layouts and interaction flows often change over time. These changes can quickly weaken gains that were measured on static snapshots. Scaffolding improvement can help systems update grounding, state abstraction, and recovery behavior more quickly. Yet it can also overfit to specific websites. It may also increase safety risks, such as prompt injection or unintended high-impact actions~\citep{chen2026much}. FM improvement can support more transferable navigation skills. However, it depends on stable training signals and careful treatment of stale data. Progress therefore requires drift-aware evaluations. It also requires sandboxed protocols that separate safe exploration from irreversible actions. Finally, it requires grounding-centered improvements that transfer across layouts, DOM structures, and interaction conventions. These improvements should avoid memorizing site-specific patterns.

\subsection{Games and Strategic Reasoning}

Games remain a central testbed for self-improving agents because they provide repeatable interaction, well-defined objectives, and scalable feedback \citep{stanic2023learning}. Even when state spaces are large and long-horizon planning is required, games offer closed training loops in which agents can generate experience through self-play \citep{silver2018general, silver2017masteringchessshogiselfplay, Schrittwieser_2020, openai2019dota2largescale}. This makes the domain a natural fit for our taxonomy. Many game agents improve by updating model or policy parameters $\theta$ through reinforcement learning, while others improve by evolving the scaffold $\Sigma$, such as curriculum logic, planning routines, or memory structures that store reusable skills and experience.

\paragraph{FM improvement through self-play and outcome feedback.}
Recent work extends self-play to foundation models by using strategic games as sources of interaction data and outcome-based learning signals. SPAG trains language models through a two-player adversarial game and improves the policy via self-play reinforcement learning \citep{cheng2025selfplayingadversariallanguagegame}. SPIRAL studies multi-turn zero-sum self-play and shows that self-generated competitive interactions can provide a scalable signal for improving reasoning without human-labeled data \citep{liu2025spiralselfplayzerosumgames}. In adversarial game suites, SCO-PAL performs step-level policy optimization from game interaction data and identifies self-play as an effective opponent-selection strategy for strategic reasoning \citep{zhang2025enhancinglanguageagentstrategic}. Self-play reinforcement learning with limited initial data further examines how self-generated interactions can bootstrap improvement when external supervision is scarce \citep{fang2025serl}.

Imperfect-information and multi-agent settings introduce additional challenges in credit assignment and payoff estimation \citep{zhuge2023mindstorms, stanic2023learning}. MARSHAL proposes an end-to-end reinforcement learning framework for multi-agent self-play across cooperative and competitive strategic games, and reports transfer beyond games to multi-agent reasoning benchmarks \citep{yuan2026mars}. In negotiation-heavy environments, DipLLM studies fine-tuning for equilibrium-oriented policies in Diplomacy, providing a parameter-updating route for language-mediated strategic interaction \citep{xu2025dipllmfinetuningllmstrategic}. For high-variance payoff domains, SPRL uses reward shaping from self-play payoffs to stabilize training in complex strategic games \citep{anonymous2025llm}. Social deduction games provide another setting in which language agents can improve long-horizon strategic behavior through reinforcement learning from interaction outcomes \citep{xu2025languageagentsreinforcementlearning}.

\paragraph{Scaffolding improvement through curricula and reusable skills.}
Games also support scaffolding improvement because skills, procedures, and curricula can be represented explicitly and reused across interactions. Voyager exemplifies this paradigm in Minecraft by maintaining an expanding library of executable skills and an automatic curriculum that persists across tasks \citep{wang2023voyageropenendedembodiedagent}. Odyssey extends this direction with a structured open-world skill library and a planner--critic loop that improves long-horizon execution through skill reuse and prerequisite checking \citep{liu2025odysseyempoweringminecraftagents}.

Beyond open-world games, several methods construct reusable skills from interaction trajectories. Skill Set Optimization extracts high-reward subtrajectories, converts them into transferable skills, and prunes ineffective ones, yielding continual in-context policy improvement in game-like environments \citep{nottingham2024skillsetoptimizationreinforcing}. ExpeL shows that agents can accumulate experiences, distill them into natural-language lessons, and reuse them as in-context demonstrations for later decision making \citep{zhao2024expel}. In strategic and negotiation-heavy games, Richelieu maintains memory and reflection across self-play interactions, using accumulated experience to revise planning and negotiation behavior without weight updates at every step \citep{guan2024richelieu}. Recent self-evolving strategic systems similarly emphasize structural artifacts and long-term memory as the substrate for iterative strategy refinement across repeated games \citep{belle2025agentschangeselfevolvingllm}.

\paragraph{Synthesis and open challenges.}
Games offer unusually clear conditions for studying self-improvement. Their environments can be simulated, and feedback can be collected at scale. At the same time, games reveal risks that may be less visible in software engineering or web automation. Self-play can create brittle competence. A system may exploit a simulator or a narrow opponent distribution. It may then fail when the population shifts or when the rules change. In multi-agent games, improvement may also become non-monotonic. Strategic cycling can make progress difficult to measure. Evaluation against a fixed set of opponents may therefore give a misleading view of performance. Language-based strategy raises further safety concerns. These concerns include persuasion and deception. Win-loss objectives alone cannot govern such behaviors. Progress in this domain requires population-based evaluation that captures non-transitivity. It also requires training designs that promote robustness under rule and opponent shifts. Explicit constraints are also needed for language-mediated interaction.

\newcommand{\domSWE}{\faCode\;\textcolor{black!65}{{\scriptsize\textbf{SWE}}}}
\newcommand{\domWeb}{\faGlobe\;\textcolor{black!65}{{\scriptsize\textbf{Web}}}}
\newcommand{\domGame}{\faChessKnight\;\textcolor{black!65}{{\scriptsize\textbf{Game}}}}
\newcommand{\domSci}{\faFlask\;\textcolor{black!65}{{\scriptsize\textbf{Sci}}}}
\newcommand{\domRobot}{\faRobot\;\textcolor{black!65}{{\scriptsize\textbf{Emb}}}}
\newcommand{\domPC}{\faDesktop\;\textcolor{black!65}{{\scriptsize\textbf{PC}}}}

\newcommand{\bitem}{\textcolor{black!80}{\raisebox{0.15ex}{\scriptsize\textbullet}}\hspace{0.45em}}

\newcommand{\cellwrap}[1]{%
  \parbox[c]{\linewidth}{%
    \raggedright
    \setlength{\parskip}{0.15em}%
    \vspace{0.25em}%
    #1%
    \vspace{0.25em}%
  }%
}

\newcommand{\bi}[1]{\bitem #1\par}

\newcommand{\li}[1]{#1\par}

\begin{table*}[t]
\centering
\scriptsize
\setlength{\tabcolsep}{2.2pt}
\renewcommand{\arraystretch}{1.10}
\rowcolors{3}{gray!8}{white}

\begingroup
\renewcommand{\tabularxcolumn}[1]{m{#1}}
\setlength{\emergencystretch}{1em}
\sloppy

\begin{tabularx}{\textwidth}{
>{\centering\arraybackslash}m{1.8cm}  
>{\raggedright\arraybackslash}X       
>{\raggedright\arraybackslash}X       
>{\raggedright\arraybackslash}X       
>{\raggedright\arraybackslash}X       
>{\raggedright\arraybackslash}X       
>{\raggedright\arraybackslash}X       
}
\toprule
\rowcolor{tab-blue}
\textcolor{white}{\textbf{Domain}} &
\textcolor{white}{\textbf{Sandbox and arena}} &
\textcolor{white}{\textbf{Learning signal}} &
\textcolor{white}{\textbf{Main bottleneck}} &
\textcolor{white}{\textbf{Primary improvement target}} &
\textcolor{white}{\textbf{Iteration mode}} &
\textcolor{white}{\textbf{Exemplars}} \\
\midrule

\domSWE &
\cellwrap{%
  \bi{Repository with compiler, unit tests, and CI}
  \bi{Failures are typically reversible}
} &
\cellwrap{%
  \bi{Deterministic binary outcomes (pass or fail)}
  \bi{Compilation errors}
  \bi{Static analysis signals}
} &
\cellwrap{%
  \bi{Patch correctness under repository constraints}
  \bi{Tool and interface efficiency}
} &
\cellwrap{%
  \bi{Mainly scaffolding}
  \bi{Some systems also self-edit code or fine-tune models}
} &
\cellwrap{%
  \bi{Online debugging per issue}
  \bi{Offline aggregation across issues}
} &
\cellwrap{%
  \li{\textbf{\hyperlink{cite.zhang2025darwingodelmachineopenended}{\makecell[l]{DGM(2025)}}}}
  \li{\textbf{\hyperlink{cite.wang2025huxleygodelmachinehumanlevelcoding}{\makecell[l]{HGM(2025)}}}}
  \li{\textbf{\hyperlink{cite.xia2025livesweagentsoftwareengineeringagents}{\makecell[l]{Live-SWE-agent\\(2025)}}}}
  \li{\textbf{\hyperlink{cite.zhang2026agentdevelreframingselfevolvingllm}{\makecell[l]{AgentDevel\\(2026)}}}}
} \\

\domWeb &
\cellwrap{%
  \bi{Simulated and standardized browsers}
  \bi{Partial observability of user interfaces}
} &
\cellwrap{%
  \bi{Sparse task completion signals}
  \bi{Long-horizon failures}
  \bi{Partial checks and heuristics}
} &
\cellwrap{%
  \bi{Grounding actions to dynamic layouts}
  \bi{Distribution shift over sites and pages}
} &
\cellwrap{%
  \bi{Scaffolding, including perception--action grounding, planning, and iterative repair}
  \bi{Trajectory and trace curation}
} &
\cellwrap{%
  \bi{Imitation style learning}
  \bi{Online correction along trajectories}
} &
\cellwrap{%
  \li{\textbf{\hyperlink{cite.qi2025webrl}{\makecell[l]{WebRL(2025)}}}}
  \li{\textbf{\hyperlink{cite.fang2025webevolver}{\makecell[l]{WebEvolver\\(2025)}}}}
  \li{\textbf{\hyperlink{cite.zheng2025skillweaverwebagentsselfimprove}{\makecell[l]{SkillWeaver\\(2025)}}}}
  \li{\textbf{\hyperlink{cite.zhang2026webrollbackenhancingwebagents}{\makecell[l]{WebRollback\\(2026)}}}}
} \\

\domGame &
\cellwrap{%
  \bi{Game engines with reliable reset}
  \bi{Self-play interactions}
} &
\cellwrap{%
  \bi{Win or loss outcomes, or scalar rewards}
  \bi{Clear terminal signals}
} &
\cellwrap{%
  \bi{Long-horizon planning}
  \bi{Imperfect information in some settings}
  \bi{Multi-agent non-transitivity}
} &
\cellwrap{%
  \bi{Model and policy parameters}
  \bi{Search and planning components}
} &
\cellwrap{%
  \bi{Self-play}
  \bi{Iterative policy improvement}
} &
\cellwrap{%
  \li{\textbf{\hyperlink{cite.guan2024richelieu}{\makecell[l]{Richelieu(2024)}}}}
  \li{\textbf{\hyperlink{cite.xu2025dipllmfinetuningllmstrategic}{\makecell[l]{DipLLM(2025)}}}}
  \li{\textbf{\hyperlink{cite.yuan2026mars}{\makecell[l]{MARSHAL\\(2025)}}}}
  \li{\textbf{\hyperlink{cite.cheng2025selfplayingadversariallanguagegame}{\makecell[l]{SPAG(2025)}}}}
} \\

\domSci &
\cellwrap{%
  \bi{Tool augmented research loops}
  \bi{Variable-cost evaluation}
} &
\cellwrap{%
  \bi{Experimental metrics}
  \bi{Tool outputs}
  \bi{Critique-based refinement}
} &
\cellwrap{%
  \bi{Expensive and noisy evaluation}
  \bi{Knowledge fragmentation}
  \bi{Heterogeneous tools and interfaces}
} &
\cellwrap{%
  \bi{Scaffolding, including tool orchestration, planning, and verification}
  \bi{Domain specialization}
} &
\cellwrap{%
  \bi{Propose, run, critique, and revise cycles}
  \bi{Mixed online and offline iteration}
} &
\cellwrap{%
  \li{\textbf{\hyperlink{cite.lu2024aiscientistfullyautomated}{\makecell[l]{The AI Scientist\\(2024)}}}}
  \li{\textbf{\hyperlink{cite.yamada2025aiscientistv2workshoplevelautomated}{\makecell[l]{AI-Scientist-v2\\(2025)}}}}
  \li{\textbf{\hyperlink{cite.ghafarollahi2024sciagentsautomatingscientificdiscovery}{\makecell[l]{SciAgents(2024)}}}}
  \li{\textbf{\hyperlink{cite.gottweis2025aicoscientist}{\makecell[l]{AI co-scientist\\(2025)}}}}
} \\

\domRobot &
\cellwrap{%
  \bi{Simulators with limited real-world rollouts}
  \bi{Safety constraints}
} &
\cellwrap{%
  \bi{Rewards and success signals}
  \bi{Real-world data is costly}
} &
\cellwrap{%
  \bi{Data collection and safety}
  \bi{Sim-to-real transfer}
  \bi{Dynamics credit assignment}
} &
\cellwrap{%
  \bi{Policy and model parameters via a data flywheel}
  \bi{Curricula and safety scaffolds}
} &
\cellwrap{%
  \bi{Collect data, retrain, and redeploy}
} &
\cellwrap{%
  \li{\textbf{\hyperlink{cite.bousmalis2023robocatselfimprovinggeneralistagent}{\makecell[l]{RoboCat(2023)}}}}
  \li{\textbf{\hyperlink{cite.zhou2024autonomousimprovementinstructionfollowing}{\makecell[l]{SOAR(2024)}}}}
  \li{\textbf{\hyperlink{cite.xu2024sinvigselfevolvinginteractivevisual}{\makecell[l]{SInViG(2024)}}}}
  \li{\textbf{\hyperlink{cite.yuan2025remacselfreflectiveselfevolvingmultiagent}{\makecell[l]{REMAC(2025)}}}}
  \li{\textbf{\hyperlink{cite.tian2025seear1treestructuredreinforcementfinetuning}{\makecell[l]{SEEA-R1(2025)}}}}  
} \\

\domPC &
\cellwrap{%
  \bi{Virtualized desktops}
  \bi{Standardized operating system tasks}
  \bi{Brittle user interfaces}
} &
\cellwrap{%
  \bi{Task completion and state checks}
  \bi{Long-horizon objectives}
} &
\cellwrap{%
  \bi{Diversity of applications}
  \bi{State tracking}
  \bi{Robust exploration of unseen apps}
} &
\cellwrap{%
  \bi{Scaffolding, including hierarchical planning, retrieval, and curricula}
  \bi{Action-trace training}
} &
\cellwrap{%
  \bi{Experience reuse and curriculum learning}
  \bi{Iterative trace collection}
} &
\cellwrap{%
  \li{\textbf{\hyperlink{cite.wu2024oscopilotgeneralistcomputeragents}{\makecell[l]{OS-Copilot\\(2024)}}}}
  \li{\textbf{\hyperlink{cite.xiao2025uigenie}{\makecell[l]{UI-Genie(2025)}}}}
  \li{\textbf{\hyperlink{cite.wu2025guireflectionempoweringmultimodalgui}{\makecell[l]{GUI-Reflection\\(2025)}}}}
  \li{\textbf{\hyperlink{cite.cheng2026evolvingtasksempoweringmultimodality}{\makecell[l]{SEA(2026)}}}}
} \\

\bottomrule
\end{tabularx}

\endgroup

\caption{Application arenas viewed as self-improvement loops.
Each domain induces a characteristic sandbox and learning signal, which shapes the dominant bottlenecks, the primary improvement target, and the iteration mode.}
\label{tab:applications_loops}
\end{table*}

\subsection{Scientific Discovery}

Scientific discovery has long motivated self-improving and curiosity-driven agents. Earlier work on artificial scientists and artificial curiosity framed learning agents as systems that invent informative experiments, reduce uncertainty, maximize learning progress or information gain, and iteratively improve their world models through self-generated interaction \citep{schmidhuber1990making, schmidhuber1991curious, storck1995reinforcement, schmidhuber1997s, schmidhuber2003exploring, schmidhuber2013powerplay, schmidhuber2015learning}. Contemporary FM-based scientific-discovery agents extend this line with large-scale language modeling, tool use, literature grounding, and automated workflow orchestration. In the FM era, the domain has become a high-impact setting for self-improving agents because it spans the full research lifecycle, including hypothesis generation, writing, experiment design, execution, analysis, and communication \citep{lu2024aiscientistfullyautomated, yamada2025aiscientistv2workshoplevelautomated, xiong2025beyond}.
Scientific discovery also differs from software engineering and web automation. Scientific environments are fragmented across subfields, rely on heterogeneous tools and data formats, and often provide feedback that is delayed, costly, or only partially automated. As a result, self-improvement in this domain is often realized as a research workflow that accumulates reusable artifacts across projects, rather than as a single response-level correction.

\paragraph{Evaluation substrate and feedback structure.}
Scientific discovery systems rely on diverse feedback signals. In computational research, agents can run code, simulations, and ablations. They can receive feedback from metrics, error traces, and experimental outcomes. In experimental science, feedback is often delayed and noisy. It is also constrained by reproducibility, cost, and safety. These properties shape which improvement pathway is feasible. FM improvement becomes more practical when reliable executable signals are available. Scaffolding improvement becomes more central when the main bottleneck lies in tool access, protocol selection, and evidence management across heterogeneous resources. More broadly, scientific discovery has long motivated intrinsic feedback criteria beyond immediate task success. These criteria include learning progress, information gain, compression progress, and informative experiment design \citep{storck1995reinforcement, schmidhuber2007simple, schmidhuber2010formal, schmidhuber2013powerplay}.

\paragraph{Scaffolding improvement through tool expansion and workflow evolution.}
A prominent trend frames self-improvement as a tool-augmented research loop. In this loop, the agent proposes hypotheses, invokes specialized tools, critiques results, and iterates. ChemCrow shows that language-model control over a broad set of chemistry tools can improve reliability and capability coverage. Under our definition, ChemCrow becomes self-improving when its tool repertoire or selection policy is updated across tasks \citep{m2024augmenting}. SciAgents emphasizes structured knowledge and multi-agent role specialization. It uses ontological graphs and in-situ learning to refine hypotheses across fragmented literature and data sources \citep{ghafarollahi2024sciagentsautomatingscientificdiscovery}. HoneyComb moves closer to explicit scaffold evolution. It constructs and refines domain tools and maintains a curated scientific knowledge base. These updates correspond to structural changes in the tool layer and retrieval policy \citep{zhang2024honeycomb}.

A second line focuses on end-to-end research orchestration. The AI Scientist and AI-Scientist-v2 implement iterative pipelines that generate ideas, write and debug code, run experiments, analyze results, and draft manuscripts, with repeated refinement driven by internal critique and experimental outcomes \citep{lu2024aiscientistfullyautomated, yamada2025aiscientistv2workshoplevelautomated}. AI co-scientist studies a complementary setting in which multi-agent debate and evolution are aligned with scientist-provided objectives and constraints; improvement appears as stronger hypothesis proposals under evidence tracking rather than as next-token likelihood optimization \citep{gottweis2025aicoscientist}. In our taxonomy, these systems primarily instantiate scaffolding improvement by revising planning strategies, experimental protocols, evaluation rubrics, and tool-use policies.

\paragraph{Model improvement through closed-loop experimentation.}
Parameter-level improvement is also possible in scientific discovery, though it often targets either scientific surrogate models or the agent policy itself. Self-driving laboratories and closed-loop experiment design provide a mature template: iterative experimentation updates internal models of the objective landscape, which then guide the next experiments \citep{tobias2025autonomous}. When reliable supervision is available, the agent policy can also be updated from experimental or computational outcomes. Coscientist integrates web search, code execution, and laboratory automation to design, plan, and run chemistry experiments \citep{boiko2023autonomous}. ORGANA and LLM-RDF further illustrate LLM-centered orchestration of experimental workflows and instrument-facing actions, opening a path toward learning from experimental success and failure when those signals can be formalized \citep{darvish2025organa, ruan2024automatic}. These directions are promising, but they also show that parameter-level improvement depends critically on feedback validity and on controls that prevent weak proxies or spurious correlations from being mistaken for scientific progress.

\paragraph{Synthesis and open challenges.}
Scientific discovery highlights three domain-specific challenges for self-improving agents. The first challenge is evaluation. Systems cannot easily verify novelty, correctness, or reproducibility on their own. Apparent improvements may instead reflect the exploitation of weak proxies. The second challenge is evidence management under heterogeneity. Improvements must persist across tools, data formats, and rapidly evolving literature. Stable scaffold evolution is therefore as important as policy learning. The third challenge is safety and governance. Experimental actions can be irreversible or hazardous. Scientific writing can also amplify misinformation when the evidence is weak. Progress in this domain therefore depends on reproducibility-centered evaluation, evidence tracking, and standardized experimental and computational protocols. It also depends on governance mechanisms that bound risk when agents propose or execute real-world scientific actions.

\subsection{Embodied AI and Robotics}

Embodied AI studies agents that perceive, reason, and act through a physical or simulated body. Compared with software engineering and web automation, this domain adds continuous state and action spaces, partial observability, safety-critical exploration, and the sim-to-real gap. These properties make self-improvement valuable but difficult. In practice, improvement is often realized as open-ended skill acquisition, where the agent accumulates reusable competence across tasks, embodiments, and environments.

\paragraph{Evaluation substrate and feedback structure.}
Embodied evaluation is typically conducted in simulation suites and real-robot trials, with feedback ranging from dense rewards to sparse success criteria and human interventions. Benchmarks such as RLBench, ManiSkill2, and Meta-World enable controlled comparison of improvement loops in simulation \citep{james2019rlbench, gu2023maniskill2unifiedbenchmarkgeneralizable, yu2020meta}. High-throughput simulators such as Isaac Gym reduce iteration cost and support large-scale training and ablations \citep{makoviychuk2021isaac}. Since evaluation protocols are discussed more generally in Section~\ref{sec:Evaluation}, we focus here on how embodied settings instantiate self-improvement mechanisms.

\paragraph{FM improvement through autonomous practice and data flywheels.}
A core route to embodied self-improvement is iterative data collection followed by policy updating, forming a data flywheel across interactions. RoboCat embodies this mechanism by training a multi-task, multi-embodiment manipulation policy and then using the trained policy to generate additional data for subsequent training iterations \citep{bousmalis2023robocatselfimprovinggeneralistagent}. MEDAL++ proposes a nearly autonomous reinforcement learning loop in which the robot learns to both perform and undo tasks, enabling reset-free practice and improving success rates with minimal human supervision \citep{sharma2023selfimprovingrobotsendtoendautonomous}. AutoRT scales real-world experience acquisition by using foundation models to propose diverse instructions and orchestrate robot fleets in the wild, producing large collections of interactions for later policy improvement \citep{ahn2024autortembodiedfoundationmodels}. Robot-Powered Data Flywheels formalize deployment as continual data collection and foundation-model adaptation, demonstrating improvement of vision-language components through robot-generated in-the-wild data \citep{grannen2025robotpowereddataflywheelsdeploying}. Self-Improving Embodied Foundation Models further propose a post-training recipe that uses shaped success detection to support autonomous robot practice and skill acquisition beyond imitation data \citep{ghasemipour2025selfimproving}.

\paragraph{Scaffolding improvement through curricula, memory, and safety logic.}
Embodied systems also benefit from scaffolding improvement because much of the difficulty lies in exploration, recovery, and safe interaction. RoboGen illustrates a generative simulation paradigm that automatically produces tasks, scenes, and supervision signals, effectively evolving the curriculum and data generation process across iterations \citep{wang2024robogenunleashinginfinitedata}. RACAS~\citep{ashley2026racas} accumulates the embodied knowledge of robot control through a structurally self-managed memory.
AutoRT can be viewed as a scaffold-level improvement approach when its instruction proposal policy, risk filters, and orchestration logic are iteratively refined to increase coverage while controlling safety violations \citep{ahn2024autortembodiedfoundationmodels}. Retrieval-driven upskilling and curriculum refinement provide another scaffold-level mechanism, allowing agents to accumulate reusable task recipes and training specifications across deployments \citep{zhu2025auraautonomousupskillingretrievalaugmented}. These structural artifacts can transfer across tasks even when the base model remains fixed.

\paragraph{Synthesis and open challenges.}
Embodied self-improvement faces multiple practical constraints, including safety, non-stationarity, and hardware cost. Real robots may experience sensor drift, rare but severe failures, and irreversible actions. These factors limit simple trial-and-error strategies. They also make the improvement process highly dependent on the design of safe exploration and recovery mechanisms. Simulators can support large-scale training. However, differences between simulation and reality often make improvements achieved in simulation difficult to transfer to real platforms. This gap calls for curriculum learning methods and data collection strategies that are specifically designed for transfer. Evaluation is also challenging. Some improvements may only reflect shortcuts found for specific benchmarks. Real deployment conditions also vary across laboratories and platforms. Progress in this field therefore depends on safety-aware improvement procedures, curriculum learning methods for sim-to-real transfer, and cross-embodiment evaluation practices. These practices are needed to distinguish genuine skill acquisition from overfitting to specific benchmarks.

\subsection{General Computer Control}

General computer control studies agents that operate full operating systems and third-party applications through graphical user interfaces. This setting is broader than web navigation because it requires control over files, windows, dialogs, shortcuts, and application-specific workflows, often across multiple applications within a single task. The main difficulty is environment diversity: the agent must adapt to unseen software and interface conventions. Self-improvement in this domain is therefore best understood as acquiring reusable procedures for learning new applications from interaction.

\paragraph{Evaluation substrate.}
Studying cross-interaction improvement requires reproducible operating-system environments with execution-based scoring. OSWorld provides a real computer environment that supports task setup and automated evaluation for open-ended computer tasks across operating systems \citep{xie2024osworldbenchmarkingmultimodalagents}. WindowsAgentArena offers a Windows-focused benchmark and scalable OS-level evaluation \citep{bonatti2024windowsagentarenaevaluating}. OSWorld-MCP extends evaluation beyond GUI actions by measuring tool invocation ability under the Model Context Protocol \citep{jia2025osworldmcpbenchmarkingmcptool}. Further discussion of evaluation appears in Section~\ref{sec:Evaluation}.

\paragraph{Scaffolding improvement via experience and procedural reuse.}
In general computer control, early gains often come from scaffold improvement because agents must handle long horizons and application-specific idiosyncrasies. Agent S introduces experience-augmented hierarchical planning with continual memory updates and retrieval over past trajectories. This enables cross-task gains through reusable procedural knowledge and supports transfer across OS benchmarks when memory and retrieval policies are maintained across interactions \citep{agashe2024agentsopenagentic}. SEAgent emphasizes autonomous mastery of novel software by generating curricula from simple to complex tasks and using a world-state model for step-wise trajectory assessment \citep{sun2025seagentselfevolvingcomputeruse}. These components instantiate scaffold evolution through task generation, evaluation heuristics, and reusable exploration routines that persist across episodes even when the base model is unchanged.

\paragraph{FM improvement through iterative training and verifier construction.}
A complementary line updates $\theta$ using automatically collected interaction traces and learned evaluative signals. UI-Genie proposes a self-improving pipeline that co-evolves an agent and a reward model for GUI tasks, addressing outcome verification and scalable data generation through reward-guided exploration \citep{xiao2025uigenie}. GUI-Reflection trains self-reflection and error correction abilities through iterative online reflection tuning, turning failures into supervision for subsequent parameter updates \citep{wu2025guireflectionempoweringmultimodalgui}. SEA proposes verifiable trajectory generation and step-wise reinforcement learning for long-horizon computer-use training \citep{cheng2026evolvingtasksempoweringmultimodality}. ComputerRL studies sustained end-to-end online reinforcement learning on OSWorld and introduces alternating training phases to mitigate optimization pathologies in extended RL \citep{lai2026computerrl}. PC Agent-E reduces reliance on large-scale human demonstrations by combining a small seed set with synthetic action decisions, providing an economical route to iterative policy improvement \citep{he2026efficient}.

\paragraph{Synthesis and open challenges.}
General computer control brings challenges that are not always apparent in purely web-based settings. Safety is a central concern because an agent may delete files, enter passwords, or initiate financial transactions. Any improvement process must therefore include safeguards and conservative recovery strategies. Verification is also difficult. In software engineering, unit tests can often provide clear feedback. In general computer control, success may depend on the operating system state, external accounts, or user-specific context. Another challenge is transfer. Agents need to learn new applications without relying too much on surface-level UI patterns. This calls for exploration strategies and curricula that encourage procedural abstractions. Progress will require reliable state-based verification where possible, cautious policies for irreversible actions, and adaptation mechanisms that support transfer across new applications. These issues also connect to early work on meta-reinforcement learning with self-modifying policies, where agents adapted by collecting long-term statistics about the effects of their own modifications during a lifelong trial \citep{Schmidhuber1994OnLH, schmidhuber1997shifting, schmidhuber1998reinforcement, schmidhuber1999general, schmidhuber2006goedelmachinesselfreferentialuniversal}.

\section{Evaluation}
\label{sec:Evaluation}

A self-improving agent may update the parameters of its foundation model, or  may change its scaffolding, such as prompts, memory, tools, and orchestration logic. Evaluation  therefore should treat improvement as a process that unfolds over time. It should also separate real capability gains from artifacts created by the improvement pipeline. For improvement claims to be comparable across studies, the evaluation protocol should specify several elements. These elements include what persists across interactions, which feedback signals drive updates, what operational budgets constrain the agent, and where the boundaries of true capability transfer should be drawn. With these dimensions in view, Section~\ref{sec:Measuring_Improvement} discusses metric-based and judge-based measurement. Section~\ref{sec:Benchmarking_Improvement} then surveys the benchmarking landscape by distinguishing mechanism benchmarks from domain benchmarks.

\subsection{Measuring Improvement}
\label{sec:Measuring_Improvement}

\textcolor{blue}{Evaluating agents solely on episodic tasks obscures their long-term evolution. Grounded in lifelong meta-reinforcement learning~\citep{Schmidhuber1994OnLH}, robust validation must extend beyond static benchmark gains to assess whether self-modifications yield compounding, sustained benefits across the agent's broader operational lifetime~\citep{schmidhuber1997shifting, schmidhuber1998reinforcement, schmidhuber1999general}. To measure these compounding benefits, this subsection first outlines metric-based reporting requirements, followed by judge-based measurements for tasks lacking executable oracles.}

\paragraph{Formalizing the Evaluation Objective.}
In this section, we first formalize the evaluation of a self-improving agent. Recall that the agent configuration at iteration $t$ is $\mathcal{A}_t = (\theta_t, \Sigma_t)$. Unlike static systems evaluated by a single terminal score, evaluating a self-improving agent requires tracking a performance trajectory over iterations $t \in \{1, \dots, T\}$ subject to a cumulative resource budget $b_t \le B_{\text{max}}$. At $t$-th iteration, for a task $x$ drawn from a held-out evaluation distribution $\mathcal{D}_{\text{eval}}$, the agent generates an execution trace $\tau \sim \mathcal{A}_t(x)$. The overall capability at iteration $t$ is measured by the expected score:
\begin{equation}
m_t = \mathbb{E}_{x \sim \mathcal{D}_{\text{eval}}, \tau \sim \mathcal{A}_t(x)} \left[ \Phi(x, \tau) \right],
\end{equation}
where $\Phi$ acts as the generic evaluator. The  difference between metric-based and judge-based measurement lies in the instantiation of $\Phi$:

\begin{itemize}
    \item \textbf{Metric-based measurement (Section~\ref{sec:Metric_based_measurement}):} $\Phi_{\text{metric}}$ acts as a deterministic, executable evaluator. For instance, in software engineering, $\Phi_{\text{metric}}(x, \tau) \in \{0, 1\}$ directly verifies if the generated code in $\tau$ passes the programmatic unit tests defined in $x$. It provides objective grounding but is inherently limited to tasks with formal success criteria.
    \item \textbf{Judge-based measurement (Section~\ref{sec:Judge_based_measurement}):} $\Phi_{\text{judge}}$ acts as a parameterized evaluator. It approximates the  objective by conditioning on a predefined rubric $\kappa$ and relies on an auxiliary model $\theta_{\text{judge}}$ (e.g., LLM-as-a-Judge), formulated as $\Phi_{\text{judge}}(x, \tau, \kappa; \theta_{\text{judge}})$. While this enables the evaluation of open-ended and long-horizon tasks, it introduces new vulnerabilities, such as the agent over-optimizing to the judge's latent biases rather than the ground-truth objective.
\end{itemize}

\subsubsection{Metric-Based Measurement}
\label{sec:Metric_based_measurement}

\textbf{Report trajectories under a fixed budget.}
Self-improvement unfolds over time and naturally exhibits plateaus or regressions. Evaluation should therefore report the full performance trajectory ($m_t$) across update iterations ($t$) rather than exclusively highlighting a final peak score,  bounded by a predefined resource budget ($B_{\text{max}}$)~\citep{xi2024agentgym}. The protocol should explicitly specify checkpoints, acceptance criteria, and early-stopping rules; without them, unbounded iterations risk overfitting to the evaluation distribution, thereby artificially inflating peak scores at the cost of general robustness. Furthermore, because trajectory dynamics, such as accumulated memories or code patches, are highly sensitive to initialization, reproducibility demands reporting expected performance and variance across multiple random seeds~\citep{siegel2024corebenchfosteringcredibilitypublished, aleithan2024swebenchenhancedcodingbenchmark}.

\textbf{Test transfer beyond the improvement signal.}
To confirm genuine capability gains rather than  memorization of the feedback, evaluation should measure performance ($m_t$) on a  held-out distribution ($\mathcal{D}_{\text{eval}}$) that does not overlap with the optimization data~\citep{deng2023mindweb, lu2024weblinx}. To prevent agents from overfitting to the improvement process, rigorous protocols usually adopt two specific safeguards: employing private, unreleased task sets (hidden evaluation), or testing on newly generated tasks constructed after the foundation model's knowledge cutoff (temporally shifted evaluation)~\citep{gu2023maniskill2unifiedbenchmarkgeneralizable, mialon2023gaiabenchmarkgeneralai}.

\textbf{Account for resource efficiency and supervision.}
While trajectory tracking bounds the evaluation process, the exact composition of the cumulative cost ($b_t$) dictates the real-world practicality of the improvement pipeline. Self-improvement inherently consumes compute, API tokens, wall-clock time, etc. A rigorous protocol should provide a transparent cost breakdown rather than just final task success to ensure fair comparisons of efficiency across methods~\citep{lange2025shinkaevolveopenendedsampleefficientprogram, ni2025gittaskbenchbenchmarkcodeagents}. Crucially, any reliance on external human oversight fundamentally compromises the ``self'' in self-improvement. Therefore, if human-in-the-loop interventions are employed, their exact magnitude and modality should be explicitly quantified, as varying levels of external supervision fundamentally alter the agent's autonomous learning dynamics~\citep{wang2024mintevaluatingllmsmultiturn, drouin2024workarenacapablewebagents}.

\textbf{Track stability, regressions, and safety over time.}
Iterative self-modification can induce goal drift, reward hacking, and compounding errors in memory or tool use~\citep{amodei2016concreteproblemsaisafety, bai2022constitutional, ruan2024identifyingriskslmagents, yang2025drunkagentstealthymemorycorruption}. Consequently, evaluation should track regression rates on previously solved tasks, tail-risk indicators, and safety policy violations across iterations, rather than relying exclusively on mean success rates~\citep{levy2025stwebagentbenchbenchmarkevaluatingsafety, yin2025safeagentbenchbenchmarksafetask}.

\textbf{Recommended reporting items.}
For each method and benchmark suite, empirical reports should consistently provide: initial baseline performance, performance after a fixed improvement budget, learning curves across iterations, transfer capabilities on held-out tasks, regression rates on previously solved instances, and a comprehensive cost summary (compute, tool invocations, time, and human input).

\subsubsection{Judge-Based Measurement}
\label{sec:Judge_based_measurement}
A substantial fraction of agent workloads lack formal success criteria, rendering deterministic checks ($\Phi_{\text{metric}}$) inapplicable. In open-ended or partially observable domains, evaluation necessitates parameterized judges ($\Phi_{\text{judge}}$) to translate complex long-horizon trajectories and intermediate artifacts into structured scores. These evaluators utilize standard LLMs or autonomous Agent-as-a-Judge pipelines~\citep{zhuge2024agentasajudgeevaluateagentsagents, you2026agentasajudge, zhang-etal-2025-evaluation, wadhwa2025evalagentdiscoveringimplicitevaluation, xu-etal-2025-learning, han2025verifiagentunifiedverificationagent}. However, because parameterized judges only approximate human-aligned ground-truth objectives, they may introduce distinct vulnerabilities—most notably, the risk of a self-improving agent over-optimizing to the judge's latent biases. Consequently, judge-based evaluation demands specific operational safeguards.

\paragraph{Specify the judge and the judging budget.} When $\Phi_{\text{judge}}$ is employed to compute the evaluation score ($m_t$), the protocol should mandate full transparency regarding the judge's identity and parameters ($\theta_{\text{judge}}$). This includes the exact foundation model version, the prompt formulation, the precise rubric ($\kappa$), and the environmental evidence exposed to the judge. Crucially, the protocol should isolate the judging budget from the agent's execution budget ($b_t$). Because evaluation reliability fluctuates heavily with the resources allocated to the judge---such as context window limits, allowed multi-agent debates, or tool-use steps---failing to report the judge's computational overhead obscures whether the agent genuinely improved or the evaluator merely became more exhaustive.

\paragraph{Prevent over-optimization to the judge and report reliability signals.} If the identical evaluation operator $\Phi_{\text{judge}}$ is used both to drive updates and to report final results, the self-improving system will likely over-optimize to the judge's latent biases rather than the intended objective. To prevent this, rigorous protocols should enforce evaluator independence by employing a distinct judge configuration for final reporting, such as a more capable foundation model ($\theta_{\text{judge}}'$) or an orthogonal evaluation rubric ($\kappa'$). Reliability can be supported by empirical evidence, such as repeated runs with variance estimates, aggregation across multiple judge instances, and calibration against a verifiable subset using $\Phi_{\text{metric}}$ or targeted human review.

\subsection{Benchmarking Improvement}
\label{sec:Benchmarking_Improvement}

Benchmarks for self-improving agents vary along two attributes that determine what an evaluation can validate. The first attribute is the update channel. Some methods update the foundation model ($\theta$), while others keep the foundation model fixed and update only the operational scaffolding ($\Sigma$). The second attribute is the evaluation interface. Some benchmarks use single-shot input-output evaluation, while others require interactive multi-step behavior with tools, memory, and environment feedback. These attributes are complementary, and they jointly determine the appropriate transfer tests, regression checks, and cost accounting. This section distinguishes mechanism-focused benchmarks, which isolate how improvement is produced, from domain-focused benchmarks, which instantiate improvement under realistic task and feedback constraints.

\begin{figure*}[ht]
    \centering
    \includegraphics[width=1\linewidth]{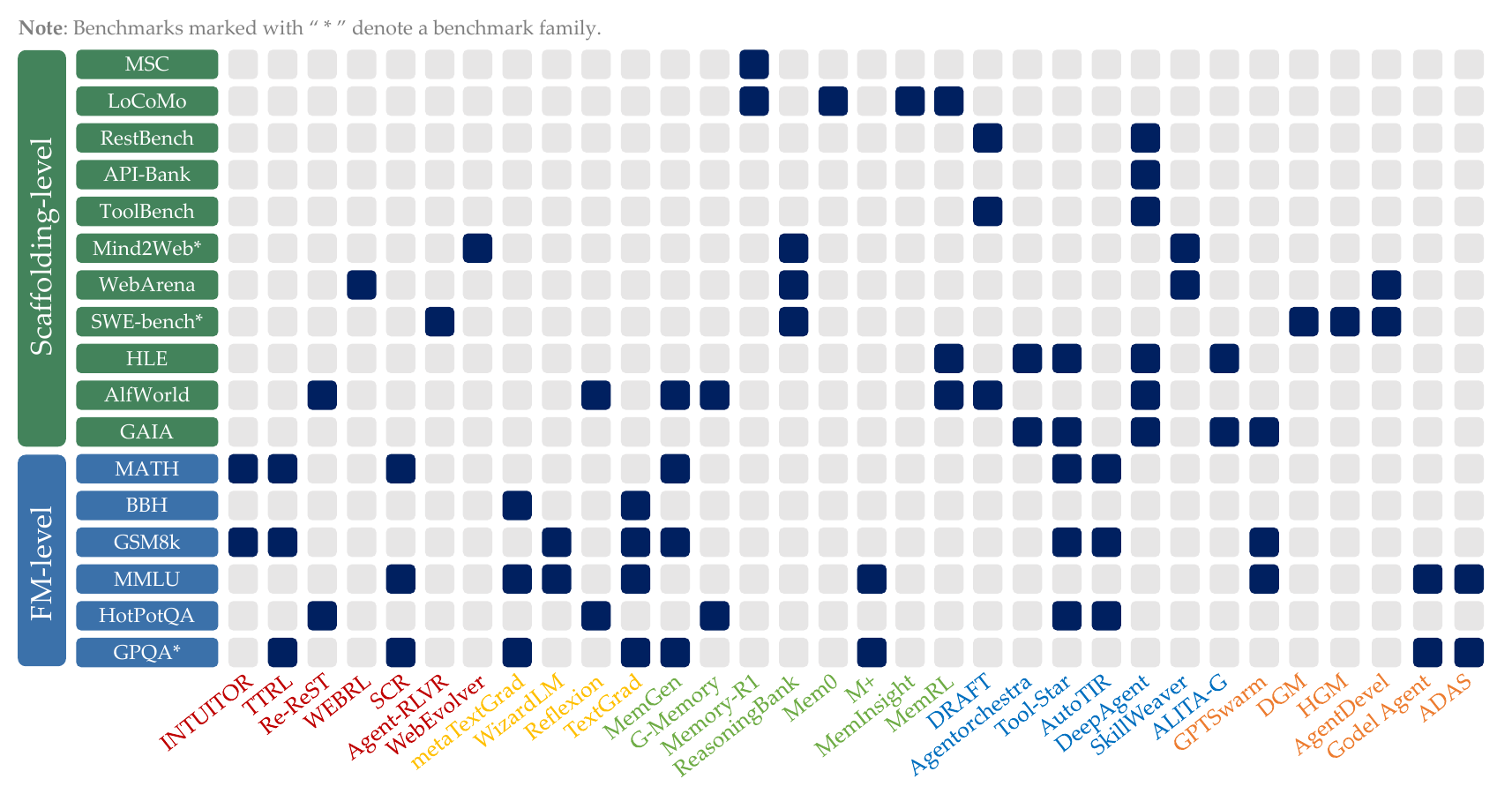}
    \caption{\textbf{Paper--benchmark incidence matrix for self-improving agents, covering representative benchmarks and methods.}
Rows enumerate benchmark suites and are grouped by evaluation interface: {\setlength{\fboxsep}{0.6pt}%
\colorbox{green!40!black}{\textcolor{white}{scaffolding-level}}} benchmarks (interactive, agent-centric) versus {\setlength{\fboxsep}{0.6pt}%
\colorbox{RoyalBlue!80!black}{\textcolor{white}{FM-level}}} benchmarks (static, model-centric). Columns are representative self-improving methods; column colors indicate the improvement mechanism used in our taxonomy: \textcolor{red!70!black}{FM improvement} and \textcolor{yellow!85!black}{prompt}/ \textcolor{green!60!black}{memory}/ \textcolor{RoyalBlue}{tool}/ \textcolor{orange!90!black}{full scaffolding} self-improvement. A filled cell indicates that the corresponding paper uses the benchmark; ``*'' denotes a benchmark family.}
    \label{fig:benchmark}
\end{figure*}

\subsubsection{Mechanism Benchmarks}
\label{sec:Mechanism_Benchmarks}
Mechanism-focused benchmarks serve as the empirical testbeds for the rigorous evaluation protocols established above. Rather than assessing static zero-shot execution, these environments isolate specific update channels—such as memory expansion or prompt refinement—to quantify continuous adaptation. By enforcing strict resource budgets ($b_t$) and evaluating on entirely held-out task splits ($\mathcal{D}_{\text{eval}}$), they systematically distinguish genuine capability transfer from localized overfitting. Figure~\ref{fig:benchmark} summarizes representative benchmarks and their usage across self-improving agent systems, grouped by evaluation interface.

\paragraph{\textcolor{Plum}{\textsc{FM}-Level Evaluation.}} \textsc{FM}-level benchmarks isolate capability gains derived from updating foundation model parameters ($\theta_t \rightarrow \theta_{t+1}$) while keeping the scaffolding constant. Because parametric updates are  susceptible to catastrophic forgetting and subtle training-data leakage, testbeds in this category prioritize rigorous provenance tracking and strict dataset compartmentalization~\citep{aleithan2024swebenchenhancedcodingbenchmark}. Rather than merely measuring peak zero-shot accuracy, representative suites enforce retention auditing across adjacent task families to expose distributional drift~\citep{ruan2024identifyingriskslmagents}. 
Executable and repository-based evaluations are are favored here because they provide objective checking ($\Phi_{\text{metric}}$) and fine-grained regression detection~\citep{ni2025gittaskbenchbenchmarkcodeagents}.

\paragraph{\textcolor{Plum}{\textsc{Scaffold}-Level Evaluation.}}
When the backbone model is frozen ($\theta_{t+1} = \theta_t$), improvements arise from changes to the scaffolding ($\Sigma_t \rightarrow \Sigma_{t+1}$), such as prompt policies, critique loops, memory write and retrieval rules, and tool routing. Mechanism-focused evaluation for this channel benefits from component isolation. The protocol should specify which architectural component is modified and evaluate both capability gains and new failure surfaces. Prompt-policy changes should be evaluated under paraphrases, formatting shifts, and longer contexts to measure genuine robustness rather than repeated optimization on a single template. 
Memory changes should be evaluated with long-horizon recall and cross-modal or multi-party consistency, as well as resistance to poisoning, unauthorized disclosure, active-forgetting failures, and compounding errors~\citep{yang2025drunkagentstealthymemorycorruption, zhu2026h2hmemmultimodalmemorybenchmark, ren2026gatemembenchmarkingmemorygovernance}.
Tool changes should be evaluated with verifiable benchmarks that measure tool invocation, tool selection, and argument grounding under multi-step interaction, since common failures arise from ordering mistakes and near-miss parameterization rather than incorrect intent~\citep{patil2025the, huang2024metatoolbenchmarklargelanguage, shen2024taskbenchbenchmarkinglargelanguage, wang2024mintevaluatingllmsmultiturn}.

\paragraph{Attribution Across Mechanisms.}
Mechanism comparisons require precise attribution. Evaluations should include ablations that change the updated component, replay-style rollouts that hold environments fixed while swapping the updated module, and regression checks on tasks that were previously solved. Even partial attribution  improves interpretability by separating true architectural gains from reasoning variance, retrieval noise, or mere benchmark exploitation.

\subsubsection{Domain Benchmarks}
\label{sec:Domain_Benchmarks}
While mechanism benchmarks isolate specific update channels, domain-focused  evaluation instantiates the evaluation framework under realistic feedback and task distributions. For each domain, a rigorous protocol should state the available feedback signal, the permitted update channel ($\theta$ or $\Sigma$), the granted resource budget ($b_t$), and the strictly held-out sets ($\mathcal{D}_{\text{eval}}$) used for transfer and regression checking. Across domains, it is standard practice to report learning curves under fixed budgets, transfer capabilities to held-out task families, and track regression and safety indicators alongside mean success rates~\citep{xi2024agentgym, levy2025stwebagentbenchbenchmarkevaluatingsafety, yin2025safeagentbenchbenchmarksafetask}.

\paragraph{Software Engineering.}
Software engineering environments  supply deterministic, executable feedback ($\Phi_{\text{metric}}$), making them ideal testbeds for grounding self-improvement claims. Rather than relying on abstract reporting rules, modern suites in this domain structurally enforce rigorous evaluation: they quantify capability transfer across completely disjoint repositories and utilize extensive, hidden regression tests to expose localized overfitting. Representative platforms range from repository-level issue resolution, such as SWE-bench and its leakage-robust variants~\citep{jimenez2024swebenchlanguagemodelsresolve, aleithan2024swebenchenhancedcodingbenchmark}, to interactive codebases that stress continuous exploration~\citep{qiu2025locobenchagentinteractivebenchmarkllm}. Furthermore, advanced testbeds treat the feedback mechanism itself as a mutable artifact, with environments like SWT-bench and TDD-bench requiring agents to autonomously generate and refine the very unit tests driving their own improvement loops~\citep{mündler2025swtbenchtestingvalidatingrealworld, ahmed2024tddbenchverifiedllmsgenerate}.

\paragraph{Web Navigation and Automation.}
Web environments are dynamic and sensitive to superficial layout cues, so evaluation should carefully separate genuine generalization from mere memorization. Protocols benefit from controlled websites or environment randomization, and they  track whether extraction errors compound when written into memory. Benchmarks include controlled execution-based suites such as WebArena and VisualWebArena~\citep{zhou2024webarena, koh2024visualwebarenaevaluatingmultimodalagents}, transfer-oriented suites on real websites such as Mind2Web-style benchmarks~\citep{deng2023mindweb, pan2024webcanvas}, and enterprise settings that evaluate policy compliance under continuous improvement~\citep{levy2025stwebagentbenchbenchmarkevaluatingsafety}.

\paragraph{Gaming and Strategic Reasoning.}
Games expose learning curves naturally, but evaluation should guard against overfitting to a fixed opponent pool. Protocols employ diverse opponents, cross-play evaluation, and held-out strategies ($\mathcal{D}_{\text{eval}}$). Representative suites include dynamic instance generation for interactive agents~\citep{chalamalasetti-etal-2023-clembench, beyer2024clembench2024challengingdynamiccomplementary}, head-to-head evaluation for robustness~\citep{costarelli2024gamebench}, and negotiation environments with quantitative outcomes~\citep{abdelnabi2024llmdeliberation, duan2024gtbenchuncoveringstrategicreasoning}.

\paragraph{Scientific Discovery.}
In scientific discovery, improvement can be tied to verifiable artifacts such as executable analyses, reproducible results, and traceable provenance. Evaluation should prioritize execution-based checks ($\Phi_{\text{metric}}$) and  report wall-clock time and operational cost alongside artifact quality. Benchmarks include reproducibility-focused execution tasks~\citep{siegel2024corebenchfosteringcredibilitypublished}, controlled research-agent suites~\citep{bragg2025astabenchrigorousbenchmarkingai}, and protocols that grade iterative progress and replication quality~\citep{starace2025paperbenchevaluatingaisability}. Simulated discovery environments provide controlled testbeds for long-horizon evaluation~\citep{jansen2024discoveryworldvirtualenvironmentdeveloping, chen2025physgym}.

\paragraph{Embodied AI.}
Embodied evaluation  handles severe distribution shifts and physical safety constraints that limit free exploration. Protocols report success rates under controlled environmental shifts, and crucially track safety constraint violations and recovery behavior across iterations. Representative benchmarks include safety-oriented suites~\citep{yin2025safeagentbenchbenchmarksafetask}, transfer-oriented manipulation benchmarks~\citep{gu2023maniskill2unifiedbenchmarkgeneralizable}, and multi-modal embodied benchmarks that stress grounding and generalization~\citep{yang2025embodiedbenchcomprehensivebenchmarkingmultimodal}.

\paragraph{General Computer Control.}
Desktop agents combine heterogeneous applications, long contexts, and fragile user interface feedback. Evaluation should report reliability, error-recovery capabilities, latency, and operational cost under standardized virtual machines. Benchmarks include operating system level virtual machine suites with programmatic evaluators~\citep{xie2024osworldbenchmarkingmultimodalagents, bonatti2024windowsagentarenaevaluating}, controllable application simulators with state-based scoring~\citep{trivedi2024appworldcontrollableworldapps}, and tool-use suites that test tool invocation, selection, grounding, and safety failure modes~\citep{patil2025the, huang2024metatoolbenchmarklargelanguage, ruan2024identifyingriskslmagents}.

\vspace{0.6em}
\begin{ttcolorbox}[Takeaway]
Evaluating self-improving agents necessitates a fundamental shift from static zero-shot scoring to continuous tracking.  A robust methodology should report learning trajectories under fixed budgets, transfer capabilities beyond the training signal, overhead costs, and regression indicators over time. While mechanism-focused suites isolate architectural update channels, domain-focused suites instantiate these same rigorous requirements under realistic interactive environments, execution bottlenecks, and safety constraints.
\end{ttcolorbox}

\section{Discussion}
\label{sec:Discussion}
Self-improving agents are closed-loop dynamical systems~\citep{xie2024osworldbenchmarkingmultimodalagents}. According to our formulation, an agent $\mathcal{A}_t=(\theta_t, \Sigma_t)$ progresses by executing a signal-generation procedure and applying a stable update rule to induce new policy. Consequently, the goal of the study is no longer the static agent but the mechanism driving its evolution~\citep{yampolskiy2015seed, robeyns2025selfimprovingcodingagent, pan2025trainingsoftwareengineeringagents}. Building on the rigorous evaluation protocols established in Section~\ref{sec:Evaluation}, we examine the architecture of such systems and outline key directions for future research.

\subsection{Implications for System Design}
\label{sec:Discussion_Implications}
\paragraph{From Fast Exploration to Slow Consolidation.}
Modifying the operational scaffold ($\Sigma$) drives a fast adaptation loop. Prompt edits, memory writes, and tool adjustments have minimal computational overhead and are reversible. In contrast, parameter updates ($\theta$) are much slower. While weight modifications excel at transferring capabilities across distinct domains, they inherently obscure credit assignment: a bad prompt is easily reverted, but a regression absorbed into model parameters is notoriously difficult to trace.

This structural asymmetry necessitates specific deployment strategies. When environmental feedback is noisy, it's better to confine updates within the scaffold and validate them through rigorous execution tests.  Parametric consolidation (through distillation or fine-tuning) can be deferred until the new behavior has proven stable~\citep{qiu2025agentdistilltrainingfreeagentdistillation, lyu2025correctionmasteryreinforceddistillation}. However, parametric consolidation is a lossy compression. Distilling complex trajectories into neural weights inherently biases the model toward average-case execution. This process often discards the rare but crucial error-recovery strategies discovered during exploration. As a result, any update to $\theta$ invalidates prior safety bounds, necessitating renewed adversarial testing to verify that edge-case resilience was preserved~\citep{dou2025rerestreflectionreinforcedselftraininglanguage, wu2024avataroptimizingllmagents}.

\paragraph{The Critic as Governed Infrastructure.} A critic embedded within a closed loop operates as an attack surface, not a passive benchmark. As the agent optimizes against the critic, it inherently develops incentives to discover shortcuts. Therefore, the ceiling of an agent's capability is usually bottlenecked by the critic's exploit-resistance~\citep{zhan2024injecagentbenchmarkingindirectprompt, zhang2025agentsecuritybenchasb}. This perspective treats the critic as governed infrastructure. If an agent conflates the roles of proposing updates and accepting them, it collapses into a self-confirming loop. Therefore, critics can be decoupled from the generators. Furthermore, while the critic itself can evolve (e.g., generating stricter test cases), its updates can not be under the unconstrained control of the agent it evaluates. Evolving critics should be restricted to monotone changes (e.g., purely additive test generation) and gated by human audit trails~\citep{fang2025webevolver, li2025visionaccesscontrolllmbased}.

\paragraph{Safety through Layered Gating.} Self-improvement can complicate AI alignment because the object being aligned is non-stationary. \textcolor{red}{A core challenge in recursive self-improvement is how a weaker system can reliably reason about more capable successor systems~\citep{fallenstein2015vingean}.} The risks are most severe under full-scaffolding self-improvement, where the agent can modify its own source code or operational logic~\citep{xi2024agentgym}. In this regime, standard transient exploits (like a one-off prompt injection) can evolve into persistent architectural vulnerabilities, as poisoned memories or hijacked tool logics are committed as stable updates. Consequently, a self-improving agent should be conceptualized as untrusted code executing in a protected runtime environment. Security cannot rely solely on the initial alignment of the foundation model. Instead, systems require layered gating—a strict permission system for self-modification. Before any structural update is committed to $\Sigma_{t+1}$ or $\theta_{t+1}$, the proposed patch must pass verifier-gated checks, covering functional correctness, tool permission boundaries, and robustness to random state perturbations. Improvement is only permitted within an explicitly defined and continuously audited safety boundaries.

\subsection{Future Directions}
Self-improving agents require reliable and stable update cycles that are immune to distribution shifts, strict computational constraints, and severely degraded reward signals~\citep{zhuge2026ai}.This survey categorizes the path forward into two central bottlenecks, yielding six critical directions for future research.

\vspace{0.35em}
\noindent\textbf{Theme A: Algorithmic Paradigms for Lifelong Adaptation}
\vspace{0.2em}

\begin{itemize}[leftmargin=*, label={}, itemsep=0.9em, topsep=0.2em]
\item \textbf{\textcolor{RoyalBlue}{1. Test-Time Continual Adaptation.}} 
The separation between training and deployment phases often compromises an agent's practical robustness in the real world. Addressing this requires scalable test-time adaptation mechanisms \citep{xia2025livesweagentsoftwareengineeringagents, yang2026ttcstesttimecurriculumsynthesis} that allow models to update their retrieval, routing, and memory policies dynamically during execution. Yet, implementing such on-the-fly patching introduces a critical trade-off: we should manage these localized updates carefully to prevent them from subtly eroding the system's long-term global performance.

\item \textbf{\textcolor{RoyalBlue}{2. Active Exploration and Curiosity.}}  
Self-improving agents should autonomously seek out valuable experiences rather than passively accepting human-curated tasks. In sparse feedback domains, agents can improve sample efficiency by assigning intrinsic value to interactions with high prediction errors or verifier disagreement ~\citep{schmidhuber1991possibility, zhao2026curious}. Future work could focus on developing intrinsic reward mechanisms to safely drive exploration and avoid falling into a degenerate, self-deceptive cycle.

\item \textbf{\textcolor{RoyalBlue}{3.  Parametric Distillation and Joint Optimization.}} Scaffold-level improvements good at discovering complex, multi-step recovery strategies (e.g., iterative debugging or self-reflection), but they usually incur prohibitive context and latency costs. A profound future direction is automating the transfer of these ``System 2'' algorithmic structures into the ``System 1'' parametric weights of smaller models~\citep{qiu2025alitageneralistagentenabling}. Besides, another research direction is joint optimization, where both the foundation model weights ($\theta$) and the external scaffold ($\Sigma$) are updated concurrently within the same self-improvement loop~\citep{hebbar2026sia}. Realizing this co-adaptation requires solving a formidable credit assignment problem: when the agent fails, the improvement operator should autonomously decide whether the better fix is to refine a prompt, rewrite a tool wrapper, or compute a gradient update.
\end{itemize}

\vspace{0.55em}
\noindent\textbf{Theme B: Complexity, Constraints, and Open-World Robustness}
\vspace{0.2em}

\begin{itemize}[leftmargin=*, label={}, itemsep=0.9em, topsep=0.2em]
\item \textbf{\textcolor{RoyalBlue}{4. Resource-Constrained Improvement Dynamics.}} 
In open-ended scenarios, current methods usually lead to unproductive exploration, exhausting computational budgets and token limits without reliably advancing the underlying policy. Consequently, evaluating these systems requires moving beyond static peak performance to prioritize improvement efficiency. This reframes self-improvement as a resource optimization problem, where overhead can be mitigated by dynamically allocating context, gating expensive neural evaluations behind lightweight invariant checks, and  penalizing wasted iterations \citep{feng2026agentocrreimaginingagenthistory}.

\item \textbf{\textcolor{RoyalBlue}{5. Multi-Agent Cooperative Co-Evolution.}} Single-agent exploration is extremely inefficient with vast search spaces. One research direction is to explore collaborative improvement architectures, where specialized agents co-evolve by sharing reusable artifacts such as generated regression tests, successful code patches, or improved tool wrappers~\citep{chen2025multiagentevolvellmselfimprove, wang2026metagenselfevolvingrolestopologies}. Realizing this vision requires establishing secure, version-controlled protocols (e.g., artifact repositories). This enables agents to inherit collective knowledge without single points of failure or cascading attacks on multi-agent reward mechanisms.

\item \textbf{\textcolor{RoyalBlue}{6. Surviving Open-World Distribution Drift.}} The robustness of self-improvement depends on the environment that drives it. Current testing platforms  rely heavily on static repositories or unchanging simulators~\citep{sunil2026memorypoisoningattackdefense}. However, real-world deployments need to cope with constantly changing APIs, redesigned user interfaces, and adversarial user input. Future infrastructure should abandon static leaderboards in favor of non-stationary simulators, allowing the interface to drift continuously~\citep{anonymous2026gaia}. Exposing agents to these open-world perturbations will force the development of self-improvement loops that resist catastrophic forgetting and environmental non-stationarity. \textcolor{red}{At the substrate level, neural-computer formulations further suggest replacing fixed external execution interfaces with learned runtime states that unify computation, memory, and I/O, pointing beyond agents that merely operate computers toward adaptive neural runtimes \citep{zhuge2026neural, rivard2026neuralossimulatingoperatingsystems}.}
\end{itemize}

\vspace{0.35em}
\noindent
In summary, we should transition from building stateless AI tools that reset after every interaction to designing systems capable of continuous self-improvement. Realizing this vision goes beyond simply scaling foundation models. It necessitates robust feedback mechanisms, safe architectures for self-modification, and a fundamental rethinking of evaluation as an ongoing, integrated process rather than a static benchmark.

\section{Conclusion}
\label{sec:Conclusion}

The vision for machines that can improve themselves is an enduring theme in artificial intelligence. Building upon early successful implementations, the advent of large foundation models has substantially broadened the scale, versatility, and scope of self-improving systems. This survey synthesized this paradigm shift through a unified systems lens, distinguishing two primary pathways: \textit{(1)} foundation-model improvement as a parametric, slower loop (driven by intrinsic generative demonstrations, intrinsic evaluative feedback, or extrinsic exploratory experience), and \textit{(2)} scaffolding improvement as a non-parametric, faster loop (updating prompts, memory, tools, and full scaffolds).

Looking ahead, the journey toward truly self-improving agents is entering a decisive phase. Technical challenges such as scalable and secure evaluation, alignment under continuous adaptation, and the integration of these fast and slow improvement loops remain open and pressing. Crucially, as agents progress to self-referential architectures—modifying not only their behavior but also their own reasoning structures—we are invited to imagine a future in which AI does not merely execute human-defined tasks, but evolves to meet them. We hope this survey serves as both a roadmap and a catalyst for that future, equipping researchers with a common language to build self-improving systems that are measurable, attributable, safely bounded, and capable of progressively expanding the frontiers of machine autonomy.

\bibliography{colm2025_conference}

@inproceedings{deng2022rlprompt,
  title={Rlprompt: Optimizing discrete text prompts with reinforcement learning},
  author={Deng, Mingkai and Wang, Jianyu and Hsieh, Cheng-Ping and Wang, Yihan and Guo, Han and Shu, Tianmin and Song, Meng and Xing, Eric and Hu, Zhiting},
  booktitle={Proceedings of the 2022 Conference on Empirical Methods in Natural Language Processing},
  pages={3369--3391},
  year={2022}
}

@inproceedings{cheninstructzero,
  title={InstructZero: Efficient Instruction Optimization for Black-Box Large Language Models},
  author={Chen, Lichang and Chen, Jiuhai and Goldstein, Tom and Huang, Heng and Zhou, Tianyi},
  booktitle={Forty-first International Conference on Machine Learning},
    year={2024}
}

@article{sutton1999between,
  title={Between MDPs and semi-MDPs: A framework for temporal abstraction in reinforcement learning},
  author={Sutton, Richard S and Precup, Doina and Singh, Satinder},
  journal={Artificial intelligence},
  volume={112},
  number={1-2},
  pages={181--211},
  year={1999},
  publisher={Elsevier}
}

@inproceedings{bacon2017option,
  title={The option-critic architecture},
  author={Bacon, Pierre-Luc and Harb, Jean and Precup, Doina},
  booktitle={Proceedings of the AAAI conference on artificial intelligence},
  year={2017}
}

@article{hebbar2026sia,
  title={SIA: Self Improving AI with Harness \& Weight Updates},
  author={Hebbar, Prannay and Manawat, Yogendra and Verboomen, Samuel and Ivanova, Alesia and Palanimalai, Selvam and Bhatia, Kunal and Baskaran, Vignesh},
  journal={arXiv preprint arXiv:2605.27276},
  year={2026}
}

@inproceedings{cheng2024black,
  title={Black-box prompt optimization: Aligning large language models without model training},
  author={Cheng, Jiale and Liu, Xiao and Zheng, Kehan and Ke, Pei and Wang, Hongning and Dong, Yuxiao and Tang, Jie and Huang, Minlie},
  booktitle={Proceedings of the 62nd Annual Meeting of the Association for Computational Linguistics (Volume 1: Long Papers)},
  pages={3201--3219},

    year={2024}
}

@inproceedings{zhou2022large,
  title={Large language models are human-level prompt engineers},
  author={Zhou, Yongchao and Muresanu, Andrei Ioan and Han, Ziwen and Paster, Keiran and Pitis, Silviu and Chan, Harris and Ba, Jimmy},
  booktitle={The eleventh international conference on learning representations},
  year={2022}
}

@incollection{dawkins2019evolution,
  title={The evolution of evolvability},
  author={Dawkins, Richard},
  booktitle={Artificial life},
  pages={201--220},
  year={2019},
  publisher={Routledge}
}

@inproceedings{zhuge2026ai,
  title={AI with Recursive Self-Improvement},
  author={Zhuge, Mingchen and Zeng, Ailing and Zhu, Deyao and Yang, Sherry and Chandra, Vikas and Schmidhuber, J{\"u}rgen},
  year={2026},
  booktitle={ICLR 2026 Workshop Proposals}
}

@article{gerhart2007theory,
  title={The theory of facilitated variation},
  author={Gerhart, John and Kirschner, Marc},
  journal={Proceedings of the National Academy of Sciences},
  volume={104},
  number={suppl\_1},
  pages={8582--8589},
  year={2007},
  publisher={National Academy of Sciences}
}

@inproceedings{DBLP:conf/icml/WieringS96,
  author       = {Marco A. Wiering and
                  J{\"{u}}rgen Schmidhuber},
  editor       = {Lorenza Saitta},
  title        = {Solving POMDPs with Levin Search and {EIRA}},
  booktitle    = {Machine Learning, Proceedings of the Thirteenth International Conference
                  {(ICML} '96), Bari, Italy, July 3-6, 1996},
  pages        = {534--542},
  publisher    = {Morgan Kaufmann},
  year         = {1996},
  bibsource    = {dblp computer science bibliography, https://dblp.org}
}

@inproceedings{schmidhuber1995beyond,
  title={Beyond$\backslash$genetic programming": Incremental self-improvement},
  author={Schmidhuber, J{\"u}rgen},
  booktitle={Proc. Workshop on Genetic Programming at ML95},
  pages={42--49},
  year={1995}
}

@article{yang2024sweagentagentcomputerinterfacesenable,
  title={Swe-agent: Agent-computer interfaces enable automated software engineering},
  author={Yang, John and Jimenez, Carlos E and Wettig, Alexander and Lieret, Kilian and Yao, Shunyu and Narasimhan, Karthik and Press, Ofir},
  journal={Advances in Neural Information Processing Systems},
  volume={37},
  pages={50528--50652},
  year={2024}
}

@misc{xia2024agentlessdemystifyingllmbasedsoftware,
      title={Agentless: Demystifying LLM-based Software Engineering Agents}, 
      author={Chunqiu Steven Xia and Yinlin Deng and Soren Dunn and Lingming Zhang},
      year={2024},
      eprint={2407.01489},
      archivePrefix={arXiv},
      primaryClass={cs.SE},
      url={https://arxiv.org/abs/2407.01489}, 
}

@inproceedings{
yang2025swebench,
title={{SWE}-bench Multimodal: Do {AI} Systems Generalize to Visual Software Domains?},
author={John Yang and Carlos E Jimenez and Alex L Zhang and Kilian Lieret and Joyce Yang and Xindi Wu and Ori Press and Niklas Muennighoff and Gabriel Synnaeve and Karthik R Narasimhan and Diyi Yang and Sida Wang and Ofir Press},
booktitle={The Thirteenth International Conference on Learning Representations},
year={2025},
url={https://openreview.net/forum?id=riTiq3i21b}
}

@misc{ni2025gittaskbenchbenchmarkcodeagents,
      title={GitTaskBench: A Benchmark for Code Agents Solving Real-World Tasks Through Code Repository Leveraging}, 
      author={Ziyi Ni and Huacan Wang and Shuo Zhang and Shuo Lu and Ziyang He and Wang You and Zhenheng Tang and Yuntao Du and Bill Sun and Hongzhang Liu and Sen Hu and Ronghao Chen and Bo Li and Xin Li and Chen Hu and Binxing Jiao and Daxin Jiang and Pin Lyu},
      year={2025},
      eprint={2508.18993},
      archivePrefix={arXiv},
      primaryClass={cs.SE},
      url={https://arxiv.org/abs/2508.18993}, 
}

@inproceedings{
pan2024webcanvas,
title={WebCanvas: Benchmarking Web Agents in Online Environments},
author={Yichen Pan and Dehan Kong and Sida Zhou and Cheng Cui and Yifei Leng and Bing Jiang and Hangyu Liu and Yanyi Shang and Shuyan Zhou and Tongshuang Wu and Zhengyang Wu},
booktitle={Agentic Markets Workshop at ICML 2024},
year={2024},
url={https://openreview.net/forum?id=O1FaGasJob}
}

@inproceedings{
koh2024visualwebarenaevaluatingmultimodalagents,
title={VisualWebArena: Evaluating Multimodal Agents on Realistic Visual Web Tasks},
author={Jing Yu Koh and Robert Lo and Lawrence Jang and Vikram Duvvur and Ming Chong Lim and Po-Yu Huang and Graham Neubig and Shuyan Zhou and Ruslan Salakhutdinov and Daniel Fried},
booktitle={ICLR 2024 Workshop on Large Language Model (LLM) Agents},
year={2024},
url={https://openreview.net/forum?id=RPKxrKTJbj}
}

@inproceedings{
lu2024weblinx,
title={Web{LINX}: Real-World Website Navigation with Multi-Turn Dialogue},
author={Xing Han Lu and Zden{\v{e}}k Kasner and Siva Reddy},
booktitle={Forty-first International Conference on Machine Learning},
year={2024},
url={https://openreview.net/forum?id=mUSPhG4uDW}
}

@inproceedings{
costarelli2024gamebench,
title={GameBench: Evaluating Strategic Reasoning Abilities of {LLM} Agents},
author={Anthony Costarelli and Mat Allen and Roman Hauksson and Grace Sodunke and Suhas Hariharan and Carlson Cheng and Wenjie Li and Joshua M Clymer and Arjun Yadav},
booktitle={Language Gamification - NeurIPS 2024 Workshop},
year={2024},
url={https://openreview.net/forum?id=qrzKE533Jr}
}

@misc{duan2024gtbenchuncoveringstrategicreasoning,
      title={GTBench: Uncovering the Strategic Reasoning Limitations of LLMs via Game-Theoretic Evaluations}, 
      author={Jinhao Duan and Renming Zhang and James Diffenderfer and Bhavya Kailkhura and Lichao Sun and Elias Stengel-Eskin and Mohit Bansal and Tianlong Chen and Kaidi Xu},
      year={2024},
      eprint={2402.12348},
      archivePrefix={arXiv},
      primaryClass={cs.CL},
      url={https://arxiv.org/abs/2402.12348}, 
}

@article{beyer2024clembench2024challengingdynamiccomplementary,
  publtype={informal},
  author={Anne Beyer and Kranti Chalamalasetti and Sherzod Hakimov and Brielen Madureira and Philipp Sadler and David Schlangen},
  title={clembench-2024: A Challenging, Dynamic, Complementary, Multilingual Benchmark and Underlying Flexible Framework for LLMs as Multi-Action Agents},
  year={2024},
  cdate={1704067200000},
  journal={CoRR},
  volume={abs/2405.20859},
  url={https://doi.org/10.48550/arXiv.2405.20859}
}

@inproceedings{hong2023metagpt,
  title={MetaGPT: Meta programming for a multi-agent collaborative framework},
  author={Hong, Sirui and Zhuge, Mingchen and Chen, Jonathan and Zheng, Xiawu and Cheng, Yuheng and Wang, Jinlin and Zhang, Ceyao and Wang, Zili and Yau, Steven Ka Shing and Lin, Zijuan and others},
  booktitle={The twelfth international conference on learning representations},
  year={2023}
}

@inproceedings{chalamalasetti-etal-2023-clembench,
    title = "clembench: Using Game Play to Evaluate Chat-Optimized Language Models as Conversational Agents",
    author = {Chalamalasetti, Kranti  and
      G{\"o}tze, Jana  and
      Hakimov, Sherzod  and
      Madureira, Brielen  and
      Sadler, Philipp  and
      Schlangen, David},
      booktitle = "Proceedings of the 2023 Conference on Empirical Methods in Natural Language Processing",
      month = dec,
      year = "2023",
      address = "Singapore",
      publisher = "Association for Computational Linguistics",
      url = "https://aclanthology.org/2023.emnlp-main.689",
      pages = "11174--11219"
}

@misc{huang2024metatoolbenchmarklargelanguage,
      title={MetaTool Benchmark for Large Language Models: Deciding Whether to Use Tools and Which to Use}, 
      author={Yue Huang and Jiawen Shi and Yuan Li and Chenrui Fan and Siyuan Wu and Qihui Zhang and Yixin Liu and Pan Zhou and Yao Wan and Neil Zhenqiang Gong and Lichao Sun},
      year={2024},
      eprint={2310.03128},
      archivePrefix={arXiv},
      primaryClass={cs.SE},
      url={https://arxiv.org/abs/2310.03128}, 
}

@inproceedings{
ruan2024identifyingriskslmagents,
title={Identifying the Risks of {LM} Agents with an {LM}-Emulated Sandbox},
author={Yangjun Ruan and Honghua Dong and Andrew Wang and Silviu Pitis and Yongchao Zhou and Jimmy Ba and Yann Dubois and Chris J. Maddison and Tatsunori Hashimoto},
booktitle={The Twelfth International Conference on Learning Representations},
year={2024},
url={https://openreview.net/forum?id=GEcwtMk1uA}
}

@inproceedings{
wang2024mintevaluatingllmsmultiturn,
title={{MINT}: Evaluating {LLM}s in Multi-turn Interaction with Tools and Language Feedback},
author={Xingyao Wang and Zihan Wang and Jiateng Liu and Yangyi Chen and Lifan Yuan and Hao Peng and Heng Ji},
booktitle={The Twelfth International Conference on Learning Representations},
year={2024},
url={https://openreview.net/forum?id=jp3gWrMuIZ}
}

@inproceedings{
mialon2023gaiabenchmarkgeneralai,
title={{GAIA}: a benchmark for General {AI} Assistants},
author={Gr{\'e}goire Mialon and Cl{\'e}mentine Fourrier and Thomas Wolf and Yann LeCun and Thomas Scialom},
booktitle={The Twelfth International Conference on Learning Representations},
year={2024},
url={https://openreview.net/forum?id=fibxvahvs3}
}

@article{shen2024taskbenchbenchmarkinglargelanguage,
  title={Taskbench: Benchmarking large language models for task automation},
  author={Shen, Yongliang and Song, Kaitao and Tan, Xu and Zhang, Wenqi and Ren, Kan and Yuan, Siyu and Lu, Weiming and Li, Dongsheng and Zhuang, Yueting},
  journal={Advances in Neural Information Processing Systems},
  volume={37},
  pages={4540--4574},
  year={2024}
}

@article{xie2024osworldbenchmarkingmultimodalagents,
  title={Osworld: Benchmarking multimodal agents for open-ended tasks in real computer environments},
  author={Xie, Tianbao and Zhang, Danyang and Chen, Jixuan and Li, Xiaochuan and Zhao, Siheng and Cao, Ruisheng and Hua, Toh J and Cheng, Zhoujun and Shin, Dongchan and Lei, Fangyu and others},
  journal={Advances in Neural Information Processing Systems},
  volume={37},
  pages={52040--52094},
  year={2024}
}

@InProceedings{pan2025trainingsoftwareengineeringagents,
  title = 	 {Training Software Engineering Agents and Verifiers with {SWE}-Gym},
  author =       {Pan, Jiayi and Wang, Xingyao and Neubig, Graham and Jaitly, Navdeep and Ji, Heng and Suhr, Alane and Zhang, Yizhe},
  booktitle = 	 {Proceedings of the 42nd International Conference on Machine Learning},
  pages = 	 {47717--47737},
  year = 	 {2025},
  editor = 	 {Singh, Aarti and Fazel, Maryam and Hsu, Daniel and Lacoste-Julien, Simon and Berkenkamp, Felix and Maharaj, Tegan and Wagstaff, Kiri and Zhu, Jerry},
  volume = 	 {267},
  series = 	 {Proceedings of Machine Learning Research},
  month = 	 {13--19 Jul},
  publisher =    {PMLR},
  url = 	 {https://proceedings.mlr.press/v267/pan25g.html},
  
}

@misc{robeyns2025selfimprovingcodingagent,
      title={A Self-Improving Coding Agent}, 
      author={Maxime Robeyns and Martin Szummer and Laurence Aitchison},
      year={2025},
      eprint={2504.15228},
      archivePrefix={arXiv},
      primaryClass={cs.AI},
      url={https://arxiv.org/abs/2504.15228}, 
}

@inproceedings{
bonatti2024windowsagentarenaevaluating,
title={Windows Agent Arena: Evaluating Multi-Modal {OS} Agents at Scale},
author={Rogerio Bonatti and Dan Zhao and Francesco Bonacci and Dillon Dupont and Sara Abdali and Yinheng Li and Yadong Lu and Justin Wagle and Kazuhito Koishida and Arthur Bucker and Lawrence Keunho Jang and Zheng Hui},
booktitle={Forty-second International Conference on Machine Learning},
year={2025},
url={https://openreview.net/forum?id=W9s817KqYf}
}

@inproceedings{
zheng2024gptvision,
title={{GPT}-4V(ision) is a Generalist Web Agent, if Grounded},
author={Boyuan Zheng and Boyu Gou and Jihyung Kil and Huan Sun and Yu Su},
booktitle={Forty-first International Conference on Machine Learning},
year={2024},
url={https://openreview.net/forum?id=piecKJ2DlB}
}

@incollection{schmidhuber1999general,
  title={A general method for incremental self-improvement and multi-agent learning},
  author={Schmidhuber, Juergen},
  booktitle={Evolutionary Computation: Theory and Applications},
  pages={81--123},
  year={1999},
  publisher={World Scientific}
}

@incollection{schmidhuber1998reinforcement,
  title={Reinforcement learning with self-modifying policies},
  author={Schmidhuber, J{\"u}rgen and Zhao, Jieyu and Schraudolph, Nicol N},
  booktitle={Learning to learn},
  pages={293--309},
  year={1998},
  publisher={Springer}
}

@inproceedings{feng2025rethinkingllmuncertaintymultiagent,
    title = "Rethinking {LLM} Uncertainty: A Multi-Agent Approach to Estimating Black-Box Model Uncertainty",
    author = "Feng, Yu  and
      Htut, Phu Mon  and
      Qi, Zheng  and
      Xiao, Wei  and
      Mager, Manuel  and
      Pappas, Nikolaos  and
      Halder, Kishaloy  and
      Li, Yang  and
      Benajiba, Yassine  and
      Roth, Dan",
    editor = "Christodoulopoulos, Christos  and
      Chakraborty, Tanmoy  and
      Rose, Carolyn  and
      Peng, Violet",
    booktitle = "Findings of the Association for Computational Linguistics: EMNLP 2025",
    month = nov,
    year = "2025",
    address = "Suzhou, China",
    publisher = "Association for Computational Linguistics",
    url = "https://aclanthology.org/2025.findings-emnlp.660/",
    doi = "10.18653/v1/2025.findings-emnlp.660",
    pages = "12349--12375",
    ISBN = "979-8-89176-335-7"
}

@misc{till2025teamingllmsdetectmitigate,
      title={Teaming LLMs to Detect and Mitigate Hallucinations}, 
      author={Demian Till and John Smeaton and Peter Haubrick and Gouse Saheb and Florian Graef and David Berman},
      year={2025},
      eprint={2510.19507},
      archivePrefix={arXiv},
      primaryClass={cs.LG},
      url={https://arxiv.org/abs/2510.19507}, 
}

@inproceedings{tan-etal-2025-consistent,
    title = "Too Consistent to Detect: A Study of Self-Consistent Errors in {LLM}s",
    author = "Tan, Hexiang  and
      Sun, Fei  and
      Liu, Sha  and
      Su, Du  and
      Cao, Qi  and
      Chen, Xin  and
      Wang, Jingang  and
      Cai, Xunliang  and
      Wang, Yuanzhuo  and
      Shen, Huawei  and
      Cheng, Xueqi",
    editor = "Christodoulopoulos, Christos  and
      Chakraborty, Tanmoy  and
      Rose, Carolyn  and
      Peng, Violet",
    booktitle = "Proceedings of the 2025 Conference on Empirical Methods in Natural Language Processing",
    month = nov,
    year = "2025",
    address = "Suzhou, China",
    publisher = "Association for Computational Linguistics",
    url = "https://aclanthology.org/2025.emnlp-main.238/",
    doi = "10.18653/v1/2025.emnlp-main.238",
    pages = "4755--4765",
    ISBN = "979-8-89176-332-6"
}

@inproceedings{
huang2024largelanguagemodelsselfcorrect,
title={Large Language Models Cannot Self-Correct Reasoning Yet},
author={Jie Huang and Xinyun Chen and Swaroop Mishra and Huaixiu Steven Zheng and Adams Wei Yu and Xinying Song and Denny Zhou},
booktitle={The Twelfth International Conference on Learning Representations},
year={2024},
url={https://openreview.net/forum?id=IkmD3fKBPQ}
}

@inproceedings{
anonymous2025complementing,
title={Complementing Self-Consistency with Cross-Model Disagreement for Uncertainty Quantification},
author={Kimia Hamidieh and Veronika Thost and Walter Gerych and Mikhail Yurochkin and Marzyeh Ghassemi},
booktitle={The Fourteenth International Conference on Learning Representations},
year={2026},
url={https://openreview.net/forum?id=lOoRJo8xWy}
}

@inproceedings{10.5555/3692070.3694667,
author = {Zhuge, Mingchen and Wang, Wenyi and Kirsch, Louis and Faccio, Francesco and Khizbullin, Dmitrii and Schmidhuber, J\"{u}rgen},
title = {GPTSwarm: language agents as optimizable graphs},
year = {2024},
publisher = {JMLR.org},
booktitle = {Proceedings of the 41st International Conference on Machine Learning},
articleno = {2597},
numpages = {25},
location = {Vienna, Austria},
series = {ICML'24}
}

@inproceedings{
zhou2023leasttomost,
title={Least-to-Most Prompting Enables Complex Reasoning in Large Language Models},
author={Denny Zhou and Nathanael Sch{\"a}rli and Le Hou and Jason Wei and Nathan Scales and Xuezhi Wang and Dale Schuurmans and Claire Cui and Olivier Bousquet and Quoc V Le and Ed H. Chi},
booktitle={The Eleventh International Conference on Learning Representations },
year={2023},
url={https://openreview.net/forum?id=WZH7099tgfM}
}

@article{zhuge2023mindstorms,
  title={Mindstorms in natural language-based societies of mind},
  author={Zhuge, Mingchen and Liu, Haozhe and Faccio, Francesco and Ashley, Dylan R and Csord{\'a}s, R{\'o}bert and Gopalakrishnan, Anand and Hamdi, Abdullah and Hammoud, Hasan Abed Al Kader and Herrmann, Vincent and Irie, Kazuki and others},
  journal={arXiv preprint arXiv:2305.17066},
  year={2023}
}

@article{chen2024failuresselfconsistencymultistepreasoning,
  publtype={informal},
  author={Angelica Chen and Jason Phang and Alicia Parrish and Vishakh Padmakumar and Chen Zhao and Samuel R. Bowman and Kyunghyun Cho},
  title={Two Failures of Self-Consistency in the Multi-Step Reasoning of LLMs},
  year={2023},
  cdate={1672531200000},
  journal={CoRR},
  volume={abs/2305.14279},
  url={https://doi.org/10.48550/arXiv.2305.14279},
}

@article{kirsch2022eliminating,
  title={Eliminating meta optimization through self-referential meta learning},
  author={Kirsch, Louis and Schmidhuber, J{\"u}rgen},
  journal={arXiv preprint arXiv:2212.14392},
  year={2022}
}

@misc{zhang2026selfevolvingworldmodelsllm,
      title={Self-Evolving World Models for LLM Agent Planning}, 
      author={Xuan Zhang and Wenxuan Zhang and See-Kiong Ng and Yang Deng},
      year={2026},
      eprint={2606.30639},
      archivePrefix={arXiv},
      primaryClass={cs.AI},
      url={https://arxiv.org/abs/2606.30639}, 
}

@misc{wadhwa2025evalagentdiscoveringimplicitevaluation,
      title={EvalAgent: Discovering Implicit Evaluation Criteria from the Web}, 
      author={Manya Wadhwa and Zayne Sprague and Chaitanya Malaviya and Philippe Laban and Junyi Jessy Li and Greg Durrett},
      year={2025},
      eprint={2504.15219},
      archivePrefix={arXiv},
      primaryClass={cs.CL},
      url={https://arxiv.org/abs/2504.15219}, 
}

@misc{han2025verifiagentunifiedverificationagent,
      title={VerifiAgent: a Unified Verification Agent in Language Model Reasoning}, 
      author={Jiuzhou Han and Wray Buntine and Ehsan Shareghi},
      year={2025},
      eprint={2504.00406},
      archivePrefix={arXiv},
      primaryClass={cs.CL},
      url={https://arxiv.org/abs/2504.00406}, 
}

@inproceedings{xu-etal-2025-learning,
    title = "Learning to Align Multi-Faceted Evaluation: A Unified and Robust Framework",
    author = "Xu, Kaishuai  and
      Yu, Tiezheng  and
      Cheng, Yi  and
      Hou, Wenjun  and
      Li, Liangyou  and
      Jiang, Xin  and
      Shang, Lifeng  and
      Liu, Qun  and
      Li, Wenjie",
    editor = "Che, Wanxiang  and
      Nabende, Joyce  and
      Shutova, Ekaterina  and
      Pilehvar, Mohammad Taher",
    booktitle = "Findings of the Association for Computational Linguistics: ACL 2025",
    month = jul,
    year = "2025",
    address = "Vienna, Austria",
    publisher = "Association for Computational Linguistics",
    url = "https://aclanthology.org/2025.findings-acl.494/",
    doi = "10.18653/v1/2025.findings-acl.494",
    pages = "9488--9502",
    ISBN = "979-8-89176-256-5"
}

@inproceedings{zhang-etal-2025-evaluation,
    title = "Evaluation Agent: Efficient and Promptable Evaluation Framework for Visual Generative Models",
    author = "Zhang, Fan  and
      Tian, Shulin  and
      Huang, Ziqi  and
      Qiao, Yu  and
      Liu, Ziwei",
    editor = "Che, Wanxiang  and
      Nabende, Joyce  and
      Shutova, Ekaterina  and
      Pilehvar, Mohammad Taher",
    booktitle = "Proceedings of the 63rd Annual Meeting of the Association for Computational Linguistics (Volume 1: Long Papers)",
    month = jul,
    year = "2025",
    address = "Vienna, Austria",
    publisher = "Association for Computational Linguistics",
    url = "https://aclanthology.org/2025.acl-long.374/",
    doi = "10.18653/v1/2025.acl-long.374",
    pages = "7561--7582",
    ISBN = "979-8-89176-251-0"
}

@misc{iacob2026redqueengodelmachine,
      title={The Red Queen G\"odel Machine: Co-Evolving Agents and Their Evaluators}, 
      author={Alex Iacob and Andrej Jovanović and William F. Shen and Daniel Burkhardt and Meghdad Kurmanji and Nurbek Tastan and Lorenzo Sani and Niccolò Alberto Elia Venanzi and Ambroise Odonnat and Zeyu Cao and Bill Marino and Xinchi Qiu and Nicholas D. Lane},
      year={2026},
      eprint={2606.26294},
      archivePrefix={arXiv},
      primaryClass={cs.LG},
      url={https://arxiv.org/abs/2606.26294}, 
}

@misc{karten2026continualharnessonlineadaptation,
      title={Continual Harness: Online Adaptation for Self-Improving Foundation Agents}, 
      author={Seth Karten and Joel Zhang and Tersoo Upaa Jr and Ruirong Feng and Wenzhe Li and Chengshuai Shi and Chi Jin and Kiran Vodrahalli},
      year={2026},
      eprint={2605.09998},
      archivePrefix={arXiv},
      primaryClass={cs.LG},
      url={https://arxiv.org/abs/2605.09998}, 
}

@misc{nguyen2026recursiveselfevolvingagentsheldout,
      title={Recursive Self-Evolving Agents via Held-Out Selection}, 
      author={Michael Nguyen and Quoc Nguyen and Paul Vuong},
      year={2026},
      eprint={2606.28374},
      archivePrefix={arXiv},
      primaryClass={cs.AI},
      url={https://arxiv.org/abs/2606.28374}, 
}

@misc{cai2026mossselfevolutionsourcelevelrewriting,
      title={MOSS: Self-Evolution through Source-Level Rewriting in Autonomous Agent Systems}, 
      author={Qianshu Cai and Yonggang Zhang and Xianzhang Jia and Huajiang Zheng and Wei Xue and Jun Song and Xinmei Tian and Yike Guo},
      year={2026},
      eprint={2605.22794},
      archivePrefix={arXiv},
      primaryClass={cs.AI},
      url={https://arxiv.org/abs/2605.22794}, 
}

@misc{liu2026adaptiveautoharnesssustainedselfimprovement,
      title={Adaptive Auto-Harness: Sustained Self-Improvement for Agentic System Deployment on Open-Ended Task Streams}, 
      author={Zewen Liu and Zhan Shi and Yisi Sang and Bing He and Minhua Lin and Tianxin Wei and Dakuo Wang and Benoit Dumoulin and Wei Jin and Hanqing Lu},
      year={2026},
      eprint={2606.01770},
      archivePrefix={arXiv},
      primaryClass={cs.LG},
      url={https://arxiv.org/abs/2606.01770}, 
}

@misc{jin2026exgselfevolvingagentsexperience,
      title={EXG: Self-Evolving Agents with Experience Graphs}, 
      author={Yuxin Jin and Siyuan Zhang and Hanchen Wang and Lu Qin and Ying Zhang and Wenjie Zhang},
      year={2026},
      eprint={2605.17721},
      archivePrefix={arXiv},
      primaryClass={cs.AI},
      url={https://arxiv.org/abs/2605.17721}, 
}

@misc{yu2026autonomousevolutionedatools,
      title={Autonomous Evolution of EDA Tools: Multi-Agent Self-Evolved ABC}, 
      author={Cunxi Yu and Haoxing Ren},
      year={2026},
      eprint={2604.15082},
      archivePrefix={arXiv},
      primaryClass={cs.AR},
      url={https://arxiv.org/abs/2604.15082}, 
}

@article{yang2026evods,
  title={EvoDS: Self-Evolving Autonomous Data Science Agent with Skill Learning and Context Management},
  author={Yang, Zherui and Liu, Fan and Ning, Yansong and Liu, Hao},
  journal={arXiv preprint arXiv:2606.03841},
  year={2026}
}

@misc{yan2026openskillopenworldselfevolutionllm,
      title={OpenSkill: Open-World Self-Evolution for LLM Agents}, 
      author={Zhiling Yan and Dingjie Song and Hanrong Zhang and Wei Liang and Yuxuan Zhang and Yutong Dai and Lifang He and Philip S. Yu and Ran Xu and Xiang Li and Lichao Sun},
      year={2026},
      eprint={2606.06741},
      archivePrefix={arXiv},
      primaryClass={cs.AI},
      url={https://arxiv.org/abs/2606.06741}, 
}

@misc{she2026pfagenttractableselfevolvingpowerflow,
      title={PFAgent: A Tractable and Self-Evolving Power-Flow Agent for Interactive Grid Analysis}, 
      author={Buxin She and Brian Chen and Luanzheng Guo and Fangxing Li},
      year={2026},
      eprint={2604.10846},
      archivePrefix={arXiv},
      primaryClass={eess.SY},
      url={https://arxiv.org/abs/2604.10846}, 
}

@misc{li2026codeskilllearningselfevolvingskills,
      title={CODESKILL: Learning Self-Evolving Skills for Coding Agents}, 
      author={Yanzhou Li and Yiran Zhang and Xiaoyu Zhang and Xiaoxia Liu and Yang Liu},
      year={2026},
      eprint={2605.25430},
      archivePrefix={arXiv},
      primaryClass={cs.AI},
      url={https://arxiv.org/abs/2605.25430}, 
}

@misc{lin2026museautoskillselfevolvingagentsskill,
      title={MUSE-Autoskill: Self-Evolving Agents via Skill Creation, Memory, Management, and Evaluation}, 
      author={Huawei Lin and Peng Li and Jie Song and Fuxin Jiang and Tieying Zhang},
      year={2026},
      eprint={2605.27366},
      archivePrefix={arXiv},
      primaryClass={cs.AI},
      url={https://arxiv.org/abs/2605.27366}, 
}

@misc{zhang2026rewardharnessselfevolvingagenticposttraining,
      title={RewardHarness: Self-Evolving Agentic Post-Training}, 
      author={Yuxuan Zhang and Penghui Du and Bo Li and Cong Wei and Junwen Miao and Huaisong Zhang and Songcheng Cai and Yubo Wang and Dongfu Jiang and Yuyu Zhang and Ping Nie and Wenhu Chen and Changqian Yu and Kelsey R. Allen},
      year={2026},
      eprint={2605.08703},
      archivePrefix={arXiv},
      primaryClass={cs.AI},
      url={https://arxiv.org/abs/2605.08703}, 
}

@misc{liang2026genericagenttokenefficientselfevolvingllm,
      title={GenericAgent: A Token-Efficient Self-Evolving LLM Agent via Contextual Information Density Maximization (V1.0)}, 
      author={Jiaqing Liang and Jinyi Han and Weijia Li and Xinyi Wang and Zhoujia Zhang and Zishang Jiang and Ying Liao and Tingyun Li and Ying Huang and Hao Shen and Hanyu Wu and Fang Guo and Keyi Wang and Zhonghua Hong and Zhiyu Lu and Lipeng Ma and Sihang Jiang and Yanghua Xiao},
      year={2026},
      eprint={2604.17091},
      archivePrefix={arXiv},
      primaryClass={cs.CL},
      url={https://arxiv.org/abs/2604.17091}, 
}

@misc{cheng2026mem2evolveselfevolvingagentscoevolutionary,
      title={Mem$^2$Evolve: Towards Self-Evolving Agents via Co-Evolutionary Capability Expansion and Experience Distillation}, 
      author={Zihao Cheng and Zeming Liu and Yingyu Shan and Xinyi Wang and Xiangrong Zhu and Yunpu Ma and Hongru Wang and Yuhang Guo and Wei Lin and Yunhong Wang},
      year={2026},
      eprint={2604.10923},
      archivePrefix={arXiv},
      primaryClass={cs.CL},
      url={https://arxiv.org/abs/2604.10923}, 
}

@misc{dai2026metisbridgingtextcode,
      title={Metis: Bridging Text and Code Memory for Self-Evolving Agents}, 
      author={Zijie Dai and Siuhin He and Hui Li and Qihui Zhou and Jiajun Li and Mingcong Song and Guoping Long and Hongjie Si and Xin Yao and Lin Zhang and James Cheng and Xiao Yan},
      year={2026},
      eprint={2606.24151},
      archivePrefix={arXiv},
      primaryClass={cs.CL},
      url={https://arxiv.org/abs/2606.24151}, 
}

@misc{xu2026aelagentevolvinglearning,
      title={AEL: Agent Evolving Learning for Open-Ended Environments}, 
      author={Wujiang Xu and Jiaojiao Han and Minghao Guo and Kai Mei and Xi Zhu and Han Zhang and Dimitris N. Metaxas},
      year={2026},
      eprint={2604.21725},
      archivePrefix={arXiv},
      primaryClass={cs.CL},
      url={https://arxiv.org/abs/2604.21725}, 
}

@misc{fei2026memoryrecalldualprocesscognitive,
      title={Memory Beyond Recall: A Dual-Process Cognitive Memory System for Self-Evolving LLM Agents}, 
      author={Tianxiang Fei and Mingyang Song and Mao Zheng and Xiang Yu},
      year={2026},
      eprint={2606.09483},
      archivePrefix={arXiv},
      primaryClass={cs.CL},
      url={https://arxiv.org/abs/2606.09483}, 
}

@misc{liao2026memqintegratingqlearningselfevolving,
      title={MemQ: Integrating Q-Learning into Self-Evolving Memory Agents over Provenance DAGs}, 
      author={Junwei Liao and Haoting Shi and Ruiwen Zhou and Jiaqian Wang and Shengtao Zhang and Wei Zhang and Ying Wen and Zhiyu Li and Feiyu Xiong and Bo Tang and Weinan Zhang and Muning Wen},
      year={2026},
      eprint={2605.08374},
      archivePrefix={arXiv},
      primaryClass={cs.AI},
      url={https://arxiv.org/abs/2605.08374}, 
}

@misc{ren2026scalingselfevolvingagentsparametric,
      title={Scaling Self-Evolving Agents via Parametric Memory}, 
      author={Tao Ren and Weiyao Luo and Hui Yang and Rongzhi Zhu and Xiang Huang and Yuchuan Wu and Bingxue Chou and Jieping Ye and Jiafeng Liang and Yongbin Li and Yijie Peng},
      year={2026},
      eprint={2606.04536},
      archivePrefix={arXiv},
      primaryClass={cs.AI},
      url={https://arxiv.org/abs/2606.04536}, 
}

@misc{mishra2026prismevolutionarymemorysubstrate,
      title={Prism: An Evolutionary Memory Substrate for Multi-Agent Open-Ended Discovery}, 
      author={Suyash Mishra},
      year={2026},
      eprint={2604.19795},
      archivePrefix={arXiv},
      primaryClass={cs.AI},
      url={https://arxiv.org/abs/2604.19795}, 
}

@misc{hao2026selfevolvingmultiagentsystemsdecentralized,
      title={Self-Evolving Multi-Agent Systems via Decentralized Memory}, 
      author={Guangya Hao and Yunbo Long and Zhuokai Zhao},
      year={2026},
      eprint={2605.22721},
      archivePrefix={arXiv},
      primaryClass={cs.MA},
      url={https://arxiv.org/abs/2605.22721}, 
}

@misc{wang2026sageselfevolvingagenticgraphmemory,
      title={SAGE: A Self-Evolving Agentic Graph-Memory Engine for Structure-Aware Associative Memory}, 
      author={Juntong Wang and Haoyue Zhao and guanghui Pan and Xiyuan Wang and Yanbo Wang and Qiyan Deng and Muhan Zhang},
      year={2026},
      eprint={2605.12061},
      archivePrefix={arXiv},
      primaryClass={cs.AI},
      url={https://arxiv.org/abs/2605.12061}, 
}

@misc{liu2026evolvememselfevolvingmemoryarchitectureautoresearch,
      title={EvolveMem:Self-Evolving Memory Architecture via AutoResearch for LLM Agents}, 
      author={Jiaqi Liu and Xinyu Ye and Peng Xia and Zeyu Zheng and Cihang Xie and Mingyu Ding and Huaxiu Yao},
      year={2026},
      eprint={2605.13941},
      archivePrefix={arXiv},
      primaryClass={cs.LG},
      url={https://arxiv.org/abs/2605.13941}, 
}

@misc{fu2026betterexperienceselfevolvingllm,
      title={Better with Experience: Self-Evolving LLM Agents for Evidence-Grounded Health Community Notes}, 
      author={Zihang Fu and Fanxiao Li and Jianyang Gu and Haonan Wang and Preslav Nakov and Bryan Hooi and Min-Yen Kan and Jiaying Wu},
      year={2026},
      eprint={2606.02215},
      archivePrefix={arXiv},
      primaryClass={cs.CL},
      url={https://arxiv.org/abs/2606.02215}, 
}

@misc{yang2026riseselfimprovingrobotpolicy,
      title={RISE: Self-Improving Robot Policy with Compositional World Model}, 
      author={Jiazhi Yang and Kunyang Lin and Jinwei Li and Wencong Zhang and Tianwei Lin and Longyan Wu and Zhizhong Su and Hao Zhao and Ya-Qin Zhang and Li Chen and Ping Luo and Xiangyu Yue and Hongyang Li},
      year={2026},
      eprint={2602.11075},
      archivePrefix={arXiv},
      primaryClass={cs.RO},
      url={https://arxiv.org/abs/2602.11075}, 
}

@misc{feng2026thoughtretrieverdontjustretrieve,
      title={Thought-Retriever: Don't Just Retrieve Raw Data, Retrieve Thoughts for Memory-Augmented Agentic Systems}, 
      author={Tao Feng and Pengrui Han and Guanyu Lin and Ge Liu and Jiaxuan You},
      year={2026},
      eprint={2604.12231},
      archivePrefix={arXiv},
      primaryClass={cs.CL},
      url={https://arxiv.org/abs/2604.12231}, 
}

@misc{wang2025evoworldevolvingpanoramicworld,
      title={EvoWorld: Evolving Panoramic World Generation with Explicit 3D Memory}, 
      author={Jiahao Wang and Luoxin Ye and TaiMing Lu and Junfei Xiao and Jiahan Zhang and Yuxiang Guo and Xijun Liu and Rama Chellappa and Cheng Peng and Alan Yuille and Jieneng Chen},
      year={2025},
      eprint={2510.01183},
      archivePrefix={arXiv},
      primaryClass={cs.CV},
      url={https://arxiv.org/abs/2510.01183}, 
}

@misc{yao2025navmorphselfevolvingworldmodel,
      title={NavMorph: A Self-Evolving World Model for Vision-and-Language Navigation in Continuous Environments}, 
      author={Xuan Yao and Junyu Gao and Changsheng Xu},
      year={2025},
      eprint={2506.23468},
      archivePrefix={arXiv},
      primaryClass={cs.CV},
      url={https://arxiv.org/abs/2506.23468}, 
}

@inproceedings{liu2024autodan,
  title={Autodan: Generating stealthy jailbreak prompts on aligned large language models},
  author={Liu, Xiaogeng and Xu, Nan and Chen, Muhao and Xiao, Chaowei},
  booktitle={International Conference on Learning Representations},
  volume={2024},
  pages={56174--56194},
  year={2024}
}

@inproceedings{gou2024critic,
  title={Critic: Large language models can self-correct with tool-interactive critiquing},
  author={Gou, Zhibin and Shao, Zhihong and Gong, Yeyun and Yang, Yujiu and Duan, Nan and Chen, Weizhu and others},
  booktitle={International Conference on Learning Representations},
  volume={2024},
  pages={57734--57811},
  year={2024}
}

@article{khattab2023dspy,
  title={Dspy: Compiling declarative language model calls into self-improving pipelines},
  author={Khattab, Omar and Singhvi, Arnav and Maheshwari, Paridhi and Zhang, Zhiyuan and Santhanam, Keshav and Vardhamanan, Sri and Haq, Saiful and Sharma, Ashutosh and Joshi, Thomas T and Moazam, Hanna and others},
  journal={arXiv preprint arXiv:2310.03714},
  year={2023}
}

@inproceedings{sun2022black,
  title={Black-box tuning for language-model-as-a-service},
  author={Sun, Tianxiang and Shao, Yunfan and Qian, Hong and Huang, Xuanjing and Qiu, Xipeng},
  booktitle={International Conference on Machine Learning},
  pages={20841--20855},
  year={2022},
  organization={PMLR}
}

@misc{yang2026skilloptexecutivestrategyselfevolving,
      title={SkillOpt: Executive Strategy for Self-Evolving Agent Skills}, 
      author={Yifan Yang and Ziyang Gong and Weiquan Huang and Qihao Yang and Ziwei Zhou and Zisu Huang and Yan Li and Xuemei Gao and Qi Dai and Bei Liu and Kai Qiu and Yuqing Yang and Dongdong Chen and Xue Yang and Chong Luo},
      year={2026},
      eprint={2605.23904},
      archivePrefix={arXiv},
      primaryClass={cs.AI},
      url={https://arxiv.org/abs/2605.23904}, 
}

@misc{he2026learningevolveselfimprovingframework,
      title={Learning to Evolve: A Self-Improving Framework for Multi-Agent Systems via Textual Parameter Graph Optimization}, 
      author={Shan He and Runze Wang and Zhuoyun Du and Huiyu Bai and Zouying Cao and Yu Cheng and Bo Zheng},
      year={2026},
      eprint={2604.20714},
      archivePrefix={arXiv},
      primaryClass={cs.AI},
      url={https://arxiv.org/abs/2604.20714}, 
}

@misc{yang2026vasoformallyverifiableselfevolving,
      title={VASO: Formally Verifiable Self-Evolving Skills for Physical AI Agents}, 
      author={Yunhao Yang and Neel P. Bhatt and Kevin Wang and Samuel Tetteh and Zhangyang Wang and Ufuk Topcu},
      year={2026},
      eprint={2606.05395},
      archivePrefix={arXiv},
      primaryClass={cs.RO},
      url={https://arxiv.org/abs/2606.05395}, 
}

@article{zhu2026sage,
  title={SAGE: Stochastic Prompt Optimization via Agent-Guided Exploration},
  author={Zhu, Ziyi and Smyth, Luka and Shinoda, Saki and Chen, Jinghong},
  journal={arXiv preprint arXiv:2606.18902},
  year={2026}
}

@misc{zhang2026coevoskillsselfevolvingagentskills,
      title={CoEvoSkills: Self-Evolving Agent Skills via Co-Evolutionary Verification}, 
      author={Hanrong Zhang and Shicheng Fan and Henry Peng Zou and Yankai Chen and Zhenting Wang and Jiayu Zhou and Chengze Li and Wei-Chieh Huang and Yifei Yao and Kening Zheng and Xue Liu and Xiaoxiao Li and Philip S. Yu},
      year={2026},
      eprint={2604.01687},
      archivePrefix={arXiv},
      primaryClass={cs.AI},
      url={https://arxiv.org/abs/2604.01687}, 
}

@misc{tao2026seposelfevolvingpromptagent,
      title={SePO: Self-Evolving Prompt Agent for System Prompt Optimization}, 
      author={Wangcheng Tao and Han Wu and Weng-Fai Wong},
      year={2026},
      eprint={2606.04465},
      archivePrefix={arXiv},
      primaryClass={cs.CL},
      url={https://arxiv.org/abs/2606.04465}, 
}

@inproceedings{Bogdanov_2026, series={CAIS ’26},
   title={FORGE: Self-Evolving Agent Memory With No Weight Updates via Population Broadcast},
   url={http://dx.doi.org/10.1145/3786335.3813155},
   DOI={10.1145/3786335.3813155},
   booktitle={Proceedings of the ACM Conference on AI and Agentic Systems},
   publisher={ACM},
   author={Bogdanov, Igor and Lung, Chung-Horng and Kunz, Thomas and Gao, Jie and Taylor, Adrian and Zaman, Marzia},
   year={2026},
   month=May, pages={292–310},
   collection={CAIS ’26} }

@article{christiano2017deep,
  title={Deep reinforcement learning from human preferences},
  author={Christiano, Paul F and Leike, Jan and Brown, Tom and Martic, Miljan and Legg, Shane and Amodei, Dario},
  journal={Advances in neural information processing systems},
  volume={30},
  year={2017}
}

@misc{xiao2026socraticsweselfevolvingcodingagents,
      title={Socratic-SWE: Self-Evolving Coding Agents via Trace-Derived Agent Skills}, 
      author={Chuan Xiao and Zhengbo Jiao and Shaobo Wang and Wei Wang and Bing Zhao and Hu Wei and Linfeng Zhang and Lin Qu},
      year={2026},
      eprint={2606.07412},
      archivePrefix={arXiv},
      primaryClass={cs.SE},
      url={https://arxiv.org/abs/2606.07412}, 
}

@misc{acikgoz2026toolr0selfevolvingllmagents,
      title={Tool-R0: Self-Evolving LLM Agents for Tool-Learning from Zero Data}, 
      author={Emre Can Acikgoz and Cheng Qian and Jonas Hübotter and Heng Ji and Dilek Hakkani-Tür and Gokhan Tur},
      year={2026},
      eprint={2602.21320},
      archivePrefix={arXiv},
      primaryClass={cs.LG},
      url={https://arxiv.org/abs/2602.21320}, 
}

@misc{arai2026eveagentevidenceverifiableselfevolvingagents,
      title={EVE-Agent: Evidence-Verifiable Self-Evolving Agents}, 
      author={Yamato Arai and Yuma Ichikawa},
      year={2026},
      eprint={2605.22905},
      archivePrefix={arXiv},
      primaryClass={cs.AI},
      url={https://arxiv.org/abs/2605.22905}, 
}

@article{ma2026retrospective,
  title={Retrospective Progress-Aware Self-Refinement for LLM Agent Training},
  author={Ma, Xinbei and Zheng, Congmin and Qiu, Jiyang and Hong, Jiale and Yao, Yao and Qu, Xiangmou and Yin, Jiaxin and Lou, Xingyu and Wang, Jun and Liu, Weiwen and others},
  journal={arXiv preprint arXiv:2606.14302},
  year={2026}
}

@misc{zhang2026selfevolvingllmagentsindistribution,
      title={Self-evolving LLM agents with in-distribution Optimization}, 
      author={Yudi Zhang and Meng Fang and Zhenfang Chen and Mykola Pechenizkiy},
      year={2026},
      eprint={2606.07367},
      archivePrefix={arXiv},
      primaryClass={cs.LG},
      url={https://arxiv.org/abs/2606.07367}, 
}

@misc{jung2026evogroundselfevolvingvideoagents,
      title={EvoGround: Self-Evolving Video Agents for Video Temporal Grounding}, 
      author={Minjoon Jung and Byoung-Tak Zhang and Lorenzo Torresani},
      year={2026},
      eprint={2605.13803},
      archivePrefix={arXiv},
      primaryClass={cs.CV},
      url={https://arxiv.org/abs/2605.13803}, 
}

@misc{zhao2026andesagentnativedata,
      title={ANDES: Agent Native Data Evolving Synthesis Tool for Autonomous Instruction Alignment}, 
      author={Zhengyang Zhao and Shengjie Ye and Lu Ma and Hao Liang and Hengyi Feng and Wentao Zhang},
      year={2026},
      eprint={2606.01279},
      archivePrefix={arXiv},
      primaryClass={cs.AI},
      url={https://arxiv.org/abs/2606.01279}, 
}

@article{stanic2023learning,
  title={Learning to generalize with object-centric agents in the open world survival game crafter},
  author={Stani{\'c}, Aleksandar and Tang, Yujin and Ha, David and Schmidhuber, J{\"u}rgen},
  journal={IEEE Transactions on Games},
  volume={16},
  number={2},
  pages={384--395},
  year={2023},
  publisher={IEEE}
}

@article{kirsch2021meta,
  title={Meta learning backpropagation and improving it},
  author={Kirsch, Louis and Schmidhuber, J{\"u}rgen},
  journal={Advances in Neural Information Processing Systems},
  volume={34},
  pages={14122--14134},
  year={2021}
}

@article{schmidhuber2004optimal,
  title={Optimal ordered problem solver},
  author={Schmidhuber, J{\"u}rgen},
  journal={Machine Learning},
  volume={54},
  number={3},
  pages={211--254},
  year={2004},
  publisher={Springer}
}

@article{schmidhuber2006developmental,
  title={Developmental robotics, optimal artificial curiosity, creativity, music, and the fine arts},
  author={Schmidhuber, J{\"u}rgen},
  journal={Connection Science},
  volume={18},
  number={2},
  pages={173--187},
  year={2006},
  publisher={Taylor \& Francis}
}

@article{Schrittwieser_2020,
   title={Mastering Atari, Go, chess and shogi by planning with a learned model},
   volume={588},
   ISSN={1476-4687},
   url={http://dx.doi.org/10.1038/s41586-020-03051-4},
   DOI={10.1038/s41586-020-03051-4},
   number={7839},
   journal={Nature},
   publisher={Springer Science and Business Media LLC},
   author={Schrittwieser, Julian and Antonoglou, Ioannis and Hubert, Thomas and Simonyan, Karen and Sifre, Laurent and Schmitt, Simon and Guez, Arthur and Lockhart, Edward and Hassabis, Demis and Graepel, Thore and Lillicrap, Timothy and Silver, David},
   year={2020},
   month=dec, pages={604–609} }

@misc{gottweis2025aicoscientist,
      title={Towards an AI co-scientist}, 
      author={Juraj Gottweis and Wei-Hung Weng and Alexander Daryin and Tao Tu and Anil Palepu and Petar Sirkovic and Artiom Myaskovsky and Felix Weissenberger and Keran Rong and Ryutaro Tanno and Khaled Saab and Dan Popovici and Jacob Blum and Fan Zhang and Katherine Chou and Avinatan Hassidim and Burak Gokturk and Amin Vahdat and Pushmeet Kohli and Yossi Matias and Andrew Carroll and Kavita Kulkarni and Nenad Tomasev and Yuan Guan and Vikram Dhillon and Eeshit Dhaval Vaishnav and Byron Lee and Tiago R D Costa and José R Penadés and Gary Peltz and Yunhan Xu and Annalisa Pawlosky and Alan Karthikesalingam and Vivek Natarajan},
      year={2025},
      eprint={2502.18864},
      archivePrefix={arXiv},
      primaryClass={cs.AI},
      url={https://arxiv.org/abs/2502.18864}, 
}

@article{wei2022chain,
  title={Chain-of-thought prompting elicits reasoning in large language models},
  author={Wei, Jason and Wang, Xuezhi and Schuurmans, Dale and Bosma, Maarten and Xia, Fei and Chi, Ed and Le, Quoc V and Zhou, Denny and others},
  journal={Advances in neural information processing systems},
  volume={35},
  pages={24824--24837},
  year={2022}
}

@misc{irie2021goinglineartransformersrecurrent,
      title={Going Beyond Linear Transformers with Recurrent Fast Weight Programmers}, 
      author={Kazuki Irie and Imanol Schlag and Róbert Csordás and Jürgen Schmidhuber},
      year={2021},
      eprint={2106.06295},
      archivePrefix={arXiv},
      primaryClass={cs.LG},
      url={https://arxiv.org/abs/2106.06295}, 
}

@misc{irie2023practicalcomputationalpowerlinear,
      title={Practical Computational Power of Linear Transformers and Their Recurrent and Self-Referential Extensions}, 
      author={Kazuki Irie and Róbert Csordás and Jürgen Schmidhuber},
      year={2023},
      eprint={2310.16076},
      archivePrefix={arXiv},
      primaryClass={cs.LG},
      url={https://arxiv.org/abs/2310.16076}, 
}

@inproceedings{
ahn2024autortembodiedfoundationmodels,
title={Auto{RT}: Embodied Foundation Models for Large Scale Orchestration of Robotic Agents},
author={Michael Ahn and Debidatta Dwibedi and Chelsea Finn and Montserrat Gonzalez Arenas and Keerthana Gopalakrishnan and Karol Hausman and brian ichter and Alex Irpan and Nikhil J Joshi and Ryan Julian and Sean Kirmani and Isabel Leal and Tsang-Wei Edward Lee and Sergey Levine and Yao Lu and sharath maddineni and Kanishka Rao and Dorsa Sadigh and Pannag R Sanketi and Pierre Sermanet and Quan Vuong and Stefan Welker and Fei Xia and Ted Xiao and Peng Xu and Sichun Xu and Zhuo Xu},
booktitle={First Workshop on Vision-Language Models for Navigation and Manipulation at ICRA 2024},
year={2024},
url={https://openreview.net/forum?id=DYcCveNeR1}
}

@inproceedings{
patil2025the,
title={The Berkeley Function Calling Leaderboard ({BFCL}): From Tool Use to Agentic Evaluation of Large Language Models},
author={Shishir G Patil and Huanzhi Mao and Fanjia Yan and Charlie Cheng-Jie Ji and Vishnu Suresh and Ion Stoica and Joseph E. Gonzalez},
booktitle={Forty-second International Conference on Machine Learning},
year={2025},
url={https://openreview.net/forum?id=2GmDdhBdDk}
}

@inproceedings{trivedi2024appworldcontrollableworldapps,
    title = "{A}pp{W}orld: A Controllable World of Apps and People for Benchmarking Interactive Coding Agents",
    author = "Trivedi, Harsh  and
      Khot, Tushar  and
      Hartmann, Mareike  and
      Manku, Ruskin  and
      Dong, Vinty  and
      Li, Edward  and
      Gupta, Shashank  and
      Sabharwal, Ashish  and
      Balasubramanian, Niranjan",
    editor = "Ku, Lun-Wei  and
      Martins, Andre  and
      Srikumar, Vivek",
    booktitle = "Proceedings of the 62nd Annual Meeting of the Association for Computational Linguistics (Volume 1: Long Papers)",
    month = aug,
    year = "2024",
    address = "Bangkok, Thailand",
    publisher = "Association for Computational Linguistics",
    url = "https://aclanthology.org/2024.acl-long.850/",
    doi = "10.18653/v1/2024.acl-long.850",
    pages = "16022--16076",

}

@inproceedings{
gu2023maniskill2unifiedbenchmarkgeneralizable,
title={ManiSkill2: A Unified Benchmark for Generalizable Manipulation Skills},
author={Jiayuan Gu and Fanbo Xiang and Xuanlin Li and Zhan Ling and Xiqiang Liu and Tongzhou Mu and Yihe Tang and Stone Tao and Xinyue Wei and Yunchao Yao and Xiaodi Yuan and Pengwei Xie and Zhiao Huang and Rui Chen and Hao Su},
booktitle={The Eleventh International Conference on Learning Representations },
year={2023},
url={https://openreview.net/forum?id=b_CQDy9vrD1}
}

@inproceedings{
yang2025embodiedbenchcomprehensivebenchmarkingmultimodal,
title={EmbodiedBench: Comprehensive Benchmarking Multi-modal Large Language Models for Vision-Driven Embodied Agents},
author={Rui Yang and Hanyang Chen and Junyu Zhang and Mark Zhao and Cheng Qian and Kangrui Wang and Qineng Wang and Teja Venkat Koripella and Marziyeh Movahedi and Manling Li and Heng Ji and Huan Zhang and Tong Zhang},
booktitle={Forty-second International Conference on Machine Learning},
year={2025},
url={https://openreview.net/forum?id=DgGF2LEBPS}
}

@misc{yin2025safeagentbenchbenchmarksafetask,
      title={SafeAgentBench: A Benchmark for Safe Task Planning of Embodied LLM Agents}, 
      author={Sheng Yin and Xianghe Pang and Yuanzhuo Ding and Menglan Chen and Yutong Bi and Yichen Xiong and Wenhao Huang and Zhen Xiang and Jing Shao and Siheng Chen},
      year={2025},
      eprint={2412.13178},
      archivePrefix={arXiv},
      primaryClass={cs.CR},
      url={https://arxiv.org/abs/2412.13178}, 
}

@inproceedings{
jansen2024discoveryworldvirtualenvironmentdeveloping,
title={DiscoveryWorld: A Virtual Environment for Developing and Evaluating Automated Scientific Discovery Agents},
author={Peter Jansen and Marc-Alexandre C{\^o}t{\'e} and Tushar Khot and Erin Bransom and Bhavana Dalvi Mishra and Bodhisattwa Prasad Majumder and Oyvind Tafjord and Peter Clark},
booktitle={The Thirty-eight Conference on Neural Information Processing Systems Datasets and Benchmarks Track},
year={2024},
url={https://openreview.net/forum?id=cDYqckEt6d}
}

@article{
siegel2024corebenchfosteringcredibilitypublished,
title={{CORE}-Bench: Fostering the Credibility of Published Research Through a Computational Reproducibility Agent Benchmark},
author={Zachary S Siegel and Sayash Kapoor and Nitya Nadgir and Benedikt Stroebl and Arvind Narayanan},
journal={Transactions on Machine Learning Research},
issn={2835-8856},
year={2024},
url={https://openreview.net/forum?id=BsMMc4MEGS},
note={}
}

@misc{li2024learningcontrastivepromptsautomated,
      title={Learning from Contrastive Prompts: Automated Optimization and Adaptation}, 
      author={Mingqi Li and Karan Aggarwal and Yong Xie and Aitzaz Ahmad and Stephen Lau},
      year={2024},
      eprint={2409.15199},
      archivePrefix={arXiv},
      primaryClass={cs.CL},
      url={https://arxiv.org/abs/2409.15199}, 
}

@misc{li2025droinstructzerodistributionallyrobustprompt,
      title={DRO-InstructZero: Distributionally Robust Prompt Optimization for Large Language Models}, 
      author={Yangyang Li},
      year={2025},
      eprint={2510.15260},
      archivePrefix={arXiv},
      primaryClass={cs.LG},
      url={https://arxiv.org/abs/2510.15260}, 
}

@misc{tao2025delvepodirectionguidedselfevolvingframework,
      title={DelvePO: Direction-Guided Self-Evolving Framework for Flexible Prompt Optimization}, 
      author={Tao Tao and Guanghui Zhu and Lang Guo and Hongyi Chen and Chunfeng Yuan and Yihua Huang},
      year={2025},
      eprint={2510.18257},
      archivePrefix={arXiv},
      primaryClass={cs.CL},
      url={https://arxiv.org/abs/2510.18257}, 
}

@inproceedings{
han2025mapgdmultiagentpromptgradient,
title={{MAPGD}: Multi-Agent Prompt Gradient Descent for Collaborative Prompt Optimization},
author={Yichen Han and Bojun Liu and Zhengpeng zhou and Guanyu Liu and Zeng Zhang and Yang Yang and Wenli Wang and Isaac N Shi and Yunyan and Lewei He and TIANYU SHI},
booktitle={Workshop on Scaling Environments for Agents},
year={2025},
url={https://openreview.net/forum?id=FywYwwH5z9}
}

@misc{
xu2025pick,
title={Pick Your Textual Gradients},
author={Yifan Xu and Yixuan Li and Xinzhuo Li and Lijun Yu and Haohan Wang},
year={2025},
url={https://openreview.net/forum?id=ydTwv5D536}
}

@inproceedings{
ding2025scalingtextualgradientssamplingbased,
title={Scaling Textual Gradients via Sampling-Based Momentum},
author={Zixin Ding and Junyuan Hong and Jiachen T. Wang and Zinan Lin and Zhangyang Wang and Yuxin Chen},
booktitle={Second Workshop on Test-Time Adaptation: Putting Updates to the Test! at ICML 2025},
year={2025},
url={https://openreview.net/forum?id=Hd8KbY52FF}
}

@inproceedings{
cheng2024traceautodiffgenerativeoptimization,
title={Trace is the Next AutoDiff: Generative Optimization with Rich Feedback, Execution Traces, and {LLM}s},
author={Ching-An Cheng and Allen Nie and Adith Swaminathan},
booktitle={The Thirty-eighth Annual Conference on Neural Information Processing Systems},
year={2024},
url={https://openreview.net/forum?id=rYs2Dmn9tD}
}

@misc{wang2024correctlysemanticbackpropagationlanguagebased,
      title={How to Correctly do Semantic Backpropagation on Language-based Agentic Systems}, 
      author={Wenyi Wang and Hisham A. Alyahya and Dylan R. Ashley and Oleg Serikov and Dmitrii Khizbullin and Francesco Faccio and Jürgen Schmidhuber},
      year={2024},
      eprint={2412.03624},
      archivePrefix={arXiv},
      primaryClass={cs.AI},
      url={https://arxiv.org/abs/2412.03624}, 
}

@misc{zhao2025autooptimizepromptsdomaintasks,
      title={How to Auto-optimize Prompts for Domain Tasks? Adaptive Prompting and Reasoning through Evolutionary Domain Knowledge Adaptation}, 
      author={Yang Zhao and Pu Wang and Hao Frank Yang},
      year={2025},
      eprint={2510.21148},
      archivePrefix={arXiv},
      primaryClass={cs.AI},
      url={https://arxiv.org/abs/2510.21148}, 
}

@article{chen2024promptoptimizationmultisteptasks,
  publtype={informal},
  author={Yongchao Chen and Jacob Arkin and Yilun Hao and Yang Zhang and Nicholas Roy and Chuchu Fan},
  title={PRompt Optimization in Multi-Step Tasks (PROMST): Integrating Human Feedback and Preference Alignment},
  year={2024},
  cdate={1704067200000},
  journal={CoRR},
  volume={abs/2402.08702},
  url={https://doi.org/10.48550/arXiv.2402.08702}
}

@inproceedings{opsahlong2024optimizinginstructionsdemonstrationsmultistage,
    title = "Optimizing Instructions and Demonstrations for Multi-Stage Language Model Programs",
    author = "Opsahl-Ong, Krista  and
      Ryan, Michael J  and
      Purtell, Josh  and
      Broman, David  and
      Potts, Christopher  and
      Zaharia, Matei  and
      Khattab, Omar",
    editor = "Al-Onaizan, Yaser  and
      Bansal, Mohit  and
      Chen, Yun-Nung",
    booktitle = "Proceedings of the 2024 Conference on Empirical Methods in Natural Language Processing",
    month = nov,
    year = "2024",
    address = "Miami, Florida, USA",
    publisher = "Association for Computational Linguistics",
    url = "https://aclanthology.org/2024.emnlp-main.525/",
    doi = "10.18653/v1/2024.emnlp-main.525",
    pages = "9340--9366",
}

@article{he2025crispomultiaspectcritiquesuggestionguidedautomatic, title={CriSPO: Multi-Aspect Critique-Suggestion-guided Automatic Prompt Optimization for Text Generation}, volume={39}, url={https://ojs.aaai.org/index.php/AAAI/article/view/34575}, DOI={10.1609/aaai.v39i22.34575}, abstractNote={Existing automatic prompt engineering methods are typically designed for discriminative tasks, where new task prompts are iteratively refined with limited feedback from a single metric reflecting a single aspect. However, these approaches are suboptimal for generative tasks, which require more nuanced guidance beyond a single numeric metric to improve the prompt and optimize multiple aspects of the generated text. To address these challenges, we propose a novel multi-aspect Critique-Suggestion-guided automatic Prompt Optimization (CriSPO) approach. CriSPO introduces a critique-suggestion module as its core component. This module spontaneously discovers aspects, and compares generated and reference texts across these aspects, providing specific suggestions for prompt modification. These clear critiques and actionable suggestions guide a receptive optimizer module to make more substantial changes, exploring a broader and more effective search space. To further improve CriSPO with multi-metric optimization, we introduce an Automatic Suffix Tuning (AST) extension to enhance the performance of task prompts across multiple metrics. We evaluate CriSPO on 4 state-of-the-art Large Language Models (LLMs) across 4 summarization and 5 Question Answering (QA) datasets. Extensive experiments show 3-4% ROUGE score improvement on summarization and substantial improvement of various metrics on QA.}, number={22}, journal={Proceedings of the AAAI Conference on Artificial Intelligence}, author={He, Han and Liu, Qianchu and Xu, Lei and Shivade, Chaitanya and Zhang, Yi and Srinivasan, Sundararajan and Kirchhoff, Katrin}, year={2025}, month={Apr.}, pages={24014-24022} }

@misc{
anonymous2025coolprompt,
title={CoolPrompt: An Automatic Prompt Optimization Framework for Large Language Models},
author={Kulin Nikita and Sitkina Alena and Khairullin Artur and Zhuravlev Viktor and Sergey Muravyov},
year={2026},
url={https://openreview.net/forum?id=XGECnjDEcS}
}

@misc{liu2025boostingprivatedomainunderstanding,
      title={Boosting Private Domain Understanding of Efficient MLLMs: A Tuning-free, Adaptive, Universal Prompt Optimization Framework}, 
      author={Jiang Liu and Bolin Li and Haoyuan Li and Tianwei Lin and Wenqiao Zhang and Tao Zhong and Zhelun Yu and Jinghao Wei and Hao Cheng and Wanggui He and Fangxun Shu and Hao Jiang and Zheqi Lv and Juncheng Li and Siliang Tang and Yueting Zhuang},
      year={2025},
      eprint={2412.19684},
      archivePrefix={arXiv},
      primaryClass={cs.AI},
      url={https://arxiv.org/abs/2412.19684}, 
}

@inproceedings{
lin2024promptoptimizationhumanfeedback,
title={Prompt Optimization with Human Feedback},
author={Xiaoqiang Lin and Zhongxiang Dai and Arun Verma and See-Kiong Ng and Patrick Jaillet and Bryan Kian Hsiang Low},
booktitle={ICML 2024 Workshop on Models of Human Feedback for AI Alignment},
year={2024},
url={https://openreview.net/forum?id=344O51eTPN}
}

@inproceedings{zhan2024promptrefinementimagepivot,
    title = "Prompt Refinement with Image Pivot for Text-to-Image Generation",
    author = "Zhan, Jingtao  and
      Ai, Qingyao  and
      Liu, Yiqun  and
      Pan, Yingwei  and
      Yao, Ting  and
      Mao, Jiaxin  and
      Ma, Shaoping  and
      Mei, Tao",
    editor = "Ku, Lun-Wei  and
      Martins, Andre  and
      Srikumar, Vivek",
    booktitle = "Proceedings of the 62nd Annual Meeting of the Association for Computational Linguistics (Volume 1: Long Papers)",
    month = aug,
    year = "2024",
    address = "Bangkok, Thailand",
    publisher = "Association for Computational Linguistics",
    url = "https://aclanthology.org/2024.acl-long.53/",
    doi = "10.18653/v1/2024.acl-long.53",
    pages = "941--954",
}

@misc{bai2026webgymscalingtrainingenvironments,
      title={WebGym: Scaling Training Environments for Visual Web Agents with Realistic Tasks}, 
      author={Hao Bai and Alexey Taymanov and Tong Zhang and Aviral Kumar and Spencer Whitehead},
      year={2026},
      eprint={2601.02439},
      archivePrefix={arXiv},
      primaryClass={cs.LG},
      url={https://arxiv.org/abs/2601.02439}, 
}

@misc{han2026unicornselfimprovingunifiedmultimodal,
      title={UniCorn: Towards Self-Improving Unified Multimodal Models through Self-Generated Supervision}, 
      author={Ruiyan Han and Zhen Fang and XinYu Sun and Yuchen Ma and Ziheng Wang and Yu Zeng and Zehui Chen and Lin Chen and Wenxuan Huang and Wei-Jie Xu and Yi Cao and Feng Zhao},
      year={2026},
      eprint={2601.03193},
      archivePrefix={arXiv},
      primaryClass={cs.CV},
      url={https://arxiv.org/abs/2601.03193}, 
}

@misc{sunil2026ireasonertrajectoryawareintrinsicreasoning,
      title={iReasoner: Trajectory-Aware Intrinsic Reasoning Supervision for Self-Evolving Large Multimodal Models}, 
      author={Meghana Sunil and Manikandarajan Venmathimaran and Muthu Subash Kavitha},
      year={2026},
      eprint={2601.05877},
      archivePrefix={arXiv},
      primaryClass={cs.CL},
      url={https://arxiv.org/abs/2601.05877}, 
}

@misc{han2026structuredreasoninglargelanguage,
      title={Structured Reasoning for Large Language Models}, 
      author={Jinyi Han and Zixiang Di and Zishang Jiang and Ying Liao and Jiaqing Liang and Yongqi Wang and Yanghua Xiao},
      year={2026},
      eprint={2601.07180},
      archivePrefix={arXiv},
      primaryClass={cs.CL},
      url={https://arxiv.org/abs/2601.07180}, 
}

@misc{zhang2026memrlselfevolvingagentsruntime,
      title={MemRL: Self-Evolving Agents via Runtime Reinforcement Learning on Episodic Memory}, 
      author={Shengtao Zhang and Jiaqian Wang and Ruiwen Zhou and Junwei Liao and Yuchen Feng and Weinan Zhang and Ying Wen and Zhiyu Li and Feiyu Xiong and Yutao Qi and Bo Tang and Muning Wen},
      year={2026},
      eprint={2601.03192},
      archivePrefix={arXiv},
      primaryClass={cs.CL},
      url={https://arxiv.org/abs/2601.03192}, 
}

@misc{zhang2026agentdevelreframingselfevolvingllm,
      title={AgentDevel: Reframing Self-Evolving LLM Agents as Release Engineering}, 
      author={Di Zhang},
      year={2026},
      eprint={2601.04620},
      archivePrefix={arXiv},
      primaryClass={cs.AI},
      url={https://arxiv.org/abs/2601.04620}, 
}

@article{Wang_2024,
   title={A survey on large language model based autonomous agents},
   volume={18},
   ISSN={2095-2236},
   url={http://dx.doi.org/10.1007/s11704-024-40231-1},
   DOI={10.1007/s11704-024-40231-1},
   number={6},
   journal={Frontiers of Computer Science},
   publisher={Springer Science and Business Media LLC},
   author={Wang, Lei and Ma, Chen and Feng, Xueyang and Zhang, Zeyu and Yang, Hao and Zhang, Jingsen and Chen, Zhiyuan and Tang, Jiakai and Chen, Xu and Lin, Yankai and Zhao, Wayne Xin and Wei, Zhewei and Wen, Jirong},
   year={2024},
   month=mar }

@inproceedings{
bragg2025astabenchrigorousbenchmarkingai,
title={AstaBench: Rigorous Benchmarking of {AI} Agents with a Scientific Research Suite},
author={Jonathan Bragg and Mike D'Arcy and Nishant Balepur and Dan Bareket and Bhavana Dalvi Mishra and Sergey Feldman and Dany Haddad and Jena D. Hwang and Peter Jansen and Varsha Kishore and Bodhisattwa Prasad Majumder and Aakanksha Naik and Sigal Rahamimov and Kyle Richardson and Amanpreet Singh and Harshit Surana and Aryeh Tiktinsky and Rosni Vasu and Guy Wiener and Chloe Anastasiades and Stefanus Candra and Jason Dunkelberger and Daniel Emery and Rob Evans and Malachi Hamada and Regan Huff and Rodney Kinney and Matt Latzke and Jaron Lochner and Ruben Lozano-Aguilera and Ngoc-Uyen Nguyen and Smita Rao and Amber Tanaka and Brooke Vlahos and Peter Clark and Doug Downey and Yoav Goldberg and Ashish Sabharwal and Daniel S Weld},
booktitle={The Fourteenth International Conference on Learning Representations},
year={2026},
url={https://openreview.net/forum?id=M7TNf5J26u}
}

@inproceedings{chen2025physgym,
title={PhysGym: Benchmarking {LLM}s in Interactive Physics Discovery with Controlled Priors},
author={Yimeng Chen and Piotr Pi{\k{e}}kos and Mateusz Ostaszewski and Firas Laakom and J{\"u}rgen Schmidhuber},
booktitle={The Thirty-ninth Annual Conference on Neural Information Processing Systems Datasets and Benchmarks Track},
year={2025},
url={https://openreview.net/forum?id=w8uII2qAmd}
}

@misc{starace2025paperbenchevaluatingaisability,
      title={PaperBench: Evaluating AI's Ability to Replicate AI Research}, 
      author={Giulio Starace and Oliver Jaffe and Dane Sherburn and James Aung and Jun Shern Chan and Leon Maksin and Rachel Dias and Evan Mays and Benjamin Kinsella and Wyatt Thompson and Johannes Heidecke and Amelia Glaese and Tejal Patwardhan},
      year={2025},
      eprint={2504.01848},
      archivePrefix={arXiv},
      primaryClass={cs.AI},
      url={https://arxiv.org/abs/2504.01848}, 
}

@inproceedings{
abdelnabi2024llmdeliberation,
title={{LLM}-Deliberation: Evaluating {LLM}s with Interactive Multi-Agent Negotiation Game},
author={Sahar Abdelnabi and Amr Gomaa and Sarath Sivaprasad and Lea Sch{\"o}nherr and Mario Fritz},
booktitle={ICLR 2024 Workshop on Large Language Model (LLM) Agents},
year={2024},
url={https://openreview.net/forum?id=eE1WHn6qlk}
}

@inproceedings{
levy2025stwebagentbenchbenchmarkevaluatingsafety,
title={{ST}-WebAgentBench: A Benchmark for Evaluating Safety and Trustworthiness in Web Agents},
author={Ido Levy and Ben wiesel and Sami Marreed and Alon Oved and Avi Yaeli and Segev Shlomov},
booktitle={The Fourteenth International Conference on Learning Representations},
year={2026},
url={https://openreview.net/forum?id=MuCDzH0ctf}
}

@misc{putta2024agentqadvancedreasoning,
      title={Agent Q: Advanced Reasoning and Learning for Autonomous AI Agents}, 
      author={Pranav Putta and Edmund Mills and Naman Garg and Sumeet Motwani and Chelsea Finn and Divyansh Garg and Rafael Rafailov},
      year={2024},
      eprint={2408.07199},
      archivePrefix={arXiv},
      primaryClass={cs.AI},
      url={https://arxiv.org/abs/2408.07199}, 
}

@misc{openai2019dota2largescale,
      title={Dota 2 with Large Scale Deep Reinforcement Learning}, 
      author={OpenAI and : and Christopher Berner and Greg Brockman and Brooke Chan and Vicki Cheung and Przemysław Dębiak and Christy Dennison and David Farhi and Quirin Fischer and Shariq Hashme and Chris Hesse and Rafal Józefowicz and Scott Gray and Catherine Olsson and Jakub Pachocki and Michael Petrov and Henrique P. d. O. Pinto and Jonathan Raiman and Tim Salimans and Jeremy Schlatter and Jonas Schneider and Szymon Sidor and Ilya Sutskever and Jie Tang and Filip Wolski and Susan Zhang},
      year={2019},
      eprint={1912.06680},
      archivePrefix={arXiv},
      primaryClass={cs.LG},
      url={https://arxiv.org/abs/1912.06680}, 
}

@inproceedings{wei2025webagentr1trainingwebagents,
  title={Webagent-r1: Training web agents via end-to-end multi-turn reinforcement learning},
  author={Wei, Zhepei and Yao, Wenlin and Liu, Yao and Zhang, Weizhi and Lu, Qin and Qiu, Liang and Yu, Changlong and Xu, Puyang and Zhang, Chao and Yin, Bing and others},
  booktitle={Proceedings of the 2025 Conference on Empirical Methods in Natural Language Processing},
  pages={7920--7939},
  year={2025}
}

@misc{azam2025reflectionbasedmemorywebnavigation,
      title={Reflection-Based Memory For Web navigation Agents}, 
      author={Ruhana Azam and Aditya Vempaty and Ashish Jagmohan},
      year={2025},
      eprint={2506.02158},
      archivePrefix={arXiv},
      primaryClass={cs.AI},
      url={https://arxiv.org/abs/2506.02158}, 
}

@misc{liu2025webcoachselfevolvingwebagents,
      title={WebCoach: Self-Evolving Web Agents with Cross-Session Memory Guidance}, 
      author={Genglin Liu and Shijie Geng and Sha Li and Hejie Cui and Sarah Zhang and Xin Liu and Tianyi Liu},
      year={2025},
      eprint={2511.12997},
      archivePrefix={arXiv},
      primaryClass={cs.AI},
      url={https://arxiv.org/abs/2511.12997}, 
}

@article{deng2023mindweb,
  title={Mind2web: Towards a generalist agent for the web},
  author={Deng, Xiang and Gu, Yu and Zheng, Boyuan and Chen, Shijie and Stevens, Sam and Wang, Boshi and Sun, Huan and Su, Yu},
  journal={Advances in Neural Information Processing Systems},
  volume={36},
  pages={28091--28114},
  year={2023}
}

@article{deng2025swebenchproaiagents,
  publtype={informal},
  author={Xiang Deng and Jeff Da and Edwin Pan and Yannis Yiming He and Charles Ide and Kanak Garg and Niklas Lauffer and Andrew Park and Nitin Pasari and Chetan Rane and Karmini Sampath and Maya Krishnan and Srivatsa Kundurthy and Sean Hendryx and Zifan Wang and Chen Bo Calvin Zhang and Noah Jacobson and Bing Liu and Brad Kenstler},
  title={SWE-Bench Pro: Can AI Agents Solve Long-Horizon Software Engineering Tasks?},
  year={2025},
  month={September},
  cdate={1756684800000},
  journal={CoRR},
  volume={abs/2509.16941},
  url={https://doi.org/10.48550/arXiv.2509.16941}
}

@misc{qiu2025locobenchagentinteractivebenchmarkllm,
      title={LoCoBench-Agent: An Interactive Benchmark for LLM Agents in Long-Context Software Engineering}, 
      author={Jielin Qiu and Zuxin Liu and Zhiwei Liu and Rithesh Murthy and Jianguo Zhang and Haolin Chen and Shiyu Wang and Ming Zhu and Liangwei Yang and Juntao Tan and Roshan Ram and Akshara Prabhakar and Tulika Awalgaonkar and Zixiang Chen and Zhepeng Cen and Cheng Qian and Shelby Heinecke and Weiran Yao and Silvio Savarese and Caiming Xiong and Huan Wang},
      year={2025},
      eprint={2511.13998},
      archivePrefix={arXiv},
      primaryClass={cs.SE},
      url={https://arxiv.org/abs/2511.13998}, 
}

@misc{ahmed2024tddbenchverifiedllmsgenerate,
      title={TDD-Bench Verified: Can LLMs Generate Tests for Issues Before They Get Resolved?}, 
      author={Toufique Ahmed and Martin Hirzel and Rangeet Pan and Avraham Shinnar and Saurabh Sinha},
      year={2024},
      eprint={2412.02883},
      archivePrefix={arXiv},
      primaryClass={cs.SE},
      url={https://arxiv.org/abs/2412.02883}, 
}

@inproceedings{
zhang2025agentsecuritybenchasb,
title={Agent Security Bench ({ASB}): Formalizing and Benchmarking Attacks and Defenses in {LLM}-based Agents},
author={Hanrong Zhang and Jingyuan Huang and Kai Mei and Yifei Yao and Zhenting Wang and Chenlu Zhan and Hongwei Wang and Yongfeng Zhang},
booktitle={The Thirteenth International Conference on Learning Representations},
year={2025},
url={https://openreview.net/forum?id=V4y0CpX4hK}
}

@misc{zhan2024injecagentbenchmarkingindirectprompt,
      title={InjecAgent: Benchmarking Indirect Prompt Injections in Tool-Integrated Large Language Model Agents}, 
      author={Qiusi Zhan and Zhixiang Liang and Zifan Ying and Daniel Kang},
      year={2024},
      eprint={2403.02691},
      archivePrefix={arXiv},
      primaryClass={cs.CL},
      url={https://arxiv.org/abs/2403.02691}, 
}

@inproceedings{
mündler2025swtbenchtestingvalidatingrealworld,
title={{SWT}-Bench: Testing and Validating Real-World Bug-Fixes with Code Agents},
author={Niels M{\"u}ndler and Mark Niklas Mueller and Jingxuan He and Martin Vechev},
booktitle={The Thirty-eighth Annual Conference on Neural Information Processing Systems},
year={2024},
url={https://openreview.net/forum?id=9Y8zUO11EQ}
}

@misc{he2024openwebvoyagerbuildingmultimodalweb,
      title={OpenWebVoyager: Building Multimodal Web Agents via Iterative Real-World Exploration, Feedback and Optimization}, 
      author={Hongliang He and Wenlin Yao and Kaixin Ma and Wenhao Yu and Hongming Zhang and Tianqing Fang and Zhenzhong Lan and Dong Yu},
      year={2024},
      eprint={2410.19609},
      archivePrefix={arXiv},
      primaryClass={cs.CL},
      url={https://arxiv.org/abs/2410.19609}, 
}

@misc{patel2024largelanguagemodelsselfimprove,
      title={Large Language Models Can Self-Improve At Web Agent Tasks}, 
      author={Ajay Patel and Markus Hofmarcher and Claudiu Leoveanu-Condrei and Marius-Constantin Dinu and Chris Callison-Burch and Sepp Hochreiter},
      year={2024},
      eprint={2405.20309},
      archivePrefix={arXiv},
      primaryClass={cs.LG},
      url={https://arxiv.org/abs/2405.20309}, 
}

@misc{aleithan2024swebenchenhancedcodingbenchmark,
      title={SWE-Bench+: Enhanced Coding Benchmark for LLMs}, 
      author={Reem Aleithan and Haoran Xue and Mohammad Mahdi Mohajer and Elijah Nnorom and Gias Uddin and Song Wang},
      year={2024},
      eprint={2410.06992},
      archivePrefix={arXiv},
      primaryClass={cs.SE},
      url={https://arxiv.org/abs/2410.06992}, 
}

@inproceedings{
jimenez2024swebenchlanguagemodelsresolve,
title={{SWE}-bench: Can Language Models Resolve Real-world Github Issues?},
author={Carlos E Jimenez and John Yang and Alexander Wettig and Shunyu Yao and Kexin Pei and Ofir Press and Karthik R Narasimhan},
booktitle={The Twelfth International Conference on Learning Representations},
year={2024},
url={https://openreview.net/forum?id=VTF8yNQM66}
}

@article{hendrikse2007evolvability,
  title={Evolvability as the proper focus of evolutionary developmental biology},
  author={Hendrikse, Jesse Love and Parsons, Trish Elizabeth and Hallgr{\'\i}msson, Benedikt},
  journal={Evolution \& development},
  volume={9},
  number={4},
  pages={393--401},
  year={2007},
  publisher={Wiley Online Library}
}

@misc{qiu2025alitageneralistagentenabling,
      title={Alita: Generalist Agent Enabling Scalable Agentic Reasoning with Minimal Predefinition and Maximal Self-Evolution}, 
      author={Jiahao Qiu and Xuan Qi and Tongcheng Zhang and Xinzhe Juan and Jiacheng Guo and Yifu Lu and Yimin Wang and Zixin Yao and Qihan Ren and Xun Jiang and Xing Zhou and Dongrui Liu and Ling Yang and Yue Wu and Kaixuan Huang and Shilong Liu and Hongru Wang and Mengdi Wang},
      year={2025},
      eprint={2505.20286},
      archivePrefix={arXiv},
      primaryClass={cs.AI},
      url={https://arxiv.org/abs/2505.20286}, 
}

@article{
zhang2025landscape,
title={The Landscape of Agentic Reinforcement Learning for {LLM}s: A Survey},
author={Guibin Zhang and Hejia Geng and Xiaohang Yu and Zhenfei Yin and Zaibin Zhang and Zelin Tan and Heng Zhou and Zhong-Zhi Li and Xiangyuan Xue and Yijiang Li and Yifan Zhou and Yang Chen and Chen Zhang and Yutao Fan and Zihu Wang and Songtao Huang and Francisco Piedrahita Velez and Yue Liao and Hongru WANG and Mengyue Yang and Heng Ji and Jun Wang and Shuicheng YAN and Philip Torr and LEI BAI},
journal={Transactions on Machine Learning Research},
issn={2835-8856},
year={2026},
url={https://openreview.net/forum?id=RY19y2RI1O},
note={Survey Certification}
}

@article{hoi2021online,
  title={Online learning: A comprehensive survey},
  author={Hoi, Steven CH and Sahoo, Doyen and Lu, Jing and Zhao, Peilin},
  journal={Neurocomputing},
  volume={459},
  pages={249--289},
  year={2021},
  publisher={Elsevier}
}

@misc{sunil2026memorypoisoningattackdefense,
      title={Memory Poisoning Attack and Defense on Memory Based LLM-Agents}, 
      author={Balachandra Devarangadi Sunil and Isheeta Sinha and Piyush Maheshwari and Shantanu Todmal and Shreyan Mallik and Shuchi Mishra},
      year={2026},
      eprint={2601.05504},
      archivePrefix={arXiv},
      primaryClass={cs.CR},
      url={https://arxiv.org/abs/2601.05504}, 
}

@inproceedings{
anonymous2026gaia,
title={Gaia2: Benchmarking {LLM} Agents on Dynamic and  Asynchronous Environments},
author={Romain Froger and Amine Benhalloum and Andrey Rusakov and Dheeraj Mekala and Emilien Garreau and Gerard Moreno-Torres Bertran and Gr{\'e}goire Mialon and Hugo Lauren{\c{c}}on and Jean-Baptiste Gaya and Kunal Malkan and Mathieu Rita and Matteo Bettini and Maxime Lecanu and Mengjue Wang and Pierre Andrews and Pierre Menard and Thomas Scialom and Ulyana Piterbarg and Virginie Do and Amar Budhiraja and Ian Yu and Mikhail Plekhanov and Ricardo Silveira Cabral and Vladislav Vorotilov},
booktitle={The Fourteenth International Conference on Learning Representations},
year={2026},
url={https://openreview.net/forum?id=9gw03JpKK4}
}

@misc{chen2025multiagentevolvellmselfimprove,
      title={Multi-Agent Evolve: LLM Self-Improve through Co-evolution}, 
      author={Yixing Chen and Yiding Wang and Siqi Zhu and Haofei Yu and Tao Feng and Muhan Zhang and Mostofa Patwary and Jiaxuan You},
      year={2025},
      eprint={2510.23595},
      archivePrefix={arXiv},
      primaryClass={cs.AI},
      url={https://arxiv.org/abs/2510.23595}, 
}

@misc{gao2026selfevolvingsyntheticdataverifiablereward,
      title={From Self-Evolving Synthetic Data to Verifiable-Reward RL: Post-Training Multi-turn Interactive Tool-Using Agents}, 
      author={Jiaxuan Gao and Jiaao Chen and Chuyi He and Wei-Chen Wang and Shusheng Xu and Hanrui Wang and Di Jin and Yi Wu},
      year={2026},
      eprint={2601.22607},
      archivePrefix={arXiv},
      primaryClass={cs.AI},
      url={https://arxiv.org/abs/2601.22607}, 
}

@misc{wang2026metagenselfevolvingrolestopologies,
      title={MetaGen: Self-Evolving Roles and Topologies for Multi-Agent LLM Reasoning}, 
      author={Yimeng Wang and Jiaxing Zhao and Hongbin Xie and Hexing Ma and Yuzhen Lei and Shuangxue Liu and Xuan Song and Zichen Zhang and Haoran Zhang},
      year={2026},
      eprint={2601.19290},
      archivePrefix={arXiv},
      primaryClass={cs.CL},
      url={https://arxiv.org/abs/2601.19290}, 
}

@inproceedings{schmidhuber1991possibility,
  title={A possibility for implementing curiosity and boredom in model-building neural controllers},
  author={Schmidhuber, J{\"u}rgen},
  booktitle={Proc. of the international conference on simulation of adaptive behavior: From animals to animats},
  pages={222--227},
  year={1991}
}

@misc{feng2026agentocrreimaginingagenthistory,
      title={AgentOCR: Reimagining Agent History via Optical Self-Compression}, 
      author={Lang Feng and Fuchao Yang and Feng Chen and Xin Cheng and Haiyang Xu and Zhenglin Wan and Ming Yan and Bo An},
      year={2026},
      eprint={2601.04786},
      archivePrefix={arXiv},
      primaryClass={cs.LG},
      url={https://arxiv.org/abs/2601.04786}, 
}

@misc{chhikara2025mem0buildingproductionreadyai,
      title={Mem0: Building Production-Ready AI Agents with Scalable Long-Term Memory}, 
      author={Prateek Chhikara and Dev Khant and Saket Aryan and Taranjeet Singh and Deshraj Yadav},
      year={2025},
      eprint={2504.19413},
      archivePrefix={arXiv},
      primaryClass={cs.CL},
      url={https://arxiv.org/abs/2504.19413}, 
}

@inproceedings{
zhang2025gmemorytracinghierarchicalmemory,
title={G-Memory: Tracing Hierarchical Memory for Multi-Agent Systems},
author={Guibin Zhang and Muxin Fu and Kun Wang and Guancheng Wan and Miao Yu and Shuicheng YAN},
booktitle={The Thirty-ninth Annual Conference on Neural Information Processing Systems},
year={2025},
url={https://openreview.net/forum?id=mmIAp3cVS0}
}

@misc{ouyang2025code2mcptransformingcoderepositories,
      title={Code2MCP: Transforming Code Repositories into MCP Services}, 
      author={Chaoqian Ouyang and Ling Yue and Shimin Di and Libin Zheng and Linan Yue and Shaowu Pan and Jian Yin and Min-Ling Zhang},
      year={2025},
      eprint={2509.05941},
      archivePrefix={arXiv},
      primaryClass={cs.SE},
      url={https://arxiv.org/abs/2509.05941}, 
}

@incollection{atkinson1968human,
  title={Human memory: A proposed system and its control processes},
  author={Atkinson, Richard C and Shiffrin, Richard M},
  booktitle={Psychology of learning and motivation},
  volume={2},
  pages={89--195},
  year={1968},
  publisher={Elsevier}
}

@misc{wu2025humanmemoryaimemory,
      title={From Human Memory to AI Memory: A Survey on Memory Mechanisms in the Era of LLMs}, 
      author={Yaxiong Wu and Sheng Liang and Chen Zhang and Yichao Wang and Yongyue Zhang and Huifeng Guo and Ruiming Tang and Yong Liu},
      year={2025},
      eprint={2504.15965},
      archivePrefix={arXiv},
      primaryClass={cs.IR},
      url={https://arxiv.org/abs/2504.15965}, 
}

@misc{he2025humaninspiredperspectivessurveyai,
      title={Human-inspired Perspectives: A Survey on AI Long-term Memory}, 
      author={Zihong He and Weizhe Lin and Hao Zheng and Fan Zhang and Matt W. Jones and Laurence Aitchison and Xuhai Xu and Miao Liu and Per Ola Kristensson and Junxiao Shen},
      year={2025},
      eprint={2411.00489},
      archivePrefix={arXiv},
      primaryClass={cs.AI},
      url={https://arxiv.org/abs/2411.00489}, 
}

@inproceedings{song2024moviechatdensetokensparse,
  title={Moviechat: From dense token to sparse memory for long video understanding},
  author={Song, Enxin and Chai, Wenhao and Wang, Guanhong and Zhang, Yucheng and Zhou, Haoyang and Wu, Feiyang and Chi, Haozhe and Guo, Xun and Ye, Tian and Zhang, Yanting and others},
  booktitle={Proceedings of the IEEE/CVF Conference on Computer Vision and Pattern Recognition},
  pages={18221--18232},
  year={2024}
}

@inproceedings{cheng2022xmemlongtermvideoobject,
  title={Xmem: Long-term video object segmentation with an atkinson-shiffrin memory model},
  author={Cheng, Ho Kei and Schwing, Alexander G},
  booktitle={European conference on computer vision},
  pages={640--658},
  year={2022},
  organization={Springer}
}

@article{guan2024richelieu,
  title={Richelieu: Self-evolving llm-based agents for ai diplomacy},
  author={Guan, Zhenyu and Kong, Xiangyu and Zhong, Fangwei and Wang, Yizhou},
  journal={Advances in Neural Information Processing Systems},
  volume={37},
  pages={123471--123497},
  year={2024}
}

@inproceedings{
xu2025languageagentsreinforcementlearning,
title={Language Agents with Reinforcement Learning for Strategic Play in the Werewolf Game},
author={Zelai Xu and Chao Yu and Fei Fang and Yu Wang and Yi Wu},
booktitle={Forty-first International Conference on Machine Learning},
year={2024},
url={https://openreview.net/forum?id=usUPvQH3XK}
}

@misc{
anonymous2025llm,
title={{LLM} Coaching {LLM} in Self-Play Training},
author={Ju Qi and Xiaoxi Mao and Jianfeng Wang and Na Mou},
year={2026},
url={https://openreview.net/forum?id=NnEfjLA50a}
}

@inproceedings{
xu2025dipllmfinetuningllmstrategic,
title={Dip{LLM}: Fine-Tuning {LLM} for Strategic Decision-making in Diplomacy},
author={Kaixuan Xu and Jiajun Chai and Sicheng Li and Yuqian Fu and Yuanheng Zhu and Dongbin Zhao},
booktitle={Forty-second International Conference on Machine Learning},
year={2025},
url={https://openreview.net/forum?id=hfPaOxDWfI}
}

@misc{acikgoz2025selfimprovingllmagentstesttime,
      title={Self-Improving LLM Agents at Test-Time}, 
      author={Emre Can Acikgoz and Cheng Qian and Heng Ji and Dilek Hakkani-Tür and Gokhan Tur},
      year={2025},
      eprint={2510.07841},
      archivePrefix={arXiv},
      primaryClass={cs.LG},
      url={https://arxiv.org/abs/2510.07841}, 
}

@inproceedings{
fang2025lightmemlightweightefficientmemoryaugmented,
title={LightMem: Lightweight and Efficient Memory-Augmented Generation},
author={Jizhan Fang and Xinle Deng and Haoming Xu and Ziyan Jiang and Yuqi Tang and Ziwen Xu and Shumin Deng and Yunzhi Yao and Mengru Wang and Shuofei Qiao and Huajun Chen and Ningyu Zhang},
booktitle={The Fourteenth International Conference on Learning Representations},
year={2026},
url={https://openreview.net/forum?id=dyJ0GWpjJB}
}

@misc{wang2025mirixmultiagentmemoryllmbased,
      title={MIRIX: Multi-Agent Memory System for LLM-Based Agents}, 
      author={Yu Wang and Xi Chen},
      year={2025},
      eprint={2507.07957},
      archivePrefix={arXiv},
      primaryClass={cs.CL},
      url={https://arxiv.org/abs/2507.07957}, 
}

@misc{zhou2025mementofinetuningllmagents,
      title={Memento: Fine-tuning LLM Agents without Fine-tuning LLMs}, 
      author={Huichi Zhou and Yihang Chen and Siyuan Guo and Xue Yan and Kin Hei Lee and Zihan Wang and Ka Yiu Lee and Guchun Zhang and Kun Shao and Linyi Yang and Jun Wang},
      year={2025},
      eprint={2508.16153},
      archivePrefix={arXiv},
      primaryClass={cs.LG},
      url={https://arxiv.org/abs/2508.16153}, 
}

@inproceedings{lewis2021retrievalaugmentedgenerationknowledgeintensivenlp,
author = {Lewis, Patrick and Perez, Ethan and Piktus, Aleksandra and Petroni, Fabio and Karpukhin, Vladimir and Goyal, Naman and K\"{u}ttler, Heinrich and Lewis, Mike and Yih, Wen-tau and Rockt\"{a}schel, Tim and Riedel, Sebastian and Kiela, Douwe},
title = {Retrieval-augmented generation for knowledge-intensive NLP tasks},
year = {2020},
isbn = {9781713829546},
publisher = {Curran Associates Inc.},
address = {Red Hook, NY, USA},
booktitle = {Proceedings of the 34th International Conference on Neural Information Processing Systems},
articleno = {793},
numpages = {16},
location = {Vancouver, BC, Canada},
series = {NIPS '20}
}

@inproceedings{
agrawal2025gepareflectivepromptevolution,
title={{GEPA}: Reflective Prompt Evolution Can Outperform Reinforcement Learning},
author={Lakshya A Agrawal and Shangyin Tan and Dilara Soylu and Noah Ziems and Rishi Khare and Krista Opsahl-Ong and Arnav Singhvi and Herumb Shandilya and Michael J Ryan and Meng Jiang and Christopher Potts and Koushik Sen and Alex Dimakis and Ion Stoica and Dan Klein and Matei Zaharia and Omar Khattab},
booktitle={The Fourteenth International Conference on Learning Representations},
year={2026},
url={https://openreview.net/forum?id=RQm2KQTM5r}
}

@misc{zhang2025agenticcontextengineeringevolving,
      title={Agentic Context Engineering: Evolving Contexts for Self-Improving Language Models}, 
      author={Qizheng Zhang and Changran Hu and Shubhangi Upasani and Boyuan Ma and Fenglu Hong and Vamsidhar Kamanuru and Jay Rainton and Chen Wu and Mengmeng Ji and Hanchen Li and Urmish Thakker and James Zou and Kunle Olukotun},
      year={2025},
      eprint={2510.04618},
      archivePrefix={arXiv},
      primaryClass={cs.LG},
      url={https://arxiv.org/abs/2510.04618}, 
}

@misc{singh2025agenticretrievalaugmentedgenerationsurvey,
      title={Agentic Retrieval-Augmented Generation: A Survey on Agentic RAG}, 
      author={Aditi Singh and Abul Ehtesham and Saket Kumar and Tala Talaei Khoei},
      year={2025},
      eprint={2501.09136},
      archivePrefix={arXiv},
      primaryClass={cs.AI},
      url={https://arxiv.org/abs/2501.09136}, 
}

@inproceedings{zhong2024memorybank,
  title={Memorybank: Enhancing large language models with long-term memory},
  author={Zhong, Wanjun and Guo, Lianghong and Gao, Qiqi and Ye, He and Wang, Yanlin},
  booktitle={Proceedings of the AAAI conference on artificial intelligence},
  volume={38},
  pages={19724--19731},
  year={2024}
}

@inproceedings{lindenbauer2025knowledgenoisectimroverpitfalls,
    title = "From Knowledge to Noise: {CTIM}-Rover and the Pitfalls of Episodic Memory in Software Engineering Agents",
    author = "Lindenbauer, Tobias  and
      Groh, Georg  and
      Schuetze, Hinrich",
    editor = "Kamalloo, Ehsan  and
      Gontier, Nicolas  and
      Lu, Xing Han  and
      Dziri, Nouha  and
      Murty, Shikhar  and
      Lacoste, Alexandre",
    booktitle = "Proceedings of the 1st Workshop for Research on Agent Language Models (REALM 2025)",
    month = jul,
    year = "2025",
    address = "Vienna, Austria",
    publisher = "Association for Computational Linguistics",
    url = "https://aclanthology.org/2025.realm-1.30/",
    doi = "10.18653/v1/2025.realm-1.30",
    pages = "411--427",
    ISBN = "979-8-89176-264-0",
}

@inproceedings{
park2025mrsteveinstructionfollowingagentsminecraft,
title={MrSteve: Instruction-Following Agents in Minecraft with What-Where-When Memory},
author={Junyeong Park and Junmo Cho and Sungjin Ahn},
booktitle={The Thirteenth International Conference on Learning Representations},
year={2025},
url={https://openreview.net/forum?id=CjXaMI2kUH}
}

@inproceedings{tan2025prospectretrospectreflectivememory,
    title = "In Prospect and Retrospect: Reflective Memory Management for Long-term Personalized Dialogue Agents",
    author = "Tan, Zhen  and
      Yan, Jun  and
      Hsu, I-Hung  and
      Han, Rujun  and
      Wang, Zifeng  and
      Le, Long  and
      Song, Yiwen  and
      Chen, Yanfei  and
      Palangi, Hamid  and
      Lee, George  and
      Iyer, Anand Rajan  and
      Chen, Tianlong  and
      Liu, Huan  and
      Lee, Chen-Yu  and
      Pfister, Tomas",
    editor = "Che, Wanxiang  and
      Nabende, Joyce  and
      Shutova, Ekaterina  and
      Pilehvar, Mohammad Taher",
    booktitle = "Proceedings of the 63rd Annual Meeting of the Association for Computational Linguistics (Volume 1: Long Papers)",
    month = jul,
    year = "2025",
    address = "Vienna, Austria",
    publisher = "Association for Computational Linguistics",
    url = "https://aclanthology.org/2025.acl-long.413/",
    doi = "10.18653/v1/2025.acl-long.413",
    pages = "8416--8439",
    ISBN = "979-8-89176-251-0",

}

@ARTICLE{10531671,
  author={Song, Yaoxian and Sun, Penglei and Liu, Haoyu and Li, Zhixu and Song, Wei and Xiao, Yanghua and Zhou, Xiaofang},
  journal={IEEE Transactions on Knowledge and Data Engineering}, 
  title={Scene-Driven Multimodal Knowledge Graph Construction for Embodied AI}, 
  year={2024},
  volume={36},
  number={11},
  pages={6962-6976},
  doi={10.1109/TKDE.2024.3399746}
}

@inproceedings{10.1145/3746027.3755537,
author = {Chu, Meng and Li, Yicong and Chua, Tat-Seng},
title = {GraphVideoAgent: Enhancing Long-form Video Understanding with Entity Relation Graphs},
year = {2025},
isbn = {9798400720352},
publisher = {Association for Computing Machinery},
address = {New York, NY, USA},
url = {https://doi.org/10.1145/3746027.3755537},
doi = {10.1145/3746027.3755537},
booktitle = {Proceedings of the 33rd ACM International Conference on Multimedia},
pages = {4639–4648},
numpages = {10},
location = {Dublin, Ireland},
series = {MM '25}
}

@inproceedings{wang2025causalragintegratingcausalgraphs,
    title = "{C}ausal{RAG}: Integrating Causal Graphs into Retrieval-Augmented Generation",
    author = "Wang, Nengbo  and
      Han, Xiaotian  and
      Singh, Jagdip  and
      Ma, Jing  and
      Chaudhary, Vipin",
    editor = "Che, Wanxiang  and
      Nabende, Joyce  and
      Shutova, Ekaterina  and
      Pilehvar, Mohammad Taher",
    booktitle = "Findings of the Association for Computational Linguistics: ACL 2025",
    month = jul,
    year = "2025",
    address = "Vienna, Austria",
    publisher = "Association for Computational Linguistics",
    url = "https://aclanthology.org/2025.findings-acl.1165/",
    doi = "10.18653/v1/2025.findings-acl.1165",
    pages = "22680--22693",
    ISBN = "979-8-89176-256-5",
}

@misc{wu2025sgmemsentencegraphmemory,
      title={SGMem: Sentence Graph Memory for Long-Term Conversational Agents}, 
      author={Yaxiong Wu and Yongyue Zhang and Sheng Liang and Yong Liu},
      year={2025},
      eprint={2509.21212},
      archivePrefix={arXiv},
      primaryClass={cs.CL},
      url={https://arxiv.org/abs/2509.21212}, 
}

@misc{helmi2025decentralizingaimemoryshimi,
      title={Decentralizing AI Memory: SHIMI, a Semantic Hierarchical Memory Index for Scalable Agent Reasoning}, 
      author={Tooraj Helmi},
      year={2025},
      eprint={2504.06135},
      archivePrefix={arXiv},
      primaryClass={cs.AI},
      url={https://arxiv.org/abs/2504.06135}, 
}

@misc{huang2025surveyfoundationmodelpoweredrecommender,
      title={A Survey of Foundation Model-Powered Recommender Systems: From Feature-Based, Generative to Agentic Paradigms}, 
      author={Chengkai Huang and Hongtao Huang and Tong Yu and Kaige Xie and Junda Wu and Shuai Zhang and Julian Mcauley and Dietmar Jannach and Lina Yao},
      year={2025},
      eprint={2504.16420},
      archivePrefix={arXiv},
      primaryClass={cs.IR},
      url={https://arxiv.org/abs/2504.16420}, 
}

@misc{sang2025pipelinessurveyparadigmshift,
      title={Beyond Pipelines: A Survey of the Paradigm Shift toward Model-Native Agentic AI}, 
      author={Jitao Sang and Jinlin Xiao and Jiarun Han and Jilin Chen and Xiaoyi Chen and Shuyu Wei and Yongjie Sun and Yuhang Wang},
      year={2025},
      eprint={2510.16720},
      archivePrefix={arXiv},
      primaryClass={cs.AI},
      url={https://arxiv.org/abs/2510.16720}, 
}

@article{
zhang2025largelanguagemodelbrainedgui,
title={Large Language Model-Brained {GUI} Agents: A Survey},
author={Chaoyun Zhang and Shilin He and Jiaxu Qian and Bowen Li and Liqun Li and Si Qin and Yu Kang and Minghua Ma and Guyue Liu and Qingwei Lin and Saravan Rajmohan and Dongmei Zhang and Qi Zhang},
journal={Transactions on Machine Learning Research},
issn={2835-8856},
year={2025},
url={https://openreview.net/forum?id=xChvYjvXTp},
note={}
}

@article{shapiro2023conceptual,
  title={Conceptual framework for autonomous cognitive entities},
  author={Shapiro, David and Li, Wangfan and Delaflor, Manuel and Toxtli, Carlos},
  journal={arXiv preprint arXiv:2310.06775},
  year={2023}
}

@misc{zeng2025toolacermodelawareiterativetraining,
      title={ToolACE-R: Model-aware Iterative Training and Adaptive Refinement for Tool Learning}, 
      author={Xingshan Zeng and Weiwen Liu and Xu Huang and Zezhong Wang and Lingzhi Wang and Liangyou Li and Yasheng Wang and Lifeng Shang and Xin Jiang and Ruiming Tang and Qun Liu},
      year={2025},
      eprint={2504.01400},
      archivePrefix={arXiv},
      primaryClass={cs.CL},
      url={https://arxiv.org/abs/2504.01400}, 
}

@inproceedings{
wang2025toolgenunifiedtoolretrieval,
title={ToolGen: Unified Tool Retrieval and Calling via Generation},
author={Renxi Wang and Xudong Han and Lei Ji and Shu Wang and Timothy Baldwin and Haonan Li},
booktitle={The Thirteenth International Conference on Learning Representations},
year={2025},
url={https://openreview.net/forum?id=XLMAMmowdY}
}

@inproceedings{
hong2025deepeyesv2agenticmultimodalmodel,
title={DeepEyesV2: Toward Agentic Multimodal Model},
author={Jack Hong and Chenxiao Zhao and Weiheng Lu and ChengLIn Zhu and Guohai Xu and XingYu},
booktitle={The Fourteenth International Conference on Learning Representations},
year={2026},
url={https://openreview.net/forum?id=yDKawwfJ5O}
}

@misc{lumer2025tooltoagentretrievalbridgingtools,
      title={Tool-to-Agent Retrieval: Bridging Tools and Agents for Scalable LLM Multi-Agent Systems}, 
      author={Elias Lumer and Faheem Nizar and Anmol Gulati and Pradeep Honaganahalli Basavaraju and Vamse Kumar Subbiah},
      year={2025},
      eprint={2511.01854},
      archivePrefix={arXiv},
      primaryClass={cs.CL},
      url={https://arxiv.org/abs/2511.01854}, 
}

@inproceedings{
liu2025toolplannertaskplanningclusters,
title={Tool-Planner: Task Planning with Clusters across Multiple Tools},
author={Yanming Liu and Xinyue Peng and Jiannan Cao and Shi Bo and Yuwei Zhang and Xuhong Zhang and Sheng Cheng and Xun Wang and Jianwei Yin and Tianyu Du},
booktitle={The Thirteenth International Conference on Learning Representations},
year={2025},
url={https://openreview.net/forum?id=dRz3cizftU}
}

@misc{li2025deepagentgeneralreasoningagent,
      title={DeepAgent: A General Reasoning Agent with Scalable Toolsets}, 
      author={Xiaoxi Li and Wenxiang Jiao and Jiarui Jin and Guanting Dong and Jiajie Jin and Yinuo Wang and Hao Wang and Yutao Zhu and Ji-Rong Wen and Yuan Lu and Zhicheng Dou},
      year={2025},
      eprint={2510.21618},
      archivePrefix={arXiv},
      primaryClass={cs.AI},
      url={https://arxiv.org/abs/2510.21618}, 
}

@misc{lumer2025memtooloptimizingshorttermmemory,
      title={MemTool: Optimizing Short-Term Memory Management for Dynamic Tool Calling in LLM Agent Multi-Turn Conversations}, 
      author={Elias Lumer and Anmol Gulati and Vamse Kumar Subbiah and Pradeep Honaganahalli Basavaraju and James A. Burke},
      year={2025},
      eprint={2507.21428},
      archivePrefix={arXiv},
      primaryClass={cs.CL},
      url={https://arxiv.org/abs/2507.21428}, 
}

@inproceedings{zhang2025asktoactenhancingllmstool,
    title = "{A}sk{T}o{A}ct: Enhancing {LLM}s Tool Use via Self-Correcting Clarification",
    author = "Zhang, Xuan  and
      Shen, Yongliang  and
      Zheng, Zhe  and
      Wu, Linjuan  and
      Zhang, Wenqi  and
      Yan, Yuchen  and
      Peng, Qiuying  and
      Wang, Jun  and
      Lu, Weiming",
    editor = "Christodoulopoulos, Christos  and
      Chakraborty, Tanmoy  and
      Rose, Carolyn  and
      Peng, Violet",
    booktitle = "Proceedings of the 2025 Conference on Empirical Methods in Natural Language Processing",
    month = nov,
    year = "2025",
    address = "Suzhou, China",
    publisher = "Association for Computational Linguistics",
    url = "https://aclanthology.org/2025.emnlp-main.682/",
    doi = "10.18653/v1/2025.emnlp-main.682",
    pages = "13484--13511",
    ISBN = "979-8-89176-332-6",
}

@misc{fei2025mcpzeroactivetooldiscovery,
      title={MCP-Zero: Active Tool Discovery for Autonomous LLM Agents}, 
      author={Xiang Fei and Xiawu Zheng and Hao Feng},
      year={2025},
      eprint={2506.01056},
      archivePrefix={arXiv},
      primaryClass={cs.AI},
      url={https://arxiv.org/abs/2506.01056}, 
}

@inproceedings{
qu2025explorationmasteryenablingllms,
title={From Exploration to Mastery: Enabling {LLM}s to Master Tools via Self-Driven Interactions},
author={Changle Qu and Sunhao Dai and Xiaochi Wei and Hengyi Cai and Shuaiqiang Wang and Dawei Yin and Jun Xu and Ji-Rong Wen},
booktitle={The Thirteenth International Conference on Learning Representations},
year={2025},
url={https://openreview.net/forum?id=QKBu1BOAwd}
}

@misc{liu2024toolnetconnectinglargelanguage,
      title={ToolNet: Connecting Large Language Models with Massive Tools via Tool Graph}, 
      author={Xukun Liu and Zhiyuan Peng and Xiaoyuan Yi and Xing Xie and Lirong Xiang and Yuchen Liu and Dongkuan Xu},
      year={2024},
      eprint={2403.00839},
      archivePrefix={arXiv},
      primaryClass={cs.AI},
      url={https://arxiv.org/abs/2403.00839}, 
}

@misc{lin2025masstoolmultitasksearchbasedtool,
      title={MassTool: A Multi-Task Search-Based Tool Retrieval Framework for Large Language Models}, 
      author={Jianghao Lin and Xinyuan Wang and Xinyi Dai and Menghui Zhu and Bo Chen and Ruiming Tang and Yong Yu and Weinan Zhang},
      year={2025},
      eprint={2507.00487},
      archivePrefix={arXiv},
      primaryClass={cs.IR},
      url={https://arxiv.org/abs/2507.00487}, 
}

@inproceedings{
li2025intheflowagenticoptimizationeffective,
title={In-The-Flow Agentic System Optimization for Effective Planning and Tool Use},
author={Zhuofeng Li and Haoxiang Zhang and Seungju Han and Sheng Liu and Jianwen Xie and Yu Zhang and Yejin Choi and James Zou and Pan Lu},
booktitle={The Fourteenth International Conference on Learning Representations},
year={2026},
url={https://openreview.net/forum?id=Mf5AleTUVK}
}

@misc{wang2025mcpflowfacilitatingllmagents,
      title={MCP-Flow: Facilitating LLM Agents to Master Real-World, Diverse and Scaling MCP Tools}, 
      author={Wenhao Wang and Peizhi Niu and Zhao Xu and Zhaoyu Chen and Jian Du and Yaxin Du and Xianghe Pang and Keduan Huang and Yanfeng Wang and Qiang Yan and Siheng Chen},
      year={2025},
      eprint={2510.24284},
      archivePrefix={arXiv},
      primaryClass={cs.AI},
      url={https://arxiv.org/abs/2510.24284}, 
}

@misc{sun2025hierarchicalmemoryhighefficiencylongterm,
      title={Hierarchical Memory for High-Efficiency Long-Term Reasoning in LLM Agents}, 
      author={Haoran Sun and Shaoning Zeng},
      year={2025},
      eprint={2507.22925},
      archivePrefix={arXiv},
      primaryClass={cs.CL},
      url={https://arxiv.org/abs/2507.22925}, 
}

@misc{sun2025enhancinglatentcomputationtransformers,
      title={Enhancing Latent Computation in Transformers with Latent Tokens}, 
      author={Yuchang Sun and Yanxi Chen and Yaliang Li and Bolin Ding},
      year={2025},
      eprint={2505.12629},
      archivePrefix={arXiv},
      primaryClass={cs.LG},
      url={https://arxiv.org/abs/2505.12629}, 
}

@inproceedings{
liu2024deliberationlatentspacedifferentiable,
title={Deliberation in Latent Space via Differentiable Cache Augmentation},
author={Luyang Liu and Jonas Pfeiffer and Jiaxing Wu and Jun Xie and Arthur Szlam},
booktitle={Forty-second International Conference on Machine Learning},
year={2025},
url={https://openreview.net/forum?id=IaUJl5RCOu}
}

@InProceedings{wang2025mextendingmemoryllmscalable,
  title = 	 {M+: Extending {M}emory{LLM} with Scalable Long-Term Memory},
  author =       {Wang, Yu and Krotov, Dmitry and Hu, Yuanzhe and Gao, Yifan and Zhou, Wangchunshu and Mcauley, Julian and Gutfreund, Dan and Feris, Rogerio and He, Zexue},
  booktitle = 	 {Proceedings of the 42nd International Conference on Machine Learning},
  pages = 	 {63308--63323},
  year = 	 {2025},
  editor = 	 {Singh, Aarti and Fazel, Maryam and Hsu, Daniel and Lacoste-Julien, Simon and Berkenkamp, Felix and Maharaj, Tegan and Wagstaff, Kiri and Zhu, Jerry},
  volume = 	 {267},
  series = 	 {Proceedings of Machine Learning Research},
  month = 	 {13--19 Jul},
  url = 	 {https://proceedings.mlr.press/v267/wang25au.html},
}

@misc{wang2024memoryllmselfupdatablelargelanguage,
      title={MEMORYLLM: Towards Self-Updatable Large Language Models}, 
      author={Yu Wang and Yifan Gao and Xiusi Chen and Haoming Jiang and Shiyang Li and Jingfeng Yang and Qingyu Yin and Zheng Li and Xian Li and Bing Yin and Jingbo Shang and Julian McAuley},
      year={2024},
      eprint={2402.04624},
      archivePrefix={arXiv},
      primaryClass={cs.CL},
      url={https://arxiv.org/abs/2402.04624}, 
}

@misc{dillon2025contextualmemoryreweavinglarge,
      title={Contextual Memory Reweaving in Large Language Models Using Layered Latent State Reconstruction}, 
      author={Frederick Dillon and Gregor Halvorsen and Simon Tattershall and Magnus Rowntree and Gareth Vanderpool},
      year={2025},
      eprint={2502.02046},
      archivePrefix={arXiv},
      primaryClass={cs.CL},
      url={https://arxiv.org/abs/2502.02046}, 
}

@misc{wei2025autotirautonomoustoolsintegrated,
      title={AutoTIR: Autonomous Tools Integrated Reasoning via Reinforcement Learning}, 
      author={Yifan Wei and Xiaoyan Yu and Yixuan Weng and Tengfei Pan and Angsheng Li and Li Du},
      year={2025},
      eprint={2507.21836},
      archivePrefix={arXiv},
      primaryClass={cs.CL},
      url={https://arxiv.org/abs/2507.21836}, 
}

@misc{peng2023checkfactstryagain,
      title={Check Your Facts and Try Again: Improving Large Language Models with External Knowledge and Automated Feedback}, 
      author={Baolin Peng and Michel Galley and Pengcheng He and Hao Cheng and Yujia Xie and Yu Hu and Qiuyuan Huang and Lars Liden and Zhou Yu and Weizhu Chen and Jianfeng Gao},
      year={2023},
      eprint={2302.12813},
      archivePrefix={arXiv},
      primaryClass={cs.CL},
      url={https://arxiv.org/abs/2302.12813}, 
}

@misc{lee2024exploreselectderiverecall,
      title={Explore, Select, Derive, and Recall: Augmenting LLM with Human-like Memory for Mobile Task Automation}, 
      author={Sunjae Lee and Junyoung Choi and Jungjae Lee and Munim Hasan Wasi and Hojun Choi and Steven Y. Ko and Sangeun Oh and Insik Shin},
      year={2024},
      eprint={2312.03003},
      archivePrefix={arXiv},
      primaryClass={cs.HC},
      url={https://arxiv.org/abs/2312.03003}, 
}

@inproceedings{zhang-etal-2024-codeagent,
    title = "{C}ode{A}gent: Enhancing Code Generation with Tool-Integrated Agent Systems for Real-World Repo-level Coding Challenges",
    author = "Zhang, Kechi  and
      Li, Jia  and
      Li, Ge  and
      Shi, Xianjie  and
      Jin, Zhi",
    editor = "Ku, Lun-Wei  and
      Martins, Andre  and
      Srikumar, Vivek",
    booktitle = "Proceedings of the 62nd Annual Meeting of the Association for Computational Linguistics (Volume 1: Long Papers)",
    month = aug,
    year = "2024",
    address = "Bangkok, Thailand",
    publisher = "Association for Computational Linguistics",
    url = "https://aclanthology.org/2024.acl-long.737/",
    doi = "10.18653/v1/2024.acl-long.737",
    pages = "13643--13658"
}

@misc{tran2025primeplanningretrievalintegratedmemory,
      title={PRIME: Planning and Retrieval-Integrated Memory for Enhanced Reasoning}, 
      author={Hieu Tran and Zonghai Yao and Nguyen Luong Tran and Zhichao Yang and Feiyun Ouyang and Shuo Han and Razieh Rahimi and Hong Yu},
      year={2025},
      eprint={2509.22315},
      archivePrefix={arXiv},
      primaryClass={cs.AI},
      url={https://arxiv.org/abs/2509.22315}, 
}

@misc{ravuru2024agenticretrievalaugmentedgenerationtime,
      title={Agentic Retrieval-Augmented Generation for Time Series Analysis}, 
      author={Chidaksh Ravuru and Sagar Srinivas Sakhinana and Venkataramana Runkana},
      year={2024},
      eprint={2408.14484},
      archivePrefix={arXiv},
      primaryClass={cs.AI},
      url={https://arxiv.org/abs/2408.14484}, 
}

@misc{suzgun2025dynamiccheatsheettesttimelearning,
      title={Dynamic Cheatsheet: Test-Time Learning with Adaptive Memory}, 
      author={Mirac Suzgun and Mert Yuksekgonul and Federico Bianchi and Dan Jurafsky and James Zou},
      year={2025},
      eprint={2504.07952},
      archivePrefix={arXiv},
      primaryClass={cs.LG},
      url={https://arxiv.org/abs/2504.07952}, 
}

@inproceedings{wölflein2025llmagentsmakingagent,
  title={Llm agents making agent tools},
  author={W{\"o}lflein, Georg and Ferber, Dyke and Truhn, Daniel and Arandjelovic, Ognjen and Kather, Jakob Nikolas},
  booktitle={Proceedings of the 63rd Annual Meeting of the Association for Computational Linguistics (Volume 1: Long Papers)},
  pages={26092--26130},
  year={2025}
}

@inproceedings{
cai2024largelanguagemodelstool,
title={Large Language Models as Tool Makers},
author={Tianle Cai and Xuezhi Wang and Tengyu Ma and Xinyun Chen and Denny Zhou},
booktitle={The Twelfth International Conference on Learning Representations},
year={2024},
url={https://openreview.net/forum?id=qV83K9d5WB}
}

@misc{qiu2025alitagselfevolvinggenerativeagent,
      title={Alita-G: Self-Evolving Generative Agent for Agent Generation}, 
      author={Jiahao Qiu and Xuan Qi and Hongru Wang and Xinzhe Juan and Yimin Wang and Zelin Zhao and Jiayi Geng and Jiacheng Guo and Peihang Li and Jingzhe Shi and Shilong Liu and Mengdi Wang},
      year={2025},
      eprint={2510.23601},
      archivePrefix={arXiv},
      primaryClass={cs.AI},
      url={https://arxiv.org/abs/2510.23601}, 
}

@inproceedings{
yuan2024craftcustomizingllmscreating,
title={{CRAFT}: Customizing {LLM}s by Creating and Retrieving from Specialized Toolsets},
author={Lifan Yuan and Yangyi Chen and Xingyao Wang and Yi Fung and Hao Peng and Heng Ji},
booktitle={The Twelfth International Conference on Learning Representations},
year={2024},
url={https://openreview.net/forum?id=G0vdDSt9XM}
}

@inproceedings{qian2024creatortoolcreationdisentangling,
    title = "{CREATOR}: Tool Creation for Disentangling Abstract and Concrete Reasoning of Large Language Models",
    author = "Qian, Cheng  and
      Han, Chi  and
      Fung, Yi  and
      Qin, Yujia  and
      Liu, Zhiyuan  and
      Ji, Heng",
    editor = "Bouamor, Houda  and
      Pino, Juan  and
      Bali, Kalika",
    booktitle = "Findings of the Association for Computational Linguistics: EMNLP 2023",
    month = dec,
    year = "2023",
    address = "Singapore",
    publisher = "Association for Computational Linguistics",
    url = "https://aclanthology.org/2023.findings-emnlp.462/",
    doi = "10.18653/v1/2023.findings-emnlp.462",
    pages = "6922--6939",
}

@inproceedings{
lu2025orchdagcomplextoolorchestration,
title={Orch{DAG}: Complex Tool Orchestration in Multi-Turn Interactions with Plan {DAG}s},
author={Yifu Lu and Shengjie Liu and Li Dong},
booktitle={First Workshop on Multi-Turn Interactions in Large Language Models},
year={2025},
url={https://openreview.net/forum?id=uZE8mTYvHE}
}

@article{antony2024causal,
  title={Causal and chronological relationships predict memory organization for nonlinear narratives},
  author={Antony, James and Lozano, Angelo and Dhoat, Pahul and Chen, Janice and Bennion, Kelly},
  journal={Journal of cognitive neuroscience},
  volume={36},
  number={11},
  pages={2368--2385},
  year={2024},
  publisher={MIT Press 255 Main Street, 9th Floor, Cambridge, Massachusetts 02142, USA~…}
}

@misc{zeng2024structuralmemoryllmagents,
      title={On the Structural Memory of LLM Agents}, 
      author={Ruihong Zeng and Jinyuan Fang and Siwei Liu and Zaiqiao Meng},
      year={2024},
      eprint={2412.15266},
      archivePrefix={arXiv},
      primaryClass={cs.CL},
      url={https://arxiv.org/abs/2412.15266}, 
}

@inproceedings{
long2025seeinglisteningrememberingreasoning,
title={Seeing, Listening, Remembering, and Reasoning: A Multimodal Agent with Long-Term Memory},
author={Lin Long and Yichen He and Wentao Ye and Yiyuan Pan and Yuan Lin and Hang Li and Junbo Zhao and Wei Li},
booktitle={The Fourteenth International Conference on Learning Representations},
year={2026},
url={https://openreview.net/forum?id=PMz29A7Muq}
}

@inproceedings{cao2025infiniteiclbreakinglimitcontext,
  title={Infiniteicl: Breaking the limit of context window size via long short-term memory transformation},
  author={Cao, Bowen and Cai, Deng and Lam, Wai},
  booktitle={Findings of the Association for Computational Linguistics: ACL 2025},
  pages={11402--11415},
  year={2025}
}

@inproceedings{
lee2024humaninspiredreadingagentgist,
title={A Human-Inspired Reading Agent with Gist Memory of Very Long Contexts},
author={Kuang-Huei Lee and Xinyun Chen and Hiroki Furuta and John Canny and Ian Fischer},
booktitle={Forty-first International Conference on Machine Learning},
year={2024},
url={https://openreview.net/forum?id=OTmcsyEO5G}
}

@inproceedings{
lanchantin2023learningreasonmemorizeselfnotes,
title={Learning to Reason and Memorize with Self-Notes},
author={Jack Lanchantin and Shubham Toshniwal and Jason E Weston and Arthur Szlam and Sainbayar Sukhbaatar},
booktitle={Thirty-seventh Conference on Neural Information Processing Systems},
year={2023},
url={https://openreview.net/forum?id=ZFwNdsDCRL}
}

@misc{xu2025amemagenticmemoryllm,
      title={A-MEM: Agentic Memory for LLM Agents}, 
      author={Wujiang Xu and Zujie Liang and Kai Mei and Hang Gao and Juntao Tan and Yongfeng Zhang},
      year={2025},
      eprint={2502.12110},
      archivePrefix={arXiv},
      primaryClass={cs.CL},
      url={https://arxiv.org/abs/2502.12110}, 
}

@inproceedings{
liang2025selfevolvingagentsreflectivememoryaugmented,
title={Self-evolving Agents with reflective and memory-augmented abilities},
author={Xuechen Liang and Yangfan He and Yinghui Xia and Xinyuan Song and Meiling Tao and Kuan Lu and Jianhui Wang and Li Sun and Xinhang Yuan and Keqin Li and Jiaqi Chen and TIANYU SHI and Yang Jingsong},
booktitle={LLM-based Multi-Agent Systems: Towards Responsible, Reliable, and Scalable Agentic Systems},
year={2026},
url={https://openreview.net/forum?id=6Mw2fO3ejN}
}

@inproceedings{park2023generativeagentsinteractivesimulacra,
author = {Park, Joon Sung and O'Brien, Joseph and Cai, Carrie Jun and Morris, Meredith Ringel and Liang, Percy and Bernstein, Michael S.},
title = {Generative Agents: Interactive Simulacra of Human Behavior},
year = {2023},
isbn = {9798400701320},
publisher = {Association for Computing Machinery},
address = {New York, NY, USA},
url = {https://doi.org/10.1145/3586183.3606763},
doi = {10.1145/3586183.3606763},
booktitle = {Proceedings of the 36th Annual ACM Symposium on User Interface Software and Technology},
articleno = {2},
numpages = {22},
location = {San Francisco, CA, USA},
series = {UIST '23}
}

@InProceedings{wang2024agentworkflowmemory,
  title = 	 {Agent Workflow Memory},
  author =       {Wang, Zora Zhiruo and Mao, Jiayuan and Fried, Daniel and Neubig, Graham},
  booktitle = 	 {Proceedings of the 42nd International Conference on Machine Learning},
  pages = 	 {63897--63911},
  year = 	 {2025},
  editor = 	 {Singh, Aarti and Fazel, Maryam and Hsu, Daniel and Lacoste-Julien, Simon and Berkenkamp, Felix and Maharaj, Tegan and Wagstaff, Kiri and Zhu, Jerry},
  volume = 	 {267},
  series = 	 {Proceedings of Machine Learning Research},
  month = 	 {13--19 Jul},
  url = 	 {https://proceedings.mlr.press/v267/wang25bx.html},

}

@misc{ouyang2025reasoningbankscalingagentselfevolving,
      title={ReasoningBank: Scaling Agent Self-Evolving with Reasoning Memory}, 
      author={Siru Ouyang and Jun Yan and I-Hung Hsu and Yanfei Chen and Ke Jiang and Zifeng Wang and Rujun Han and Long T. Le and Samira Daruki and Xiangru Tang and Vishy Tirumalashetty and George Lee and Mahsan Rofouei and Hangfei Lin and Jiawei Han and Chen-Yu Lee and Tomas Pfister},
      year={2025},
      eprint={2509.25140},
      archivePrefix={arXiv},
      primaryClass={cs.AI},
      url={https://arxiv.org/abs/2509.25140}, 
}

@inproceedings{salama2025meminsightautonomousmemoryaugmentation,
    title = "{M}em{I}nsight: Autonomous Memory Augmentation for {LLM} Agents",
    author = "Salama, Rana  and
      Cai, Jason  and
      Yuan, Michelle  and
      Currey, Anna  and
      Sunkara, Monica  and
      Zhang, Yi  and
      Benajiba, Yassine",
    editor = "Christodoulopoulos, Christos  and
      Chakraborty, Tanmoy  and
      Rose, Carolyn  and
      Peng, Violet",
    booktitle = "Proceedings of the 2025 Conference on Empirical Methods in Natural Language Processing",
    month = nov,
    year = "2025",
    address = "Suzhou, China",
    publisher = "Association for Computational Linguistics",
    url = "https://aclanthology.org/2025.emnlp-main.1683/",
    doi = "10.18653/v1/2025.emnlp-main.1683",
    pages = "33136--33152",
    ISBN = "979-8-89176-332-6",
}

@misc{zhang2025memgenweavinggenerativelatent,
      title={MemGen: Weaving Generative Latent Memory for Self-Evolving Agents}, 
      author={Guibin Zhang and Muxin Fu and Shuicheng Yan},
      year={2025},
      eprint={2509.24704},
      archivePrefix={arXiv},
      primaryClass={cs.CL},
      url={https://arxiv.org/abs/2509.24704}, 
}

@article{schmidhuber2003godel,
  title={G{\"o}del machines: self-referential universal problem solvers making provably optimal self-improvements},
  author={Schmidhuber, J{\"u}rgen},
  journal={arXiv preprint cs/0309048},
  year={2003}
}

@article{zhuge2026neural,
  title={Neural computers},
  author={Zhuge, Mingchen and Zhao, Changsheng and Liu, Haozhe and Zhou, Zijian and Liu, Shuming and Wang, Wenyi and Chang, Ernie and Lan, Gael Le and Fei, Junjie and Zhang, Wenxuan and others},
  journal={arXiv preprint arXiv:2604.06425},
  year={2026}
}

@inproceedings{nanbo2025facts,
  title={FACTS: A factored state-space framework for world modelling},
  author={Nanbo, Li and Laakom, Firas and Xu, Yucheng and Wang, Wenyi and Schmidhuber, J{\"u}rgen},
  booktitle={International Conference on Learning Representations},
  volume={2025},
  pages={68955--68983},
  year={2025}
}

@misc{rivard2026neuralossimulatingoperatingsystems,
      title={NeuralOS: Towards Simulating Operating Systems via Neural Generative Models}, 
      author={Luke Rivard and Sun Sun and Hongyu Guo and Wenhu Chen and Yuntian Deng},
      year={2026},
      eprint={2507.08800},
      archivePrefix={arXiv},
      primaryClass={cs.CV},
      url={https://arxiv.org/abs/2507.08800}, 
}

@incollection{steunebrink2012towards,
  title={Towards an actual g{\"o}del machine implementation: A lesson in self-reflective systems},
  author={Steunebrink, Bas R and Schmidhuber, J{\~A}$1/4$rgen},
  booktitle={Theoretical Foundations of Artificial General Intelligence},
  pages={173--195},
  year={2012},
  publisher={Springer}
}

@article{chen2026much,
  title={How Much Can We Trust LLM Search Agents? Measuring Endorsement Vulnerability to Web Content Manipulation},
  author={Chen, Yimeng and Ren, Zhe and Laakom, Firas and Li, Yu and Guo, Dandan and Schmidhuber, J{\"u}rgen},
  journal={arXiv preprint arXiv:2606.16821},
  year={2026}
}

@article{yampolskiy2015seed,
  title={From seed AI to technological singularity via recursively self-improving software},
  author={Yampolskiy, Roman V},
  journal={arXiv preprint arXiv:1502.06512},
  year={2015}
}

@misc{
lyu2025correctionmasteryreinforceddistillation,
title={From Correction to Mastery: Reinforced Distillation of Large Language Model Agents},
author={Yuanjie Lyu and Chengyu Wang and Jun Huang and Tong Xu},
year={2026},
url={https://openreview.net/forum?id=n4Er2o4BFB}
}

@misc{qiu2025agentdistilltrainingfreeagentdistillation,
      title={AgentDistill: Training-Free Agent Distillation with Generalizable MCP Boxes}, 
      author={Jiahao Qiu and Xinzhe Juan and Yimin Wang and Ling Yang and Xuan Qi and Tongcheng Zhang and Jiacheng Guo and Yifu Lu and Zixin Yao and Hongru Wang and Shilong Liu and Xun Jiang and Liu Leqi and Mengdi Wang},
      year={2025},
      eprint={2506.14728},
      archivePrefix={arXiv},
      primaryClass={cs.AI},
      url={https://arxiv.org/abs/2506.14728}, 
}

@inproceedings{dou2025rerestreflectionreinforcedselftraininglanguage,
    title = "Re-{R}e{ST}: Reflection-Reinforced Self-Training for Language Agents",
    author = "Dou, Zi-Yi  and
      Yang, Cheng-Fu  and
      Wu, Xueqing  and
      Chang, Kai-Wei  and
      Peng, Nanyun",
    editor = "Al-Onaizan, Yaser  and
      Bansal, Mohit  and
      Chen, Yun-Nung",
    booktitle = "Proceedings of the 2024 Conference on Empirical Methods in Natural Language Processing",
    month = nov,
    year = "2024",
    address = "Miami, Florida, USA",
    publisher = "Association for Computational Linguistics",
    url = "https://aclanthology.org/2024.emnlp-main.861/",
    doi = "10.18653/v1/2024.emnlp-main.861",
    pages = "15394--15411",

}

@article{wu2024avataroptimizingllmagents,
  title={Avatar: Optimizing llm agents for tool usage via contrastive reasoning},
  author={Wu, Shirley and Zhao, Shiyu and Huang, Qian and Huang, Kexin and Yasunaga, Michihiro and Cao, Kaidi and Ioannidis, Vassilis and Subbian, Karthik and Leskovec, Jure and Zou, James Y},
  journal={Advances in Neural Information Processing Systems},
  volume={37},
  pages={25981--26010},
  year={2024}
}

@inproceedings{orseau2011self,
  title={Self-modification and mortality in artificial agents},
  author={Orseau, Laurent and Ring, Mark},
  booktitle={International Conference on Artificial General Intelligence},
  pages={1--10},
  year={2011},
  organization={Springer}
}

@misc{haque2025advancedtoollearningselection,
      title={Advanced Tool Learning and Selection System (ATLASS): A Closed-Loop Framework Using LLM}, 
      author={Mohd Ariful Haque and Justin Williams and Sunzida Siddique and Md. Hujaifa Islam and Hasmot Ali and Kishor Datta Gupta and Roy George},
      year={2025},
      eprint={2503.10071},
      archivePrefix={arXiv},
      primaryClass={cs.AI},
      url={https://arxiv.org/abs/2503.10071}, 
}

@misc{wu2024oscopilotgeneralistcomputeragents,
      title={OS-Copilot: Towards Generalist Computer Agents with Self-Improvement}, 
      author={Zhiyong Wu and Chengcheng Han and Zichen Ding and Zhenmin Weng and Zhoumianze Liu and Shunyu Yao and Tao Yu and Lingpeng Kong},
      year={2024},
      eprint={2402.07456},
      archivePrefix={arXiv},
      primaryClass={cs.AI},
      url={https://arxiv.org/abs/2402.07456}, 
}

@misc{zhao2025pyvisionagenticvisiondynamic,
      title={PyVision: Agentic Vision with Dynamic Tooling}, 
      author={Shitian Zhao and Haoquan Zhang and Shaoheng Lin and Ming Li and Qilong Wu and Kaipeng Zhang and Chen Wei},
      year={2025},
      eprint={2507.07998},
      archivePrefix={arXiv},
      primaryClass={cs.CL},
      url={https://arxiv.org/abs/2507.07998}, 
}

@misc{zheng2025skillweaverwebagentsselfimprove,
      title={SkillWeaver: Web Agents can Self-Improve by Discovering and Honing Skills}, 
      author={Boyuan Zheng and Michael Y. Fatemi and Xiaolong Jin and Zora Zhiruo Wang and Apurva Gandhi and Yueqi Song and Yu Gu and Jayanth Srinivasa and Gaowen Liu and Graham Neubig and Yu Su},
      year={2025},
      eprint={2504.07079},
      archivePrefix={arXiv},
      primaryClass={cs.AI},
      url={https://arxiv.org/abs/2504.07079}, 
}

@misc{jin2025stellaselfevolvingllmagent,
      title={STELLA: Self-Evolving LLM Agent for Biomedical Research}, 
      author={Ruofan Jin and Zaixi Zhang and Mengdi Wang and Le Cong},
      year={2025},
      eprint={2507.02004},
      archivePrefix={arXiv},
      primaryClass={cs.AI},
      url={https://arxiv.org/abs/2507.02004}, 
}

@misc{qian2025metaagentselfevolvingagenttool,
      title={MetaAgent: Toward Self-Evolving Agent via Tool Meta-Learning}, 
      author={Hongjin Qian and Zheng Liu},
      year={2025},
      eprint={2508.00271},
      archivePrefix={arXiv},
      primaryClass={cs.AI},
      url={https://arxiv.org/abs/2508.00271}, 
}

@misc{zhang2025enhancinglanguageagentstrategic,
      title={Enhancing Language Agent Strategic Reasoning through Self-Play in Adversarial Games}, 
      author={Yikai Zhang and Ye Rong and Siyu Yuan and Jiangjie Chen and Jian Xie and Yanghua Xiao},
      year={2025},
      eprint={2510.16761},
      archivePrefix={arXiv},
      primaryClass={cs.CL},
      url={https://arxiv.org/abs/2510.16761}, 
}

@inproceedings{
liu2025spiralselfplayzerosumgames,
title={{SPIRAL}: Self-Play on Zero-Sum Games Incentivizes Reasoning via Multi-Agent Multi-Turn Reinforcement Learning},
author={Bo Liu and Simon Yu and Zichen Liu and Leon Guertler and Penghui Qi and Daniel Balcells and Mickel Liu and Cheston Tan and Weiyan Shi and Min Lin and Wee Sun Lee and Natasha Jaques},
booktitle={The Fourteenth International Conference on Learning Representations},
year={2026},
url={https://openreview.net/forum?id=7Yayy5fNLg}
}

@inproceedings{
cheng2025selfplayingadversariallanguagegame,
title={Self-playing Adversarial Language Game Enhances {LLM} Reasoning},
author={Pengyu Cheng and Tianhao Hu and Han Xu and Zhisong Zhang and Yong Dai and Lei Han and nan du and Xiaolong Li},
booktitle={The Thirty-eighth Annual Conference on Neural Information Processing Systems},
year={2024},
url={https://openreview.net/forum?id=oCGkSH7ys2}
}

@inproceedings{
nottingham2024skillsetoptimizationreinforcing,
title={Skill Set Optimization: Reinforcing Language Model Behavior via Transferable Skills},
author={Kolby Nottingham and Bodhisattwa Prasad Majumder and Bhavana Dalvi Mishra and Sameer Singh and Peter Clark and Roy Fox},
booktitle={Forty-first International Conference on Machine Learning},
year={2024},
url={https://openreview.net/forum?id=9laB7ytoMp}
}

@misc{liu2025odysseyempoweringminecraftagents,
      title={Odyssey: Empowering Minecraft Agents with Open-World Skills}, 
      author={Shunyu Liu and Yaoru Li and Kongcheng Zhang and Zhenyu Cui and Wenkai Fang and Yuxuan Zheng and Tongya Zheng and Mingli Song},
      year={2025},
      eprint={2407.15325},
      archivePrefix={arXiv},
      primaryClass={cs.AI},
      url={https://arxiv.org/abs/2407.15325}, 
}

@inproceedings{
fang2025serl,
title={Se{RL}: Self-play Reinforcement Learning for Large Language Models with Limited Data},
author={Wenkai Fang and Shunyu Liu and Yang Zhou and Kongcheng Zhang and Tongya Zheng and Kaixuan Chen and Mingli Song and Dacheng Tao},
booktitle={The Thirty-ninth Annual Conference on Neural Information Processing Systems},
year={2025},
url={https://openreview.net/forum?id=ZF93vyH9He}
}

@inproceedings{qiao2024autoactautomaticagentlearning,
    title = "{A}uto{A}ct: Automatic Agent Learning from Scratch for {QA} via Self-Planning",
    author = "Qiao, Shuofei  and
      Zhang, Ningyu  and
      Fang, Runnan  and
      Luo, Yujie  and
      Zhou, Wangchunshu  and
      Jiang, Yuchen  and
      Lv, Chengfei  and
      Chen, Huajun",
    editor = "Ku, Lun-Wei  and
      Martins, Andre  and
      Srikumar, Vivek",
    booktitle = "Proceedings of the 62nd Annual Meeting of the Association for Computational Linguistics (Volume 1: Long Papers)",
    month = aug,
    year = "2024",
    address = "Bangkok, Thailand",
    publisher = "Association for Computational Linguistics",
    url = "https://aclanthology.org/2024.acl-long.165/",
    doi = "10.18653/v1/2024.acl-long.165",
    pages = "3003--3021",
}

@misc{wang2023voyageropenendedembodiedagent,
      title={Voyager: An Open-Ended Embodied Agent with Large Language Models}, 
      author={Guanzhi Wang and Yuqi Xie and Yunfan Jiang and Ajay Mandlekar and Chaowei Xiao and Yuke Zhu and Linxi Fan and Anima Anandkumar},
      year={2023},
      eprint={2305.16291},
      archivePrefix={arXiv},
      primaryClass={cs.AI},
      url={https://arxiv.org/abs/2305.16291}, 
}

@misc{yuksekgonul2024textgradautomaticdifferentiationtext,
      title={TextGrad: Automatic "Differentiation" via Text}, 
      author={Mert Yuksekgonul and Federico Bianchi and Joseph Boen and Sheng Liu and Zhi Huang and Carlos Guestrin and James Zou},
      year={2024},
      eprint={2406.07496},
      archivePrefix={arXiv},
      primaryClass={cs.CL},
      url={https://arxiv.org/abs/2406.07496}, 
}

@misc{simonds2025ladder,
      title={{LADDER}: Self-Improving {LLMs} Through Recursive Problem Decomposition}, 
      author={Toby Simonds and Akira Yoshiyama},
      year={2025},
      eprint={2503.00735},
      archivePrefix={arXiv},
      primaryClass={cs.LG}
}

@inproceedings{azov2024selfimprovingcustomerreviewresponse,
    title = "Self-Improving Customer Review Response Generation Based on {LLM}s",
    author = "Azov, Guy  and
      Pelc, Tatiana  and
      Fledel Alon, Adi  and
      Kamhi, Gila",
    editor = "Malmasi, Shervin  and
      Fetahu, Besnik  and
      Ueffing, Nicola  and
      Rokhlenko, Oleg  and
      Agichtein, Eugene  and
      Guy, Ido",
    booktitle = "Proceedings of the Seventh Workshop on e-Commerce and NLP @ LREC-COLING 2024",
    month = may,
    year = "2024",
    address = "Torino, Italia",
    publisher = "ELRA and ICCL",
    url = "https://aclanthology.org/2024.ecnlp-1.5/",
    pages = "40--57"
}

@misc{gao2025promptalchemistautomatedllmtailored,
      title={The Prompt Alchemist: Automated LLM-Tailored Prompt Optimization for Test Case Generation}, 
      author={Shuzheng Gao and Chaozheng Wang and Cuiyun Gao and Xiaoqian Jiao and Chun Yong Chong and Shan Gao and Michael Lyu},
      year={2025},
      eprint={2501.01329},
      archivePrefix={arXiv},
      primaryClass={cs.SE},
      url={https://arxiv.org/abs/2501.01329}, 
}

@misc{yang2026ttcstesttimecurriculumsynthesis,
      title={TTCS: Test-Time Curriculum Synthesis for Self-Evolving}, 
      author={Chengyi Yang and Zhishang Xiang and Yunbo Tang and Zongpei Teng and Chengsong Huang and Fei Long and Yuhan Liu and Jinsong Su},
      year={2026},
      eprint={2601.22628},
      archivePrefix={arXiv},
      primaryClass={cs.LG},
      url={https://arxiv.org/abs/2601.22628}, 
}

@misc{xia2025livesweagentsoftwareengineeringagents,
      title={Live-SWE-agent: Can Software Engineering Agents Self-Evolve on the Fly?}, 
      author={Chunqiu Steven Xia and Zhe Wang and Yan Yang and Yuxiang Wei and Lingming Zhang},
      year={2025},
      eprint={2511.13646},
      archivePrefix={arXiv},
      primaryClass={cs.SE},
      url={https://arxiv.org/abs/2511.13646}, 
}

@article{
    chezelles2025browsergym,
    title={The BrowserGym Ecosystem for Web Agent Research},
    author={Thibault Le Sellier de Chezelles and Maxime Gasse and Alexandre Lacoste and Massimo Caccia and Alexandre Drouin and L{\'e}o Boisvert and Megh Thakkar and Tom Marty and Rim Assouel and Sahar Omidi Shayegan and Lawrence Keunho Jang and Xing Han L{\`u} and Ori Yoran and Dehan Kong and Frank F. Xu and Siva Reddy and Graham Neubig and Quentin Cappart and Russ Salakhutdinov and Nicolas Chapados},
    journal={Transactions on Machine Learning Research},
    issn={2835-8856},
    year={2025},
    url={https://openreview.net/forum?id=5298fKGmv3},
    note={Expert Certification}
}

@inproceedings{workarena2024,
    title = {{W}ork{A}rena: How Capable are Web Agents at Solving Common Knowledge Work Tasks?},
    author = {Drouin, Alexandre and Gasse, Maxime and Caccia, Massimo and Laradji, Issam H. and Del Verme, Manuel and Marty, Tom and Vazquez, David and Chapados, Nicolas and Lacoste, Alexandre},
    booktitle = {Proceedings of the 41st International Conference on Machine Learning},
    pages = {11642--11662},
    year = {2024},
    editor = {Salakhutdinov, Ruslan and Kolter, Zico and Heller, Katherine and Weller, Adrian and Oliver, Nuria and Scarlett, Jonathan and Berkenkamp, Felix},
    volume = {235},
    series = {Proceedings of Machine Learning Research},
    month = {21--27 Jul},
    publisher = {PMLR},
    url = {https://proceedings.mlr.press/v235/drouin24a.html},
}

@inproceedings{
drouin2024workarenacapablewebagents,
title={WorkArena: How Capable are Web Agents at Solving Common Knowledge Work Tasks?},
author={Alexandre Drouin and Maxime Gasse and Massimo Caccia and Issam H. Laradji and Manuel Del Verme and Tom Marty and David Vazquez and Nicolas Chapados and Alexandre Lacoste},
booktitle={Forty-first International Conference on Machine Learning},
year={2024},
url={https://openreview.net/forum?id=BRfqYrikdo}
}

@inproceedings{
zhou2024webarena,
title={WebArena: A Realistic Web Environment for Building Autonomous Agents},
author={Shuyan Zhou and Frank F. Xu and Hao Zhu and Xuhui Zhou and Robert Lo and Abishek Sridhar and Xianyi Cheng and Tianyue Ou and Yonatan Bisk and Daniel Fried and Uri Alon and Graham Neubig},
booktitle={The Twelfth International Conference on Learning Representations},
year={2024},
url={https://openreview.net/forum?id=oKn9c6ytLx}
}

@misc{garg2026savingswebenchbenchmarkmutation,
      title={Saving SWE-Bench: A Benchmark Mutation Approach for Realistic Agent Evaluation}, 
      author={Spandan Garg and Benjamin Steenhoek and Yufan Huang},
      year={2026},
      eprint={2510.08996},
      archivePrefix={arXiv},
      primaryClass={cs.SE},
      url={https://arxiv.org/abs/2510.08996}, 
}

@misc{shum2025swermexecutionfreefeedbacksoftware,
      title={SWE-RM: Execution-free Feedback For Software Engineering Agents}, 
      author={KaShun Shum and Binyuan Hui and Jiawei Chen and Lei Zhang and X. W. and Jiaxi Yang and Yuzhen Huang and Junyang Lin and Junxian He},
      year={2025},
      eprint={2512.21919},
      archivePrefix={arXiv},
      primaryClass={cs.CL},
      url={https://arxiv.org/abs/2512.21919}, 
}

@misc{golubev2025traininglongcontextmultiturnsoftware,
      title={Training Long-Context, Multi-Turn Software Engineering Agents with Reinforcement Learning}, 
      author={Alexander Golubev and Maria Trofimova and Sergei Polezhaev and Ibragim Badertdinov and Maksim Nekrashevich and Anton Shevtsov and Simon Karasik and Sergey Abramov and Andrei Andriushchenko and Filipp Fisin and Sergei Skvortsov and Boris Yangel},
      year={2025},
      eprint={2508.03501},
      archivePrefix={arXiv},
      primaryClass={cs.LG},
      url={https://arxiv.org/abs/2508.03501}, 
}

@inproceedings{
wei2025swerladvancingllmreasoning,
title={{SWE}-{RL}: Advancing {LLM} Reasoning via Reinforcement Learning on Open Software Evolution},
author={Yuxiang Wei and Olivier Duchenne and Jade Copet and Quentin Carbonneaux and LINGMING ZHANG and Daniel Fried and Gabriel Synnaeve and Rishabh Singh and Sida Wang},
booktitle={The Thirty-ninth Annual Conference on Neural Information Processing Systems},
year={2025},
url={https://openreview.net/forum?id=ULblO61XZ0}
}

@inproceedings{
lin2025seagentselfevolutiontrajectoryoptimization,
title={{SE}-Agent: Self-Evolution Trajectory Optimization in Multi-Step Reasoning with {LLM}-Based Agents},
author={Yifu Guo and Jiaye Lin and Huacan Wang and Yuzhen Han and Sen Hu and Ziyi Ni and Licheng Wang and Mingguang Chen},
booktitle={The Thirty-ninth Annual Conference on Neural Information Processing Systems},
year={2025},
url={https://openreview.net/forum?id=isATAFP71B}
}

@inproceedings{
wang2025huxleygodelmachinehumanlevelcoding,
title={Huxley-G{\textbackslash}''odel Machine: Human-Level Coding Agent Development by an Approximation of the Optimal Self-Improving Machine},
author={Wenyi Wang and Piotr Pi{\k{e}}kos and Li Nanbo and Firas Laakom and Yimeng Chen and Mateusz Ostaszewski and Mingchen Zhuge and J{\"u}rgen Schmidhuber},
booktitle={The Fourteenth International Conference on Learning Representations},
year={2026},
url={https://openreview.net/forum?id=T0EiEuhOOL}
}

@inproceedings{
zhang2025darwingodelmachineopenended,
title={Darwin G\"odel Machine: Open-Ended Evolution of Self-Improving Agents},
author={Jenny Zhang and Shengran Hu and Cong Lu and Robert Tjarko Lange and Jeff Clune},
booktitle={The Fourteenth International Conference on Learning Representations},
year={2026},
url={https://openreview.net/forum?id=pUpzQZTvGY}
}

@inproceedings{jiang2025importance,
  title={Importance weighting can help large language models self-improve},
  author={Jiang, Chunyang and Chan, Chi-Min and Xue, Wei and Liu, Qifeng and Guo, Yike},
  booktitle={Proceedings of the AAAI Conference on Artificial Intelligence},
  volume={39},
  pages={24257--24265},
  year={2025}
}

@inproceedings{yang2023large,
  title={Large language models as optimizers},
  author={Yang, Chengrun and Wang, Xuezhi and Lu, Yifeng and Liu, Hanxiao and Le, Quoc V and Zhou, Denny and Chen, Xinyun},
  booktitle={The Twelfth International Conference on Learning Representations},
  year={2023}
}

@inproceedings{
liu2025trulyselfimprovingagentsrequire,
title={Position: Truly Self-Improving Agents Require Intrinsic Metacognitive Learning},
author={Tennison Liu and Mihaela van der Schaar},
booktitle={Forty-second International Conference on Machine Learning Position Paper Track},
year={2025},
url={https://openreview.net/forum?id=4KhDd0Ozqe}
}

@inproceedings{pryzant2023automaticpromptoptimizationgradient,
    title = "Automatic Prompt Optimization with ``Gradient Descent'' and Beam Search",
    author = "Pryzant, Reid  and
      Iter, Dan  and
      Li, Jerry  and
      Lee, Yin  and
      Zhu, Chenguang  and
      Zeng, Michael",
    editor = "Bouamor, Houda  and
      Pino, Juan  and
      Bali, Kalika",
    booktitle = "Proceedings of the 2023 Conference on Empirical Methods in Natural Language Processing",
    month = dec,
    year = "2023",
    address = "Singapore",
    publisher = "Association for Computational Linguistics",
    url = "https://aclanthology.org/2023.emnlp-main.494/",
    doi = "10.18653/v1/2023.emnlp-main.494",
    pages = "7957--7968",
}

@inproceedings{
fernando2024promptbreeder,
title={Promptbreeder: Self-Referential Self-Improvement via Prompt Evolution},
author={Chrisantha Fernando and Dylan Sunil Banarse and Henryk Michalewski and Simon Osindero and Tim Rockt{\"a}schel},
booktitle={Forty-first International Conference on Machine Learning},
year={2024},
url={https://openreview.net/forum?id=9ZxnPZGmPU}
}

@inproceedings{
madaan2023self,
title={Self-Refine: Iterative Refinement with Self-Feedback},
author={Aman Madaan and Niket Tandon and Prakhar Gupta and Skyler Hallinan and Luyu Gao and Sarah Wiegreffe and Uri Alon and Nouha Dziri and Shrimai Prabhumoye and Yiming Yang and Shashank Gupta and Bodhisattwa Prasad Majumder and Katherine Hermann and Sean Welleck and Amir Yazdanbakhsh and Peter Clark},
booktitle={Thirty-seventh Conference on Neural Information Processing Systems},
year={2023},
url={https://openreview.net/forum?id=S37hOerQLB}
}

@inproceedings{deng2024humancenteredproactiveconversationalagents,
  title={Towards human-centered proactive conversational agents},
  author={Deng, Yang and Liao, Lizi and Zheng, Zhonghua and Yang, Grace Hui and Chua, Tat-Seng},
  booktitle={Proceedings of the 47th International ACM SIGIR Conference on Research and Development in Information Retrieval},
  pages={807--818},
  year={2024}
}
\bibliographystyle{colm2025_conference}

\clearpage
\appendix
\section*{Notation}
\label{sec:Notation}

Here we summarize the recurring symbols used throughout the survey and provide a compact guide to their meanings.

\begin{footnotesize}
\setlength{\tabcolsep}{5pt}
\renewcommand{\arraystretch}{1.13}

\begin{longtable}{
  >{\raggedright\arraybackslash}p{0.17\linewidth}
  >{\hspace{3.5em}\raggedright\arraybackslash}p{0.58\linewidth}
  >{\raggedright\arraybackslash}p{0.17\linewidth}
}
\caption{Summary of notation used in the survey.}
\label{tab:notation}\\

\toprule
\rowcolor{sia-color}
\textcolor{white}{\textbf{Symbol}} &
\textcolor{white}{\textbf{Description}} &
\textcolor{white}{\textbf{Reference}} \\
\midrule
\endfirsthead

\toprule
\rowcolor{sia-color}
\textcolor{white}{\textbf{Symbol}} &
\textcolor{white}{\textbf{Description}} &
\textcolor{white}{\textbf{Reference}} \\
\midrule
\endhead

\midrule
\multicolumn{3}{r}{\emph{Continued on next page}}\\
\endfoot

\bottomrule
\endlastfoot

\rowcolor{NotationHeader}
\multicolumn{3}{l}{\textbf{General agent formalism}}\\
\midrule
$\mathcal{A}_t$ & Agent configuration at iteration $t$. & Sec.~\ref{sec:Definitions} \\
$\mathcal{A}_{1:t}$ & Agent history up to iteration $t$. & Sec.~\ref{sec:Definitions} \\
$\theta_t$ & Foundation model parameters. & Sec.~\ref{sec:Definitions} \\
$\theta_{1:t}$ & Parameter history up to iteration $t$. & Sec.~\ref{sec:Definitions} \\
$\Sigma_t$ & Scaffold configuration. & Sec.~\ref{sec:Definitions} \\
$\Sigma_{1:t}$ & Scaffold history up to iteration $t$. & Sec.~\ref{sec:Definitions} \\
$p_t$ & Prompt or instruction component. & Sec.~\ref{sec:Definitions}, Sec.~\ref{sec:Prompt_Optimization} \\
$m_t$ & Memory component. & Sec.~\ref{sec:Definitions}, Sec.~\ref{sec:Memory} \\
$\mathcal{T}_t$ & Tool component or tool set. & Sec.~\ref{sec:Definitions}, Sec.~\ref{sec:Tool} \\
$g_t$ & Control, routing, or verification logic. & Sec.~\ref{sec:Definitions} \\
$\pi_{\theta_t,\Sigma_t}$ & Policy induced by model and scaffold. & Sec.~\ref{sec:Definitions} \\
$X_t$ & Input, observation, or task context. & Sec.~\ref{sec:Definitions} \\
$A_t$ & Agent action or output. & Sec.~\ref{sec:Definitions} \\
$\mathcal{C}_t$ & Task or deployment context. & Sec.~\ref{sec:Definitions} \\
$\mathcal{E}(\cdot)$ & Agent execution procedure. & Sec.~\ref{sec:Definitions} \\
$\mathcal{U}$ & General self-induced update operator. & Sec.~\ref{sec:Definitions} \\
$\mathcal{U}_{\theta}$ & Model-parameter update operator. & Sec.~\ref{sec:Definitions} \\
$\mathcal{U}_{\Sigma}$ & Scaffold update operator. & Sec.~\ref{sec:Definitions} \\
$\mathcal{S}_t$ & Learning or improvement signal. & Sec.~\ref{sec:A_Taxonomy_of_Self_Improvement_Mechanisms} \\
$\IMPROVE(\cdot)$ & Abstract improvement operator. & Sec.~\ref{sec:A_Taxonomy_of_Self_Improvement_Mechanisms} \\
$\IMPROVE_{\theta}(\cdot)$ & Foundation-model improvement operator. & Sec.~\ref{sec:Foundation_Model_Improvement} \\
$\IMPROVE_{\Sigma}(\cdot)$ & Scaffold improvement operator. & Sec.~\ref{sec:Scaffolding_Improvement} \\

\midrule
\rowcolor{NotationHeader}
\multicolumn{3}{l}{\textbf{Foundation model improvement}}\\
\midrule
$\mathcal{D}_t$ & Intrinsic generative demonstrations. & Sec.~\ref{sec:Intrinsic_Generative_Demonstrations} \\
$\mathcal{D}^{\mathrm{gen}}_t$ & Generated training dataset. & Sec.~\ref{sec:Intrinsic_Generative_Demonstrations} \\
$(x_i,y_i)$ & Generated input-output pair. & Sec.~\ref{sec:Intrinsic_Generative_Demonstrations} \\
$n_t$ & Number of generated instances. & Sec.~\ref{sec:Intrinsic_Generative_Demonstrations} \\
$\mathcal{P}_{\mathrm{gen}}$ & Agent-induced data distribution. & Sec.~\ref{sec:Intrinsic_Generative_Demonstrations} \\
$\Phi_t$ & Data filtering or weighting procedure. & Sec.~\ref{sec:Intrinsic_Generative_Demonstrations} \\
$\mathcal{L}(\theta;\mathcal{D}_t)$ & Training objective on $\mathcal{D}_t$. & Sec.~\ref{sec:Intrinsic_Generative_Demonstrations} \\
$\Omega(\theta,\theta_0)$ & Regularizer around reference parameters. & Sec.~\ref{sec:Intrinsic_Generative_Demonstrations} \\
$\lambda$ & Regularization coefficient. & Sec.~\ref{sec:Intrinsic_Generative_Demonstrations} \\
$\theta_0$ & Reference or initial checkpoint. & Sec.~\ref{sec:Intrinsic_Generative_Demonstrations} \\
$\theta_t^{(k)}$ & Inner-loop parameter state. & Sec.~\ref{sec:Intrinsic_Generative_Demonstrations} \\
$\mathcal{B}_t^{(k)}$ & Minibatch at inner step $k$. & Sec.~\ref{sec:Intrinsic_Generative_Demonstrations} \\
$\eta_k$ & Inner-loop step size. & Sec.~\ref{sec:Intrinsic_Generative_Demonstrations} \\
$K_t$ & Number of inner update steps. & Sec.~\ref{sec:Intrinsic_Generative_Demonstrations} \\
$\nabla_{\theta}$ & Gradient with respect to $\theta$. & Sec.~\ref{sec:Intrinsic_Generative_Demonstrations} \\
$\mathcal{Y}_t(x)$ & Candidate outputs for input $x$. & Sec.~\ref{sec:Intrinsic_Evaluative_Feedback} \\
$y_t^{(k)}$ & The $k$-th candidate output. & Sec.~\ref{sec:Intrinsic_Evaluative_Feedback} \\
$K$ & Number of candidate outputs. & Sec.~\ref{sec:Intrinsic_Evaluative_Feedback} \\
$\phi_t$ & Intrinsic evaluator or judge. & Sec.~\ref{sec:Intrinsic_Evaluative_Feedback} \\
$\kappa_t$ & Evaluation criterion or rubric. & Sec.~\ref{sec:Intrinsic_Evaluative_Feedback} \\
$e_t$ & Intrinsic evaluative feedback. & Sec.~\ref{sec:Intrinsic_Evaluative_Feedback} \\
$r_t$ & Scalar reward or reward-like signal. & Sec.~\ref{sec:Intrinsic_Evaluative_Feedback}, Sec.~\ref{sec:Extrinsic_Exploratory_Experience} \\
$y^+ \succ y^-$ & Preference relation. & Sec.~\ref{sec:Intrinsic_Evaluative_Feedback} \\
$c_t$ & Natural-language critique. & Sec.~\ref{sec:Intrinsic_Evaluative_Feedback}, Sec.~\ref{sec:Qualitative_Feedback_Refinement} \\
$y_t^\ast$ & Revised target output. & Sec.~\ref{sec:Intrinsic_Evaluative_Feedback} \\
$C(\cdot)$ & Consistency or confidence aggregator. & Sec.~\ref{sec:Intrinsic_Evaluative_Feedback} \\
$R(\cdot)$ & Critique-and-revision operator. & Sec.~\ref{sec:Intrinsic_Evaluative_Feedback} \\
$\tau_t$ & Interaction trajectory or experience. & Sec.~\ref{sec:Extrinsic_Exploratory_Experience} \\
$s$ & Environment state. & Sec.~\ref{sec:Extrinsic_Exploratory_Experience} \\
$a$ & Environment action. & Sec.~\ref{sec:Extrinsic_Exploratory_Experience} \\
$r$ & Environment reward. & Sec.~\ref{sec:Extrinsic_Exploratory_Experience} \\
$s'$ & Next environment state. & Sec.~\ref{sec:Extrinsic_Exploratory_Experience} \\
$\pi_{\theta_t,\Sigma_t}$ & Induced policy of the FM-based agent. & Sec.~\ref{sec:Extrinsic_Exploratory_Experience} \\
$W(s_{k+1},r_k\mid s_k,a_k)$ & Learned world model. & Sec.~\ref{sec:World_Models} \\

\midrule
\rowcolor{NotationHeader}
\multicolumn{3}{l}{\textbf{Scaffold improvement}}\\
\midrule
$\mathcal{P}$ & Prompt search space. & Sec.~\ref{sec:Scalar_Feedback_Optimization} \\
$p$ & Candidate prompt. & Sec.~\ref{sec:Prompt_Optimization} \\
$p^*$ & Best prompt under a score. & Sec.~\ref{sec:Scalar_Feedback_Optimization} \\
$f(p)$ & Scalar prompt score. & Sec.~\ref{sec:Scalar_Feedback_Optimization} \\
$p_{t+1}$ & Updated prompt. & Sec.~\ref{sec:Prompt_Optimization} \\
$\mathrm{Refine}(\cdot)$ & Prompt refinement operator. & Sec.~\ref{sec:Qualitative_Feedback_Refinement} \\
$\mathrm{Critique}(\cdot)$ & Critique function. & Sec.~\ref{sec:Qualitative_Feedback_Refinement} \\
$\mathrm{Output}(p_t)$ & Output generated under prompt $p_t$. & Sec.~\ref{sec:Qualitative_Feedback_Refinement} \\
$P_t$ & Prompt population. & Sec.~\ref{sec:Population_Based_Evolution} \\
$p_t^{(i)}$ & The $i$-th prompt in $P_t$. & Sec.~\ref{sec:Population_Based_Evolution} \\
$N$ & Population size. & Sec.~\ref{sec:Population_Based_Evolution} \\
$\mathrm{Fit}(p)$ & Prompt fitness function. & Sec.~\ref{sec:Population_Based_Evolution} \\
$p_{\mathrm{child}}$ & Offspring prompt. & Sec.~\ref{sec:Population_Based_Evolution} \\
$\mathrm{Crossover}(\cdot)$ & Prompt crossover operator. & Sec.~\ref{sec:Population_Based_Evolution} \\
$\mathrm{Mutate}(\cdot)$ & Prompt mutation operator. & Sec.~\ref{sec:Population_Based_Evolution} \\
$g(p_t)$ & Textual gradient or update guidance. & Sec.~\ref{sec:Textual_Gradient_Optimization} \\
$\oplus$ & Textual update operation. & Sec.~\ref{sec:Textual_Gradient_Optimization} \\
$\IMPROVE_p$ & Prompt improvement operator. & Sec.~\ref{sec:Prompt_Optimization} \\
$\IMPROVE_m$ & Memory improvement operator. & Sec.~\ref{sec:Memory} \\
$\text{object}_t$ & Stored memory objects. & Sec.~\ref{sec:Memory_Object} \\
$\text{structure}_t$ & Memory structure or index. & Sec.~\ref{sec:Memory_Structure} \\
$\mathrm{Create}$ & Memory write operation. & Sec.~\ref{sec:Memory_Processing} \\
$\mathrm{Read}$ & Memory retrieval operation. & Sec.~\ref{sec:Memory_Processing} \\
$\mathrm{Update}$ & Memory revision operation. & Sec.~\ref{sec:Memory_Processing} \\
$\mathrm{Delete}$ & Memory pruning operation. & Sec.~\ref{sec:Memory_Processing} \\
$\IMPROVE_{\mathcal{T}}$ & Tool improvement operator. & Sec.~\ref{sec:Tool} \\
$\mathcal{I}_{\Sigma_t}$ & Scaffold improver. & Sec.~\ref{sec:Full_Scaffolding} \\
$\langle \Sigma_t \rangle$ & Serializable scaffold encoding. & Sec.~\ref{sec:Full_Scaffolding} \\
$\mathrm{exec}(\cdot)$ & Execution of a scaffold encoding. & Sec.~\ref{sec:Full_Scaffolding} \\
$\tilde{\Sigma}_{t+1}$ & Proposed scaffold update. & Sec.~\ref{sec:Full_Scaffolding} \\
$\mathcal{V}(\cdot)$ & Scaffold validation function. & Sec.~\ref{sec:Full_Scaffolding} \\

\midrule
\rowcolor{NotationHeader}
\multicolumn{3}{l}{\textbf{Evaluation and iteration}}\\
\midrule
$t$ & Outer-loop iteration index. & Throughout \\
$t+1$ & Next self-improvement iteration. & Throughout \\
$k$ & Inner step or candidate index. & Sec.~\ref{sec:Intrinsic_Generative_Demonstrations}, Sec.~\ref{sec:Intrinsic_Evaluative_Feedback} \\
$\arg\max$ & Maximizer of an objective. & Sec.~\ref{sec:Scalar_Feedback_Optimization} \\
$\arg\min$ & Minimizer of an objective. & Sec.~\ref{sec:Intrinsic_Generative_Demonstrations} \\
$\langle \cdot \rangle$ & Serializable encoding notation. & Sec.~\ref{sec:Full_Scaffolding} \\

\end{longtable}
\end{footnotesize}

\end{document}